\documentclass[runningheads]{llncs}

\usepackage[mobile,year=2024]{eccv}

\usepackage{eccvabbrv}  %

\usepackage{graphicx}
\usepackage{booktabs}  %
\usepackage[accsupp]{axessibility}  %
\usepackage[utf8]{inputenc} %
\usepackage{amsfonts}       %
\usepackage{nicefrac}       %
\usepackage{microtype}      %
\usepackage{gensymb}
\usepackage{tabularx}
\usepackage{multirow}
\usepackage{svg}
\usepackage{cuted}
\usepackage{tabu}
\usepackage{pifont}  %
\newcommand{\cmark}{\ding{51}}%
\newcommand{\xmark}{\ding{55}}%
\usepackage{mwe}
\usepackage{bm}  %
\usepackage{textcomp}
\usepackage{makecell}  %
\usepackage{grffile} %
\usepackage{rotating}  %

\usepackage{colortbl}

\definecolor{eccvblue}{rgb}{0.12,0.49,0.85}
\usepackage[pagebackref,breaklinks,colorlinks,bookmarks=false,urlcolor=eccvblue,citecolor=eccvblue,hyperfootnotes=false]{hyperref}

\usepackage{siunitx}  %
\newcommand{\ours}{GeoCalib}  %
\newcommand{\ourdataset}{OpenPano} %

\newcommand{\comment}[1]{}

\renewcommand{\*}[1]{\bm{\mathrm{#1}}}
\renewcommand{\b}[1]{\textbf{#1}}
\newcommand{\red}[1]{\textcolor{red}{#1}}

\newcommand{\0}{\phantom{0}}
\newcommand{\mytilde}{{\raise.17ex\hbox{$\scriptstyle\sim$}}}

\renewcommand{\paragraph}[1]{\vskip4pt \noindent\textbf{#1}}

\definecolor{tabfirst}{rgb}{1, 0.7, 0.7} %
\definecolor{tabsecond}{rgb}{1, 0.85, 0.7} %
\definecolor{tabthird}{rgb}{1, 1, 0.7} %
\newcommand{\cthird}{}
\newcommand{\cfirst}{\cellcolor{tabfirst}\bfseries}
\newcommand{\csecond}{\cellcolor{tabsecond}}

\newlength{\pwidth}
\newlength{\lwidth}
\newlength{\bwidth}
\newlength{\iwidth}
    
\newcommand{\matrx}[1]{\begin{bmatrix}#1\end{bmatrix}}
\newcommand{\transp}{^{\top}}
\newcommand{\real}{\mathbb{R}}
\newcommand{\image}{\*I}
\newcommand{\pointw}{{\*P}}
\newcommand{\pointim}{{\*p}}
\newcommand{\focal}{f}
\newcommand{\ppoint}{\*c}
\newcommand{\dist}{\*k}
\newcommand{\distfn}{\mathcal{D}} %
\newcommand{\radius}{r}
\newcommand{\gravity}{\*g}
\newcommand{\camproj}{\Pi}
\newcommand{\imsize}{{H{\times}W}}
\newcommand{\camparams}{\*\theta}
\newcommand{\up}{\*u}
\newcommand{\latitude}{\varphi}
\newcommand{\ray}{\*n}
\newcommand{\confidence}{\sigma}
\newcommand{\lmstep}{\*\delta}
\newcommand{\hessian}{\*H}
\newcommand{\jacobian}{\*J}
\newcommand{\weight}{\*W}
\newcommand{\residual}{\*r}
\newcommand{\norm}[1]{\left\lVert#1\right\rVert}
\newcommand{\normsmall}[1]{\lVert#1\rVert}

\newif\ifaddsupp
\addsuppfalse
\addsupptrue

\begin{document}

\title{\ours: Learning Single-image Calibration\\with Geometric Optimization}

\author{Alexander Veicht\inst{1} \hspace{0.1in} Paul-Edouard Sarlin\inst{1} \hspace{0.1in} Philipp Lindenberger\inst{1} \\%
\vspace{0.05in}
Marc Pollefeys\inst{1,2}}

\authorrunning{A.~Veicht \etal}

\institute{
$^{1}$ ETH Zurich\hspace{0.1in}
$^{2}$ Microsoft Mixed Reality \& AI Lab
}

\newcommand{\supp}{supplemental} %

\maketitle
\begin{abstract}
From a single image, visual cues can help deduce intrinsic and extrinsic camera parameters like the focal length and the gravity direction.
This single-image calibration can benefit various downstream applications like image editing and 3D mapping.
Current approaches to this problem are based on either classical geometry with lines and vanishing points or on deep neural networks trained end-to-end. 
The learned approaches are more robust but struggle to generalize to new environments and are less accurate than their classical counterparts.
We hypothesize that they lack the constraints that 3D geometry provides.
In this work, we introduce \ours, a deep neural network that leverages universal rules of 3D geometry through an optimization process.
\ours~is trained end-to-end to estimate camera parameters and learns to find useful visual cues from the data.
Experiments on various benchmarks show that \ours~is more robust and more accurate than existing classical and learned approaches.
Its internal optimization estimates uncertainties, which help flag failure cases and benefit downstream applications like visual localization. 
The code and trained models are publicly available at \url{https://github.com/cvg/GeoCalib}.
\keywords{Camera calibration \and Deep learning \and Optimization}
\end{abstract}

\section{Introduction}
\label{sec:intro}

Camera calibration consists of estimating the intrinsic and extrinsic parameters of a camera.
This information is required for most image-based 3D applications, including metrology, 3D reconstruction, and novel view synthesis.
This problem has been extensively studied, and many tools based on 3D geometry are available~\cite{lochman2021babelcalib,maye2013self,scaramuzza2006toolbox}.
Since the process of image formation is well-understood, such tools can very accurately calibrate a camera from images taken in controlled lab conditions.
The calibration can also be estimated in uncontrolled conditions, which generally requires additional sensors or multiple images observing the same scene, using structure-from-motion~\cite{schoenberger2016sfm,agarwal2010bundle,moulon2016openmvg,opensfm} or SLAM~\cite{hagemann2023deep,zhuang2019degeneracy,keivan2015online}.

In some applications, multiple images of the same scene are not available, such as in image editing, or multi-view constraints are not sufficient to accurately estimate the camera parameters,
for example due to limited visual overlap across view.
This occurs frequently when dealing with in-the-wild, crowd-sourced imagery, where each image is captured by a different camera~\cite{snavely2008modeling,snavely2006photo,wilson_eccv2014_1dsfm,agarwal2011building}.
Visual cues visible in a single image can however help estimate some camera parameters, like the gravity direction, focal length, or distortion coefficients, without the need for multi-view cues.

Example of such geometric cues are straight lines, curves, and vanishing points.
Estimating camera parameters from them has been extensively studied~\cite{sva, Pautrat_2023_UncalibratedVP,bazin20123,wildenauer2012robust,aguilera2005new}.
Because we have a good model of projective geometry, these approaches are extremely accurate.
They are however limited to man-made environments in which straight lines are visible, and catastrophically fail when this condition is not met.
Such low robustness significantly impairs their wide adoption.

\begin{figure}[t]
    \centering
    \includegraphics[width=\linewidth]{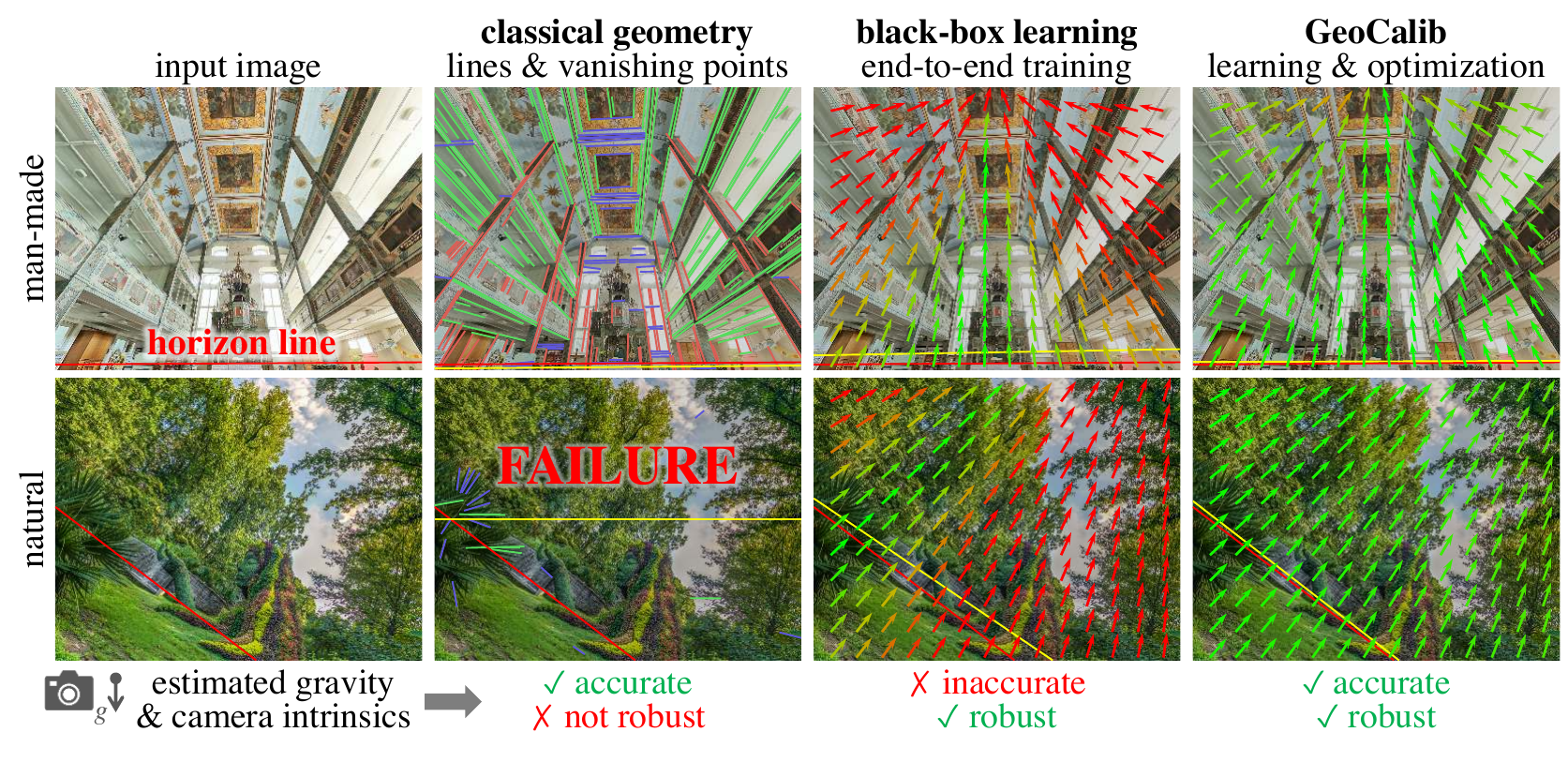} 
    \caption{\textbf{Learning \emph{vs}.\ geometry?}
    To estimate the camera calibration from a single image, classical approaches struggle with environments devoid of lines while deep networks are so far not as accurate.
    \ours~combines the best of both: its learns to steer an optimization using diverse geometric and semantic cues learned end-to-end.
    }%
    \label{fig:teaser}%
\end{figure}

Recent research has tackled the task of single-image calibration with deep neural networks (DNNs) trained in a supervised manner~\cite{lopez2019deepcalib,bogdan2018deepcalib,ctrlc,jin2022PerspectiveFields,Song2024MSCC}.
These approaches can leverage many more geometric and semantic cues and thus exhibit an impressively high robustness.
To generalize well to different environment, they however require large amounts of training data that is costly to acquire.
They are also far less accurate than their classical counterparts based on 3D geometry (\cref{fig:teaser}).
Intuitively, each deep network needs to relearn projective geometry from scratch when trained.
Given finite model capacity, this can only be approximated within the domain of the training data, without any guarantee outside.

In this work, we introduce \ours, a DNN that leverages our knowledge of projective geometry through an optimization process.
As this optimization is differentiable, \ours~learns end-to-end to estimate the vertical direction and the camera intrinsics given a single image.
Our approach can thus learn the right visual cues without explicit supervision but does not need to learn the process of estimating camera parameters, which is better achieved with knowledge of 3D geometry.
This improves the generalization to different environments with negligible overhead, which is important for practical applications.
Experiments on various benchmarks show that \ours~is both more robust and more accurate than existing classical and learned approaches.

Compared to black-box deep networks, \ours~has multiple practical benefits.
When a subset of the parameters is known, \eg the intrinsic parameters, \ours~can more accurately estimate the remaining parameters by leveraging this prior information.
This makes it possible to handle different camera models, such as pinhole and fisheye, without any retraining.
\ours~is also more interpretable: we can easily visualize the cues that it relies on, and the optimization uncertainties help flag failure cases and can benefit downstream applications.
To support this, we show that \ours~can readily improve the accuracy of visual positioning.

\section{Related work}
\label{sec:related-work}

\paragraph{Classical geometric approaches}
for single-image calibration rely on parallel lines~\cite{von2012lsd} that intersect at a vanishing point (VP) in the image. 
VP estimation in a single image can be categorized in two directions: i) unconstrained VP detection~\cite{bernard1983perspective,kluger2020consac,zhai2016detecting} which relies on coplanar lines and ii) approaches that assume a Manhattan world (3 orthogonal directions)~\cite{tong2022transformer,bazin20123,coughlan1999manhattan} that is often found in man-made environments but not in natural ones.
The vertical VP generally corresponds to the gravity direction.
VP detection has been extensively studied with approaches based on RANSAC~\cite{sva,bazin20123,wildenauer2012robust,aguilera2005new}, optimization~\cite{bazin2012globally, tretyak2012geometric}, exhaustive search~\cite{magee1984determining,qian2022reliable,bazin2012rotation}, and even deep learning~\cite{tong2022transformer,antunes2017unsupervised,liu2021vapid,zhou2019neurvps}.
Some approaches also recover the focal length~\cite{Pautrat_2023_UncalibratedVP,sva}
or the distortion parameters, using curved lines, circular arcs, or covariant regions~\cite{sva,pritts2020minimal}.
Our approach also solves an optimization based on low-level cues, including but not limited to lines, making it much more robust in natural and outdoor scenes.

\paragraph{Deep learning for single-image calibration}
has recently gained popularity.
Existing research is divided into two categories: direct and indirect methods.
Direct methods directly regress the camera parameters (focal length, distortion, horizon line) or classify them into bins~\cite{lopez2019deepcalib,perceptual,zhai2016detecting}.
This simple approach is typically less accurate than traditional methods because of missing geometric constraints.
Indirect methods aim to alleviate this issue by predicting observable, lower-level cues.
This includes classifying horizontal/vertical lines~\cite{ctrlc,Song2024MSCC} and regressing surface normals~\cite{xian2019uprightnet} or vector fields~\cite{jin2022PerspectiveFields}.
These cues are often used as auxiliary supervision~\cite{Song2024MSCC,ctrlc}
or as inputs to a decoder network that regresses camera parameters~\cite{jin2022PerspectiveFields}.
Compared to classical algorithms, DNNs are usually more robust because they do not assume any scene configuration.
They however fail to match their accuracy when sufficient geometric constraints are available.
In this work, we combine deep priors with robust optimization, which enables the network to focus on well-constrained priors, boosting the estimation accuracy.

\paragraph{End-to-end learning with differentiable optimization}
has been recently popular in geometric computer vision, including for 
pose estimation~\cite{brachmann2020dsacstar,sarlin21pixloc,chen2020bpnp,fu2023batch},
camera tracking~\cite{clark2018ls,lv2019taking,tang2018ba,xu2020deep,teed2021droid}, 
and image matching~\cite{sarlin2020superglue,campbell2020solving} or alignment~\cite{wang2018deep,pineda2022theseus}.
In these, a DNN typically predicts observations on the image level, based on which an optimization problem is solved.
It can be made differentiable by unrolling a fixed number of optimization steps or using implicit derivatives~\cite{russell2019fixing}.
The DNN can be trained by only supervising the optimized quantity.
The robustness to incorrect priors can be improved with carefully selected loss functions or by formulating the optimization through differentiable RANSAC~\cite{brachmann2019neural,bhowmik2020reinforced,wei2023generalized}.

For the task of single-image calibration, UprightNet~\cite{xian2019uprightnet} is closest to our work.
It aligns normals in camera and world frames to estimate the gravity direction by solving a simpler weighted least squares problem with closed form solution.
It cannot recover the camera intrinsics and its training requires ground-truth normals.
In contrast, we use cues that are solely derived from the camera parameters and sufficient to constraint all of them.
We show that the differentiable optimization improves both the generalization and the accuracy of \ours.

\begin{figure}[t]
    \centering
    \includegraphics[width=\linewidth]{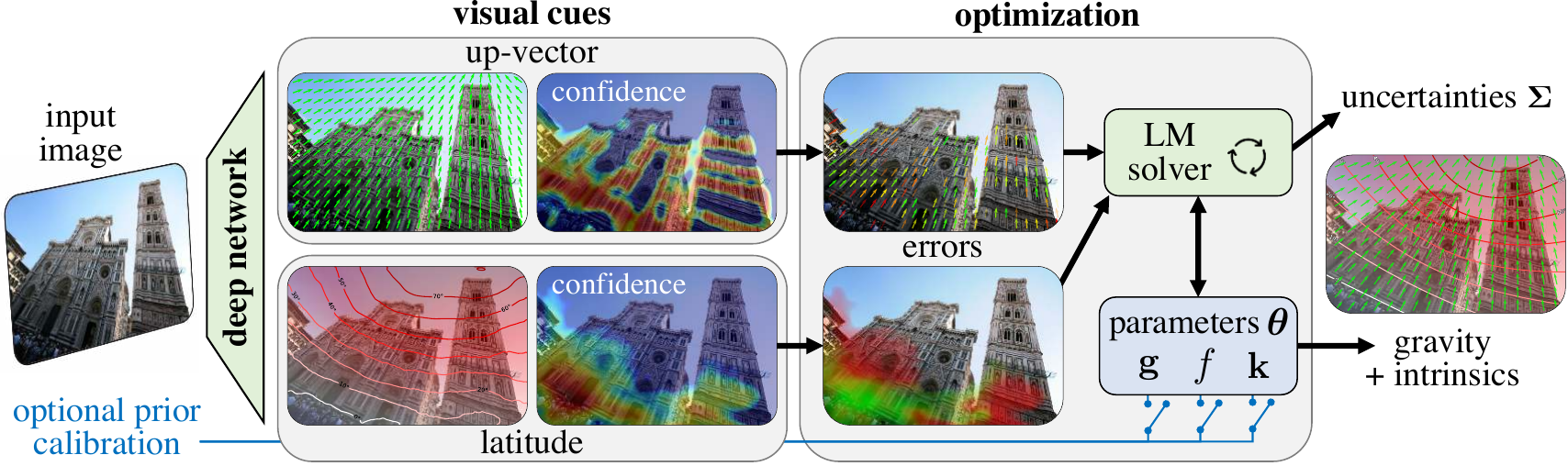}%
    \caption{\textbf{Architecture of \ours.}
    A DNN predicts a Perspectivel Field with confidences, to which camera parameters are fitted with a Levenberg-Marquardt optimization.
    \ours~is trained end-to-end by supervising the optimized parameters.
    Priors over some of them or a different distortion model can be easily included without retraining.
    }%
    \label{fig:architecture}%
\end{figure}

\section{Single-image calibration with \ours}
\label{sec:approach}

\paragraph{Camera calibration:}
A 3D point $\pointw = \matrx{X&Y&Z}\transp$, 
expressed in the coordinate frame of the camera, maps to an image point $\pointim\,{\in}\,\real^2$ with the projection function~$\camproj\left(\cdot\right)$.
Assuming square pixels and no skew, we write for a pinhole camera
$\camproj\left(\pointw\right) = \focal\matrx{u&v}\transp+\ppoint$,
where
$\matrx{u&v} = \frac{1}{Z}\matrx{X&Y}$ are the normalized image coordinates,
$\focal$ is the focal length in pixels,
and $\ppoint$ is the principal point, which can often be assumed to be at the center of the image unless it has been cropped.

To handle lens distortion, various models are commonly used~\cite{faugeras1992camera,pollefeys1999self,scaramuzza2006flexible,urban2015improved,usenko2018double,fitzgibbon2001simultaneous,kannala2006generic}.
Many of them, including those based on polynomials, transform $\matrx{u&v}$ into
$\matrx{u_d&v_d} := \distfn\left(\matrx{u&v}, \dist\right)$,
where $\dist\in\real^D$ are $D$ distortion parameters.
The undistortion function $\distfn^{-1}$ can be computed using an iterative process~\cite{ddvernay2008straight} or inverse parameters~\cite{drap2016exact}.
We generally refer to $\focal$, $\ppoint$, and $\dist$ as \emph{intrinsic} parameters.

The gravity direction in the camera coordinate frame is the unit vector $\gravity=\matrx{\gravity_x&\gravity_y&\gravity_z}$.
It can be decomposed into two Euler angles, the roll, and the pitch.
The third degree of freedom of the rotation, the yaw, is generally not observable from a single image unless one makes additional restrictive assumptions, \eg~Manhattan-world.
Our goal is to estimate $\gravity$, $\focal$, and $\dist$, collectively referred to as $\camparams$, given a single image $\image\in\real^\imsize$ as input.

\paragraph{Overview:}
\ours~first infers visual cues -- a Perspective Field, with associated confidences, from the input image with a DNN.
An iterative optimization then fits the camera parameters $\camparams$ to be consistent with these cues (\cref{fig:architecture}, \cref{sec:persp-alignment}).
The DNN is then trained end-to-end by supervising the result of the optimization.
We detail in \cref{sec:practical-benefits} the multiple practical benefits of this simple approach.

\subsection{Perspective alignment}
\label{sec:persp-alignment}

\paragraph{Visual cues:}
We choose the \emph{Perspective Field}~\cite{jin2022PerspectiveFields} as visual cues driving the optimization.
It is an over-parameterized, pixel-wise representation of the camera parameters and consists, for each pixel $\pointim$, of a unit up-vector $\up_\pointim$ and a latitude $\latitude_\pointim$.
The up-vector is the projection of the up direction at $\pointw$,
while
the latitude $\latitude_\pointim$ is the angle between the ray $\ray=\matrx{u&v&1}\transp$ at pixel $\pointim$ and the horizontal plane:
\begin{equation}
    \up_{\pointim} = \lim_{t \rightarrow 0} 
\frac{\camproj(\pointw- t\gravity) - \camproj(\pointw)}
{\norm{\camproj(\pointw- t\gravity) - \camproj(\pointw)}_2}
\quad
\quad
\quad
\latitude_{\pointim} = \arcsin \left( \frac{\ray\transp \gravity}{\norm{\ray}_2}\right)
\label{eq:up}
\end{equation}

This representation is visually observable: 
the up-vector corresponds to the direction of vertical lines, and the latitude is $0\degree$ on the horizon.
It is not only geometric, but also semantic: the up-vector can be deduced from objects known to have a canonical orientation, like trees and humans.
Unlike purely geometric cues like lines, it can be inferred in a wider range of environments.
It is applicable to arbitrary camera models, though Jin~\etal~\cite{jin2022PerspectiveFields} mainly focused on perspective projection.
Here we also apply it to distorted images.

We predict a pixel-wise Perspective Field $(\hat\up_\pointim, \hat\latitude_\pointim)$ from the input image using a DNN.
We employ a SegNeXt~\cite{guo2022segnext} encoder-decoder architecture, modified such that the outputs have the same spatial resolutions as the input.
This provides more accurate constraints for the subsequent optimization.

\begin{figure}[t]
    \centering
    
    \def\ncols{4}
    \setlength{\pwidth}{0.005\linewidth}
    \setlength{\lwidth}{0.023\linewidth}
    \setlength{\bwidth}{0.0155\linewidth}
    \setlength{\iwidth}{\dimexpr(0.999\linewidth - \ncols\pwidth -\pwidth - \lwidth - \bwidth)/\ncols \relax}

    \begin{minipage}{\dimexpr \linewidth - \bwidth \relax}
    \begin{minipage}[t]{\lwidth}
    \rotatebox[origin=c]{90}{up-vector $\confidence_{\up}$}
    \end{minipage}%
    \begin{minipage}{\dimexpr \ncols\iwidth + \ncols\pwidth \relax}
    \hspace{\pwidth}%
    \includegraphics[width=\iwidth]{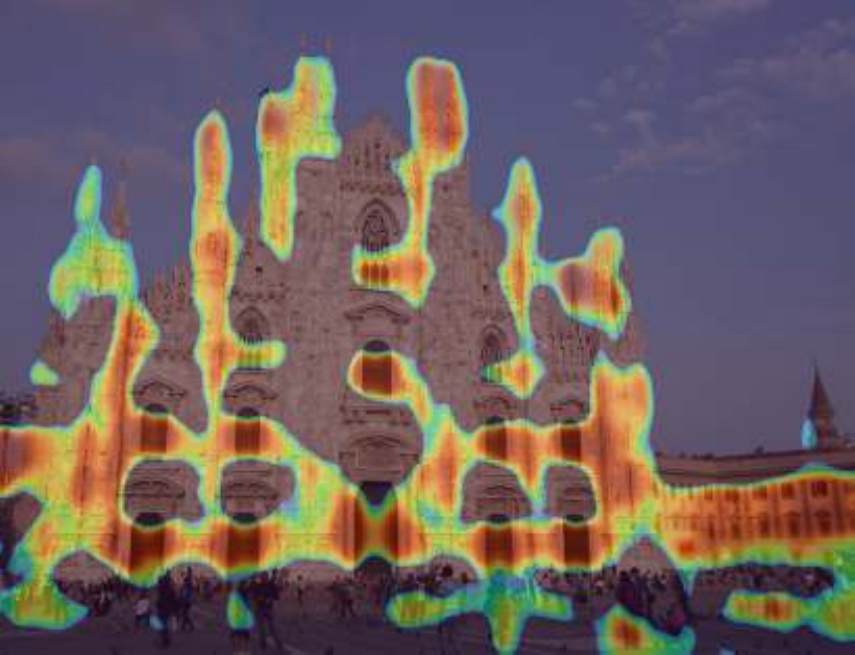}%
    \hspace{\pwidth}%
    \includegraphics[width=\iwidth]{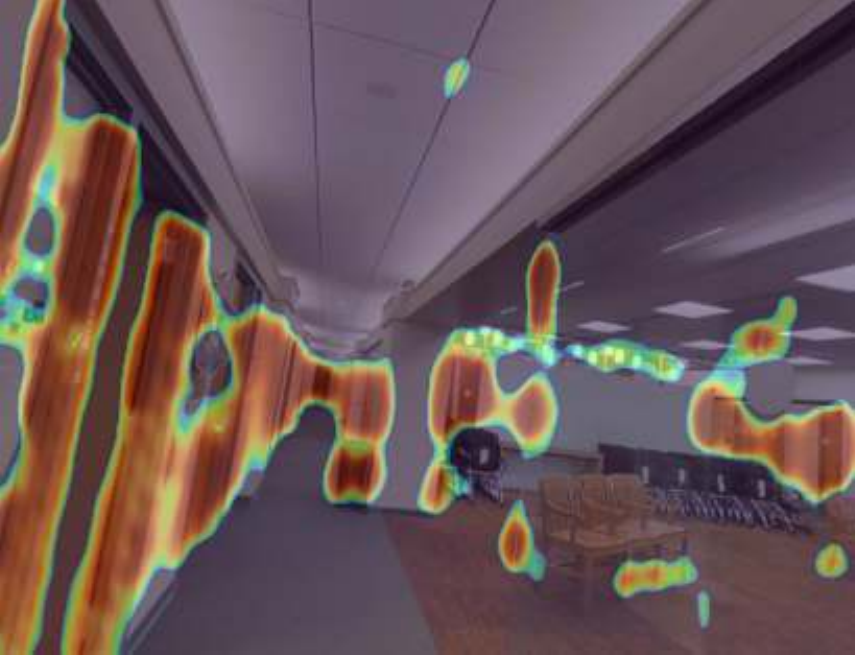}%
    \hspace{\pwidth}%
    \includegraphics[width=\iwidth]{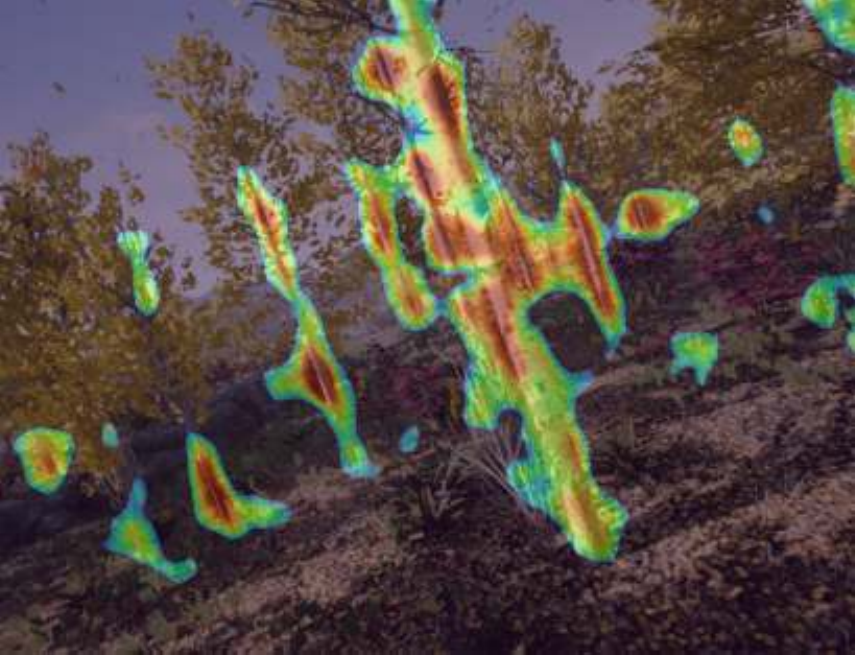}%
    \hspace{\pwidth}%
    \includegraphics[width=\iwidth]{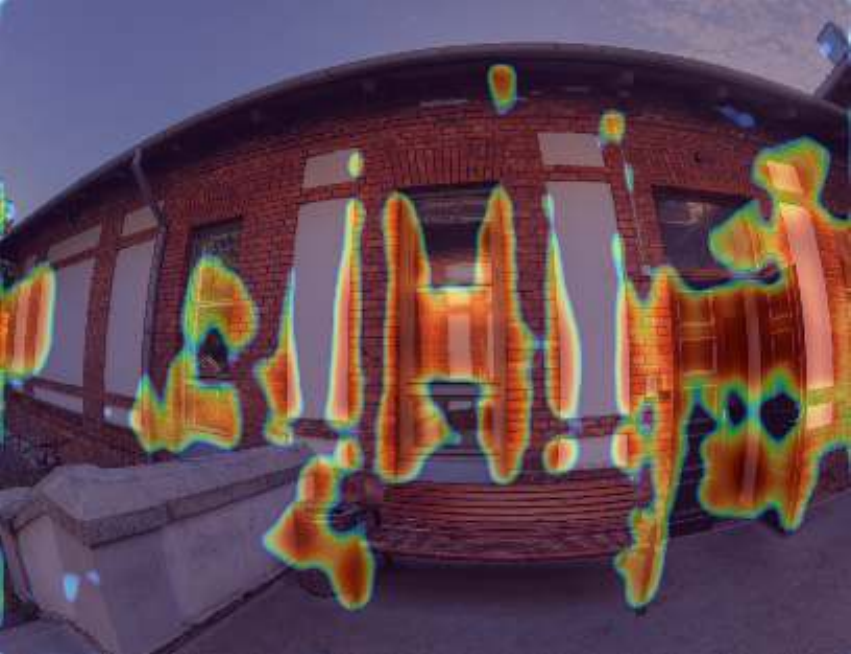}%
    \end{minipage}%
    
    \begin{minipage}[t]{\lwidth}
    \rotatebox[origin=c]{90}{latitude $\confidence_{\latitude}$}
    \end{minipage}%
    \begin{minipage}{\dimexpr \ncols\iwidth + \ncols\pwidth \relax}
    \hspace{\pwidth}%
    \includegraphics[width=\iwidth]{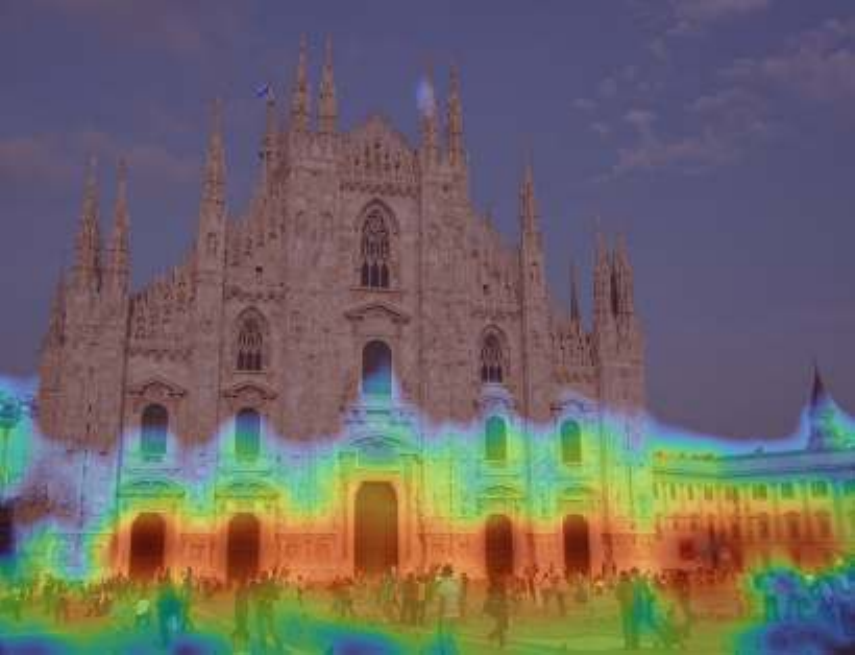}%
    \hspace{\pwidth}%
    \includegraphics[width=\iwidth]{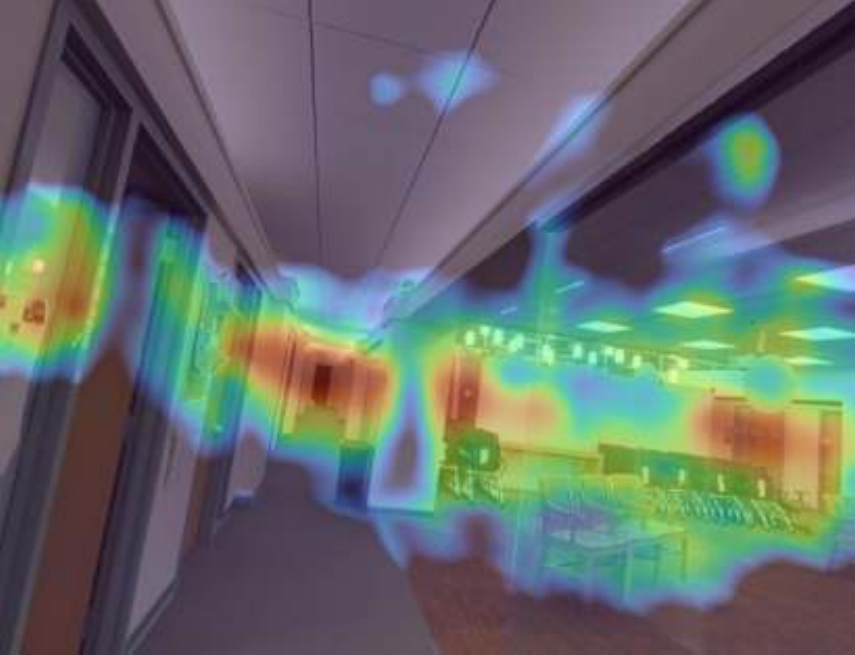}%
    \hspace{\pwidth}%
    \includegraphics[width=\iwidth]{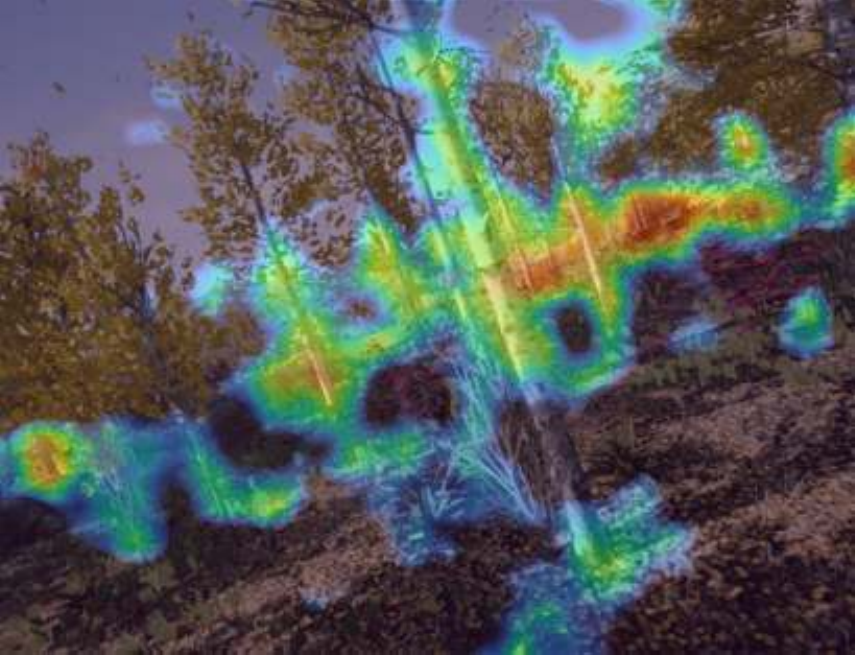}%
    \hspace{\pwidth}%
    \includegraphics[width=\iwidth]{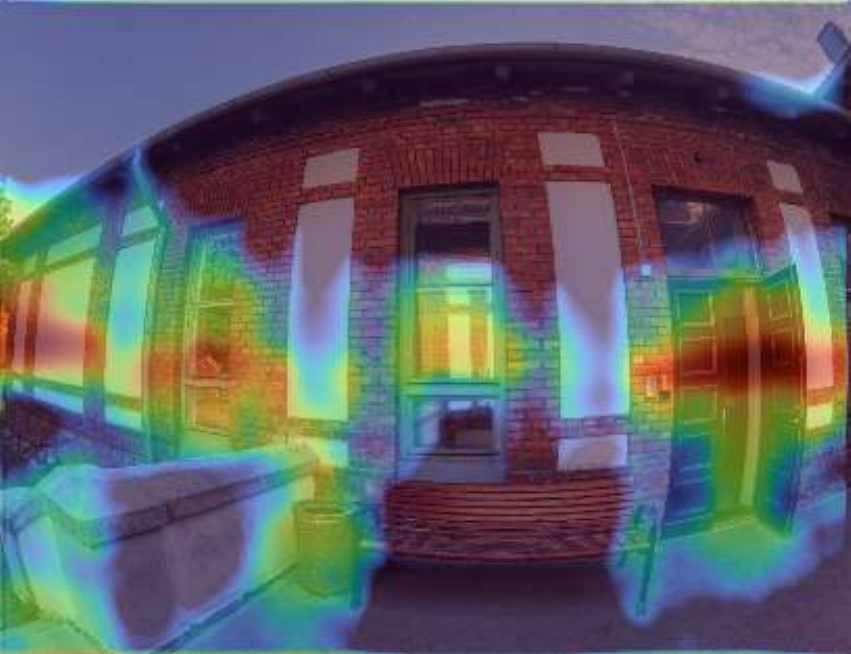}%
    \end{minipage}%
    \hspace{\pwidth}%
    \end{minipage}%
    \begin{minipage}{\bwidth}
    \includegraphics[width=\bwidth]{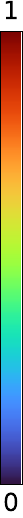}%
    \end{minipage}%
    
    \caption{\textbf{Good features to calibrate.}
    We show the confidences learned by \ours~for both components of the Perspective Field.
    The up-vector is most confident near vertical lines or upright objects like trees.
    The latitude is most confident near the horizon.
    }%
    \label{fig:confidence}%
\end{figure}

\paragraph{Confidence:}
Predicting the Perspective Field is easy in some areas, such as near the horizon or vertical lines, but difficult in others, such as in texture-less areas.
It needs to be accurate only for a subset of pixels to sufficiently constrain the camera parameters.
The neural network thus also predicts a pixel-wise confidence $(\confidence_{\up_\pointim}, \confidence_{\latitude_\pointim}) \in [0,1]$ for each component of the Perspective Field.
The confidences are not directly supervised, but rather learned implicitly to help the optimization converge to the right solution.
\Cref{fig:confidence} shows the predicted confidence after training.
The up-vector is not limited to vertical lines; it is also confidently predicted on upright objects and curves in distorted images.

\paragraph{Objective function:}
We then find the camera parameters $\camparams$ that generate a Perspective Field $\left(\up(\camparams), \latitude(\camparams)\right)$ similar to the one that is observed $(\hat\up, \hat\latitude)$.
This amounts to minimizing the objective function %
\begin{equation}
E(\camparams) =
\sum_{\pointim\in\imsize}
\confidence_{\up_\pointim}
\normsmall{
    \underbrace{
        \up_\pointim\left(\camparams\right) - \hat\up_\pointim
    }_{\residual_{\up_\pointim}}
}_2^2
+ \confidence_{\latitude_\pointim}
\normsmall{
    \underbrace{
        \sin\latitude_\pointim\left(\camparams\right) - \sin\hat\latitude_\pointim
    }_{\residual_{\latitude_\pointim}}
}_2^2
\enspace.
\label{eq:objective}
\end{equation}
This requires an explicit expression of the up-vector for the distortion camera model by solving for the limit of \cref{eq:up}.
We notice that the up-vector is collinear with the directional derivative of the projection function along the up direction, \ie 
$\up_\pointim \propto -\nabla\camproj(\pointw)\cdot\gravity$.
For a radial distortion model such that $\distfn\left(\matrx{u&v}, \dist\right) = d(u,v,\dist)\matrx{u&v}$,
the up-vector can be expressed as
\begin{equation}
\up_\pointim\left(\focal,\gravity,\dist\right) \propto
\left(1 + \frac{1}{d(u,v,\dist)}\matrx{u\\v}\frac{\partial d(u,v,\dist)}{\partial(u,v)} \right)
\matrx{u\gravity_z-\gravity_x\\v\gravity_z-\gravity_y}
\enspace.
\label{eq:up-dist}
\end{equation}
This can be derived similarly for other distortion models, as shown in \cref{sec:distortion-models}.

\paragraph{Optimization:}
Since \cref{eq:objective} describes a non-linear least-squares problem, we find its minimum with the Levenberg–Marquardt (LM) algorithm~\cite{levenberg1944method,marquardt1963algorithm}.
We parametrize the gravity $\gravity$ on the $S^2$ manifold~\cite{hartley2003multiple,hertzberg2013integrating} since it has unit norm
and the focal length by $\log\focal$ since it is strictly positive.
The distortion parameters generally live in the Euclidean space, unless constrained by the camera model.
At each iteration, we compute the update $\lmstep\in\real^{3{+}D}$ by solving the linear system
\begin{equation*}
    \lmstep =
    -\left(\hessian + \lambda\operatorname{diag}\left(\hessian\right)\right)^{-1}
    \jacobian\transp\weight\residual
    \quad \text{with} \quad
    \jacobian = \frac{\partial\residual}{\partial\camparams}
    \quad \text{and} \quad
    \hessian = \jacobian\transp\weight\jacobian
    \enspace,
\end{equation*}
where $\residual$ stacks the residuals $\residual_{\up_\pointim}$ and $\residual_{\latitude_\pointim}$
and $\weight$ is a weight matrix whose diagonal stacks the corresponding confidences $\confidence_{\up}$ and $\confidence_{\latitude}$. %
The damping factor $\lambda$ interpolates between the Gauss-Newton~($\lambda{=}0$) and gradient descent~($\lambda\!\!\to\!\!\infty$) strategies.
It is usually adjusted at each iteration using heuristics~\cite{levenberg1944method, marquardt1963algorithm, madsen2004methods}.
We derive $\jacobian$ analytically and provide it in \cref{sec:jacobians}.
The optimization stops when the update $\lmstep$ is sufficiently small.

\paragraph{Initialization:}
The LM algorithm requires an initial guess.
Since the objective function is not necessarily convex, the choice of initialization should matter.
We consider several strategies.
The trivial strategy initializes $\gravity{=}\matrx{0&1&0}$, $\dist{=}\*0$, and $\focal{=}0.7\max\left(W,H\right)$, assuming a common sensor size.
Other, more complex strategies are described in~\cref{sec:initialization}.
In practice, we have found that neither of them improves the performance over the trivial initialization.

\paragraph{Training:}
The LM optimization is differentiable, either by unrolling its steps and applying backpropagation to each of them~\cite{sarlin21pixloc,teed2021droid} or by leveraging implicit derivatives~\cite{russell2019fixing}.
\ours~is thus end-to-end differentiable, from the estimated camera parameters to the input image.
We train it by supervising the estimated parameters $\hat{\camparams}$ but also the intermediate perspective field, which accelerates the convergence of the training.
This translates into the loss function
\begin{equation}
    \mathcal{L} = \normsmall{\hat{\camparams} - \bar{\camparams}}_\gamma
    + \beta \sum_{\pointim\in\imsize}
    \confidence_{\up_{\pointim}}
        \normsmall{\hat{\up}_{\pointim} - \up(\bar{\camparams})_{\pointim}}_\gamma
    + \confidence_{\latitude_{\pointim}}
    \normsmall{\hat{\latitude}_{\pointim} - \latitude(\bar{\camparams})_{\pointim}}_\gamma
    \enspace,
    \label{eq:loss}
\end{equation}
where $\bar\camparams$ are the ground-truth parameters, $\gamma$ is the L1 loss, and $\beta$ balances the two terms.
The confidences are supervised only by the LM optimization and their gradient is detached in \cref{eq:loss}.
They cancel the second term in areas that the network deems not useful for the optimization. No capacity is thus wasted to unnecessarily improve the prediction of the Perspective Field for all pixels.

\subsection{Practical benefits}
\label{sec:practical-benefits}
Because \ours~is based on well-understood optimization processes, it can be flexibly adapted to the constraints of real-world applications.
None of the existing learned approaches provide such a degree of flexibility.

\paragraph{Arbitrary distortion model:}
Unlike most existing learned approaches, our formulation is not tied to a specific distortion model.
One can train \ours~with a given model but optimize a different one at inference time without any retraining, as long as the Perspective Field is similarly observable.
This is a significant gain given the large number of different distortion models used in practice.
Our experiments show that a version of \ours~trained only on pinhole images can more accurately estimate a radial distortion than all existing approaches that operate on a single image.
This makes it possible to handle heterogeneous datasets captured by many different cameras.

\paragraph{Partial calibration:}
A subset of the camera parameters might sometimes be already available.
This includes camera intrinsics from a factory calibration, EXIF metadata, or structure-from-motion or a gravity direction estimated by an inertial measurement unit or assumed constant when the motion is restricted to be planar.
Our optimization can leverage this information as a hard constraint, by fixing the associated parameters, or as a soft prior, by biasing the parameters with a simple regularization term, in which case the uncertainty of the priors can be leveraged.
Our experiments show that such information directly improves the estimation accuracy of the remaining parameters.

\paragraph{Multi-image optimization:}
We can also couple the optimization across multiple images.
For example, images captured by the same cameras share identical intrinsics, which should be optimized jointly, while each image has its own gravity direction.
Unlike classical geometry tools, this does not require any covisibility across the views.
Differently, images taken by a multi-camera rig have distinct intrinsics but have a common gravity direction, given known relative poses.

\begin{figure}[t]
    \centering

    \def\ncols{8}
    \setlength{\pwidth}{0.005\linewidth}
    \setlength{\iwidth}{\dimexpr(0.999\linewidth - \ncols\pwidth + \pwidth)/\ncols \relax}

    \begin{minipage}[b]{\iwidth}
    \centering{\scriptsize \red{$0.5\degree$}$/1.2\degree$}
    \end{minipage}%
    \hspace{\pwidth}%
    \begin{minipage}[b]{\iwidth}
    \centering{\scriptsize \red{$0.9\degree$}$/1.4\degree$}
    \end{minipage}%
    \hspace{\pwidth}%
    \begin{minipage}[b]{\iwidth}
    \centering{\scriptsize \red{$2.1\degree$}$/2.2\degree$}
    \end{minipage}%
    \hspace{\pwidth}%
    \begin{minipage}[b]{\iwidth}
    \centering{\scriptsize \red{$3.4\degree$}$/3.2\degree$}
    \end{minipage}%
    \hspace{\pwidth}%
    \begin{minipage}[b]{\iwidth}
    \centering{\scriptsize \red{$3.4\degree$}$/4.7\degree$}
    \end{minipage}%
    \hspace{\pwidth}%
    \begin{minipage}[b]{\iwidth}
    \centering{\scriptsize \red{$5.3\degree$}$/5.4\degree$}
    \end{minipage}%
    \hspace{\pwidth}%
    \begin{minipage}[b]{\iwidth}
    \centering{\scriptsize \red{$7.3\degree$}$/9.5\degree$}
    \end{minipage}%
    \hspace{\pwidth}%
    \begin{minipage}[b]{\iwidth}
    \centering{\scriptsize \red{$8.6\degree$}$/10.0\degree$}
    \end{minipage}%
    \hspace{\pwidth}%
    
    \includegraphics[width=\iwidth]{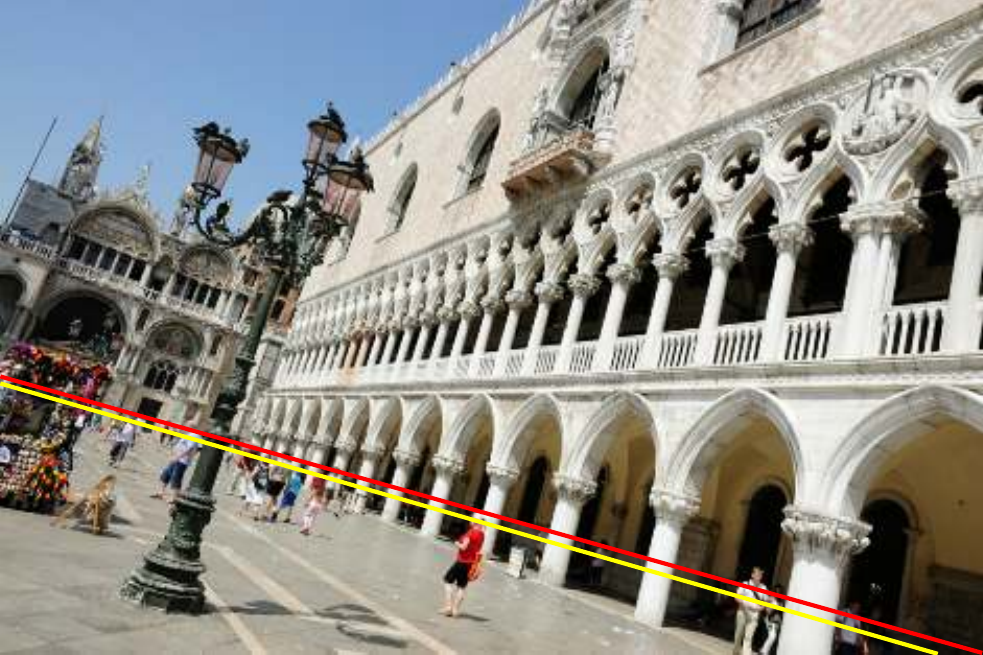}%
    \hspace{\pwidth}%
    \includegraphics[width=\iwidth]{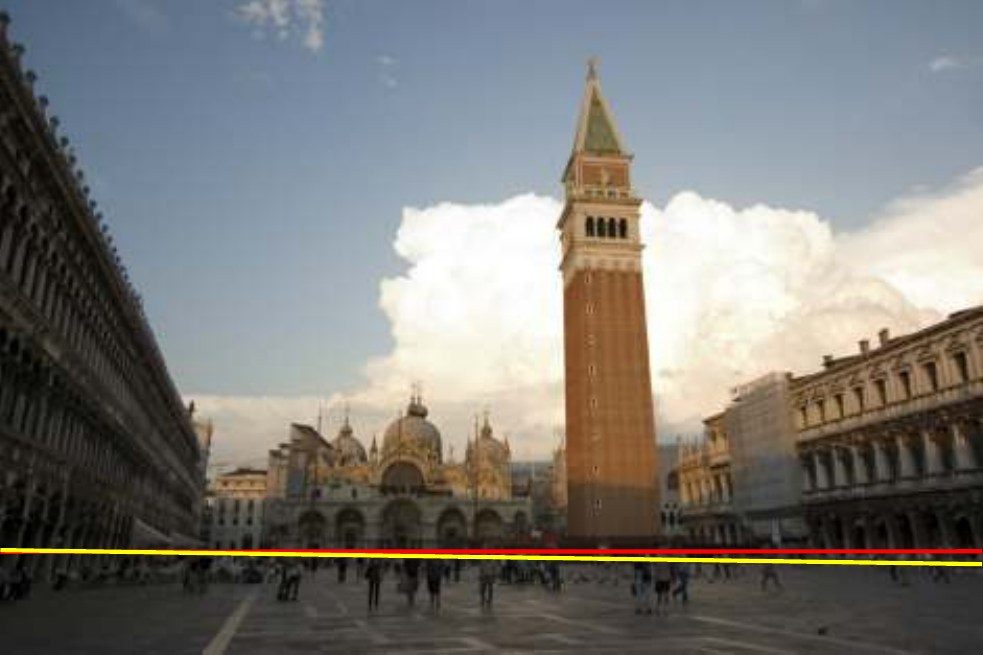}%
    \hspace{\pwidth}%
    \includegraphics[width=\iwidth]{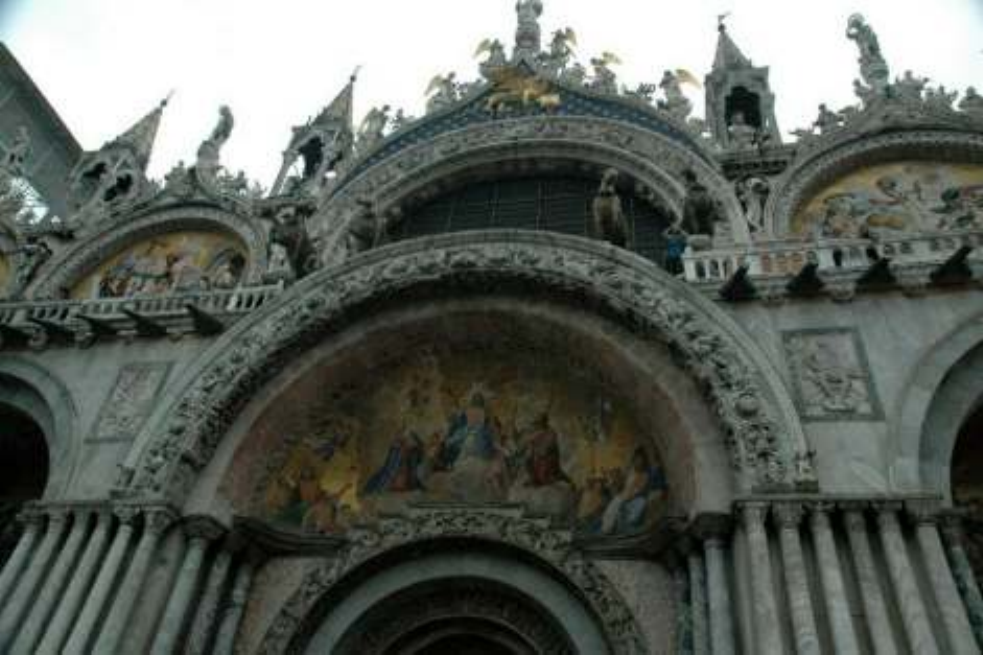}%
    \hspace{\pwidth}%
    \includegraphics[width=\iwidth]{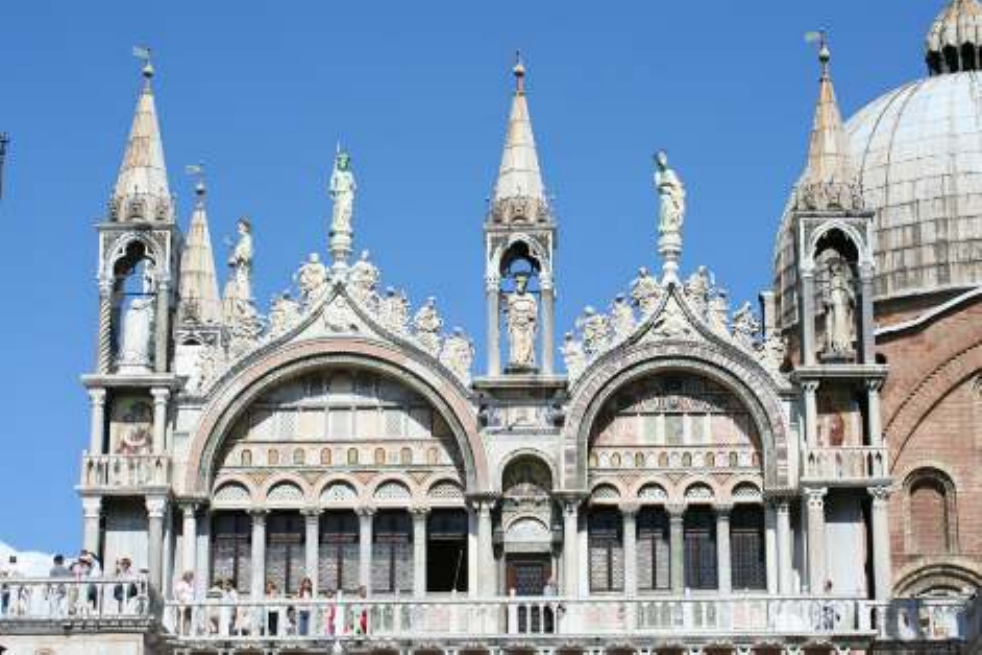}%
    \hspace{\pwidth}%
    \includegraphics[width=\iwidth]{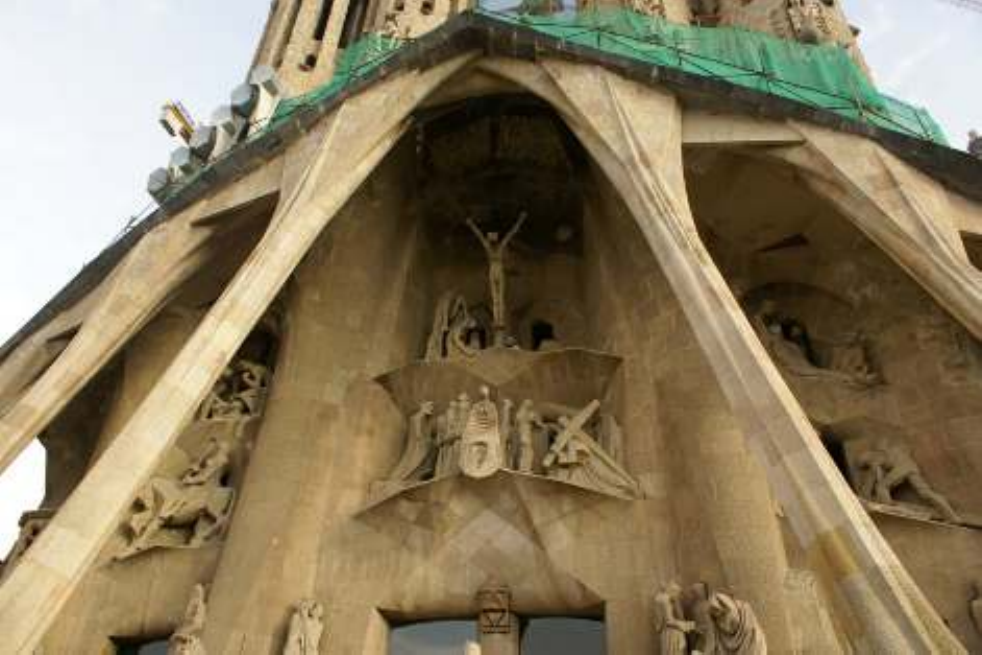}%
    \hspace{\pwidth}%
    \includegraphics[width=\iwidth]{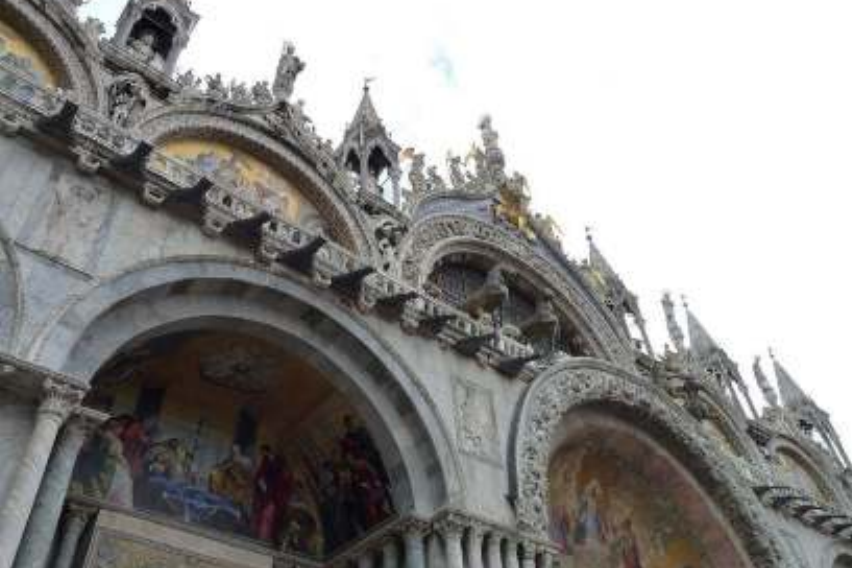}%
    \hspace{\pwidth}%
    \includegraphics[width=\iwidth]{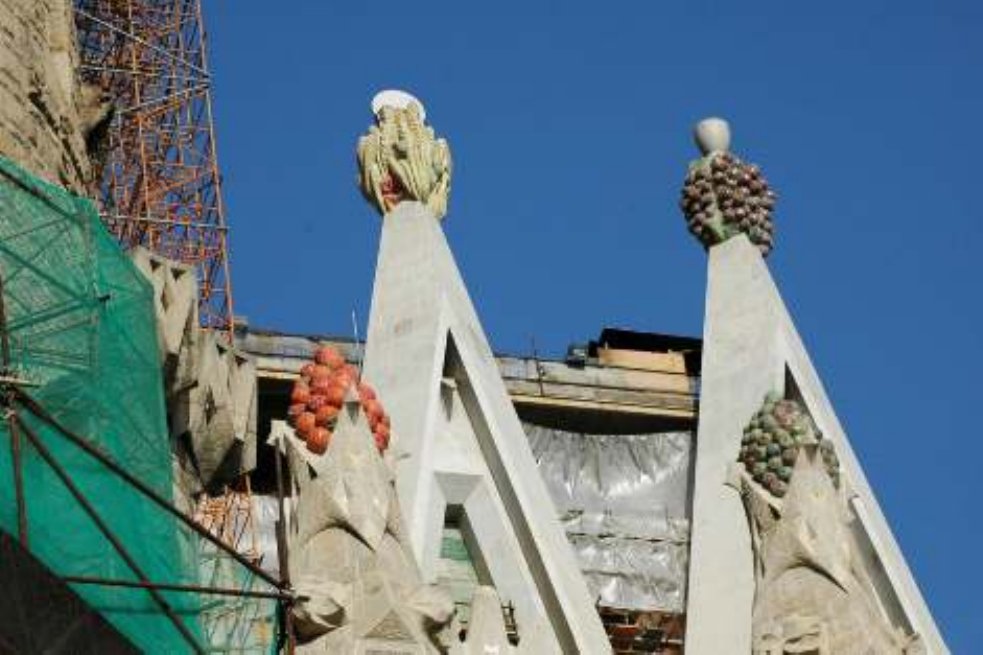}%
    \hspace{\pwidth}%
    \includegraphics[width=\iwidth]{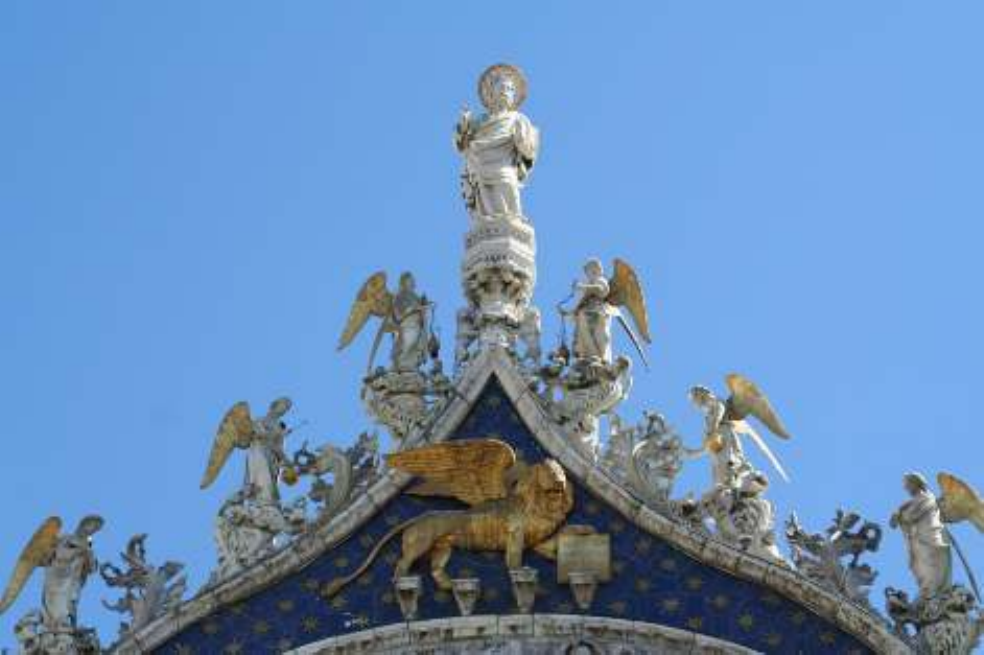}%

    \includegraphics[width=\iwidth]{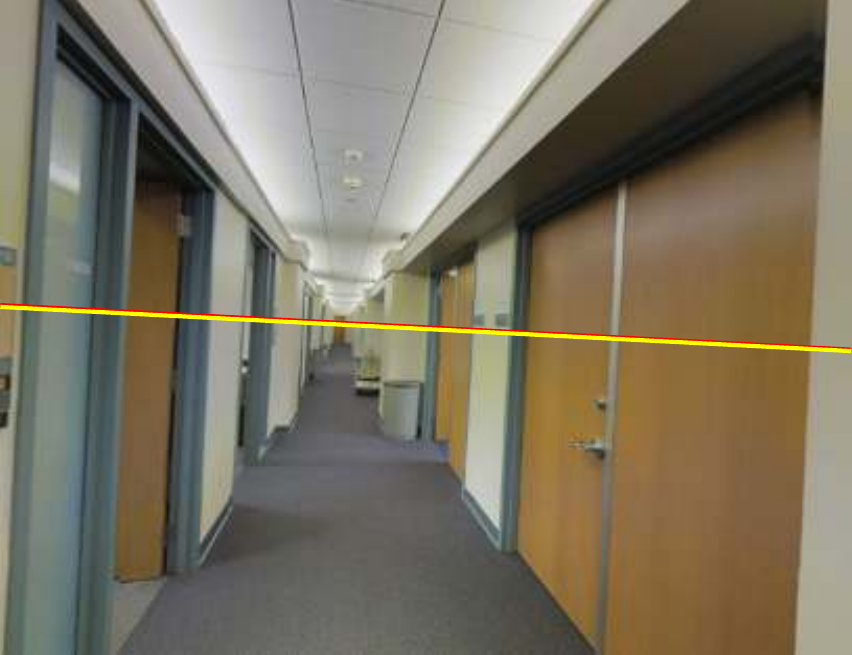}%
    \hspace{\pwidth}%
    \includegraphics[width=\iwidth]{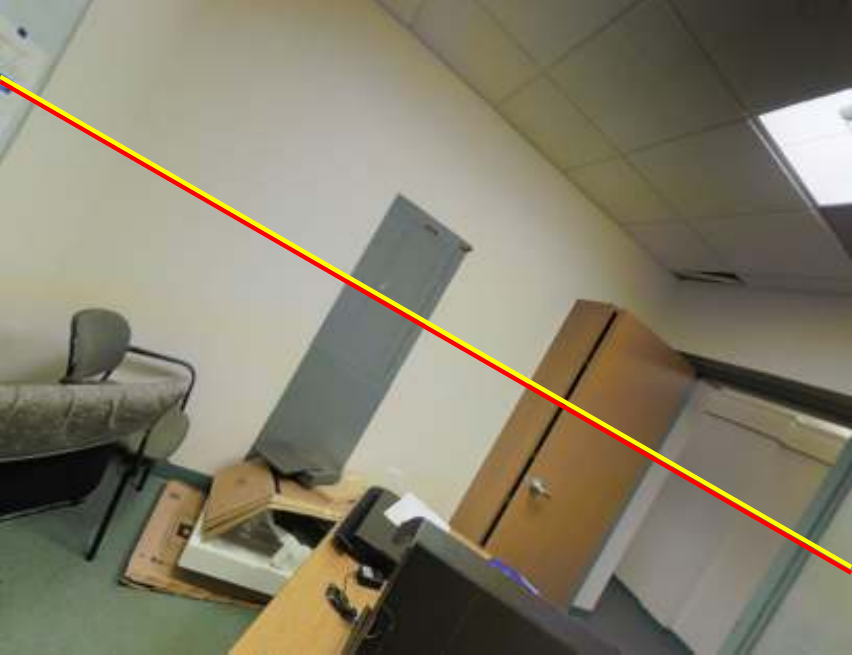}%
    \hspace{\pwidth}%
    \includegraphics[width=\iwidth]{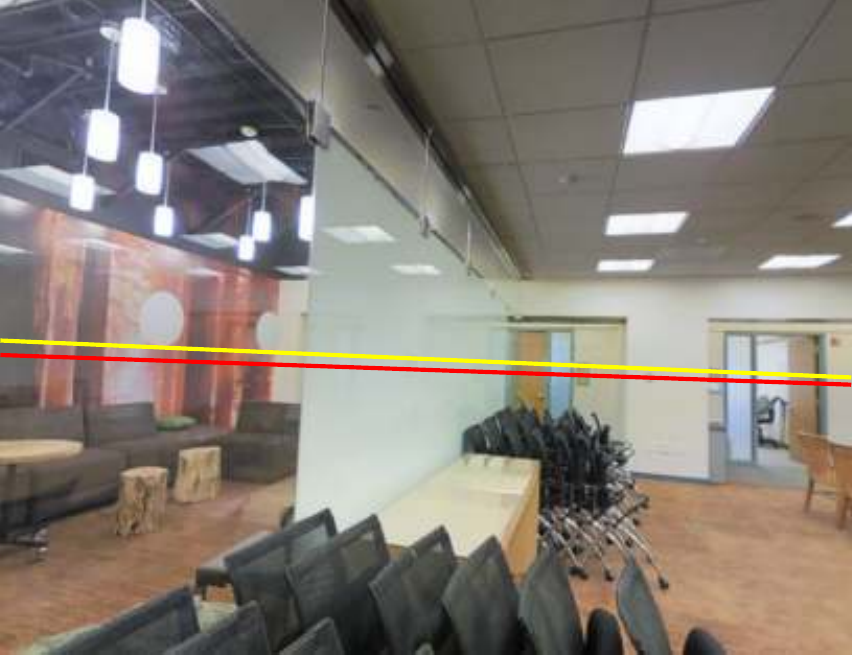}%
    \hspace{\pwidth}%
    \includegraphics[width=\iwidth]{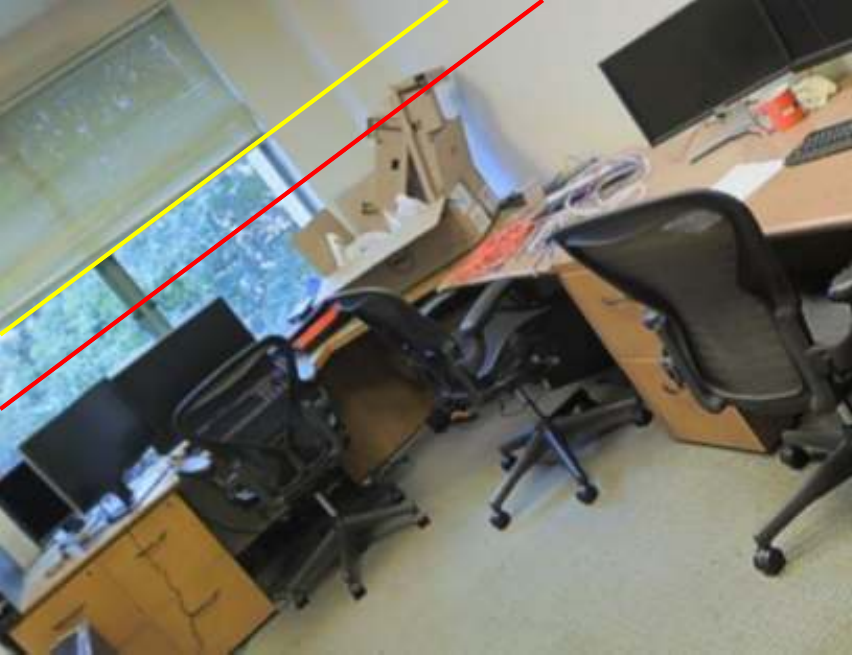}%
    \hspace{\pwidth}%
    \includegraphics[width=\iwidth]{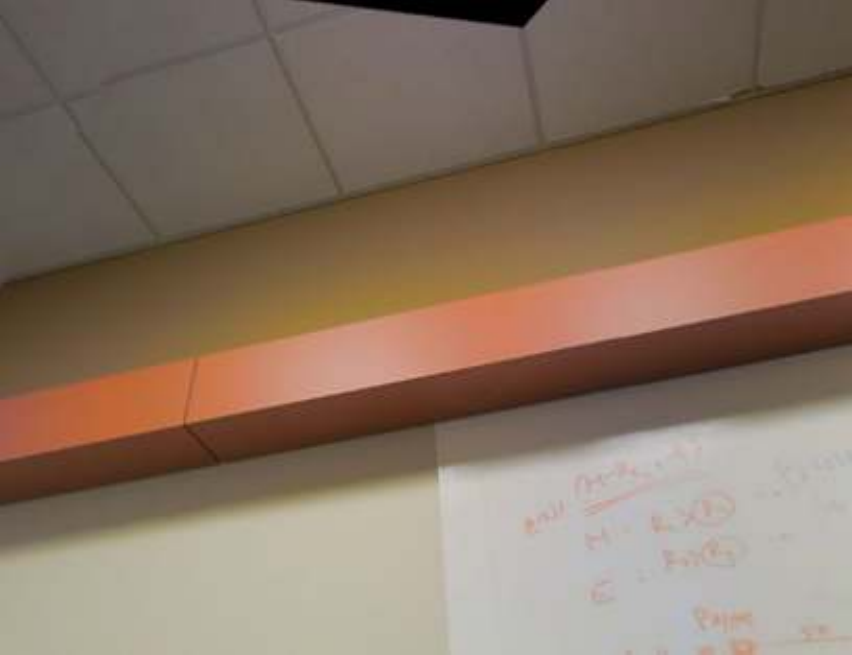}%
    \hspace{\pwidth}%
    \includegraphics[width=\iwidth]{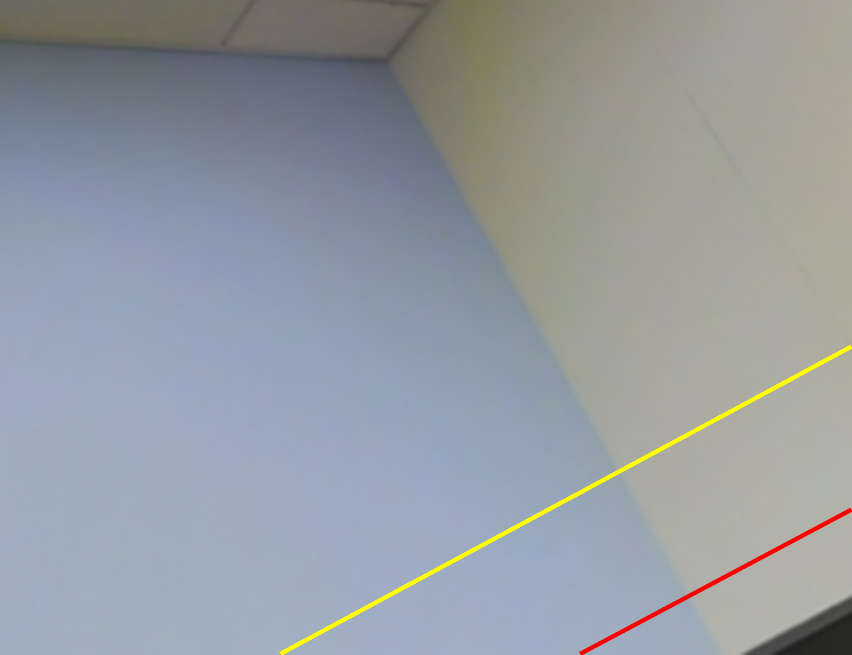}%
    \hspace{\pwidth}%
    \includegraphics[width=\iwidth]{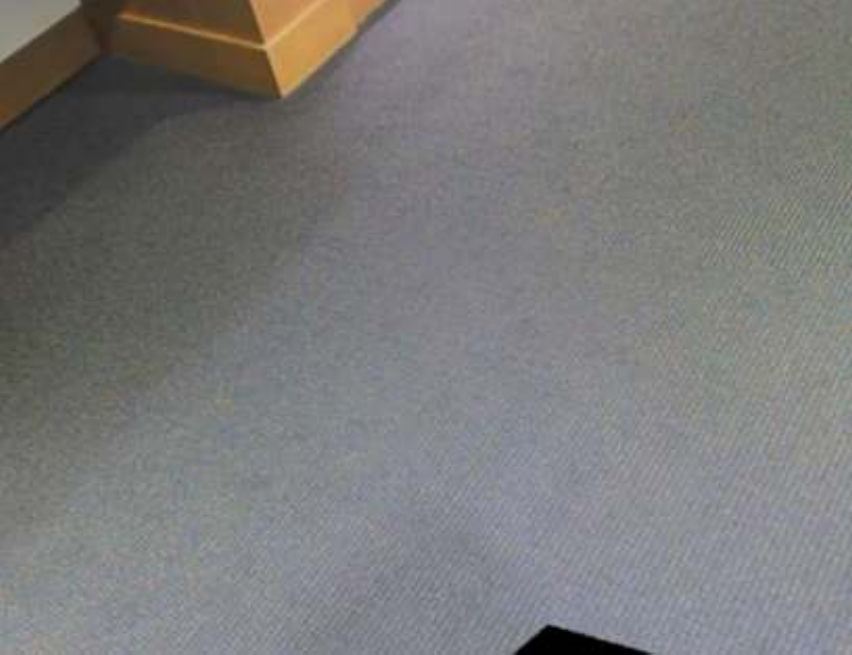}%
    \hspace{\pwidth}%
    \includegraphics[width=\iwidth]{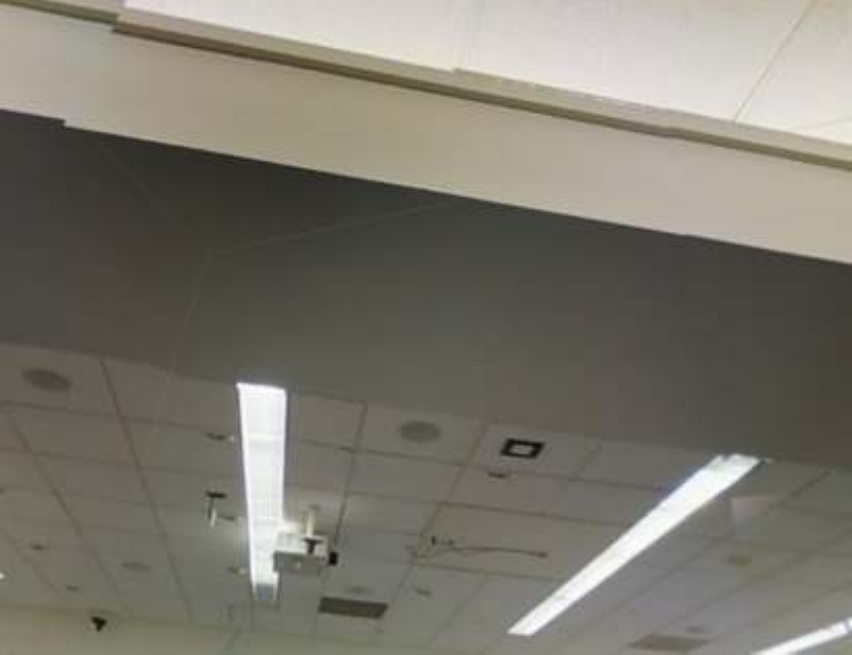}%
    \vspace{-1mm}
    
    \begin{minipage}[b]{\iwidth}
    \centering{\scriptsize \red{$0.2\degree$}$/0.7\degree$}
    \end{minipage}%
    \hspace{\pwidth}%
    \begin{minipage}[b]{\iwidth}
    \centering{\scriptsize \red{$0.7\degree$}$/1.0\degree$}
    \end{minipage}%
    \hspace{\pwidth}%
    \begin{minipage}[b]{\iwidth}
    \centering{\scriptsize \red{$1.3\degree$}$/1.5\degree$}
    \end{minipage}%
    \hspace{\pwidth}%
    \begin{minipage}[b]{\iwidth}
    \centering{\scriptsize \red{$2.4\degree$}$/2.6\degree$}
    \end{minipage}%
    \hspace{\pwidth}%
    \begin{minipage}[b]{\iwidth}
    \centering{\scriptsize \red{$4.3\degree$}$/4.8\degree$}
    \end{minipage}%
    \hspace{\pwidth}%
    \begin{minipage}[b]{\iwidth}
    \centering{\scriptsize \red{$4.5\degree$}$/6.0\degree$}
    \end{minipage}%
    \hspace{\pwidth}%
    \begin{minipage}[b]{\iwidth}
    \centering{\scriptsize \red{$6.2\degree$}$/6.0\degree$}
    \end{minipage}%
    \hspace{\pwidth}%
    \begin{minipage}[b]{\iwidth}
    \centering{\scriptsize \red{$7.6\degree$}$/9.6\degree$}
    \end{minipage}%
    
    \caption{\textbf{Ranking images by uncertainty.} We report the gravity \emph{\red{error} / uncertainty} for 8 outdoor (top) and indoor (bottom) images from left-to-right, sorted by uncertainty.
    The estimated uncertainty correlates well with the ground truth error. 
    }%
    \label{fig:uncertainty}%
\end{figure}

\paragraph{Uncertainty estimation:}
The uncertainty of parameters estimated by the LM algorithm can be easily obtained via their covariance matrix, $\*\Sigma_{\camparams} = \hessian^{-1}$ computed at convergence.
It can be further decomposed into an uncertainty per parameter $\gravity$, $\focal$, and $\dist$.
This uncertainty can flag examples for which the estimate is likely wrong, \eg due to poor geometric constraints in the optimization or excessively low-confident visual cues, for example due to occlusion, motion blur, or fully texture-less view (\cref{fig:uncertainty}).
This makes it easier to integrate the estimates of \ours~into a larger probabilistic framework, such as a factor graph~\cite{dellaert2012factor,dellaert2017factor}.

\paragraph{Comparison to Perspective Fields~\cite{jin2022PerspectiveFields}:}
Our work is inspired by Jin~\etal~\cite{jin2022PerspectiveFields}, who train a DNN to extract camera parameters from a pre-trained Perspective Field.
This black-box approach is sensitive to the training distribution and lacks the flexibility described above.
They also propose to refine its estimates with a first-order optimization based on Adam~\cite{kingma2014adam}.
We found this to converge much slower than our LM optimization.
Training it end-to-end and learning appropriate confidences is thus not practical.
Unlike our approach, this optimization is sensitive to the initialization and yields marginal accuracy gains.
We demonstrate the value of our improvements in an ablation study.%

\begin{figure}[tb]
    \centering

    \def\ncols{4}
    \setlength{\pwidth}{0.005\linewidth}
    \setlength{\iwidth}{\dimexpr(0.999\linewidth - \ncols\pwidth + \pwidth)/\ncols \relax}
    \newlength{\lamarwidth}
    \setlength{\lamarwidth}{\dimexpr(0.999\linewidth - 8\pwidth + \pwidth)/8 \relax}
    
    \begin{minipage}[b]{\iwidth}
    \centering{\footnotesize a) ground-truth}
    \end{minipage}%
    \hspace{\pwidth}%
    \begin{minipage}[b]{\iwidth}
    \centering{\footnotesize b) final prediction}
    \end{minipage}%
    \hspace{\pwidth}%
    \begin{minipage}[b]{\iwidth}
    \centering{\footnotesize c) observed latitude}
    \end{minipage}%
    \hspace{\pwidth}%
    \begin{minipage}[b]{\iwidth}
    \centering{\footnotesize d) observed up-vect.}
    \end{minipage}%
    
    \includegraphics[width=\iwidth]{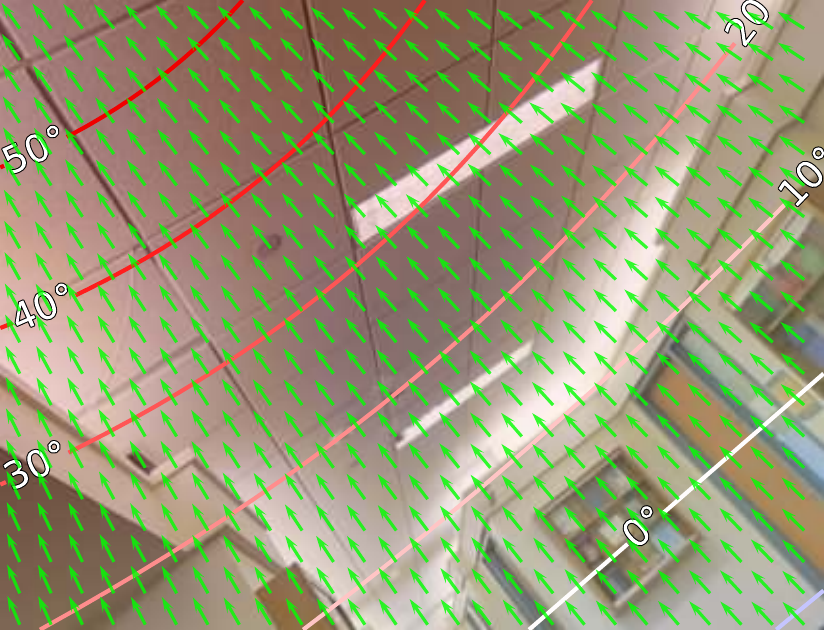}%
    \hspace{\pwidth}%
    \includegraphics[width=\iwidth]{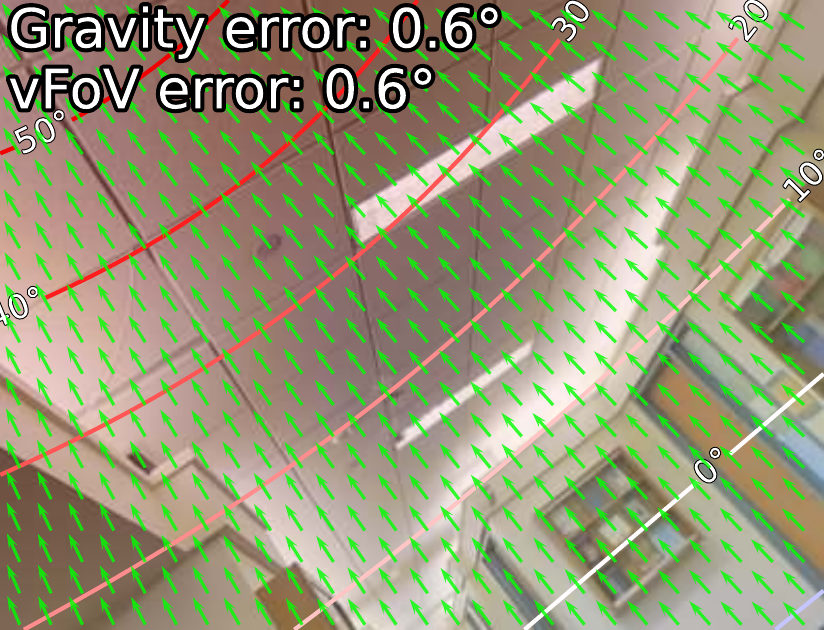}%
    \hspace{\pwidth}%
    \includegraphics[width=\iwidth]{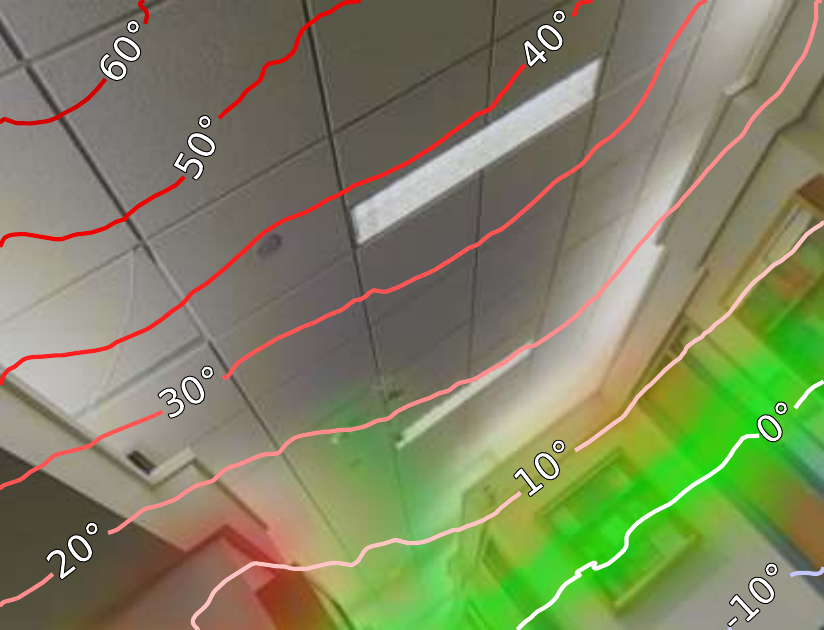}%
    \hspace{\pwidth}%
    \includegraphics[width=\iwidth]{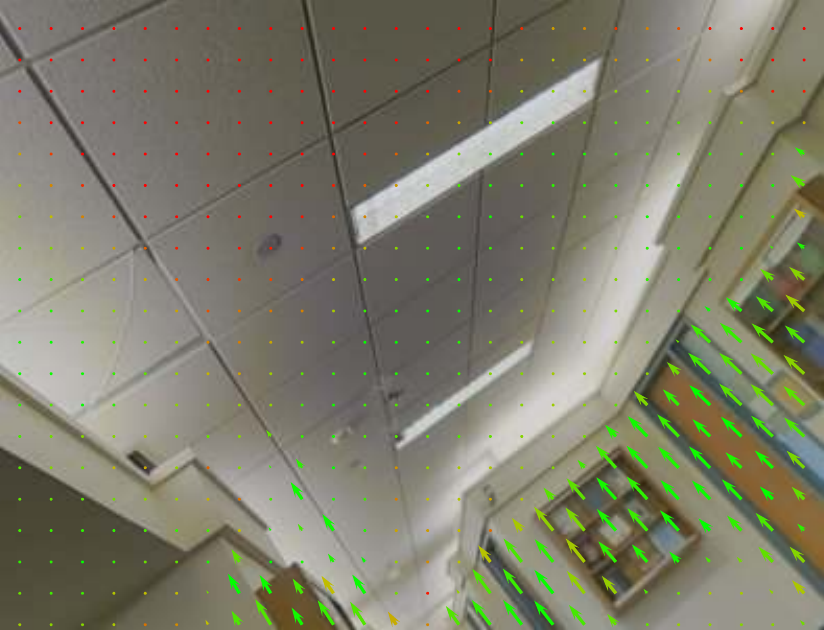}%
    
    \includegraphics[width=\iwidth]{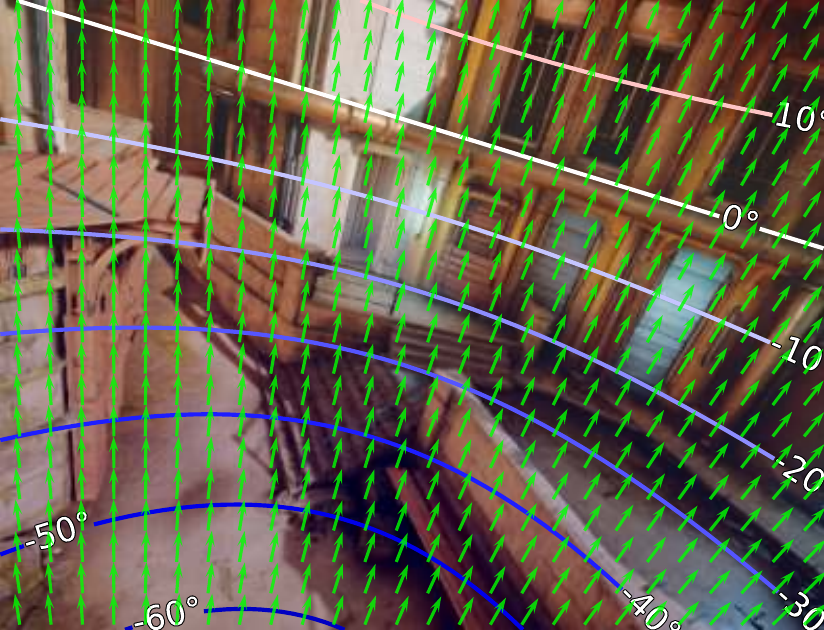}%
    \hspace{\pwidth}%
    \includegraphics[width=\iwidth]{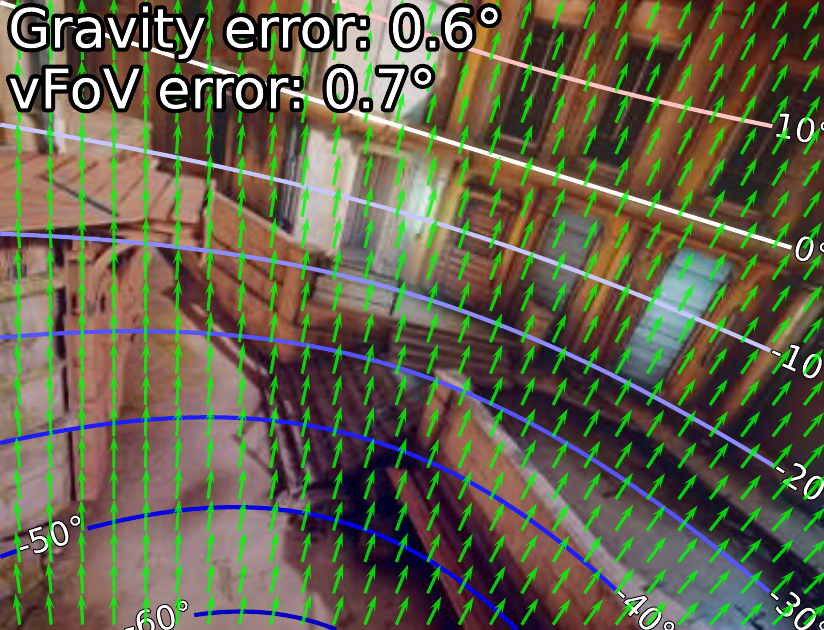}%
    \hspace{\pwidth}%
    \includegraphics[width=\iwidth]{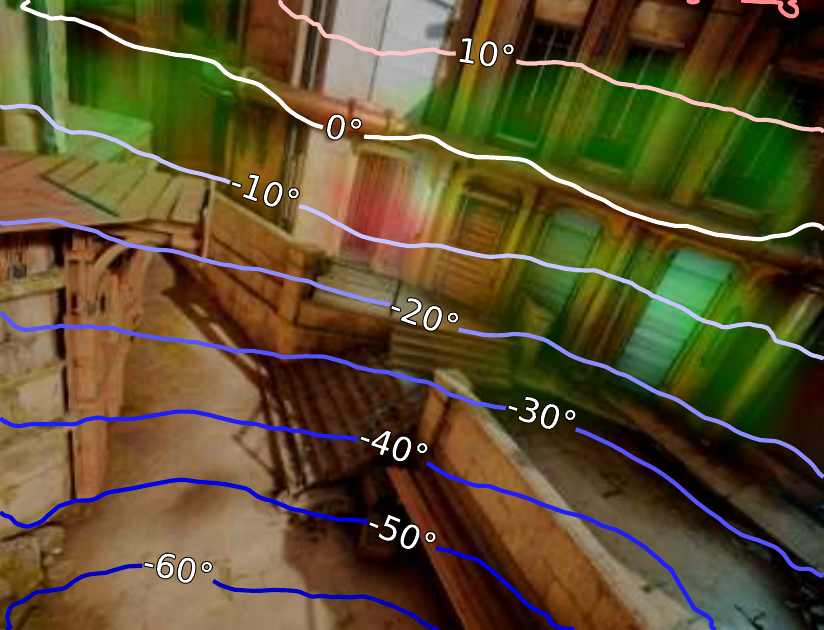}%
    \hspace{\pwidth}%
    \includegraphics[width=\iwidth]{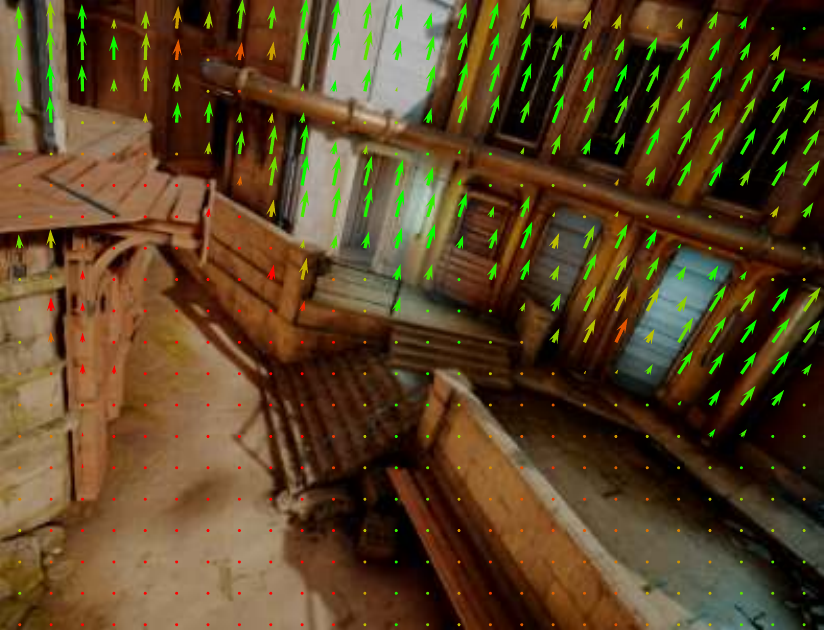}%

    \includegraphics[width=\iwidth]{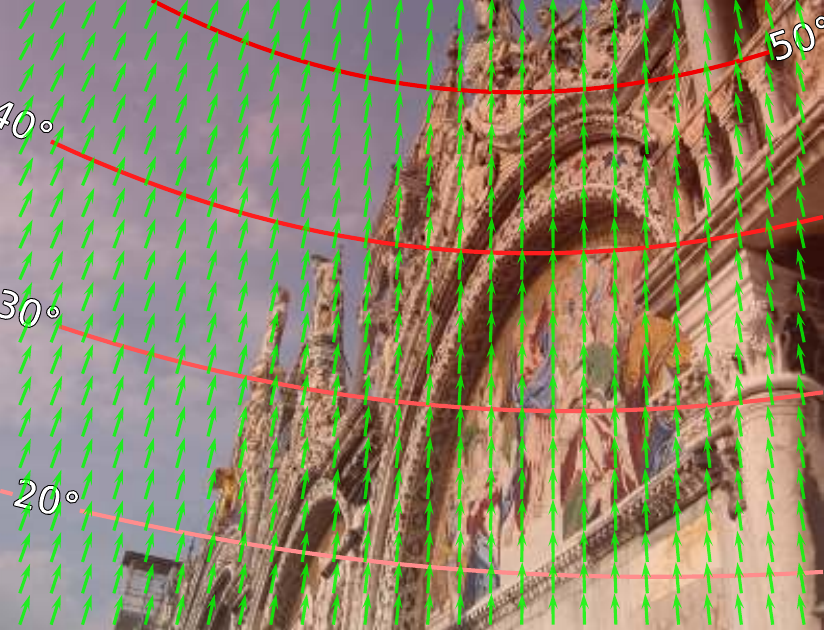}%
    \hspace{\pwidth}%
    \includegraphics[width=\iwidth]{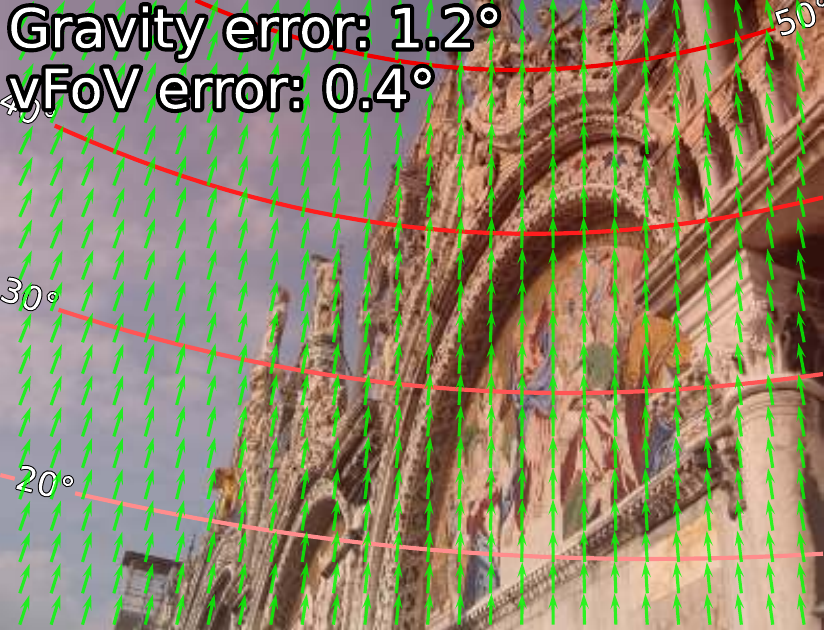}%
    \hspace{\pwidth}%
    \includegraphics[width=\iwidth]{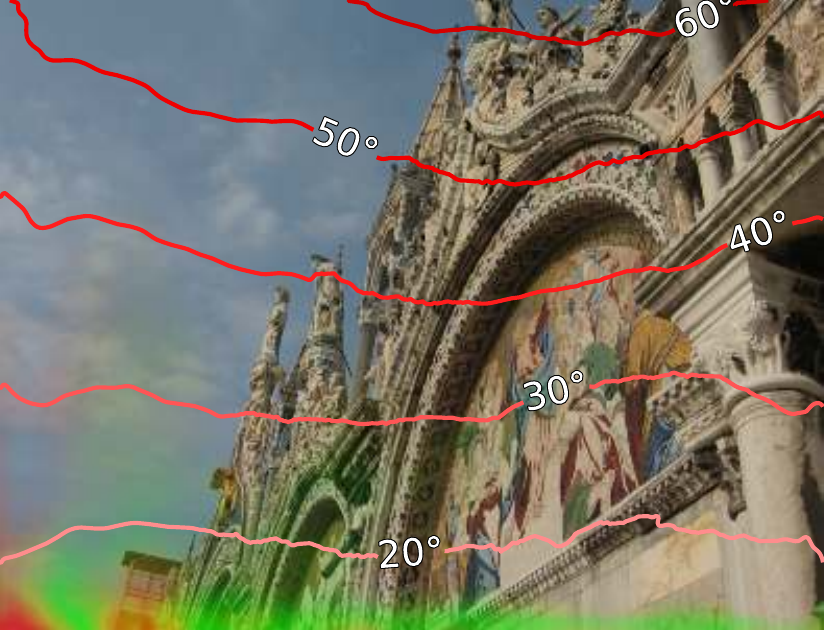}%
    \hspace{\pwidth}%
    \includegraphics[width=\iwidth]{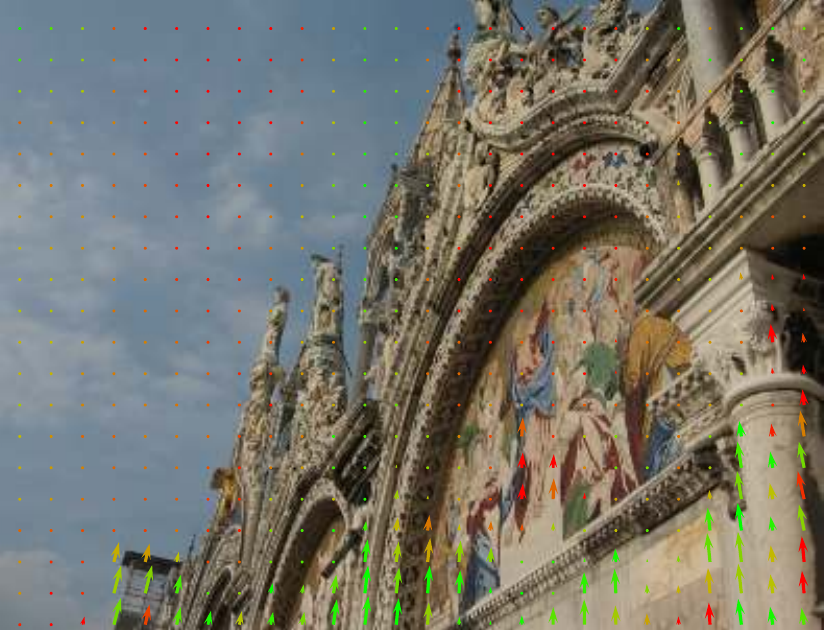}%
    
    \includegraphics[width=\lamarwidth]{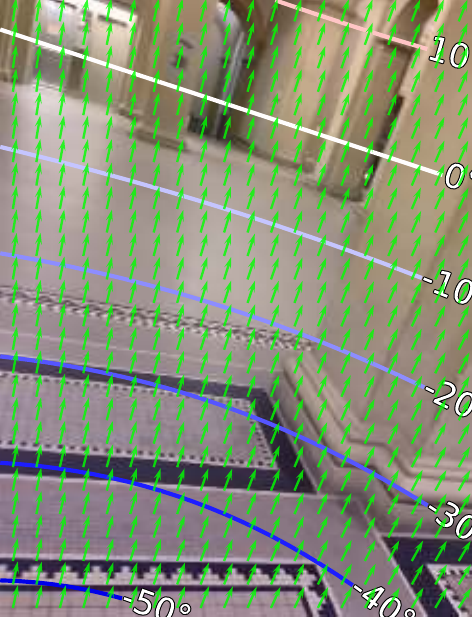}%
    \hspace{\pwidth}%
    \includegraphics[width=\lamarwidth]{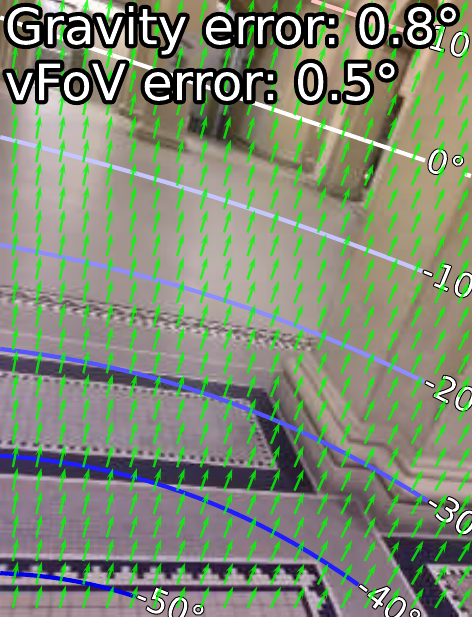}%
    \hspace{\pwidth}%
    \includegraphics[width=\lamarwidth]{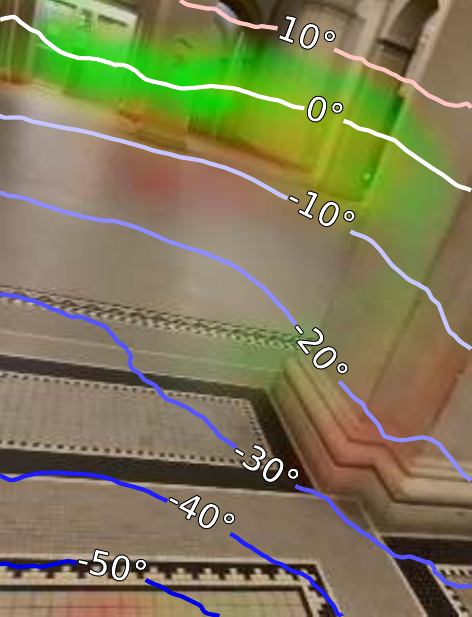}%
    \hspace{\pwidth}%
    \includegraphics[width=\lamarwidth]{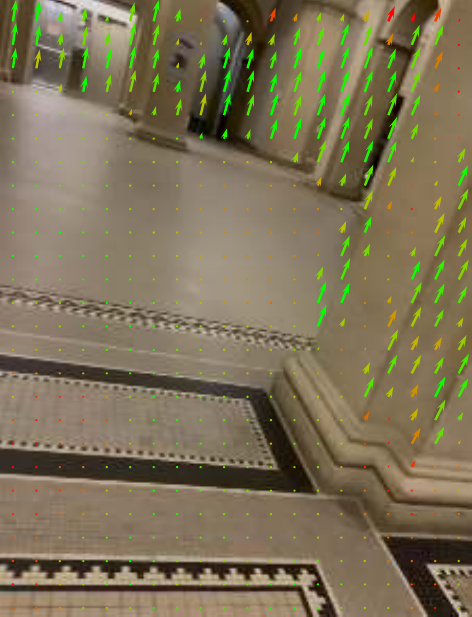}%
    \hspace{\pwidth}%
    \includegraphics[width=\lamarwidth]{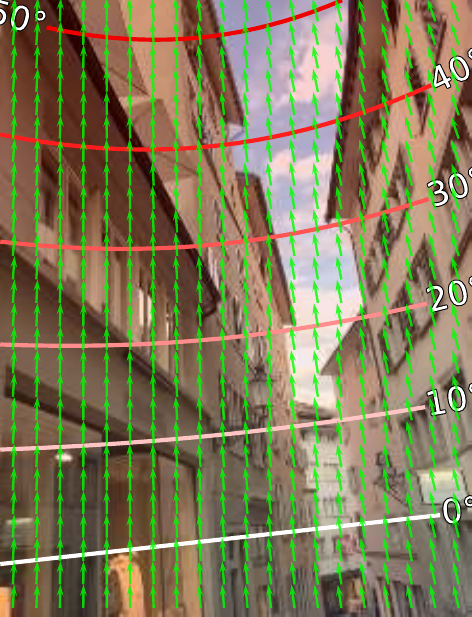}%
    \hspace{\pwidth}%
    \includegraphics[width=\lamarwidth]{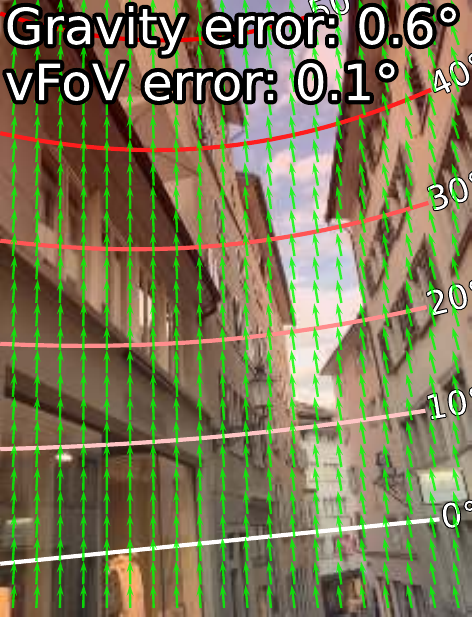}%
    \hspace{\pwidth}%
    \includegraphics[width=\lamarwidth]{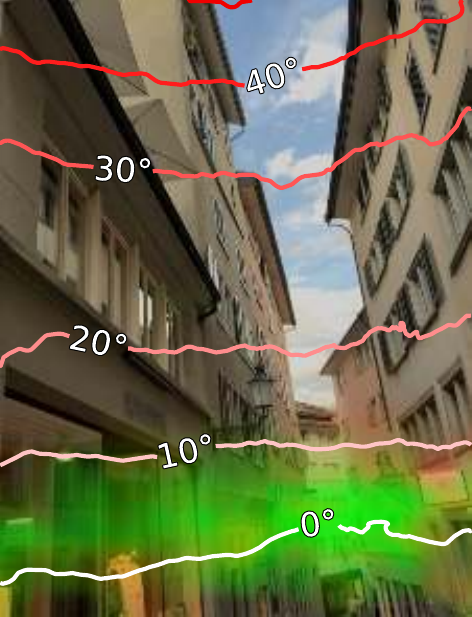}%
    \hspace{\pwidth}%
    \includegraphics[width=\lamarwidth]{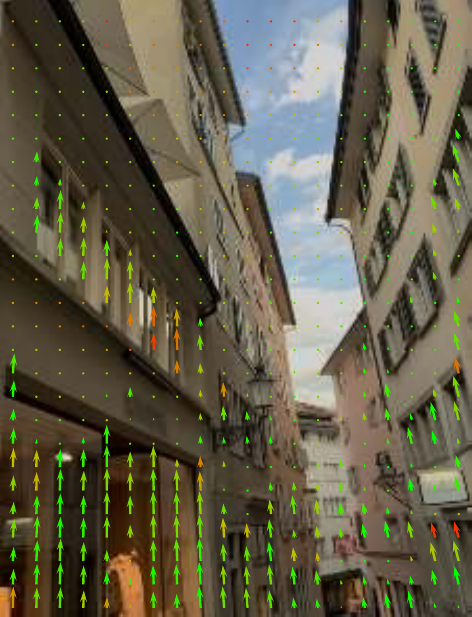}%

    \caption{\textbf{Qualitative results.} We show five examples of \ours's prediction on Stanford2D3D~\cite{armeni2017joint}, TartanAir~\cite{tartanair2020iros}, MegaDepth~\cite{li2018megadepth} and LaMAR~\cite{sarlin2022lamar} (x2). a-b) depict the generated up-vector and latitude field from ground-truth and estimated camera parameters. c-d) depict the latitude and up-vector fields observed by the DNN, with opacity from the learned confidences.
    Here, green denotes accurate and red inaccurate predicted fields w.r.t.\ ground-truth. 
    \ours~learns to predict accurate fields and to discard observations in regions that are less informative, \eg the floor.
    }%
    \label{fig:qualitative}%
\end{figure}

\section{Implementation}
\label{sec:implementation}

\paragraph{\ourdataset~dataset:}
As prior works relied on proprietary datasets~\cite{perceptual,lopez2019deepcalib,jin2022PerspectiveFields}, reproducing them has been difficult.
We have therefore assembled a new dataset, \ourdataset, of panoramas that are publicly available from various sources~\cite{hdrmaps,polyhaven,bolduc2023beyond}. %
\ourdataset~consists of 2.9k panoramas split into 2.6k/147/148 for training/validation/testing.
Sampling 16 square images per panorama yields about 42k training images.
Compared to the dataset used in~\cite{jin2022PerspectiveFields}, ours is much smaller but exhibits a better balance between indoor and outdoor scenes.
It will be made publicly available to facilitate future research. 
See \cref{sec:dataset-details} for more details.

\paragraph{Training:}
We trained variants of \ours~on the pinhole and distorted images from \ourdataset,
each for 49 epochs, with a learning rate of $10^{-4}$ and a batch size of 24. 
We linearly warm up the learning rate in the first 1000 steps and drop it after 25 and 42 epochs by a factor of 10.
The LM optimization is stopped after 10 iterations with an initial damping $\lambda{=}0.1$. The entire training takes one day on two RTX 4090 GPUs.
Calibrating a single image takes about 100ms. %

\begin{table}[th!]
\newcommand{\deepm}[1]{\textcolor{Fuchsia}{#1}}
\newcommand{\linem}[1]{\textcolor{OliveGreen}{#1}}
\centering
\scriptsize
\renewcommand{\arraystretch}{1.0}%
\setlength\tabcolsep{2.5pt}%
\begin{tabular}{clcccccccccccc}
\toprule
&\multirow{2}{*}[-.4em]{Approach} 
& \multicolumn{4}{c}{Roll [degrees]} 
& \multicolumn{4}{c}{Pitch [degrees]} 
& \multicolumn{4}{c}{FoV [degrees]}\\
\cmidrule(lr){3-6}
\cmidrule(lr){7-10}
\cmidrule(lr){11-14}
&& error\,$\downarrow$ & \multicolumn{3}{c}{AUC\,$\triangleright$\,1/5/10\degree\,$\uparrow$}
& error\,$\downarrow$ & \multicolumn{3}{c}{AUC\,$\triangleright$\,1/5/10\degree\,$\uparrow$}
& error\,$\downarrow$ & \multicolumn{3}{c}{AUC\,$\triangleright$\,1/5/10\degree\,$\uparrow$} \\
\midrule
\multirow{9}{*}{\begin{sideways}\textbf{Stanford2D3D}~\cite{armeni2017joint}\end{sideways}}
&\deepm{DeepCalib*}~\cite{lopez2019deepcalib}      &          \01.59 &          33.8 &          63.9 & \cthird  79.2 &          \02.58 &          21.6 &          46.9 &          65.7 &          \06.67 &          \08.1 &           20.6 &          37.6 \\
&\deepm{Perceptual}~\cite{perceptual}              &          \02.08 &          26.8 &          53.8 &          70.7 &          \03.17 &          21.5 &          41.8 &          57.8 &           13.84 &          \02.8 &          \07.7 &          16.1 \\
&\deepm{CTRL-C}~\cite{ctrlc}                       &          \03.04 &          23.2 &          43.0 &          56.9 &          \03.43 &          18.3 &          38.6 &          53.8 &          \08.50 &          \07.7 &           18.2 &          31.5 \\
&\deepm{MSCC}~\cite{Song2024MSCC}                  &          \03.43 &          13.5 &          36.8 &          57.3 &          \02.64 &          22.6 &          45.0 &          60.5 & \cthird  \05.81 & \cthird  \09.6 & \cthird   23.8 & \cthird  41.6 \\
&\deepm{ParamNet}~\cite{jin2022PerspectiveFields}  &          \02.52 &          20.6 &          48.5 &          68.1 &          \02.78 &          20.9 &          44.2 &          61.5 &          \07.79 &          \07.4 &           18.0 &          33.2 \\
&\deepm{ParamNet*}~\cite{jin2022PerspectiveFields} & \cthird  \01.14 & \cthird  44.6 & \cthird  73.9 & \csecond 84.8 & \cthird  \01.94 & \cthird  29.2 & \cthird  56.7 & \csecond 73.1 &          \09.01 &          \05.8 &           14.3 &          27.8 \\
&\linem{SVA}~\cite{sva}                            &              -  &          21.7 &          24.6 &          25.8 &              -  &          15.4 &          19.9 &          22.4 &              -  &          \06.2 &           11.5 &          15.2 \\
&\linem{UVP}~\cite{Pautrat_2023_UncalibratedVP}    & \csecond \00.52 & \csecond 65.3 & \csecond 74.6 &          79.1 & \csecond \00.95 & \csecond 51.2 & \csecond 63.0 & \cthird  69.2 & \csecond \03.65 & \cfirst   22.2 & \csecond  39.5 & \csecond 51.3 \\
&\textbf{\ours*}                                   & \cfirst  \00.40 & \cfirst  83.1 & \cfirst  91.8 & \cfirst  94.8 & \cfirst  \00.93 & \cfirst  52.3 & \cfirst  74.8 & \cfirst  84.6 & \cfirst  \03.21 & \csecond  17.4 & \cfirst   40.0 & \cfirst  59.4 \\
\midrule
\multirow{9}{*}{\begin{sideways}\textbf{TartanAir}~\cite{tartanair2020iros}\end{sideways}}
&\deepm{DeepCalib*}~\cite{lopez2019deepcalib}      &          \01.95 &          24.7 &          55.4 &          71.5 &          \03.27 &          16.3 &          38.8 &          58.5 &          \08.07 &          \01.5 &          \08.8 &          27.2 \\
&\deepm{Perceptual}~\cite{perceptual}              &          \02.24 &          23.2 &          48.6 &          66.7 &          \02.86 &          23.5 &          44.6 &          61.5 &           15.06 &          \05.1 &          \08.9 &          17.1 \\
&\deepm{CTRL-C}~\cite{ctrlc}                       &          \01.68 &          32.8 &          59.1 & \csecond 74.1 & \csecond \02.39 & \cthird  24.6 & \cthird  48.6 & \csecond 65.2 & \csecond \05.64 & \cthird   10.7 & \cthird   25.4 & \csecond 43.5 \\
&\deepm{MSCC}~\cite{Song2024MSCC}                  &          \03.50 &          15.0 &          37.2 &          57.7 &          \03.48 &          18.8 &          38.6 &          54.3 &           11.18 &          \04.4 &           11.8 &          23.0 \\
&\deepm{ParamNet}~\cite{jin2022PerspectiveFields}  &          \02.33 &          23.3 &          51.4 &          71.0 &          \02.87 &          19.9 &          43.8 & \cthird  62.9 & \cthird  \06.04 &          \08.5 &           22.5 & \cthird  40.8 \\
&\deepm{ParamNet*}~\cite{jin2022PerspectiveFields} & \cthird  \01.63 & \cthird  34.5 & \cthird  59.2 & \cthird  73.9 &          \03.05 &          19.4 &          42.0 &          60.3 &          \08.21 &          \06.0 &           16.8 &          31.6 \\
&\linem{SVA}~\cite{sva}                            &          \09.48 &          32.4 &          39.6 &          44.1 &           18.46 &          21.2 &          28.8 &          34.5 &           43.01 &          \08.8 &           16.1 &          21.6 \\
&\linem{UVP}~\cite{Pautrat_2023_UncalibratedVP}    & \csecond \00.89 & \csecond 52.1 & \csecond 64.8 &          71.9 & \cthird  \02.48 & \csecond 36.2 & \csecond 48.8 &          58.6 &          \09.15 & \cfirst   15.8 & \csecond  25.8 &          35.7 \\
&\textbf{\ours*}                                   & \cfirst  \00.43 & \cfirst  71.3 & \cfirst  83.8 & \cfirst  89.8 & \cfirst  \01.49 & \cfirst  38.2 & \cfirst  62.9 & \cfirst  76.6 & \cfirst  \04.90 & \csecond  14.1 & \cfirst   30.4 & \cfirst  47.6 \\
\midrule
\multirow{9}{*}{\begin{sideways}\textbf{MegaDepth}~\cite{li2018megadepth}\end{sideways}}
&\deepm{DeepCalib*}~\cite{lopez2019deepcalib}      &          \01.41 &          34.6 &          65.4 &          79.4 &          \05.19 &          11.9 &          27.8 &          44.8 &           11.14 &          \05.6 &           12.1 &          22.9 \\
&\deepm{Perceptual}~\cite{perceptual}              &          \01.07 &          47.9 &          72.4 &          83.2 & \csecond \03.49 & \cthird  19.8 & \csecond 39.1 & \cthird  54.2 &           13.40 &          \02.9 &          \08.2 &          16.8 \\
&\deepm{CTRL-C}~\cite{ctrlc}                       & \cthird  \00.88 & \cthird  54.5 & \cthird  75.0 & \cthird  84.2 &          \04.80 &          16.6 &          33.2 &          46.5 &           18.65 &          \02.0 &          \05.8 &          12.8 \\
&\deepm{MSCC}~\cite{Song2024MSCC}                  &          \00.90 &          53.1 &          72.8 &          82.1 &          \05.73 &          19.0 &          33.2 &          44.3 & \csecond  10.80 &          \06.0 &           14.6 & \cthird  26.2 \\
&\deepm{ParamNet}~\cite{jin2022PerspectiveFields}  &          \01.46 &          37.0 &          66.4 &          80.8 & \cthird  \03.53 &          15.8 & \cthird  37.3 & \csecond 57.1 &           10.98 &          \05.3 &           12.8 &          24.0 \\
&\deepm{ParamNet*}~\cite{jin2022PerspectiveFields} &          \01.17 &          43.4 &          70.7 &          82.2 &          \03.99 &          15.4 &          34.5 &          53.3 &           11.01 &          \03.2 &           10.1 &          21.3 \\
&\linem{SVA}~\cite{sva}                            &              -  &          31.9 &          35.0 &          36.2 &              -  &          13.6 &          20.6 &          24.9 &              -  & \csecond \09.4 & \cthird   16.1 &          21.1 \\
&\linem{UVP}~\cite{Pautrat_2023_UncalibratedVP}    & \csecond \00.51 & \csecond 69.2 & \csecond 81.6 & \csecond 86.9 &          \04.59 & \csecond 21.6 &          36.2 &          47.4 & \cthird   10.92 & \cthird  \08.2 & \csecond  18.7 & \csecond 29.8 \\
&\textbf{\ours*}                                   & \cfirst  \00.36 & \cfirst  82.6 & \cfirst  90.6 & \cfirst  94.0 & \cfirst  \01.94 & \cfirst  32.4 & \cfirst  53.3 & \cfirst  67.5 & \cfirst  \04.46 & \cfirst   13.6 & \cfirst   31.7 & \cfirst  48.2 \\
\midrule
\multirow{9}{*}{\begin{sideways}\textbf{LaMAR}~\cite{sarlin2022lamar}\end{sideways}}
&\deepm{DeepCalib*}~\cite{lopez2019deepcalib}      &          \01.15 &           44.1 &           73.9 &           84.8 &          \04.68 &           10.8 &           28.3 &           49.8 &           10.93 &          \00.7 &           13.0 &           24.0 \\
&\deepm{Perceptual}~\cite{perceptual}              &          \01.29 &           40.0 &           68.9 &           81.6 &          \02.83 &           21.2 &           44.7 &           62.6 &           17.78 &          \03.0 &          \05.3 &           10.7 \\
&\deepm{CTRL-C}~\cite{ctrlc}                       &          \01.20 &           43.5 &           70.9 &           82.5 & \cthird  \01.94 & \cthird   27.6 & \cthird   54.7 & \csecond  70.2 & \cthird  \05.64 & \cthird  \09.8 & \cthird   24.6 & \cthird   43.2 \\
&\deepm{MSCC}~\cite{Song2024MSCC}                  &          \01.44 &           39.6 &           60.7 &           72.8 &          \03.02 &           20.9 &           41.8 &           55.7 &           14.78 &          \03.2 &          \08.3 &           16.8 \\
&\deepm{ParamNet}~\cite{jin2022PerspectiveFields}  &          \01.30 &           38.7 &           69.4 &           82.8 &          \02.77 &           19.0 &           44.7 &           65.7 &           15.15 &          \01.8 &          \06.2 &           13.2 \\
&\deepm{ParamNet*}~\cite{jin2022PerspectiveFields} & \cthird  \00.93 & \cthird   51.7 & \cthird   77.0 & \csecond  86.0 &          \02.15 &           27.0 &           52.7 & \csecond  70.2 &           14.71 &          \02.8 &          \06.8 &           14.3 \\
&\linem{SVA}~\cite{sva}                            &              -  &          \08.6 &          \09.2 &          \09.7 &              -  &          \03.4 &          \05.7 &          \07.0 &              -  &          \01.2 &          \02.7 &          \04.1 \\
&\linem{UVP}~\cite{Pautrat_2023_UncalibratedVP}    & \csecond \00.38 & \csecond  72.7 & \csecond  81.8 & \cthird   85.7 & \csecond \01.34 & \csecond  42.3 & \csecond  59.9 & \cthird   69.4 & \csecond \05.57 & \csecond  15.6 & \csecond  30.6 & \csecond  43.5 \\
&\textbf{\ours*}                                   & \cfirst  \00.28 & \cfirst   86.4 & \cfirst   92.5 & \cfirst   95.0 & \cfirst  \00.87 & \cfirst   55.0 & \cfirst   76.9 & \cfirst   86.2 & \cfirst  \03.03 & \cfirst   19.1 & \cfirst   41.5 & \cfirst   60.0 \\

\bottomrule
\end{tabular}
\caption{\textbf{Evaluation on diverse datasets.}
Approaches marked as * were retrained on our dataset \ourdataset.
We color the \colorbox{tabfirst}{best} and \colorbox{tabsecond}{second best} results for each metric.
\ours~is more accurate than \deepm{approaches based on learning} and more robust than \linem{those based on lines and vanishing points}.
\label{tbl:generalization}
}
\end{table}

\section{Experiments}
\label{sec:experiments}

We evaluate GeoCalib on single-image calibration and radial distortion estimation in a wide range of scenarios. Experiments are performed on a diverse range of real-world images (indoor, outdoor, and natural environments), and we analyze the impacts of our design decisions. 
We show some qualitative examples in~\cref{fig:qualitative}.
We also evaluate our gravity estimation as a prior for visual localization.

\subsection{Gravity and Field-of-View estimation}
\label{sec:generalization}

We first compare \ours~to existing deep and classical approaches with pinhole images from four real-world datasets.

\paragraph{Setup:}
Following~\cite{jin2022PerspectiveFields}, we evaluate uncalibrated pinhole scenarios, \ie no lens distortion. For each image, we evaluate the gravity estimation in terms of angular roll and pitch errors (in degrees), and the focal length in terms of the vertical field-of-view error (FoV, in degrees). For each metric, we report the median error and the Area Under the recall Curve (AUC) up to 1/5/10°. 

\paragraph{Datasets:}
We conduct this experiment on four popular datasets not seen during training.
i) Stanford2D3D~\cite{armeni2017joint} consists of images samples from 360° panoramas captured inside university buildings.
ii) TartanAir~\cite{tartanair2020iros} provides images from photo-realistic simulated environments and captures dynamic objects and various appearance changes.
Its images exhibit large variations in gravity directions but share the same FoV.
For both datasets, we use the same test set as~\cite{jin2022PerspectiveFields}, with about 2k images each.
iii) The MegaDepth dataset~\cite{li2018megadepth} was built from crowd-sourced images covering popular phototourism landmarks using SfM. 
We align the respective 3D models to gravity using COLMAP~\cite{schoenberger2016sfm} and sample a total of 2k images with varying intrinsics from the scenes in the IMC 2021 test set~\cite{imwchallenge2021}.
iv) LaMAR~\cite{sarlin2022lamar} is an AR dataset for visual localization, captured over multiple years in university buildings and a city center.
We use 2k images from the phone sequences.

\paragraph{Baselines:}
We benchmark our method against the deep methods DeepCalib~\cite{lopez2019deepcalib}, CTRL-C~\cite{ctrlc}, Perceptual~\cite{perceptual}, MSCC~\cite{Song2024MSCC} and ParamNet~\cite{jin2022PerspectiveFields}. We follow the suggested evaluation setup of each method and refer to \cref{sec:impl-details} for more details. 
For fairness, we also retrain DeepCalib~\cite{lopez2019deepcalib} and ParamNet~\cite{jin2022PerspectiveFields} on our dataset.
We also evaluate two recent classical approaches based on VPs: SVA~\cite{sva} and UVP~\cite{Pautrat_2023_UncalibratedVP}, which both assume a Manhattan world (3 orthogonal vanishing points) and rely on minimal solvers for VP estimation. 

\paragraph{Results:}
\Cref{tbl:generalization} shows that GeoCalib largely improves on top of all deep single-image calibration networks, and outperforms classical methods in all metrics, except for the finest threshold on FoV on TartanAir~\cite{tartanair2020iros} and Stanford2D3D~\cite{armeni2017joint}.
UVP~\cite{Pautrat_2023_UncalibratedVP} assumes a Manhattan world, and this stronger assumption about scene configuration enables slightly more accurate predictions on easy samples, but completely fails in other scenarios. 
In contrast, GeoCalib is the first deep method that consistently matches or surpasses the accuracy of classical methods without any assumption on the scene, thus combining the accuracy of classical methods with the robustness of deep neural networks. 
We conclude that our method is more accurate in gravity estimation than FoV, likely due to the weaker constraint from latitude, which is generally harder to predict.

\begin{figure}[t]
    \centering

    \def\ncols{6}
    \setlength{\pwidth}{0.005\linewidth}
    \setlength{\iwidth}{\dimexpr(0.999\linewidth - \ncols\pwidth + \pwidth)/\ncols \relax}

    \includegraphics[width=\iwidth]{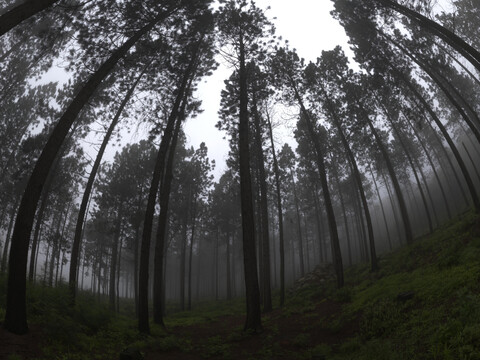}%
    \hspace{\pwidth}%
    \includegraphics[width=\iwidth]{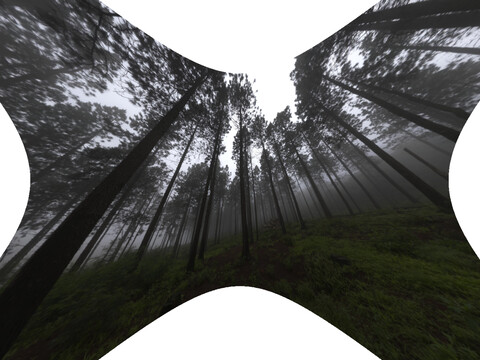}%
    \hspace{\pwidth}%
    \includegraphics[width=\iwidth]{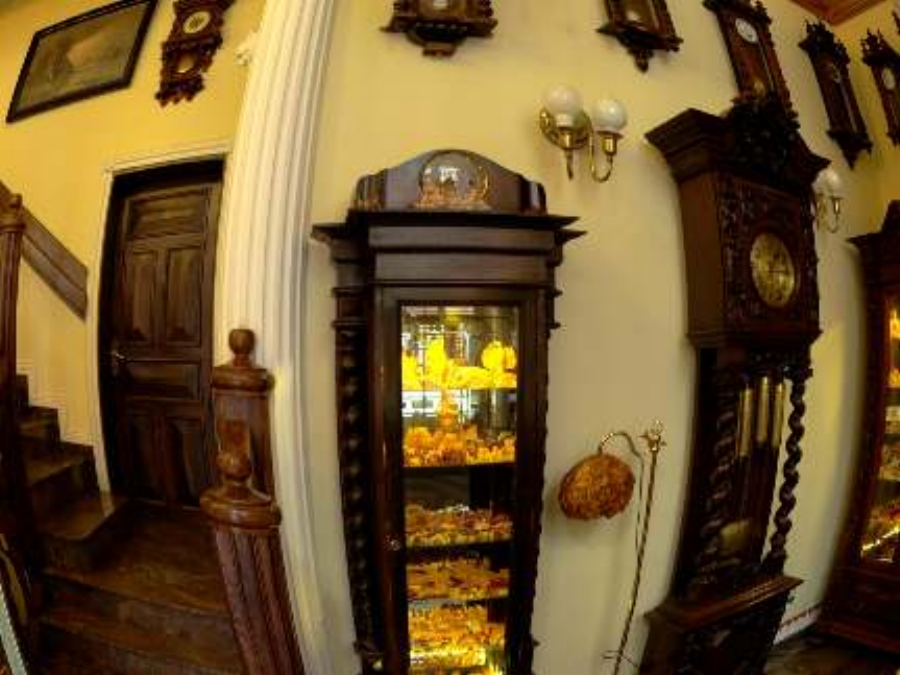}%
    \hspace{\pwidth}%
    \includegraphics[width=\iwidth]{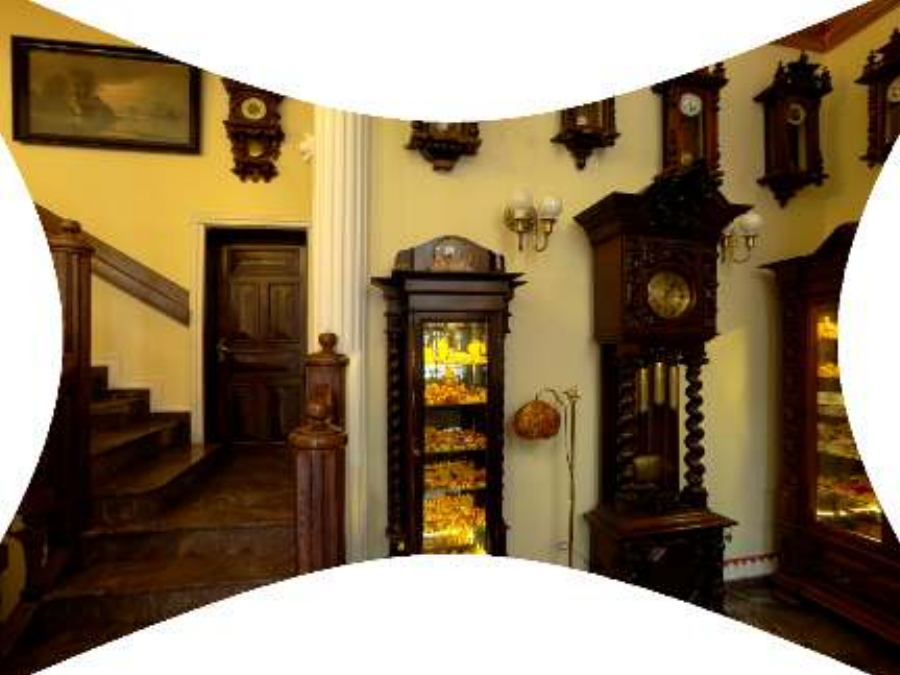}%
    \hspace{\pwidth}%
    \includegraphics[width=\iwidth]{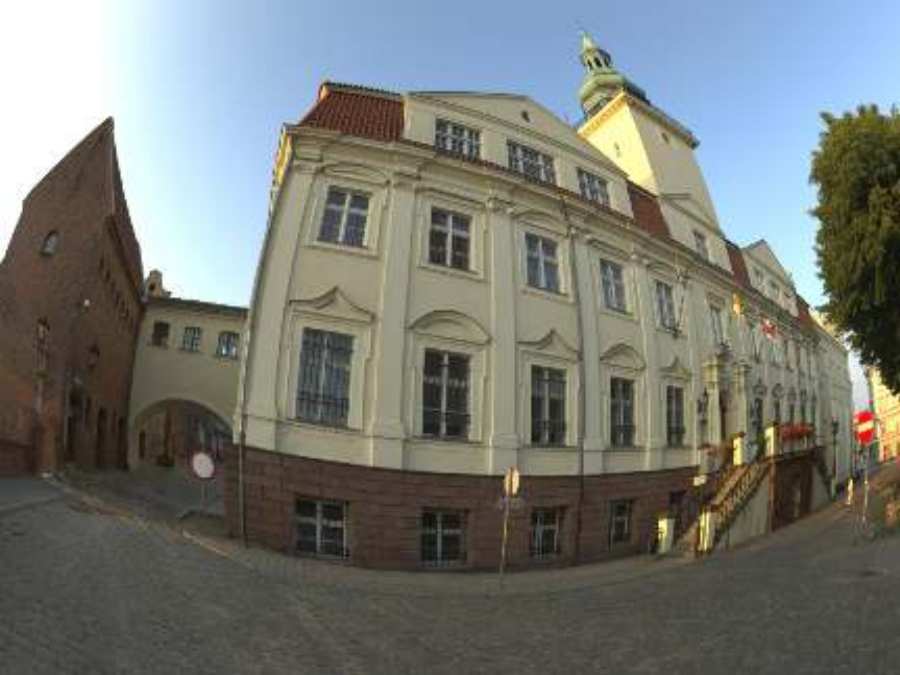}%
    \hspace{\pwidth}%
    \includegraphics[width=\iwidth]{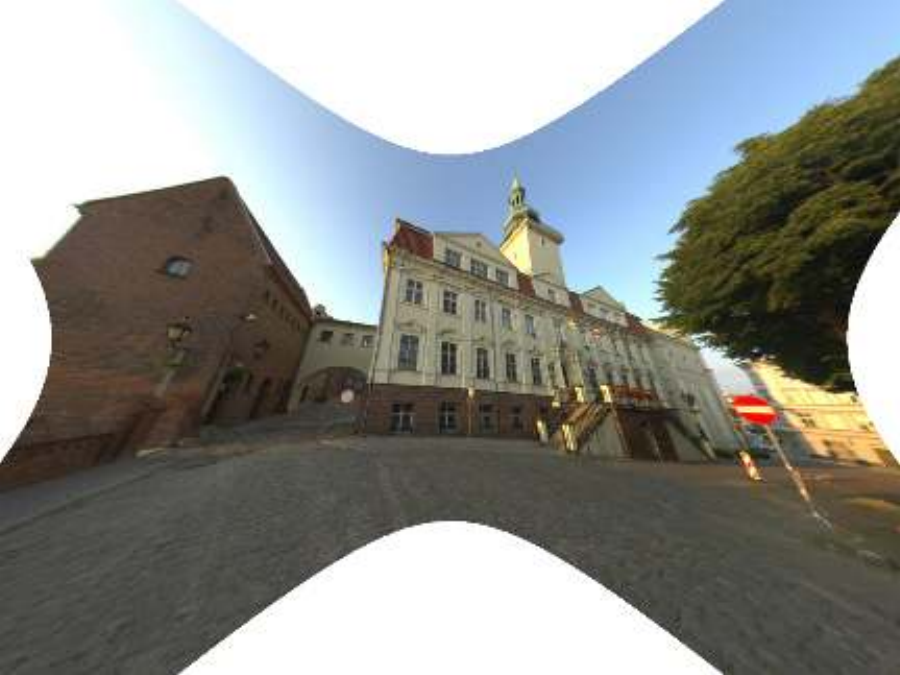}%
    
    \caption{\textbf{Image undistortion.} GeoCalib can robustly predict lens distortion from a single image (left), which can be used to rectify images (right) from any source.
    }%
    \label{fig:distortion}%
\end{figure}
\begin{table}[t]
\centering
\scriptsize
\renewcommand{\arraystretch}{1.0}%
\setlength\tabcolsep{5.0pt}%
\begin{tabular}{lcccccc}
\toprule
\multirow{2}{*}[-.4em]{Approach} 
&FoV [degrees] & $\dist_1$ & \multicolumn{4}{c}{Pixel Distortion Error} \\
\cmidrule(lr){2-2}
\cmidrule(lr){3-3}
\cmidrule(lr){4-7}
& error\,$\downarrow$& error\,$\downarrow$ & \multicolumn{4}{c}{recall\,$\triangleright$\,0.5/1/3/5px\,$\uparrow$} \\
\midrule

Perceptual~\cite{perceptual}              &           13.50 &    (spherical)  &          \00.97 &          \03.42 &          19.67 &          30.01 \\
SVA~\cite{sva}                            &              -  &   (divisional)  &           22.28 &           32.01 &          52.50 &          63.67 \\
DeepCalib*~\cite{lopez2019deepcalib}      & \cthird   10.80 & \cthird  \00.09 & \cthird   23.89 & \cthird   34.32 & \cthird  55.82 & \cthird  67.06 \\
ParamNet*~\cite{jin2022PerspectiveFields} &           11.17 &          \00.10 &           22.56 &           32.37 &          53.07 &          64.34 \\
\textbf{\ours~- pinhole*}                 & \csecond \04.55 & \csecond \00.07 & \csecond  25.92 & \csecond  37.10 & \csecond 60.40 & \csecond 72.01 \\
\textbf{\ours*}                           & \cfirst  \04.38 & \cfirst  \00.06 & \cfirst   28.36 & \cfirst   40.28 & \cfirst  64.14 & \cfirst  75.57 \\

\bottomrule
\end{tabular}
\caption{\textbf{Calibration of distorted images.} 
On Internet images from the MegaDepth~\cite{li2018megadepth} dataset with radial distortion~\cite{faugeras1992camera}, GeoCalib estimates a more accurate distortion than all other approaches, even if it is trained only on pinhole images.
\label{tbl:distortion}%
}
\end{table}

\subsection{Lens distortion}

We now evaluate how \ours~can handle radial distortion and estimate it.

\paragraph{Setup:}
We evaluate camera distortion estimation on the MegaDepth dataset~\cite{li2018megadepth} with crowd-sourced, distorted images. These images were taken from various sensors and thus exhibit a diverse distribution of distortions. The ground-truth distortion models were obtained from SfM~\cite{schoenberger2016sfm}. Similar to these GT camera models, we use a polynomial distortion model~\cite{faugeras1992camera} with one parameter and report the median error. We also report the pixel distortion error, defined as the Euclidean distance between the pixel distorted by the ground truth camera model and the distortion from the predicted model but with GT focal length. In contrast to previous works~\cite{sva}, we use the GT focal length to disentangle the distortion estimation from FoV, which can bias the evaluation if the FoV estimate is incorrect. We average the pixel distortion errors over each image and compute the recall of the average distortion error within an image over the entire dataset. For completeness, we also report the median FoV error for each method.

\paragraph{Baselines:} 
We compare our method against Perceptual~\cite{perceptual}, which uses a spherical distortion model~\cite{usenko2018double}, SVA~\cite{sva} (divisional model~\cite{fitzgibbon2001simultaneous}), as well as DeepCalib~\cite{lopez2019deepcalib} and PerspectiveNet+ParamNet~\cite{jin2022PerspectiveFields} trained on our dataset with a single parameter polynomial distortion model~\cite{faugeras1992camera}.
We evaluate both variants of \ours~trained with pinhole and distorted images.

\paragraph{Results:}
\Cref{tbl:distortion} reports the results.
\ours-pinhole already improves over all baselines, suggesting that the model can zero-shot generalize to radial distortion through optimization. 
Training our model on distorted images further boosts the accuracy.
\Cref{fig:distortion} shows that \ours~can accurately handle strong lens distortion on a diverse range of images.

\subsection{Insights}
\label{sec:insights}

\begin{figure}[bt]
    \centering
    \includegraphics[width=\linewidth]{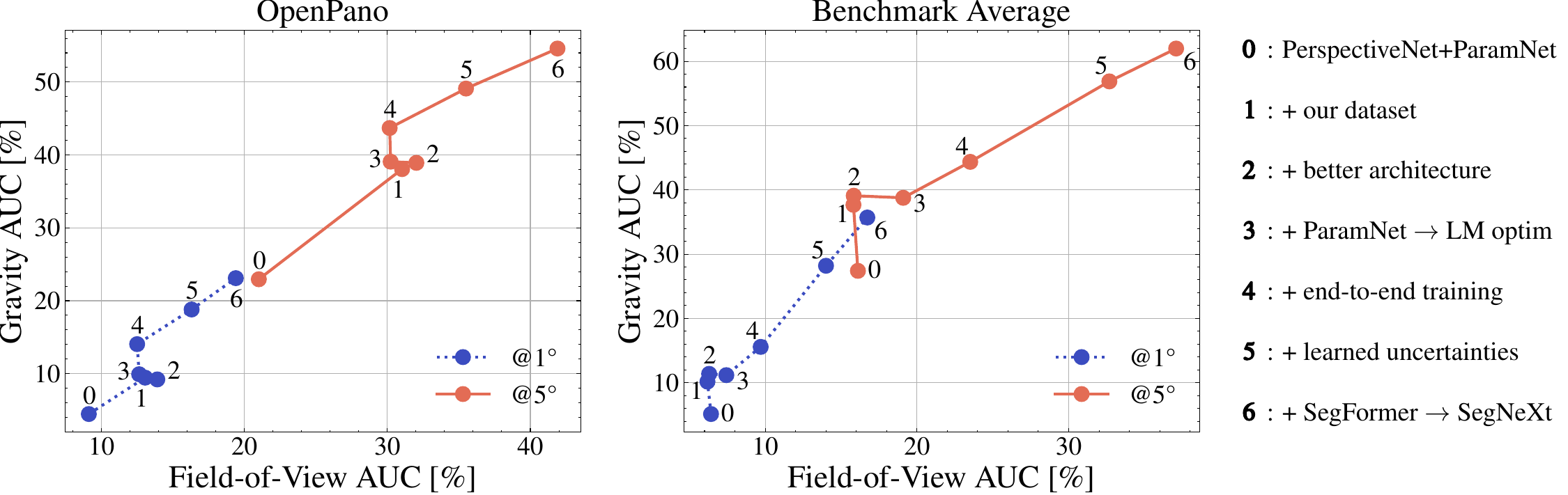}%
    \caption{\textbf{Ablation study.} Going from PerspectiveNet+ParamNet~\cite{jin2022PerspectiveFields} to \ours. 
    }%
    \label{fig:ablation}%
\end{figure}

\begin{figure}[bt]
    \centering
    \setlength{\pwidth}{0.05\linewidth}
    \setlength{\iwidth}{0.4\linewidth}
    \includegraphics[width=\linewidth]{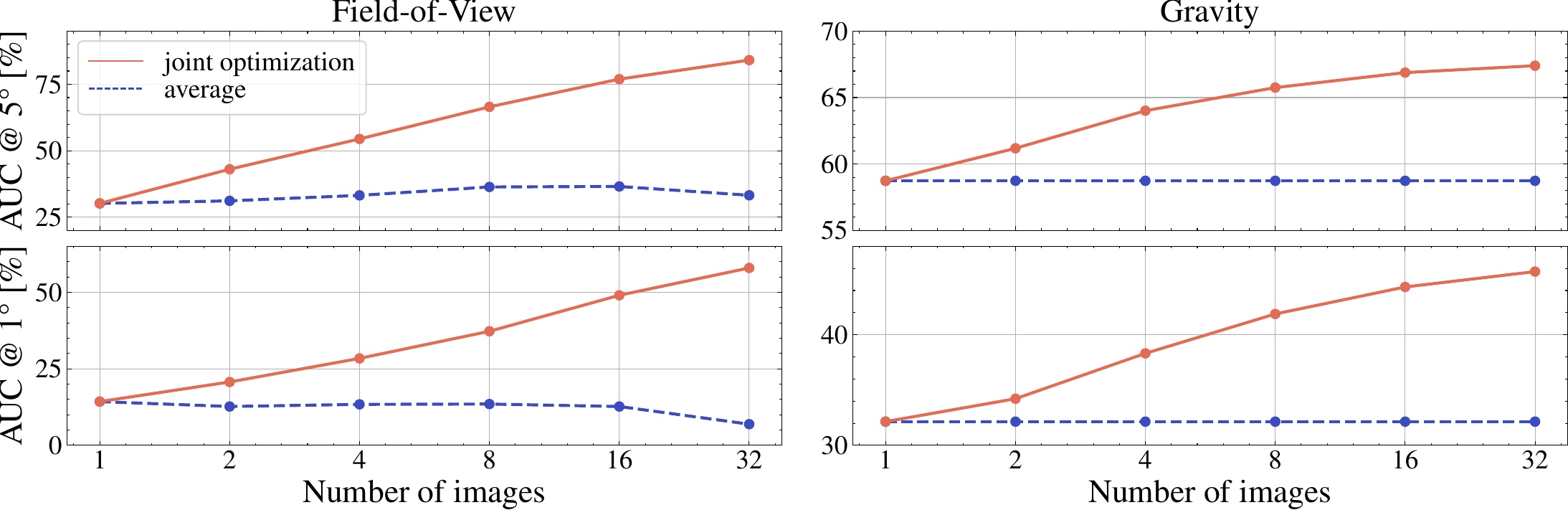}%
    \caption{\textbf{Multi-image optimization.} 
    Simultaneously optimizing multiple images with shared intrinsic parameters improves the estimation accuracy of both field of view (left) and gravity direction (right).
    This is useful for calibrating an image sequence.
    }%
    \label{fig:multi-opt}%
\end{figure}
\begin{table}[t]
\centering
\scriptsize
\renewcommand{\arraystretch}{1.0}%
\setlength\tabcolsep{5.0pt}%
\begin{tabular}{lcccccccccc}
\toprule
\multirow{2}{*}[-.4em]{Dataset} 
& \multicolumn{2}{c}{known?} 
& \multicolumn{4}{c}{Gravity [degrees]}
& \multicolumn{4}{c}{FoV [degrees]}\\
\cmidrule(lr){2-3}
\cmidrule(lr){4-7}
\cmidrule(lr){8-11}
& gravity & FoV
& error\,$\downarrow$ & \multicolumn{3}{c}{AUC\,$\triangleright$\,1/5/10\degree\,$\uparrow$}
& error\,$\downarrow$ & \multicolumn{3}{c}{AUC\,$\triangleright$\,1/5/10\degree\,$\uparrow$} \\
\midrule
\multirow{3}{*}{Stanford2D3D} 
& \xmark & \xmark &          \01.12 &          45.6 &          71.5 &          82.7 &          \03.21 &          17.4 &          40.0 &          59.4 \\
& \xmark & \cmark & \cfirst  \00.86 & \cfirst  57.0 & \cfirst  79.0 & \cfirst  87.0 &              -  &            -  &            -  &            -  \\
& \cmark & \xmark &              -  &            -  &            -  &            -  & \cfirst  \02.30 & \cfirst  24.7 & \cfirst  50.2 & \cfirst  67.5 \\
\midrule
\multirow{3}{*}{TartanAir} 
& \xmark & \xmark &          \01.69 &          31.6 &          58.6 &          73.4 &          \04.90 &          14.1 &          30.4 &          47.6 \\
& \xmark & \cmark & \cfirst  \01.18 & \cfirst  44.7 & \cfirst  66.8 & \cfirst  78.6 &              -  &            -  &            -  &            -  \\
& \cmark & \xmark &              -  &            -  &            -  &            -  & \cfirst  \03.21 & \cfirst  20.0 & \cfirst  40.4 & \cfirst  56.9 \\
\midrule
\multirow{3}{*}{MegaDepth} 
& \xmark & \xmark &          \02.10 &          28.1 &          50.8 &          65.8 &          \04.46 &          13.6 &          31.7 &          48.2 \\
& \xmark & \cmark & \cfirst  \01.41 & \cfirst  37.6 & \cfirst  62.3 & \cfirst  74.2 &              -  &            -  &            -  &            -  \\
& \cmark & \xmark &              -  &            -  &            -  &            -  & \cfirst  \02.42 & \cfirst  27.0 & \cfirst  48.8 & \cfirst  64.3 \\
\midrule
\multirow{3}{*}{LaMAR} 
& \xmark & \xmark &          \00.99 &          50.1 &          74.5 &          84.5 &          \03.03 &          19.1 &          41.5 &          60.0 \\
& \xmark & \cmark & \cfirst  \00.82 & \cfirst  58.2 & \cfirst  80.2 & \cfirst  88.2 &              -  &            -  &            -  &            -  \\
& \cmark & \xmark &              -  &            -  &            -  &            -  & \cfirst  \02.28 & \cfirst  23.5 & \cfirst  50.0 & \cfirst  67.4 \\
\bottomrule
\end{tabular}
\caption{\textbf{Partial calibration.}
When a subset of the parameters is known, the optimization is better constrained and estimates more accurately the remaining parameters.
\label{tbl:prior}%
}
\vspace{-4mm}
\end{table}

\begin{table}[t]
\centering
\scriptsize
\renewcommand{\arraystretch}{1.0}%
\setlength\tabcolsep{6.0pt}%
\begin{tabular}{clcccccc}
\toprule
&\multirow{2}{*}[-.4em]{Approach} 
& \multicolumn{3}{c}{DUC 1}
& \multicolumn{3}{c}{DUC2}\\
\cmidrule(lr){3-5}
\cmidrule(lr){6-8}
&& \multicolumn{6}{c}{Recall at (0.25m,10°) / (0.5m,10°) / (1.0m,10°)\,$\uparrow$}\\
\midrule
& Baseline (SP+SG)                                      &          43.4 &          66.7 &          78.3 &          51.9 &          74.8 & \cfirst  78.6 \\
& + gravity in refinement                           &          44.4 &          68.2 &          78.8 &          54.2 &          74.8 & \cfirst  78.6 \\
& + gravity in RANSAC                               & \cfirst  44.9 & \cfirst  69.2 & \cfirst  79.3 & \cfirst  55.0 & \cfirst  76.3 & \cfirst  78.6 \\
& gravity from UVP~\cite{Pautrat_2023_UncalibratedVP} &          43.9 &          67.7 &          78.8 &          52.7 & \cfirst  76.3 & \cfirst  78.6 \\
\bottomrule
\end{tabular}
\caption{\textbf{Visual Localization on InLoc~\cite{taira2018inloc}.} 
The gravity prior estimated by \ours~improves pose estimation and is more effective than the estimate of UVP~\cite{Pautrat_2023_UncalibratedVP}.
\label{tbl:inloc}
}
\vspace{-3mm}
\end{table}

\paragraph{Ablation study:}
We perform an extensive ablation study to verify the design decisions of our method. 
We start from the closest baseline PerspectiveNet + ParamNet~\cite{jin2022PerspectiveFields}, trained on \textit{360 cities}, and train 5 different models on our \ourdataset~dataset.
We evaluate the trained models on the 2k test images sampled from \ourdataset~as well as on the previous benchmarks, averaging their results.

We plot the results in~\cref{fig:ablation}.
\b{(1)}~While \ourdataset~is about 5 times smaller than \textit{360 Cities}, it is more balanced across different domains, resulting in improvements on the test set and on the benchmarks.
\b{(2)}~The improved architecture as described in~\cref{sec:approach} slightly increases the accuracy both in- and out-of-domain.
\b{(3)}~When switching from ParamNet~\cite{jin2022PerspectiveFields} to our second-order optimization, the FoV estimation drops in domain but gets more accurate across the benchmarks. This shows that ParamNet is unable to generalize to out-of-domain images.
\b{(4)}~Supervising the result of the optimization and \b{(5)}~learning uncertainties significantly boost the accuracy across the board as this i) allows \ours~to increase the weight for accurate predictions, improving both accuracy and robustness, and ii) enables it to focus more on well-constrained areas of the image, while saving capacity in less informative regions.
\b{(6)}~Finally, replacing the SegFormer~\cite{xie2021segformer} backbone with the more recent SegNeXt~\cite{guo2022segnext} also brings improvements.
Despite relying on classical geometry, \ours~will thus benefit from further advances in deep learning architectures.
 See \cref{sec:extended-ablation} for an extended analysis.

\paragraph{Partial Calibration:} In many real-world scenarios, we either have access to GT gravity (from IMU) or known intrinsics. We conduct an additional experiment as in~\cref{sec:generalization} to showcase how GeoCalib can leverage such priors. Here, we either fix the gravity direction or the intrinsics (FoV) in the optimization (\cref{eq:objective}) and predict the other. We report the angular error of gravity and FoV.

\Cref{tbl:prior} shows that partial calibration consistently improves the accuracy of our joint optimization. Fixing the FoV improves the gravity estimation by up to $13.1\%$ AUC@1° on TartanAir~\cite{tartanair2020iros}. On the other hand, fixing the gravity direction improves the intrinsics estimation across all thresholds equally. 

\paragraph{Multi-image optimization:}
\label{sec:multi-image}
We evaluate \ours's ability to jointly estimate camera intrinsics for multiple images captured by the same camera.
We randomly sample sets of $N$ images from the TartanAir~\cite{tartanair2020iros} dataset and report the AUC of the gravity and FoV at 1\degree~and 5\degree.

\Cref{fig:multi-opt} shows that our joint optimization estimates progressively more accurate FoVs as more images are considered simultaneously. 
This also improves the estimate of the gravity, even though it is different for each image.
In contrast, simply averaging the independently-estimated FoVs over all images is less effective and cannot benefit the gravity estimation. 
Thus, \ours~can effectively leverage geometric priors across images, even if they do not have any visual overlap.

\subsection{Indoor Visual Localization}
We evaluate the accuracy of gravity-aided visual localization 
on the InLoc~\cite{taira2018inloc} indoor dataset following hloc~\cite{sarlin2019coarse,superpoint,sarlin2020superglue}. 
We report the pose recall under three different thresholds on the two locations DUC1 and DUC2. 
The absolute pose is estimated with RANSAC and LM-refinement from 2D-3D correspondences.

Because the gravity estimates are noisy, we avoid upright solvers~\cite{kukelova2010vertical,PoseLib} and instead add soft constraints.
In pose refinement, we add a regularization on the rotation towards the estimated gravity~\cite{ceres-solver}. 
In RANSAC, we add a constant reward to the MSAC score if the estimated gravity is within $2\sigma$ of our estimated gravity uncertainty.
As a baseline, we use the gravity estimated by UVP~\cite{Pautrat_2023_UncalibratedVP}.

We report the results in~\cref{tbl:inloc}.
Adding the gravity constraint in both RANSAC and pose refinement improves the localization accuracy.
The baseline UVP~\cite{Pautrat_2023_UncalibratedVP} is less effective because its gravity estimates are less accurate.

\section{Conclusion}
\label{sec:conclusion}
This paper introduces \ours, a new approach to single-image calibration that combines the best of learning and geometry.
Thanks to its differentiable optimization, it learns strong priors that make it both more accurate and more robust than existing approaches, with a strong generalization to different environments.
\ours~offers much flexibility in terms of camera models, priors, and uncertainties and is thus easier to integrate in downstream applications.

\paragraph{Acknowledgments:} Thanks to Xu Song for helping to evaluate MSCC. Thanks to Jean-François Lalonde for allowing us to release models trained on the Laval Indoor dataset. Thanks to Shaohui Liu and Linfei Pan for their valuable feedback. Philipp Lindenberger was supported by an ETH Zurich RobotX Fellowship.

\ifaddsupp
\appendix

\section{Extended ablation study}
\label{sec:extended-ablation}
\begin{table}[ht]
\centering
\scriptsize
\renewcommand{\arraystretch}{1.0}%
\setlength\tabcolsep{4.1pt}%
\begin{tabular}{clcccccccc}
\toprule
&\multirow{2}{*}[-.4em]{Approach} 
& \multicolumn{4}{c}{Gravity [degrees]}
& \multicolumn{4}{c}{FoV [degrees]}\\
\cmidrule(lr){3-6}
\cmidrule(lr){7-10}
&& error\,$\downarrow$ & \multicolumn{3}{c}{AUC\,$\triangleright$\,1/5/10\degree\,$\uparrow$}
& error\,$\downarrow$ & \multicolumn{3}{c}{AUC\,$\triangleright$\,1/5/10\degree\,$\uparrow$} \\
\midrule
\multirow{8}{*}{\begin{sideways}\textbf{\ourdataset}\end{sideways}}
&PerspectiveNet+ParamNet~\cite{jin2022PerspectiveFields} &          \04.99 &          \04.5 &          23.0 &          43.7 &          \06.63 &          \09.1 &          21.0 &          37.7 \\
&+ our dataset (\ourdataset)                             &          \03.08 &          \09.4 &          38.1 &          60.2 & \cthird  \04.40 & \cthird   13.1 & \cthird  31.0 & \cthird  51.5 \\
&+ better architecture                                   & \cthird  \03.01 &          \09.2 &          38.9 &          60.9 & \cthird  \04.23 & \cthird   13.9 &  \cthird 32.0 & \cthird  52.5 \\
&+ ParamNet $\rightarrow$ SGD optim                      &          \03.03 & \cthird   10.3 & \cthird  39.1 & \cthird  61.0 &          \04.75 &           12.4 &          29.3 &          49.2 \\
&+ ParamNet $\rightarrow$ LM optim                       &          \03.03 &          \09.9 & \cthird  39.1 & \cthird  61.0 &          \04.58 &           12.6 &          30.2 &          50.7 \\
&+ end-to-end training                                   & \cthird  \02.65 & \cthird    14.0 & \cthird 43.7 & \cthird  64.0 &          \04.54 &           12.5 &          30.2 &          49.7 \\
&+ learned uncertainties                                 & \csecond \02.24 & \csecond  18.8 & \csecond 49.1 & \csecond 67.3 & \csecond \03.85 & \csecond  16.3 & \csecond 35.5 & \csecond 54.4 \\

&+ SegFormer $\rightarrow$ SegNext = \textbf{ours}      & \cfirst  \01.88 & \cfirst   23.1 & \cfirst  54.6 & \cfirst  71.1 & \cfirst  \03.02 & \cfirst   19.4 & \cfirst  41.9 & \cfirst  60.7 \\

\midrule
\multirow{8}{*}{\begin{sideways}\textbf{MegaDepth}\end{sideways}}
&PerspectiveNet+ParamNet~\cite{jin2022PerspectiveFields} &          \04.06 &          \05.4 &          28.3 &          51.5 &           10.98 &          \05.3 &          12.8 &          24.0 \\
&+ our dataset (\ourdataset)                             &          \04.51 &          \06.8 &          28.0 &          48.6 &           11.01 &          \03.2 &          10.1 &          21.3 \\
&+ better architecture                                   &          \04.35 &          \08.8 &          29.9 &          50.4 &           11.23 &          \03.9 &          10.0 &          20.8 \\
&+ ParamNet $\rightarrow$ SGD optim                      &          \04.22 &          \07.4 &          29.5 &          50.9 &          \08.88 &          \05.8 &          15.1 &          29.0 \\
&+ ParamNet $\rightarrow$ LM optim                       &          \04.12 &          \08.2 &          30.1 &          51.2 &          \08.70 &          \05.4 &          15.6 &          29.6 \\
&+ end-to-end training                                   & \cthird  \03.31 & \cthird   10.8 & \cthird  35.4 & \cthird  57.6 & \cthird  \07.22 & \cthird  \08.1 & \cthird  18.4 & \cthird  34.8 \\
&+ learned uncertainties                                 & \csecond \02.36 & \csecond  20.7 & \csecond 48.1 & \csecond 65.1 & \csecond \04.80 & \csecond  11.7 & \csecond 28.9 & \csecond 46.3 \\
&+ SegFormer $\rightarrow$ SegNext = \textbf{ours}      & \cfirst  \02.10 & \cfirst   28.1 & \cfirst  50.8 & \cfirst  65.8 & \cfirst  \04.46 & \cfirst   13.6 & \cfirst  31.7 & \cfirst  48.2 \\

\midrule
\multirow{8}{*}{\begin{sideways}\textbf{LaMAR}\end{sideways}}
&PerspectiveNet+ParamNet~\cite{jin2022PerspectiveFields} &          \03.46 &          \06.9 &          34.4 &          59.4 &           15.15 &          \01.8 &          \06.2 &          13.2 \\
&+ our dataset (\ourdataset)                             &          \02.57 &           13.6 &          44.7 &          65.3 &           14.71 &          \02.8 &          \06.8 &          14.3 \\
&+ better architecture                                   &          \02.53 &           14.8 &          46.0 &          66.2 &           13.93 &          \02.3 &          \06.4 &          14.4 \\
&+ ParamNet $\rightarrow$ SGD optim                      &          \02.54 &           14.2 &          45.4 &          66.0 &           11.71 &          \04.9 &           12.5 &          23.0 \\
&+ ParamNet $\rightarrow$ LM optim                       &          \02.54 &           14.4 &          45.4 &          66.1 &           11.24 &          \04.6 &           12.5 &          23.6 \\
&+ end-to-end training                                   & \cthird  \02.14 & \cthird   19.5 & \cthird  51.0 & \cthird  71.2 & \cthird  \06.34 & \cthird  \07.9 & \cthird   20.2 & \cthird  38.5 \\
&+ learned uncertainties                                 & \csecond \01.31 & \csecond  37.3 & \csecond 67.2 & \csecond 80.3 & \csecond \03.66 & \csecond  15.9 & \csecond  36.3 & \csecond 54.8 \\
&+ SegFormer $\rightarrow$ SegNeXt = \textbf{ours}      & \cfirst  \00.99 & \cfirst   50.1 & \cfirst  74.5 & \cfirst  84.5 & \cfirst  \03.03 & \cfirst   19.1 & \cfirst   41.5 & \cfirst  60.0 \\

\bottomrule
\end{tabular}
\caption{\textbf{Ablation study.}
We color the \colorbox{tabfirst}{best} and \colorbox{tabsecond}{second best} results for each metric.
\label{tbl:ablation}%
}
\end{table}
We perform an extensive ablation study to verify the design decisions of our method. 
We start from the closest baseline PerspectiveNet + ParamNet~\cite{jin2022PerspectiveFields}, trained on \textit{360 cities}, and train 5 different models on our \ourdataset~dataset.
We evaluate the trained models on the MegaDepth~\cite{li2018megadepth} and LaMAR~\cite{sarlin2022lamar} test splits and on the 2k test images sampled from \ourdataset~. We
report the median and AUC of angular gravity and FoV errors.

We report the results in~\cref{tbl:ablation}. 
\b{1)}~Re-training PerspectiveNet + ParamNet~\cite{jin2022PerspectiveFields} on \ourdataset{}, which is 5 times smaller than \textit{360 cities}~\cite{jin2022PerspectiveFields}, yields large improvements on the test split of \ourdataset~and LaMAR, but no improvements or slightly weaker generalization to MegaDepth.
We conclude that despite the smaller size of our dataset, it enables strong generalization with higher quality and more evenly sampled images across domains at a fraction of the size.
\b{2)}~The improved architecture, as described in~\cref{sec:approach} increases the accuracy both in- and out-of-domain. The improved low-level features enable a more accurate prediction of up-vectors, which results in more precise gravity estimations.
\b{3)}~Replacing ParamNet with the optimization proposed in~\cite{jin2022PerspectiveFields} drops the accuracy in-domain, but improves FoV estimation on MegaDepth and LaMAR. This suggests that ParamNet is not able to generalize well on FoV estimation.
\b{4)}~When switching from ParamNet~\cite{jin2022PerspectiveFields} to our second-order optimization, we observe that it does not improve over the optimization proposed in~\cite{jin2022PerspectiveFields} out-of-the-box. 
\b{5)}~However, propagating the gradients through the LM-optimization enables the network to learn globally consistent up-vectors and latitude, which results in more robustness and generalization, concluded from large improvements for gravity and FoV estimation both in LaMAR and MegaDepth.
\b{6)}~Letting the network predict uncertainties on the perspective fields and propagating them through the LM optimization greatly improves the accuracy. These uncertainties i) allow the network to increase the weight for accurate predictions, improving both accuracy and robustness, and ii) enable the network to focus more on well-constrained areas of the image, while saving capacity in less informative regions. This results in a much stronger generalization of our model, supported by the 2x improvements at fine thresholds on MegaDepth and LaMAR.
\b{7)}~Finally, replacing the SegFormer~\cite{xie2021segformer} architecture with the more efficient SegNeXt~\cite{guo2022segnext} increases our model's accuracy, showing that \ours~will benefit from advances in deep learning architectures despite relying on classical geometry.

\begin{table}[tb]
\centering
\scriptsize
\renewcommand{\arraystretch}{1.0}%
\setlength\tabcolsep{5pt}%
\begin{tabular}{clcccccccc}
\toprule
&\multirow{2}{*}[-.4em]{Approach} 
& \multicolumn{4}{c}{Gravity [degrees]}
& \multicolumn{4}{c}{FoV [degrees]}\\
\cmidrule(lr){3-6}
\cmidrule(lr){7-10}
&& error\,$\downarrow$ & \multicolumn{3}{c}{AUC\,$\triangleright$\,1/5/10\degree\,$\uparrow$}
& error\,$\downarrow$ & \multicolumn{3}{c}{AUC\,$\triangleright$\,1/5/10\degree\,$\uparrow$} \\
\midrule
\multirow{7}{*}{\begin{sideways}\textbf{\ourdataset}\end{sideways}}
&Trivial    &  34.36 & \00.1 & \00.3 & \01.1 &  22.60 & \01.8 & \05.5 & 11.0 \\
&Heuristic  & \04.74 & \06.0 &  26.5 &  46.2 & \06.94 & \09.1 &  21.3 & 37.1 \\
&Solver     & \06.01 &  11.4 &  30.3 &  41.5 &  11.79 &  10.4 &  21.8 & 32.0 \\
& \textbf{\ours} &&&&&&&& \\
&+Trivial   & \01.88 &  23.1 &  54.6 &  71.1 & \03.02 &  19.4 &  41.9 & 60.7 \\
&+Heuristic & \01.90 &  23.1 &  54.6 &  71.0 & \03.09 &  19.3 &  41.8 & 60.5 \\
&+Solver    & \02.03 &  21.4 &  51.8 &  67.7 & \03.31 &  18.3 &  39.7 & 57.8 \\

\midrule
\multirow{7}{*}{\begin{sideways}\textbf{MegaDepth}\end{sideways}}
&Trivial    &  21.90 & \00.7 & \02.8 & \07.4 &  21.07 & \00.5 & \02.8 & \08.3 \\
&Heuristic  & \06.93 & \04.6 &  16.7 &  35.7 & \08.45 & \05.2 &  14.8 &  29.4 \\
&Solver     & \08.69 & \07.0 &  18.8 &  31.9 &  16.67 & \04.0 &  10.5 &  19.0 \\
& \textbf{\ours} &&&&&&&& \\
&+Trivial   & \02.10 &  28.1 &  50.8 &  65.8 & \04.46 &  13.6 &  31.7 &  48.2 \\
&+Heuristic & \02.09 &  28.2 &  51.0 &  65.9 & \04.50 &  13.6 &  31.6 &  48.1 \\
&+Solver    & \02.14 &  27.8 &  50.7 &  65.6 & \04.54 &  13.4 &  31.4 &  47.8 \\

\bottomrule
\end{tabular}
\caption{\textbf{Initialization strategies.} 
While the heuristic and solver strategies are more accurate than the trivial strategy, their impact on the result of the optimization is minimal because its convergence is robust.
\label{tbl:initialization}%
}
\end{table}

\section{Initialization strategies}
\label{sec:initialization}

We describe and evaluate additional initialization strategies mentioned in~\cref{sec:persp-alignment}.

\paragraph{Trivial:}
The trivial strategy initializes $\gravity{=}\matrx{0&1&0}$, $\dist{=}\*0$, and $\focal{=}0.7\max\left(W,H\right)$, assuming a common sensor size. 

\paragraph{Heuristic:}
A heuristic strategy was proposed in~\cite{jin2022PerspectiveFields}.
We observe that, at the principal point $\ppoint$,
${\up_{\ppoint} \propto {-}\matrx{\gravity_x&\gravity_y}}$. 
Additionally, from~\cref{eq:up}, $\gravity_z{=}\sin\latitude_{\ppoint}$. 
Therefore, 
$\gravity{=}\matrx{\alpha\up_{\ppoint}\transp&\sin\latitude_{\ppoint}}$, 
where $\alpha$ is adjusted to ensure $\norm{\gravity}{=}1$. 
Notably, the vertical field of view is linked to the difference in latitude, expressed as 
$\text{vFoV}{=}\latitude_{(\ppoint_{x},0)}-\latitude_{(\ppoint_{x},H)}$, 
which can be converted into an initial estimate for \focal.

\paragraph{Solver:}
Generalizing the heuristic strategy, we derive a simple minimal solver to infer $\gravity$ and $\focal$ from any combination of two up-vectors and one latitude value. 
We embed this solver in a RANSAC~\cite{fischler1981random} loop to maximize the confidence-weighted inlier count of up-vectors and latitudes.
This can provide a more accurate estimate.

\paragraph{Results:}
We evaluate each initialization strategy and demonstrate its impact on the final estimation when used to initialize our optimization. 
We perform the evaluation on the \ourdataset~and MegaDepth~\cite{li2018megadepth} datasets and report the median and AUC of the angular gravity and vFoV errors.
The results in \cref{tbl:initialization} show that the increased accuracy of the heuristic strategy and the solver does not translate to any improvements in the final estimate of our model. 
We hypothesize that, once trained, the observed Perspective Field is sufficiently good to ensure convergence from any initial point.

\begin{figure}[t]
    \centering

    \def\ncols{4}
    \setlength{\pwidth}{0.005\linewidth}
    \setlength{\iwidth}{\dimexpr(0.999\linewidth - \ncols\pwidth + \pwidth)/\ncols \relax}

    \includegraphics[width=0.9\iwidth]{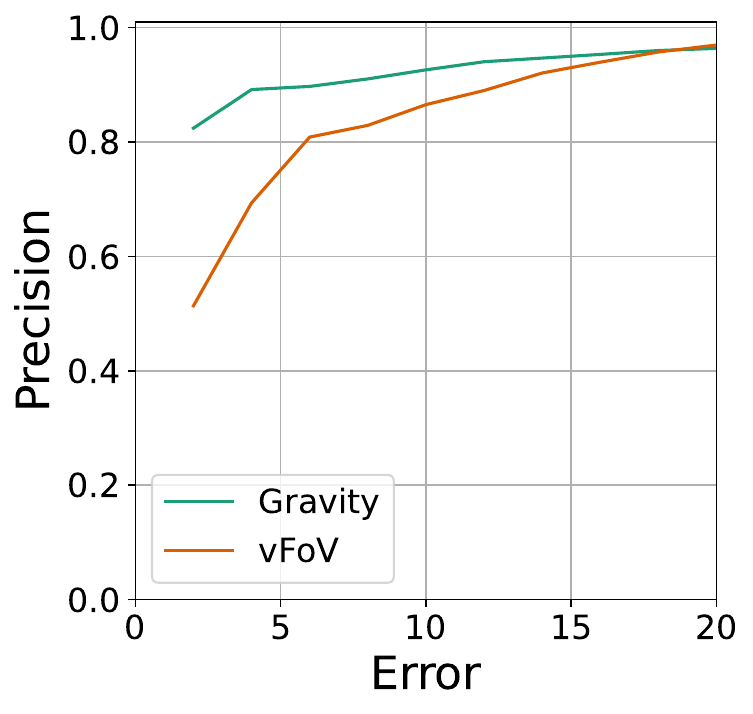}%
    \hspace{\pwidth}%
    \includegraphics[width=0.9\iwidth]{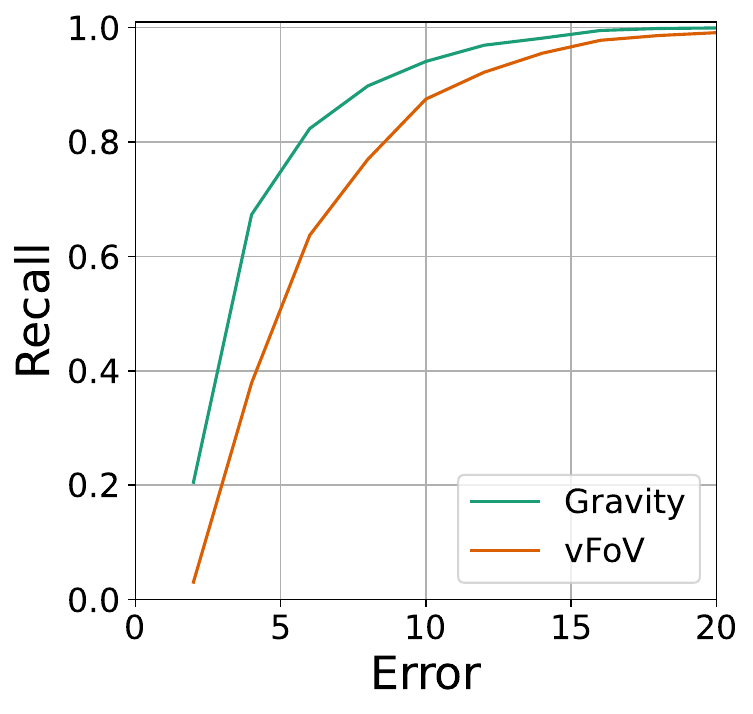}%
    \hspace{\pwidth}%
    \includegraphics[width=1.02\iwidth]{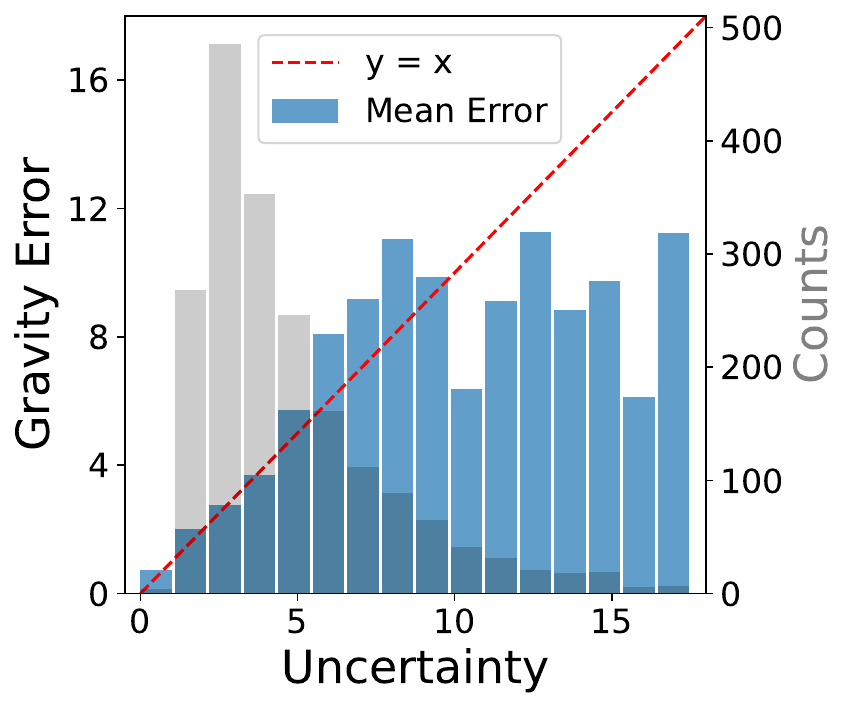}%
    \hspace{\pwidth}%
    \includegraphics[width=1.045\iwidth]{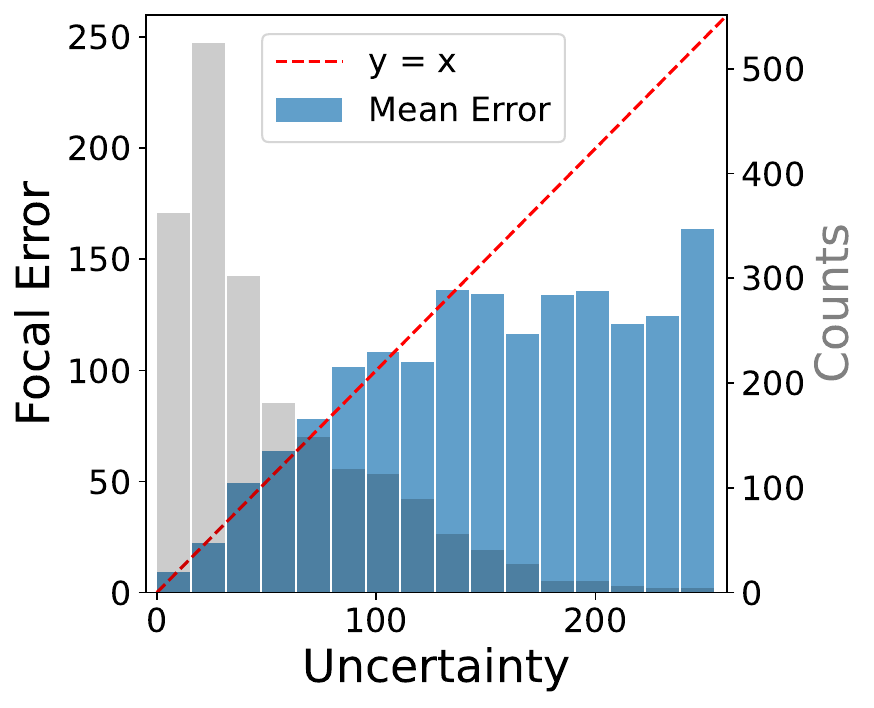}%
    
    \begin{minipage}[b]{2\iwidth}
    \centering{\footnotesize a) Precision and recall curves}
    \end{minipage}%
    \begin{minipage}[b]{2\iwidth}
    \centering{\footnotesize b) Calibration plots}
    \end{minipage}%

    \caption{\textbf{Uncertainty Evaluation.}
    a) Precision and recall curves show that the uncertainties can be used to confidently discard samples likely to be incorrect.
    b) The uncertainties follow the ideal $y{=}x$ curve and are thus fairly calibrated.
    }%
    \label{fig:uncertainty-eval}%
\end{figure}

\section{Uncertainty estimation}
\label{sec:uncertainty-eval}

We evaluate the optimization uncertainties estimated by \ours, as described in~\cref{sec:practical-benefits}, on the \ourdataset~dataset.

We show precision and recall curves for gravity and vertical Field-of-View (vFoV) in~\cref{fig:uncertainty-eval}-a).
To assess the quality of our uncertainties, we discretize the errors by increasing thresholds and determine how accurately the predicted uncertainties classify samples into their respective categories. 
This means that for each threshold, we classify samples based on whether their predicted uncertainty surpasses the threshold or not and calculate the corresponding precision and recall values.
The precision curve indicates that samples with low uncertainty generally exhibit low error rates. However, the low recall values for small errors suggest that the model is under-confident in its predictions.

We also show corresponding calibration plots in~\cref{fig:uncertainty-eval}-b).
To compute them, we bin the predictions based on the uncertainty and calculate the mean error per bin.
We observe that the calibration tightly follows the optimal calibration of $y{=}x$ for samples with low uncertainty.
For samples with high uncertainty, our model tends to be under-confident in its prediction.
The error is thus generally bounded by the predicted uncertainty.

\section{Distortion models}
\label{sec:distortion-models}
Following~\cref{eq:up}, the up-vector is collinear with the directional derivative of the projection function $\camproj(\cdot)$ along the up direction $-\gravity$, \ie $\up_\pointim \propto -\nabla\camproj(\pointw)\cdot\gravity$.
We consider a radial distortion model such that $\distfn\left(\matrx{u&v}, \dist\right) = d(u,v,\dist)\matrx{u&v}$.
Then
\begin{align}
    \nabla\camproj(\pointw)
    &\propto \frac{\partial \distfn\left(\matrx{u&v}, \dist\right)}{\partial \pointw}\\
    &\propto 
    d(u,v,\dist)
    \frac{\partial \matrx{u&v}\transp}{\partial \pointw}
    + \matrx{u\\v}
    \frac{\partial d(u,v,\dist)}{\partial \matrx{u&v}}
    \frac{\partial \matrx{u&v}\transp}{\partial \pointw}\\
    &\propto 
    \left(d(u,v,\dist) + \matrx{u\\v}\frac{\partial d(u,v,\dist)}{\partial(u,v)} \right)
    \frac{\partial \matrx{u&v}\transp}{\partial \pointw}\\
    &\propto 
    \left(1 + \frac{1}{d(u,v,\dist)}\matrx{u\\v}\frac{\partial d(u,v,\dist)}{\partial(u,v)} \right)
    \matrx{1&0&-u\\0&1&-v}
\enspace.
\end{align}
The up-vector is thus such that
\begin{equation}
 \up_\pointim\left(\focal,\gravity,\dist\right)   
 \propto \left(1 + \frac{1}{d(u,v,\dist)}\matrx{u\\v}\frac{\partial d(u,v,\dist)}{\partial(u,v)} \right)
\matrx{u\gravity_z-\gravity_x\\v\gravity_z-\gravity_y}
\enspace.
\end{equation}
For the pinhole camera model, this simplifies to 
\begin{equation}
 \up_\pointim\left(\focal,\gravity,\dist\right)   
 \propto 
\matrx{u\gravity_z-\gravity_x\\v\gravity_z-\gravity_y}
\enspace.
\end{equation}
For a polynomial radial distortion with $d(u,v,\dist)= 1 + \dist_1\radius^2 + \dist_2\radius^4$,
where $\dist = (\dist_1, \dist_2)$ are the distortion parameters and $\radius^2 = u^2+v^2$, we have
\begin{equation}
    \frac{\partial d(u,v,\dist)}{\partial\matrx{u&v}}
    = 2(\dist_1+2\dist_2r^2)\matrx{u&v}
\enspace.
\end{equation}

\section{Analytical Jacobians}
\label{sec:jacobians}
We now report the analytical expression of the jacobians used in the optimization for the pinhole camera model. 
For each pixel, we write the up-vector and latitude
\begin{equation}
\up 
= \operatorname{norm} \left( \matrx{u\gravity_z-\gravity_x\\v\gravity_z-\gravity_y} \right)
=\operatorname{norm} \left( \bar{\up} \right)
\quad
\quad
\sin \left(\latitude\right) = \operatorname{norm}\left(\ray\right)\transp \gravity
\enspace
\end{equation}
where $\ray=\matrx{u&v&1}\transp$ is the ray
and $\operatorname{norm}\left(\*x\right)=\nicefrac{\*x}{\norm{\*x}_2}$ normalizes the input vector to have unit length.
We write $\jacobian = \matrx{\jacobian_{\up} & \jacobian_{\latitude}}$ and use the general chain rule:
\begin{align}
    \jacobian_{\up} &= \frac{\partial \residual_{\up}}{\partial \lmstep} 
    = \frac{\partial \residual_{\up}}{\partial \up\left(\camparams\right)}
    \frac{\partial \up\left(\camparams\right)}{\partial \camparams}
    \frac{\partial \camparams}{\partial \lmstep}, \\
    \jacobian_{\latitude} &= \frac{\partial \residual_{\latitude}}{\partial \lmstep} 
    = \frac{\partial \residual_{\latitude}}{\partial \sin{\left(\latitude\left(\camparams\right)\right)}}
    \frac{\partial \sin{\left(\latitude\left(\camparams\right)\right)}}{\partial \camparams}
    \frac{\partial \camparams}{\partial \lmstep}.
\end{align}%
where we calculate the jacobians at $\lmstep{=}0$.
By simple derivation we can calculate the partial derivate for $\up$ as
\begin{align}
    \frac{\partial \up\left(\focal,\gravity,\dist\right)}{\partial \gravity}
    &= \left( 
        \frac{1}{\norm{\bar{\up}}}_2\mathbb{I} - \frac{\bar{\up} \bar{\up}\transp}{\norm{\bar{\up}}_2^3}
    \right)
    \matrx{-1&0&u\\0&-1&v}, \\
    \frac{\partial \up\left(\focal,\gravity,\dist\right)}{\partial \focal}
    &= \left( 
        \frac{1}{\norm{\bar{\up}}}_2\mathbb{I} - \frac{\bar{\up} \bar{\up}\transp}{\norm{\bar{\up}}_2^3}
    \right)
    \matrx{\gravity_z \\ \gravity_z} \matrx{\nicefrac{-(x-\ppoint_x)}{\focal^2} \\ \nicefrac{-(y-\ppoint_y)}{\focal^2}}.
\end{align}
Similarly, we can calculate the partial derivatives for $\sin\left(\latitude\right)$ as
\begin{align}
    \frac{\partial \sin\left(\latitude_ \pointim \left(\focal,\gravity,\dist\right)\right)}{\partial  \gravity} 
    &= \operatorname{norm}\left(\ray\right)\transp \mathbb{I}, \\
    \frac{\partial \sin\left(\latitude_ \pointim \left(\focal,\gravity,\dist\right)\right)}{\partial \focal}
    &= \gravity\transp \left( 
        \frac{1}{\norm{\ray}}_2\mathbb{I} - \frac{\ray \ray\transp}{\norm{\ray}_2^3}
    \right)
    \matrx{\nicefrac{-(x-\ppoint_x)}{\focal^2} \\ \nicefrac{-(y-\ppoint_y)}{\focal^2} \\ 0}.
\end{align}
The Jacobians for radially distorted cameras can be derived in a similar way.

\section{Implementation details}
\label{sec:impl-details}

In this section we provide more information and details on our dataset, how we train our models and evaluate our baselines.

\subsection{Dataset}
\label{sec:dataset-details}
\begin{figure}[t]
    \centering

    \def\ncols{6}
    \setlength{\pwidth}{0.005\linewidth}
    \setlength{\iwidth}{\dimexpr(0.999\linewidth - \ncols\pwidth + \pwidth)/\ncols \relax}

    \includegraphics[width=\iwidth]{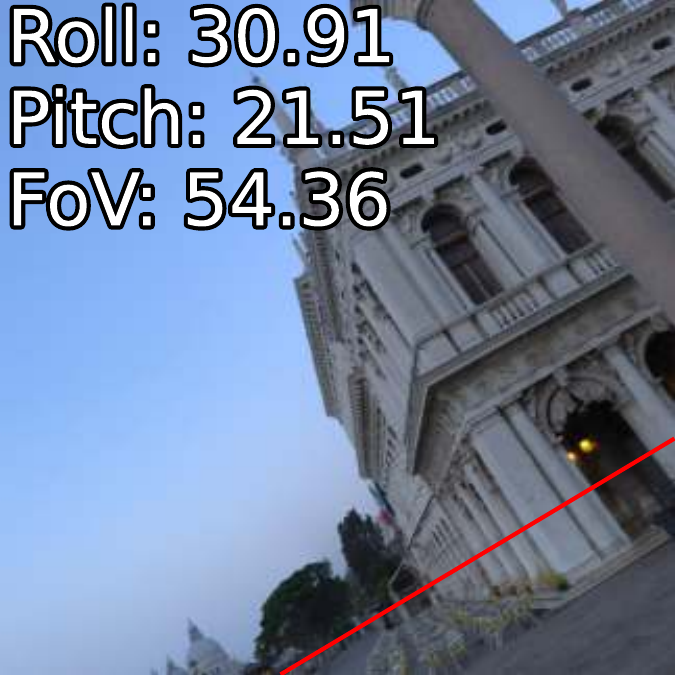}%
    \hspace{\pwidth}%
    \includegraphics[width=\iwidth]{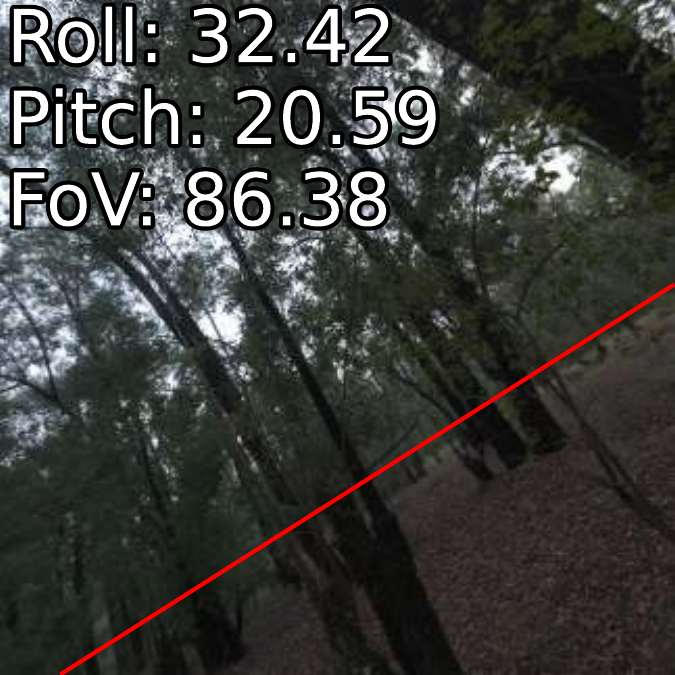}%
    \hspace{\pwidth}%
    \includegraphics[width=\iwidth]{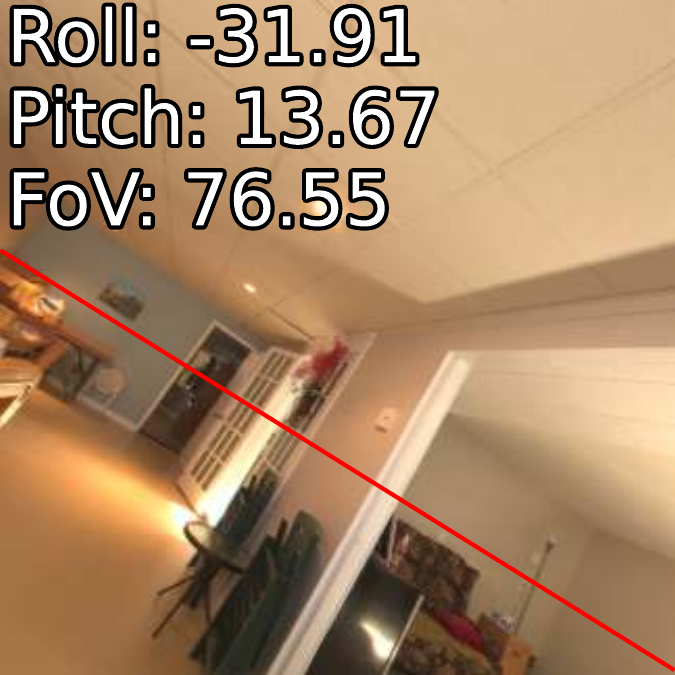}%
    \hspace{\pwidth}%
    \includegraphics[width=\iwidth]{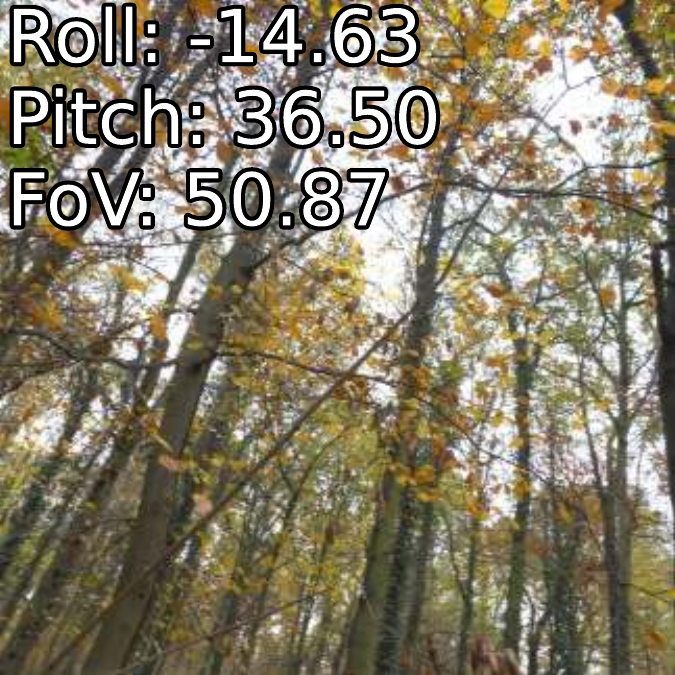}%
    \hspace{\pwidth}%
    \includegraphics[width=\iwidth]{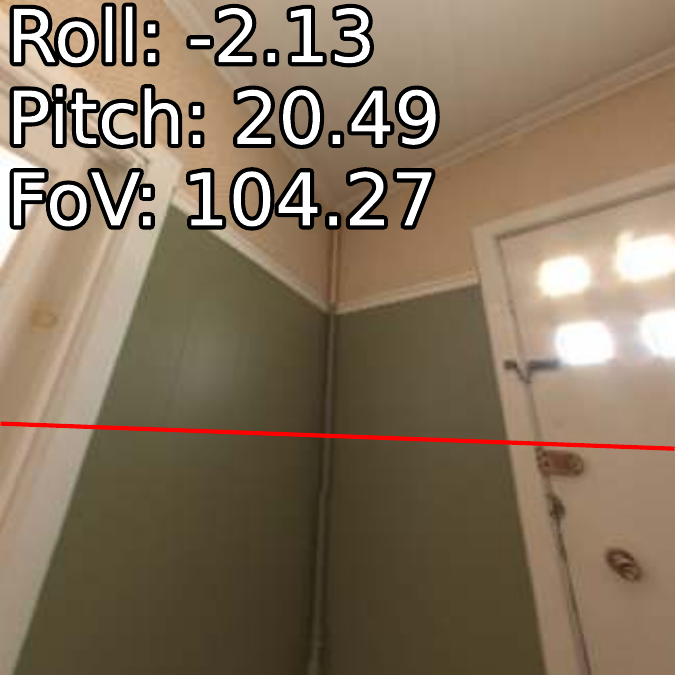}%
    \hspace{\pwidth}%
    \includegraphics[width=\iwidth]{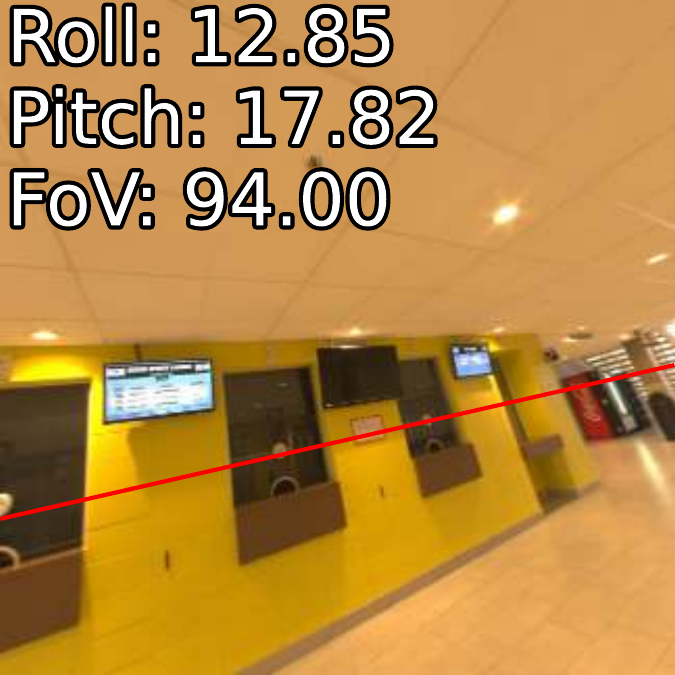}%
    
    \includegraphics[width=\iwidth]{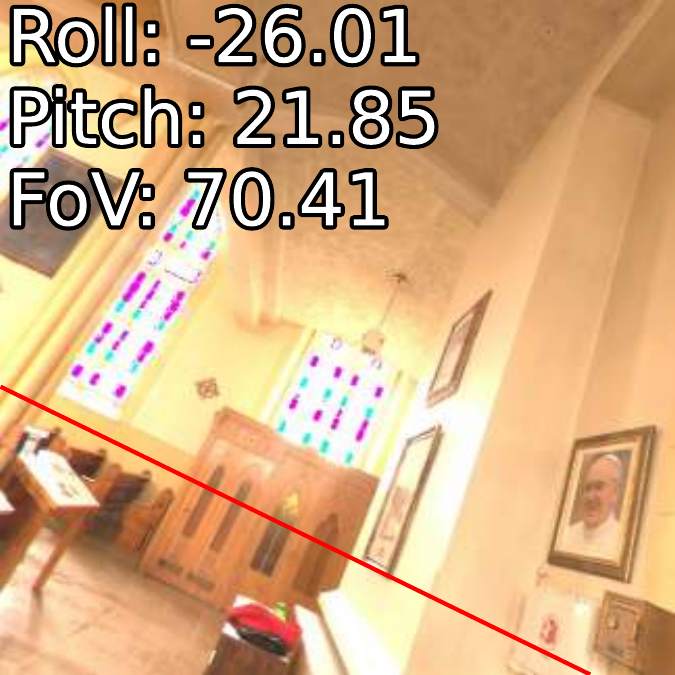}%
    \hspace{\pwidth}%
    \includegraphics[width=\iwidth]{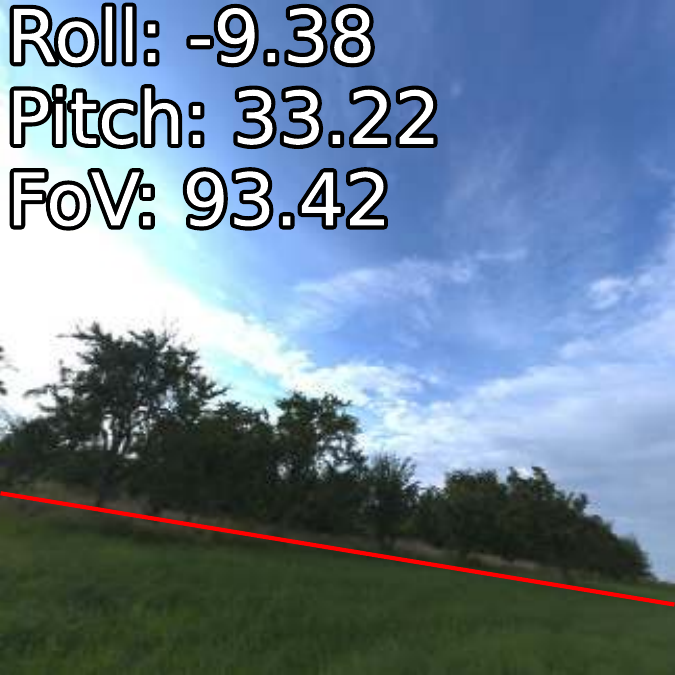}%
    \hspace{\pwidth}%
    \includegraphics[width=\iwidth]{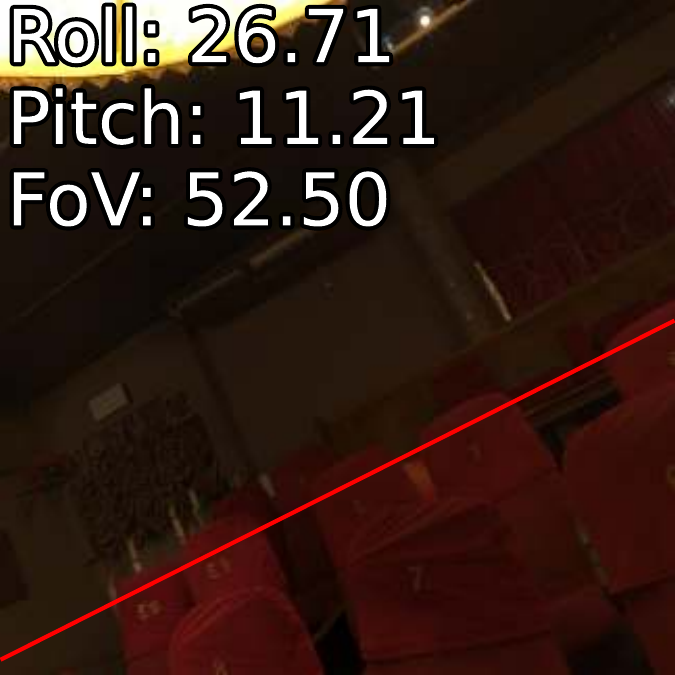}%
    \hspace{\pwidth}%
    \includegraphics[width=\iwidth]{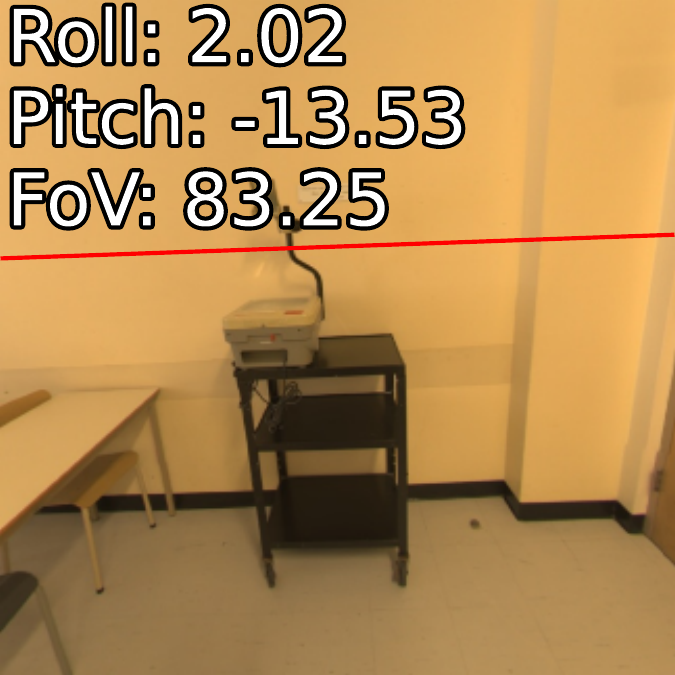}%
    \hspace{\pwidth}%
    \includegraphics[width=\iwidth]{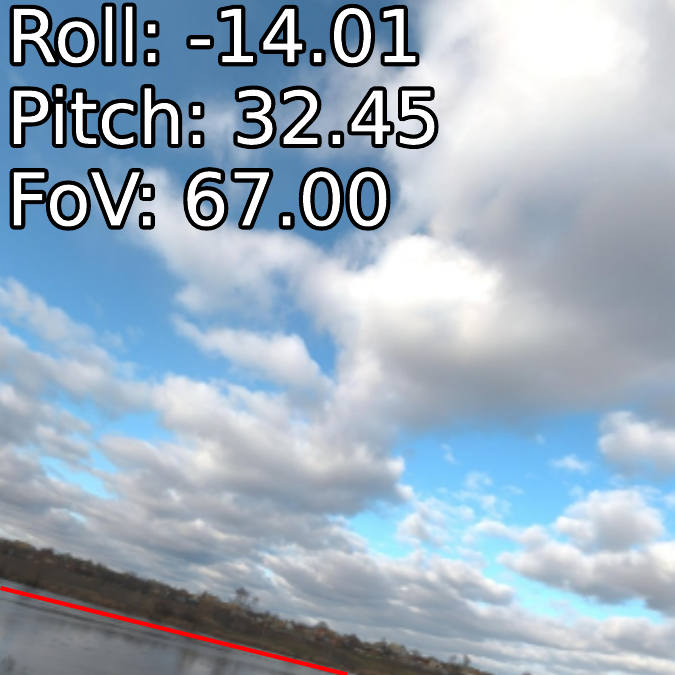}%
    \hspace{\pwidth}%
    \includegraphics[width=\iwidth]{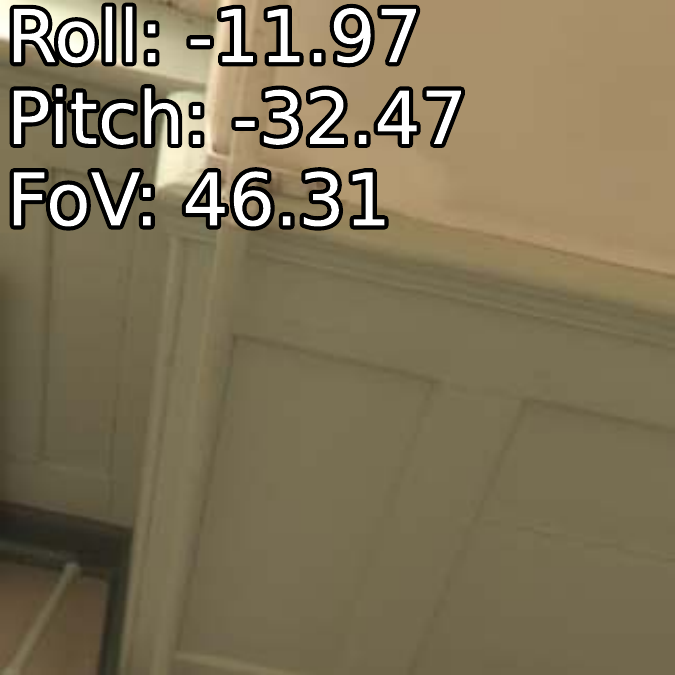}%
    
    \includegraphics[width=\iwidth]{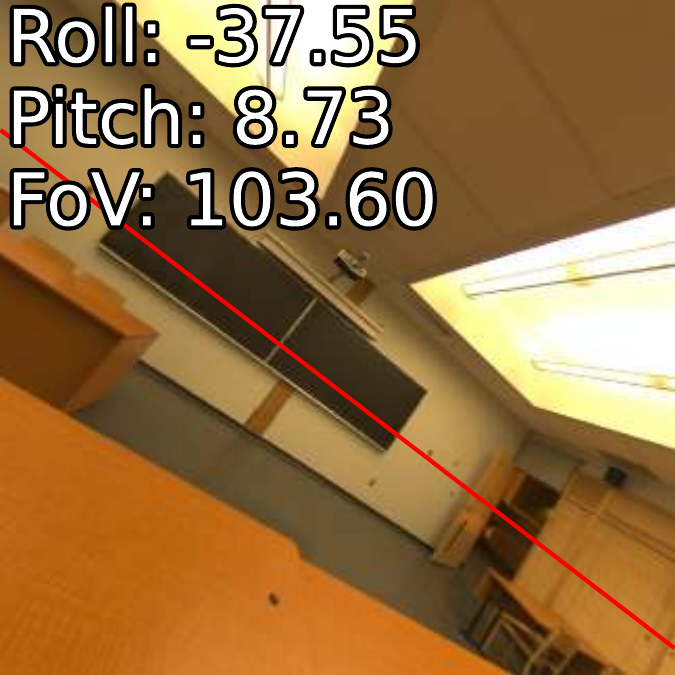}%
    \hspace{\pwidth}%
    \includegraphics[width=\iwidth]{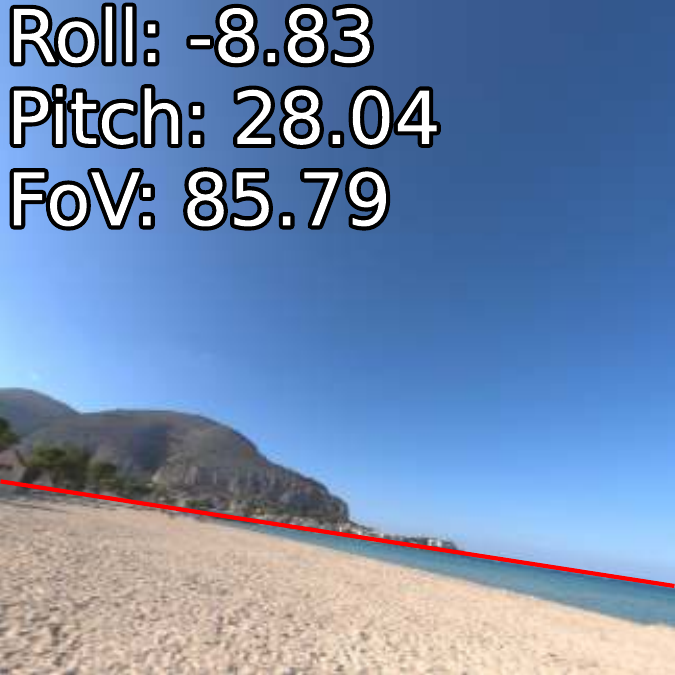}%
    \hspace{\pwidth}%
    \includegraphics[width=\iwidth]{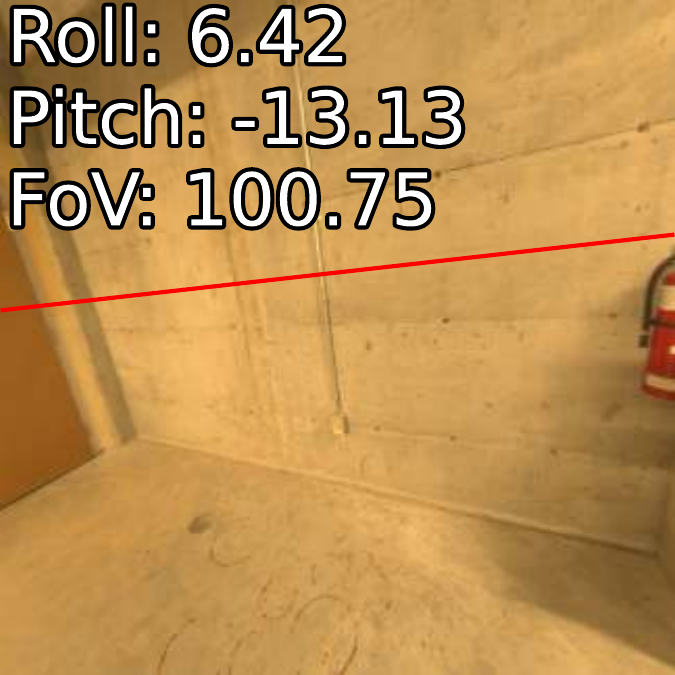}%
    \hspace{\pwidth}%
    \includegraphics[width=\iwidth]{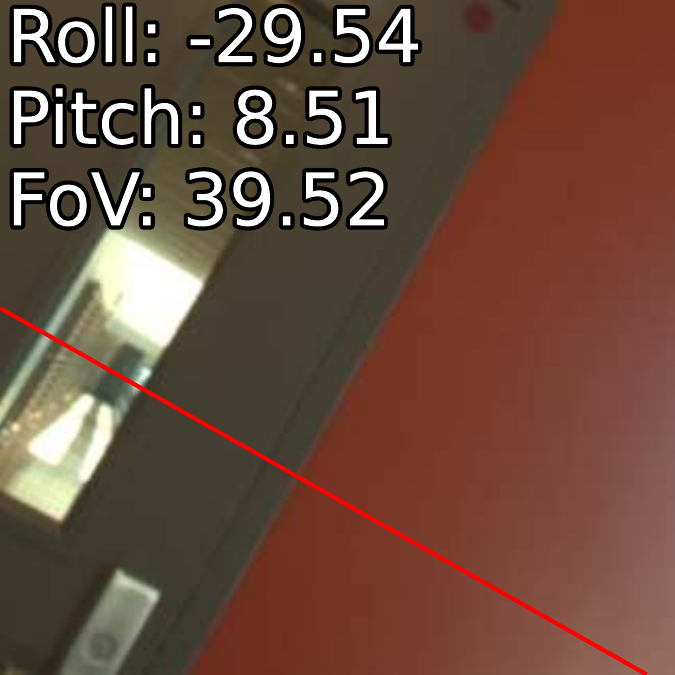}%
    \hspace{\pwidth}%
    \includegraphics[width=\iwidth]{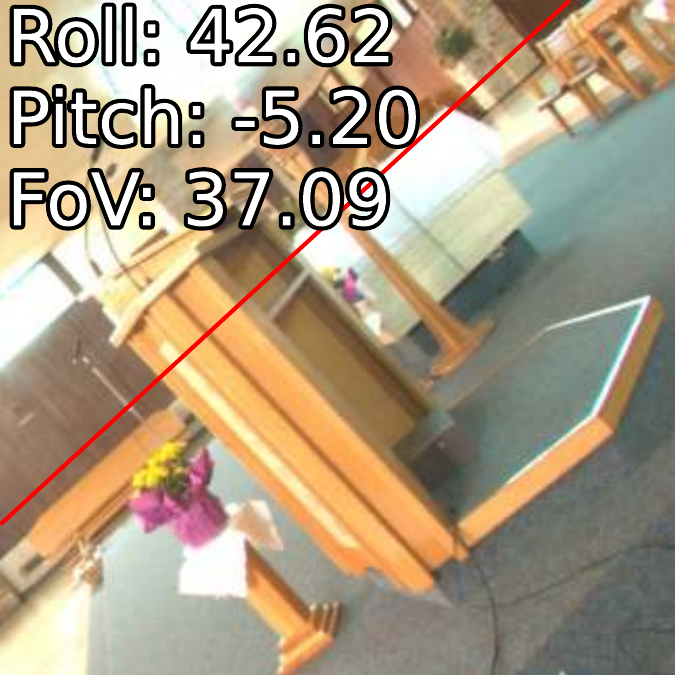}%
    \hspace{\pwidth}%
    \includegraphics[width=\iwidth]{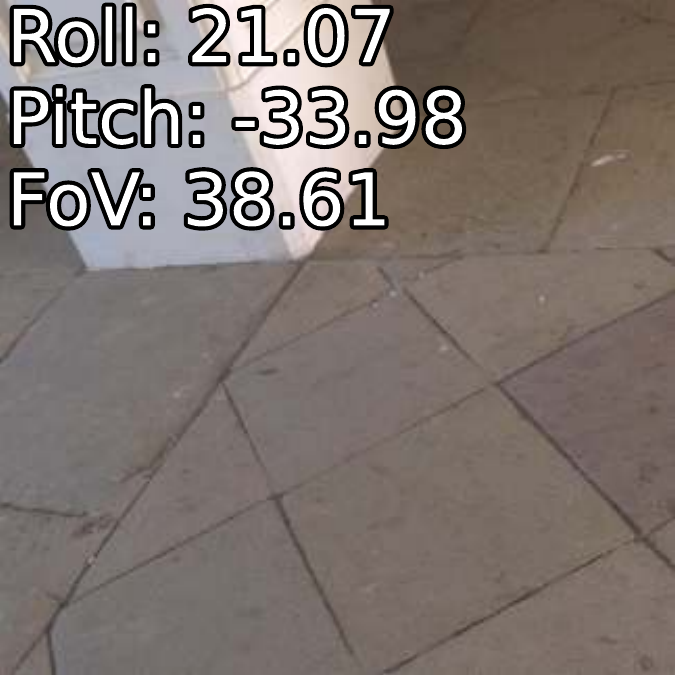}%
    
    \includegraphics[width=\iwidth]{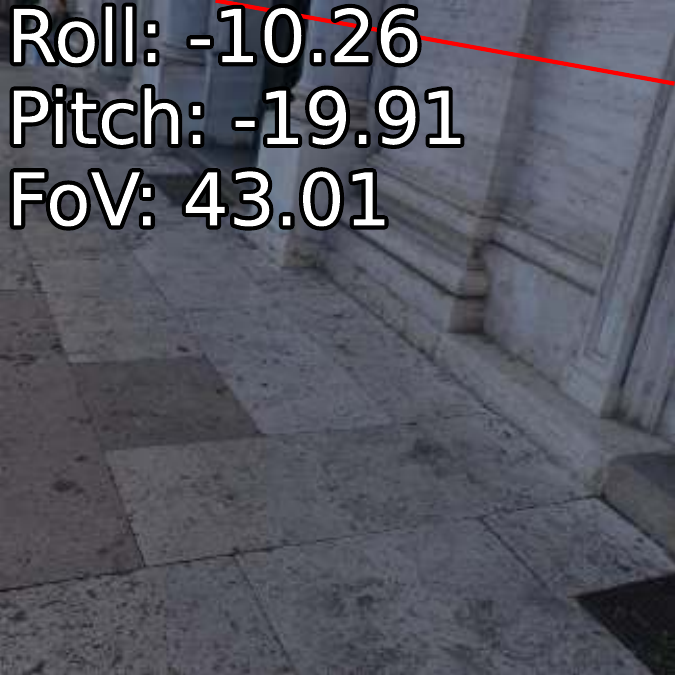}%
    \hspace{\pwidth}%
    \includegraphics[width=\iwidth]{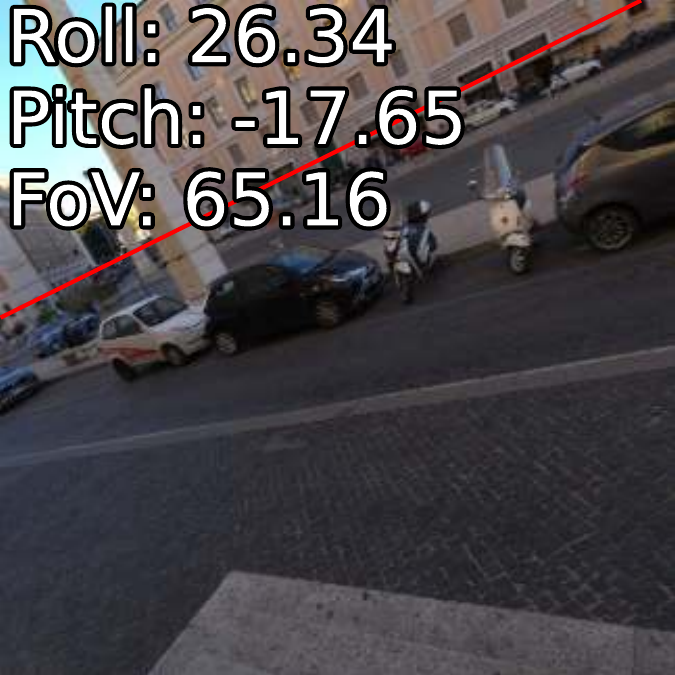}%
    \hspace{\pwidth}%
    \includegraphics[width=\iwidth]{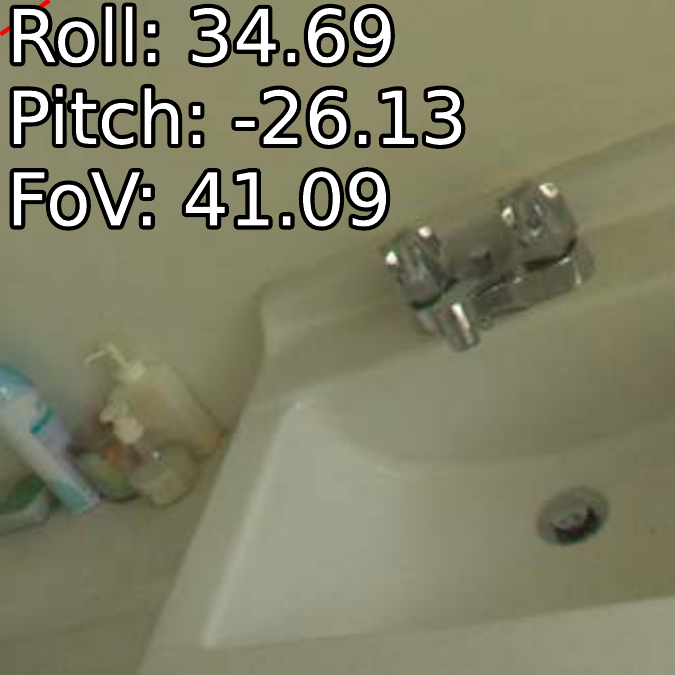}%
    \hspace{\pwidth}%
    \includegraphics[width=\iwidth]{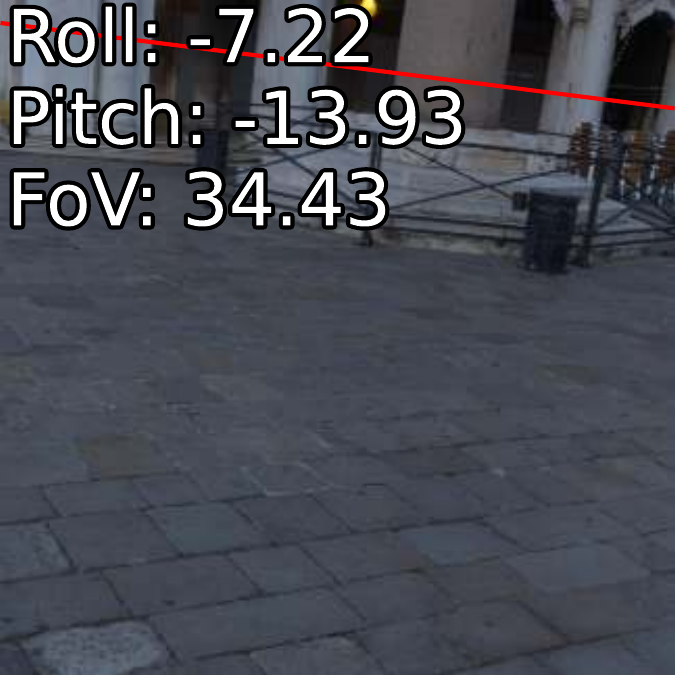}%
    \hspace{\pwidth}%
    \includegraphics[width=\iwidth]{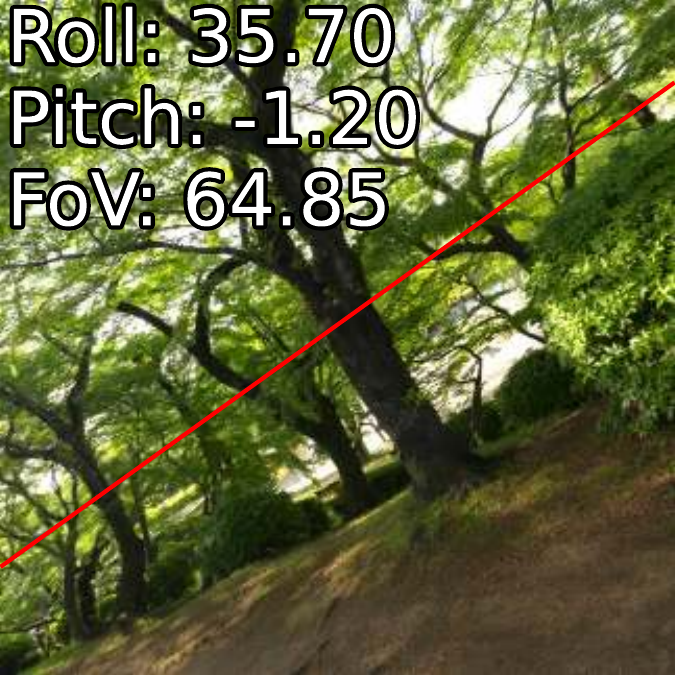}%
    \hspace{\pwidth}%
    \includegraphics[width=\iwidth]{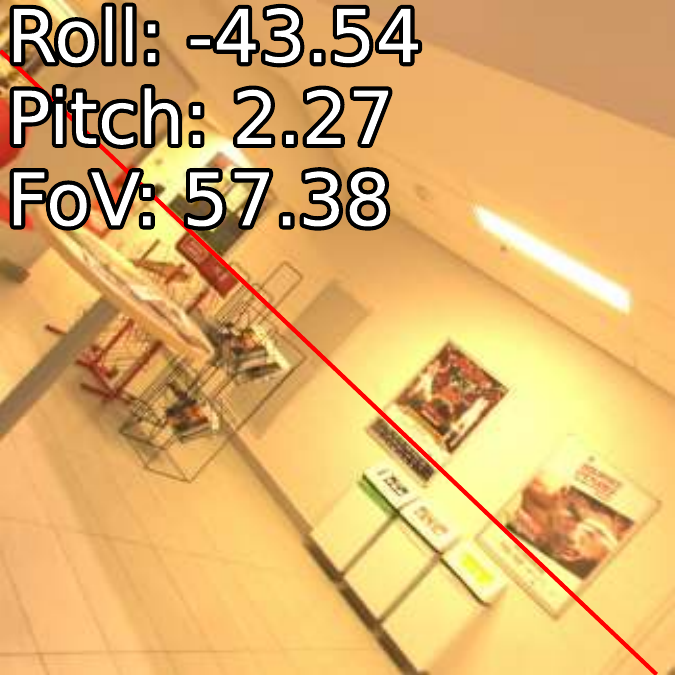}%
    
    \caption{\textbf{Sample from \ourdataset-pinhole.}
    The dataset includes images sampled from panoramas captured in diverse outdoor and indoor scenes.
    We show the ground truth horizon as a red line.
    \label{fig:dataset-samples}%
    }%
\end{figure}
For our dataset we collect 360\degree~panoramas in equirectangular format which covers 180\degree~vertically and 360\degree~horizontally. 
The dataset comprises 2912 panoramas sourced from hdrmaps~\cite{hdrmaps}, polyhaven~\cite{polyhaven}, and parts of the Laval Photometric Indoor HDR dataset~\cite{bolduc2023beyond}, where we manually select panoramas that are aligned with the gravity.
The resulting dataset contains a good balance of about 800 outdoor and 2.1k indoor panoramas which are split into 2616 training, 147 validation and 148 testing panoramas.
We create two iterations of the dataset: one featuring a simple pinhole camera and another assuming a radially distorted camera model.
For both datasets we sample 16 crops per panorama with roll, pitch and vFoV sampled uniformly within $[\pm45\degree]$, $[\pm45\degree]$, and $[20, 105]$ respectively.
To capture the relationship between the distortion parameter $\dist$~and the vFoV, we sample $\hat{\dist}=\nicefrac{\dist}{\mathrm{vFoV}}$ from a truncated normal with $\mu{=}0$ and $\sigma{=}0.07$ bounded to [-0.3, 0.3] for the distorted dataset.
Since some panoramas are padded with black pixels to be equirectangular, we remove images with $\geq 1\%$ of black pixels.
After this, we are left with 37015 images for training, 2100 for validation and 2128 for testing.
\Cref{fig:dataset-samples} shows the diversity of our dataset with difficult and easy samples, indoor and outdoor images as well as variation in gravity and camera intrinsics.

\subsection{Training}

\ours~is trained for 75k steps with a batch size of 24 and a base learning rate of $10^{-4}$ using the AdamW~\cite{loshchilov2017decoupled} optimizer.
We make use of a linear warm-up schedule in the first 1k steps and drop the learning rate after 40k and 65k steps by a factor of 10.
The input resolution is $320\times 320$ and we apply extensive color, blur and resize augmentations during training.
During inference, non-square images are first resized to $320$ along the shorter edge and afterwards cropped to fit the required multiple of 32 by our model. The resulting camera is re-scaled to fit the original image dimensions.

\subsection{Indoor Visual Localization}

\paragraph{RANSAC:} We use the P3P RANSAC solver from PoseLib~\cite{PoseLib} as a baseline, and add the gravity constraint in their framework. We add a constant, multiplicative reward to the MSAC scoring if the angle between the estimated gravity from our network and the gravity obtained from the samples' pose estimate (the second column of the rotation matrix) is smaller than twice the uncertainty predicted by our network (in degrees).

In contrast to upright solvers~\cite{kukelova2010vertical}, this weak constraint is less susceptible to noise and can directly utilize the uncertainty estimates of our model.

\paragraph{Absolute Pose Refinement:} In the non-linear absolute pose refinement, we add an L2-regularization term to the optimization, which penalizes the deviation of the rotation matrix from our estimated gravity. We use a simple L2-loss and weigh the regularization by the number of inlier correspondences. We use the Ceres-solver~\cite{ceres-solver} to build the non-linear pose refinement, similar to COLMAP~\cite{schoenberger2016sfm}.

\subsection{Evaluation}
We provide further details on our evaluation setup.

\paragraph{Metrics:}
To account for inaccuracies in the ground truth, especially for datasets derived from structure-from-motion like MegaDepth and LaMAR, we compute the AUC for curves clipped to a minimum of 1$\degree$.
When an approach fails to return a valid result~\cite{sva,Pautrat_2023_UncalibratedVP}, we set its error to $+\infty$.

\paragraph{DeepCalib~\cite{lopez2019deepcalib}:} We re-implement the model and its training following the paper.
We train the model for 20k steps using the Adam optimizer~\cite{kingma2014adam} with a learning rate of $10^{-4}$ and a batch size of 32 on our distorted dataset. The input images are resized to $320\times 320$ and we use the same augmentations as for \ours. We follow the parameterization proposed in~\cite{lopez2019deepcalib} and train the model using 256 bins per parameter. The model is trained to minimize the negative log likelihood, and we apply early stopping on the validation set.

\paragraph{Perceptual~\cite{perceptual}:}
We run inference using the online demo provided by the authors, which is based on a ConvNeXt CNN backbone~\cite{liu2022convnet}.

\paragraph{CTRL-C~\cite{ctrlc}:} We make use of the official code released by the authors to run inference on our benchmarks using the model weights trained on the SUN360~\cite{xiao2012recognizing} dataset. The input resolution is $512\times 512$.

\paragraph{MSCC~\cite{Song2024MSCC}:} We ask the authors to run inference on our benchmarks without sharing the ground-truth parameters.

\paragraph{ParamNet~\cite{jin2022PerspectiveFields}:} We re-implement ParamNet and validate our implementation by comparing its inference to the original code-base on various benchmarks. 
The model is re-trained on our pinhole dataset following the schedule proposed in the official repo: we first pre-train PerspectiveNet, then add ParamNet to train the full model. 
Each is trained for 45k steps with a base learning rate of $0.01$ using the SGD optimizer and a batch size of 64. 
After a 1k step warm-up, the learning rate drops by a factor of 10 after 30k and 40k steps. 
The input resolution is $320\times 320$, and we use the same augmentation and inference resizing patterns as in \ours.
To support radially distorted camera models (\cref{tbl:distortion}), we add another prediction head to ParamNet, using the parametrization of~\cite{lopez2019deepcalib} for the distortion value \dist. We then re-train PerspectiveNet and ParamNet on our distorted dataset.

\paragraph{SVA~\cite{sva}:} We use the official implementation released by the authors to run inference. The method fails when too few lines or coplanar repeats are visible in the image, which happens for about 58\% of the samples in MegaDepth~\cite{li2018megadepth} and about 89\% of images on Lamar~\cite{sarlin2022lamar}. Images are resized to a maximum of $3k\times 3k$.

\paragraph{UVP~\cite{Pautrat_2023_UncalibratedVP}:} To run inference using UVP, we make the upright assumption and follow the suggested configuration using DeepLSD~\cite{pautrat2023deeplsd} to extract lines.
We resize the short side of the image to $320$ as recommended by the authors, which we also found to work best.
The approach generally does not return any value when too few lines are observable in the image.

\section{Qualitative Results}
We show qualitative results in \cref{fig:stanford2d3d,fig:tartanair,fig:lamar2k,fig:megadepth2k,fig:failure}.

\begin{figure}[h!]
    \vspace{-5mm}
    \centering
    \def\ncols{4}
    \setlength{\pwidth}{0.005\linewidth}
    \setlength{\iwidth}{\dimexpr(0.999\linewidth - \ncols\pwidth + \pwidth)/\ncols \relax}
    \setlength{\lamarwidth}{\dimexpr(0.999\linewidth - 8\pwidth + \pwidth)/8 \relax}
    
    \begin{minipage}[b]{\iwidth}
    \centering{\footnotesize a) ground-truth}
    \end{minipage}%
    \hspace{\pwidth}%
    \begin{minipage}[b]{\iwidth}
    \centering{\footnotesize b) final prediction}
    \end{minipage}%
    \hspace{\pwidth}%
    \begin{minipage}[b]{\iwidth}
    \centering{\footnotesize c) observed latitude}
    \end{minipage}%
    \hspace{\pwidth}%
    \begin{minipage}[b]{\iwidth}
    \centering{\footnotesize d) observed up-vect.}
    \end{minipage}%
    
    \includegraphics[width=\iwidth]{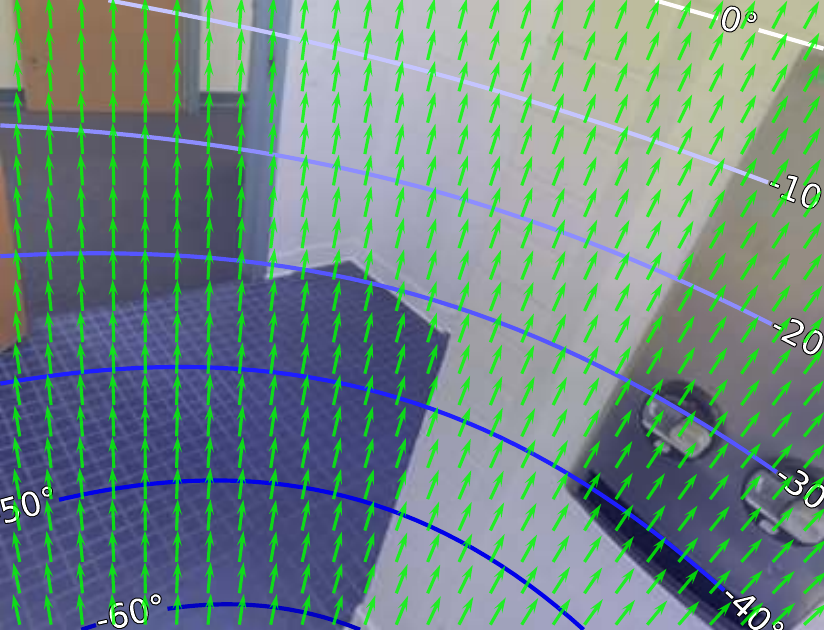}%
    \hspace{\pwidth}%
    \includegraphics[width=\iwidth]{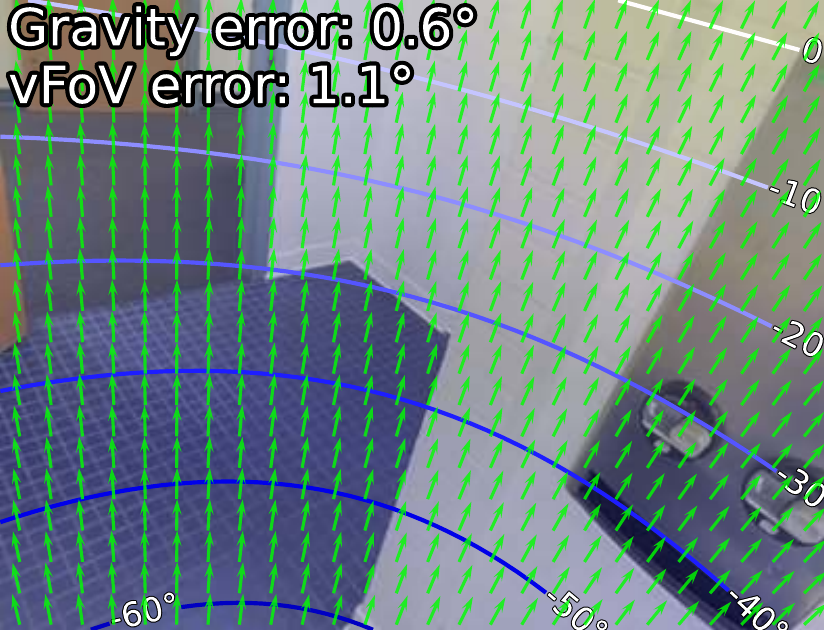}%
    \hspace{\pwidth}%
    \includegraphics[width=\iwidth]{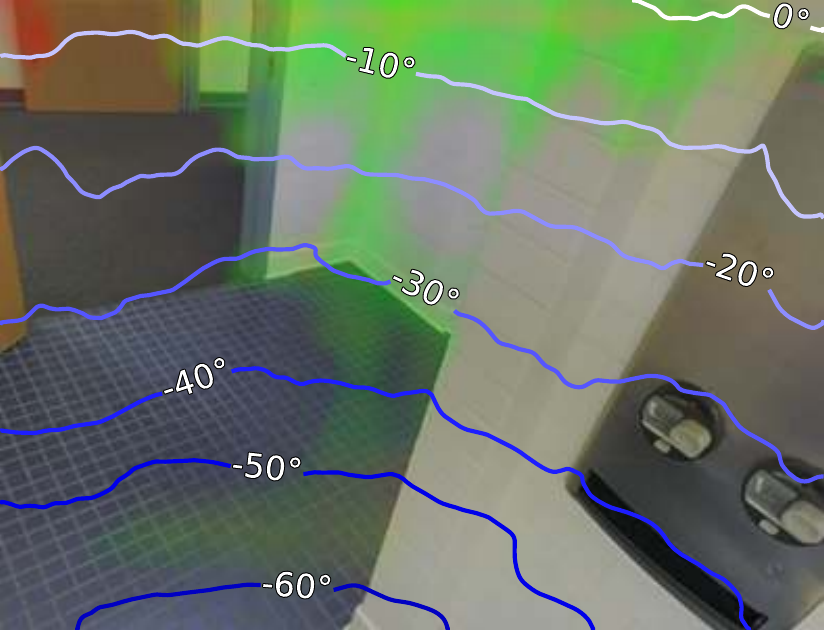}%
    \hspace{\pwidth}%
    \includegraphics[width=\iwidth]{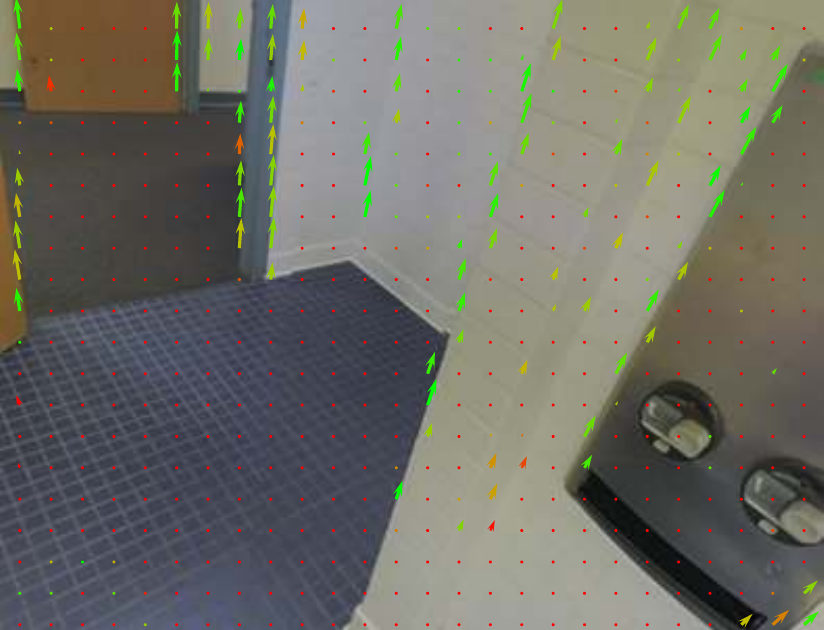}%
    
    \includegraphics[width=\iwidth]{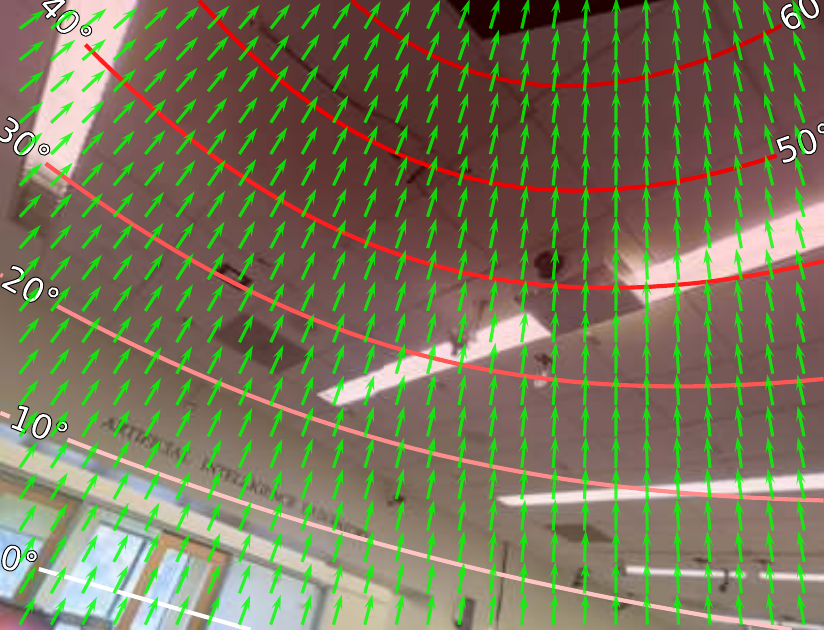}%
    \hspace{\pwidth}%
    \includegraphics[width=\iwidth]{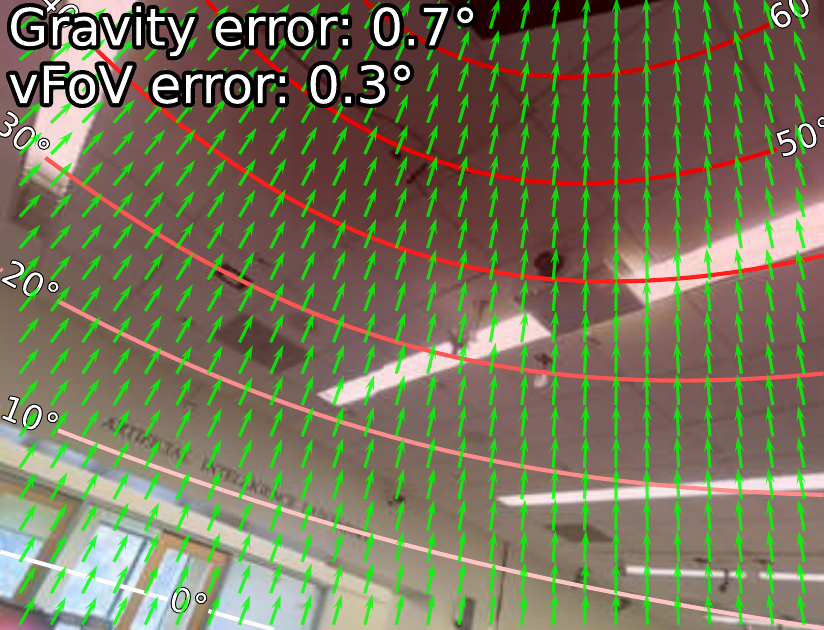}%
    \hspace{\pwidth}%
    \includegraphics[width=\iwidth]{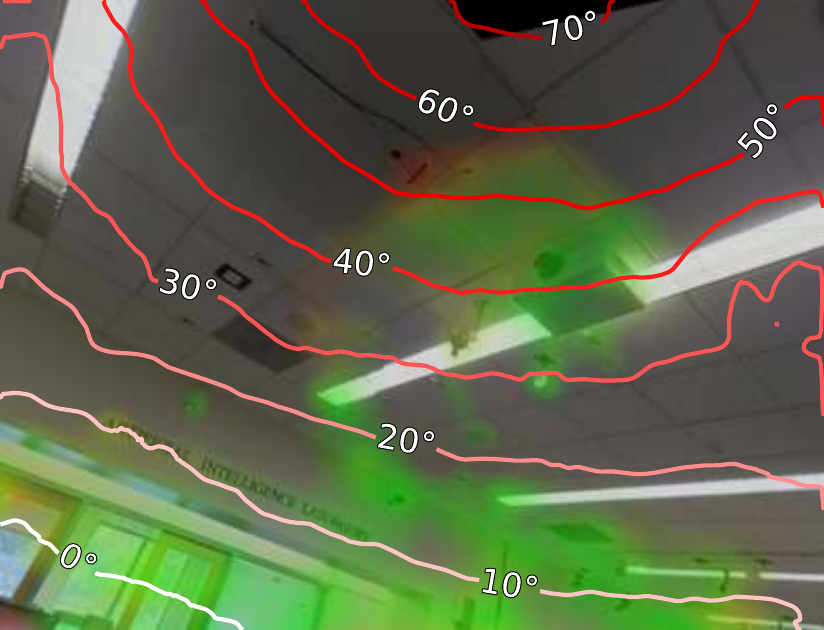}%
    \hspace{\pwidth}%
    \includegraphics[width=\iwidth]{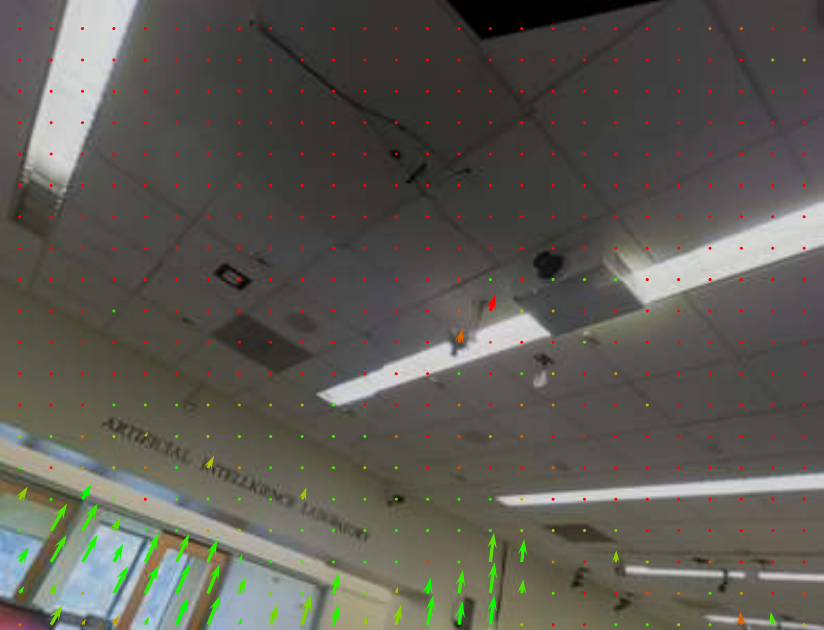}%
    
    \includegraphics[width=\iwidth]{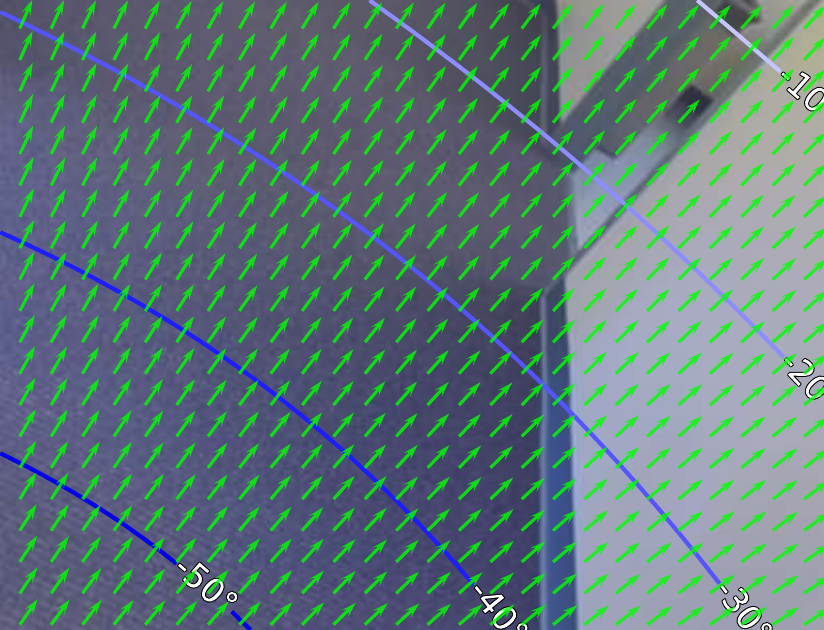}%
    \hspace{\pwidth}%
    \includegraphics[width=\iwidth]{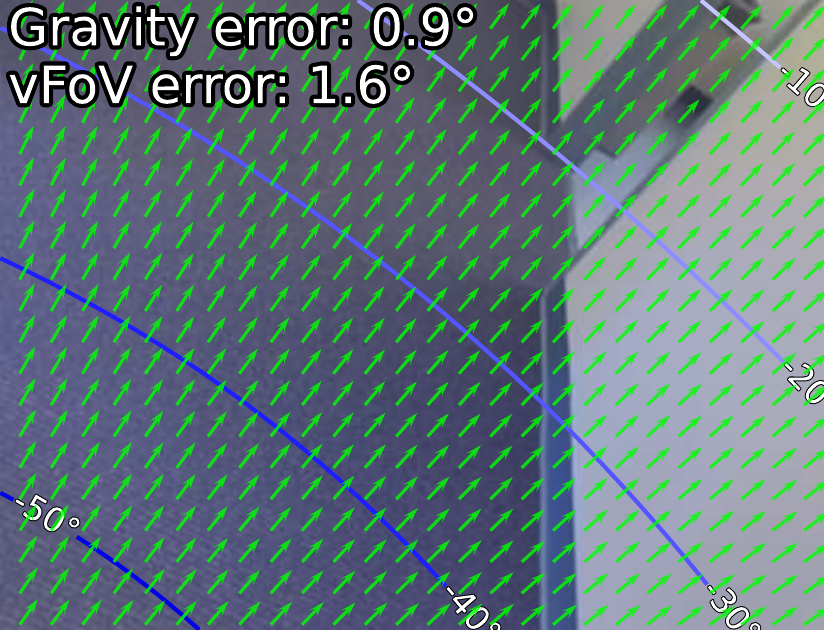}%
    \hspace{\pwidth}%
    \includegraphics[width=\iwidth]{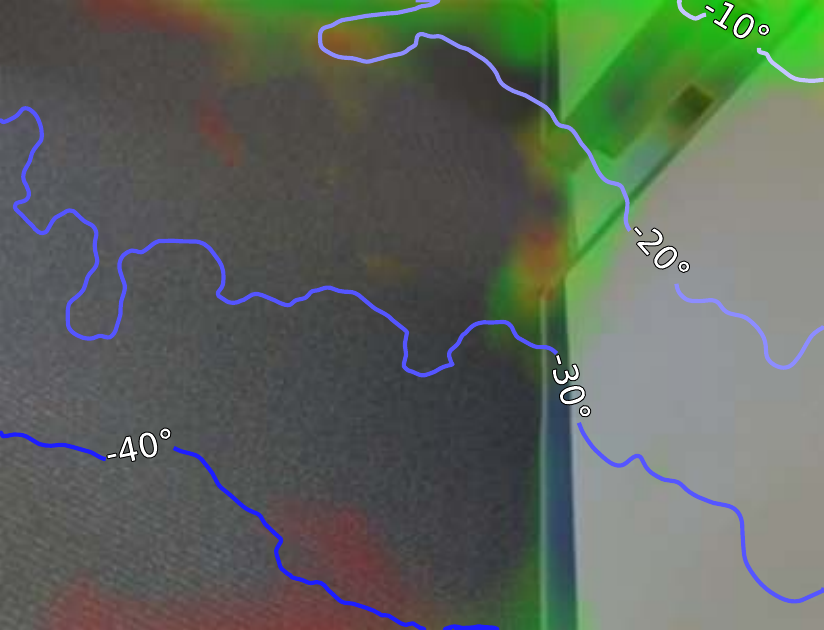}%
    \hspace{\pwidth}%
    \includegraphics[width=\iwidth]{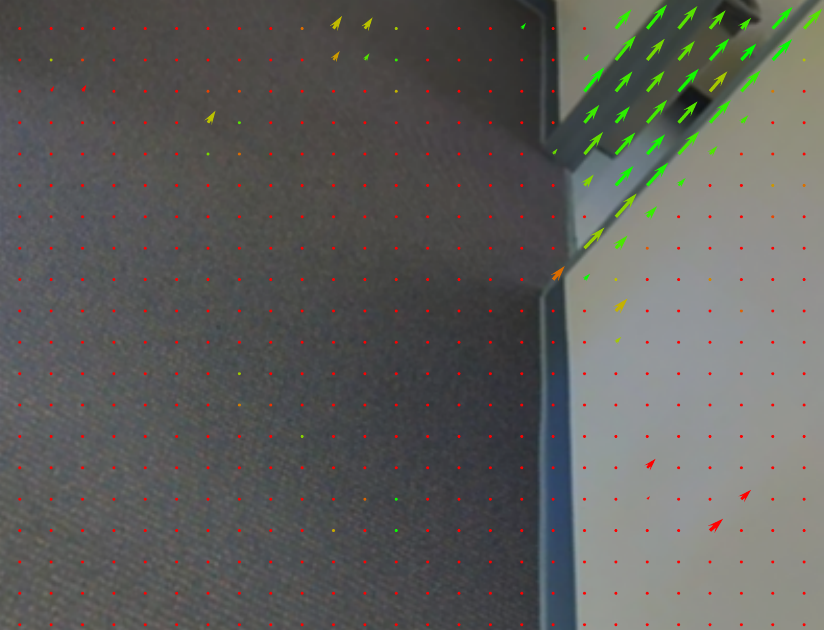}%
    
    \includegraphics[width=\iwidth]{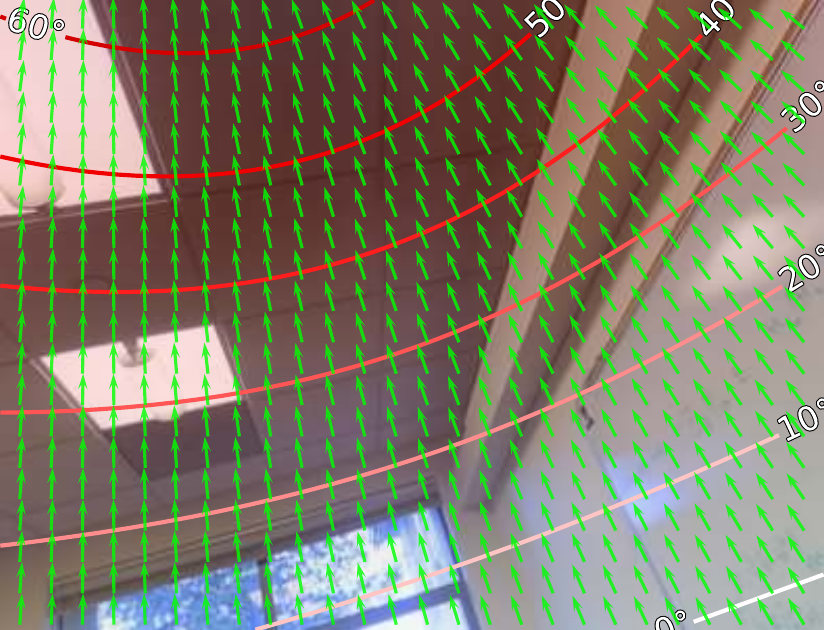}%
    \hspace{\pwidth}%
    \includegraphics[width=\iwidth]{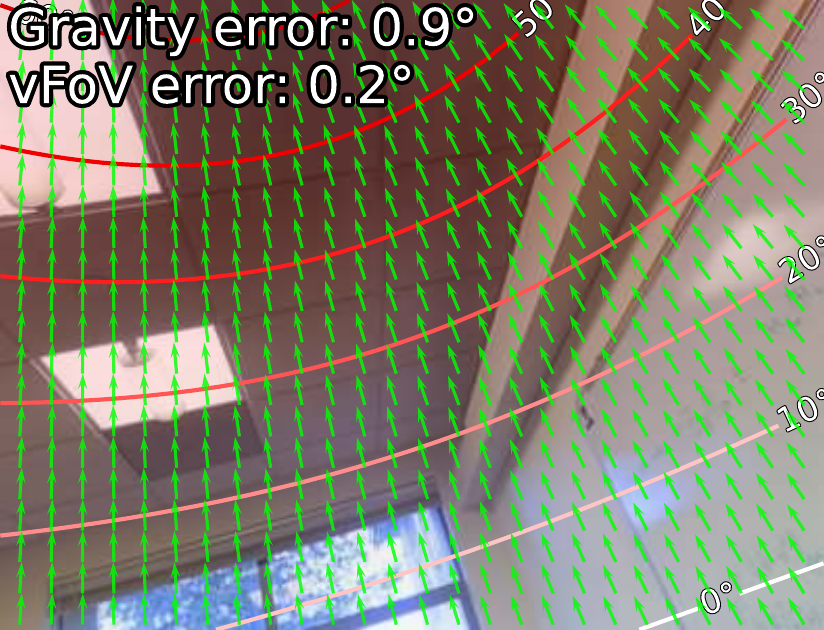}%
    \hspace{\pwidth}%
    \includegraphics[width=\iwidth]{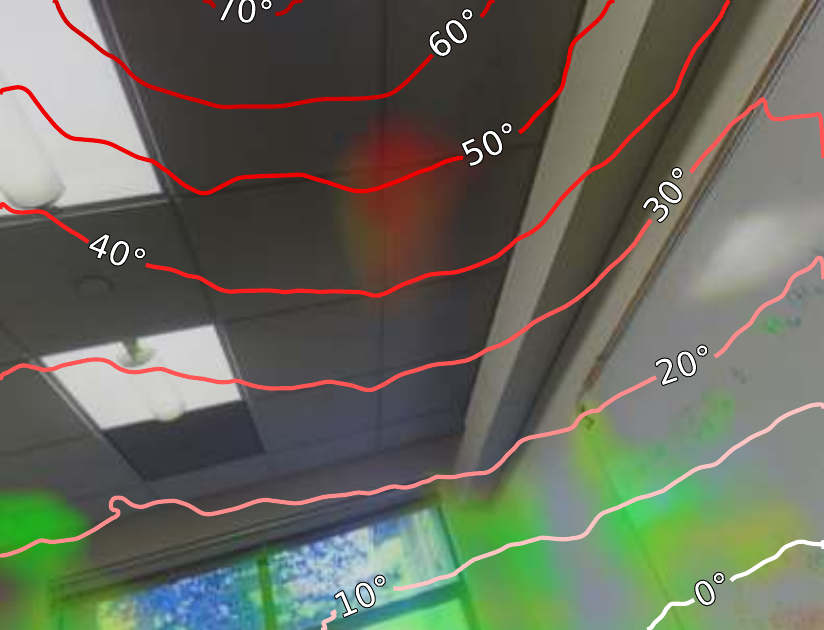}%
    \hspace{\pwidth}%
    \includegraphics[width=\iwidth]{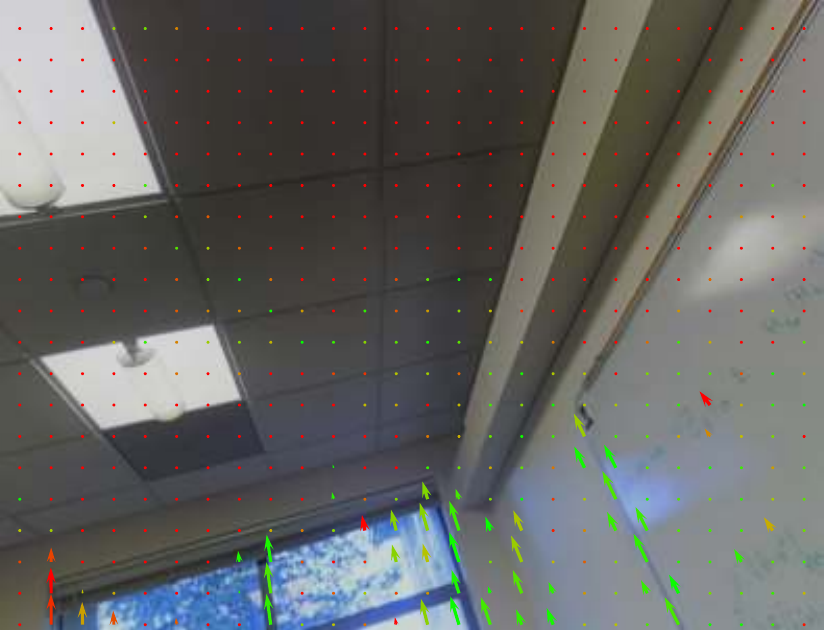}%
    
    \includegraphics[width=\iwidth]{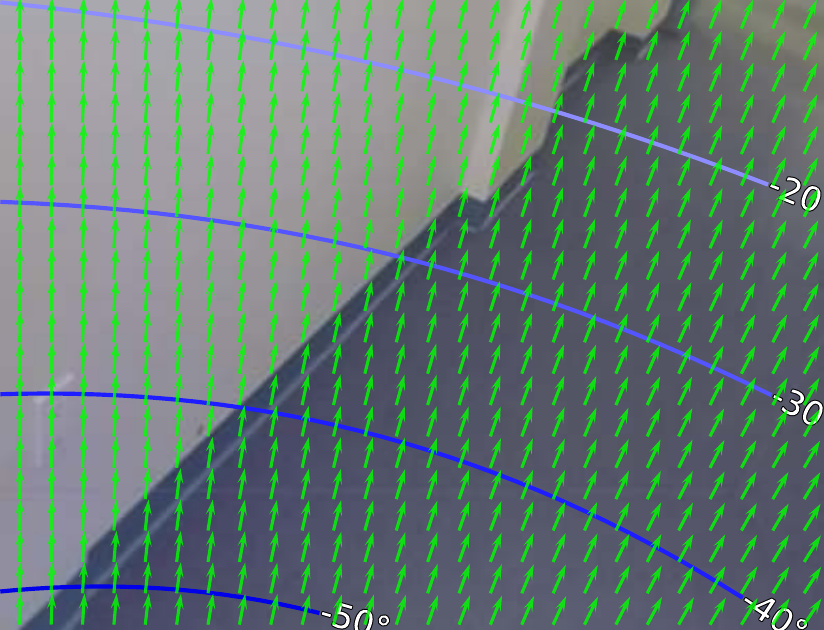}%
    \hspace{\pwidth}%
    \includegraphics[width=\iwidth]{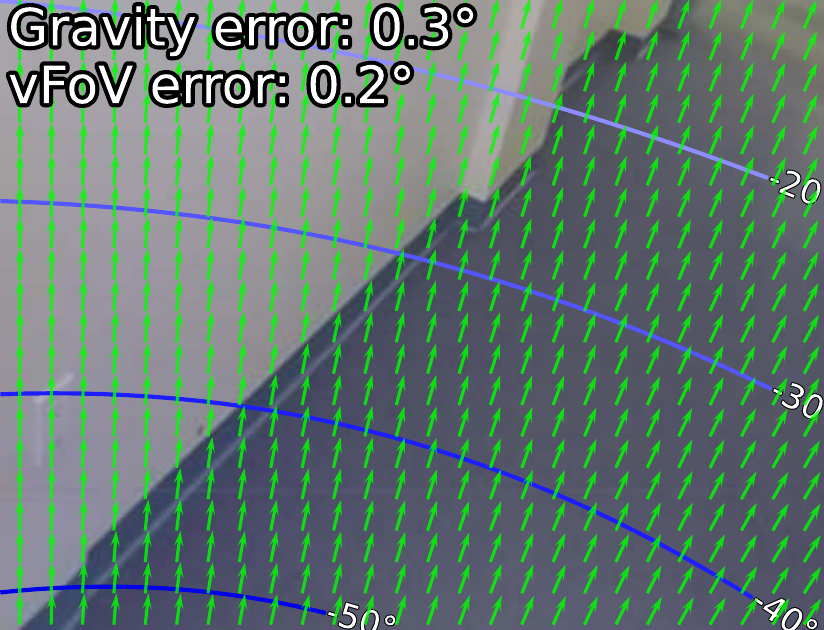}%
    \hspace{\pwidth}%
    \includegraphics[width=\iwidth]{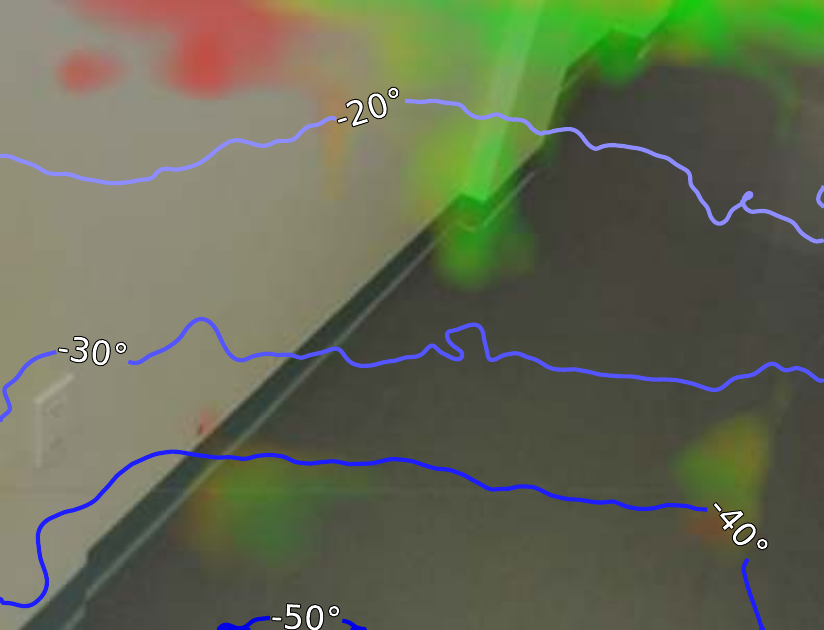}%
    \hspace{\pwidth}%
    \includegraphics[width=\iwidth]{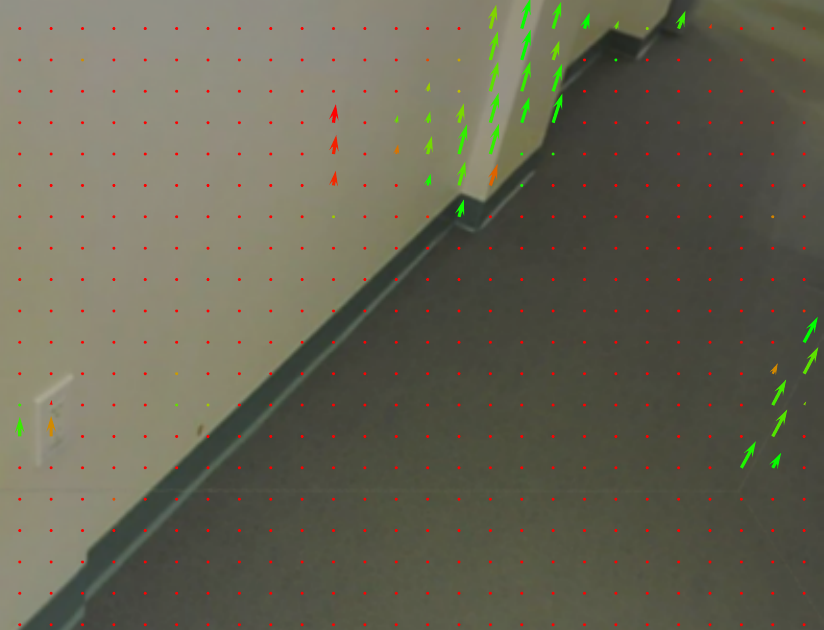}%
    
    \caption{\textbf{Qualitative examples from the Stanford2D3D dataset~\cite{armeni2017joint}.}}%
    \label{fig:stanford2d3d}%
\end{figure}

\begin{figure}[p]
    \centering
    \def\ncols{4}
    \setlength{\pwidth}{0.005\linewidth}
    \setlength{\iwidth}{\dimexpr(0.999\linewidth - \ncols\pwidth + \pwidth)/\ncols \relax}
    \setlength{\lamarwidth}{\dimexpr(0.999\linewidth - 8\pwidth + \pwidth)/8 \relax}
    
    \begin{minipage}[b]{\iwidth}
    \centering{\footnotesize a) ground-truth}
    \end{minipage}%
    \hspace{\pwidth}%
    \begin{minipage}[b]{\iwidth}
    \centering{\footnotesize b) final prediction}
    \end{minipage}%
    \hspace{\pwidth}%
    \begin{minipage}[b]{\iwidth}
    \centering{\footnotesize c) observed latitude}
    \end{minipage}%
    \hspace{\pwidth}%
    \begin{minipage}[b]{\iwidth}
    \centering{\footnotesize d) observed up-vect.}
    \end{minipage}%
    
    \includegraphics[width=\iwidth]{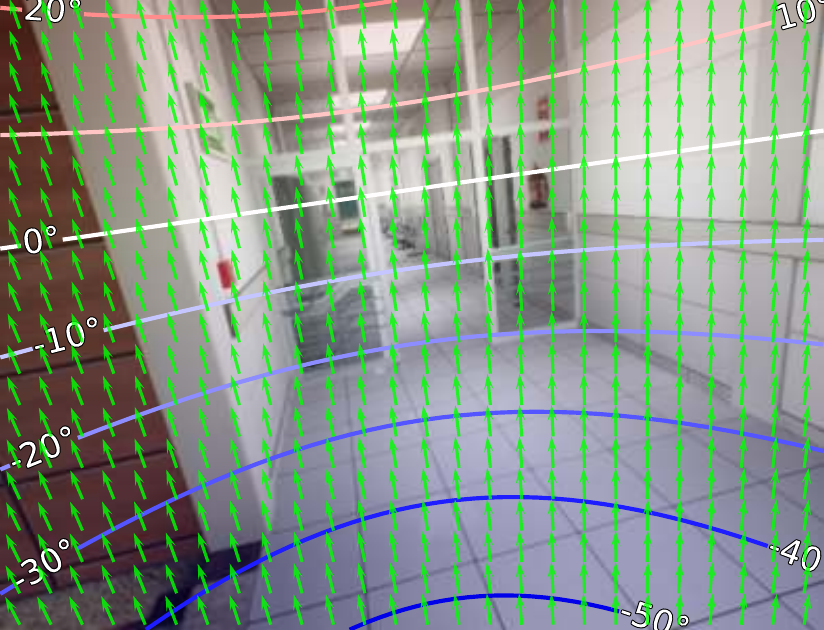}%
    \hspace{\pwidth}%
    \includegraphics[width=\iwidth]{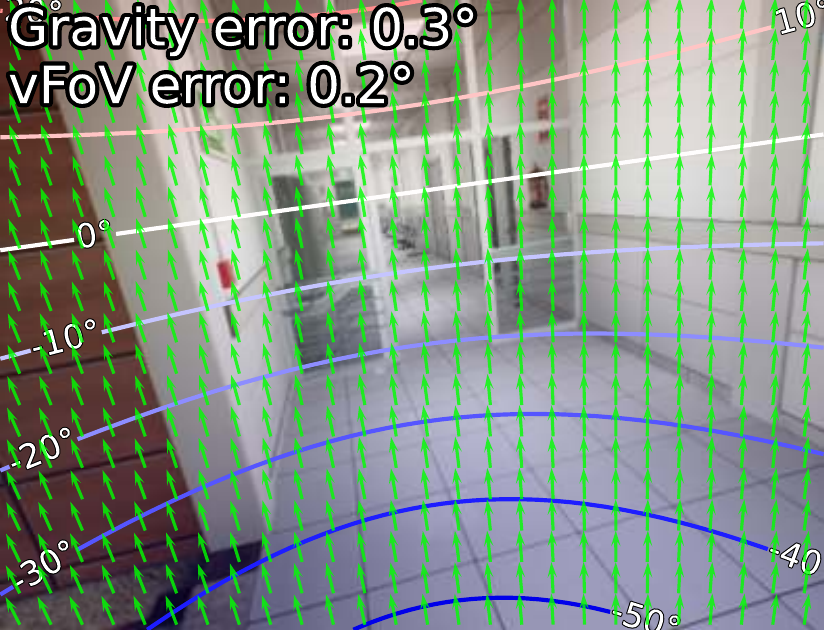}%
    \hspace{\pwidth}%
    \includegraphics[width=\iwidth]{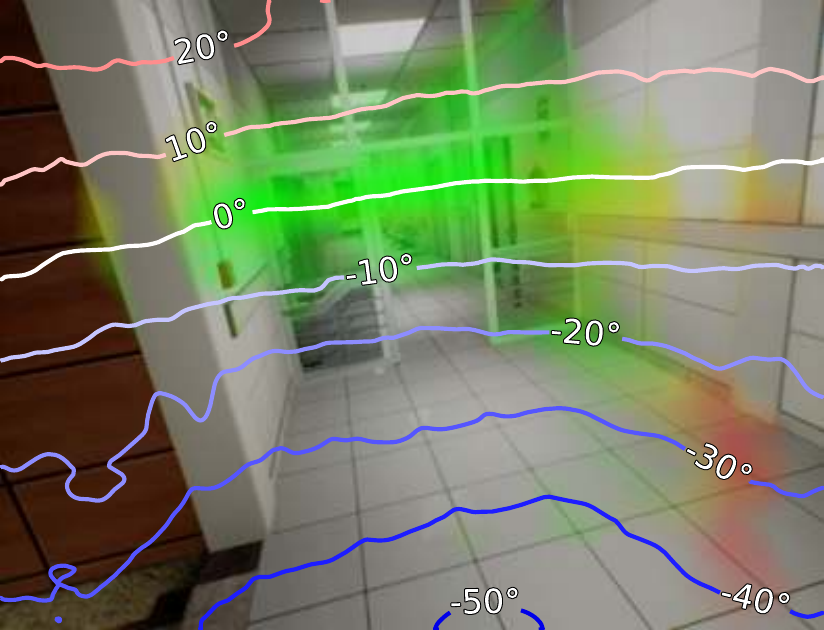}%
    \hspace{\pwidth}%
    \includegraphics[width=\iwidth]{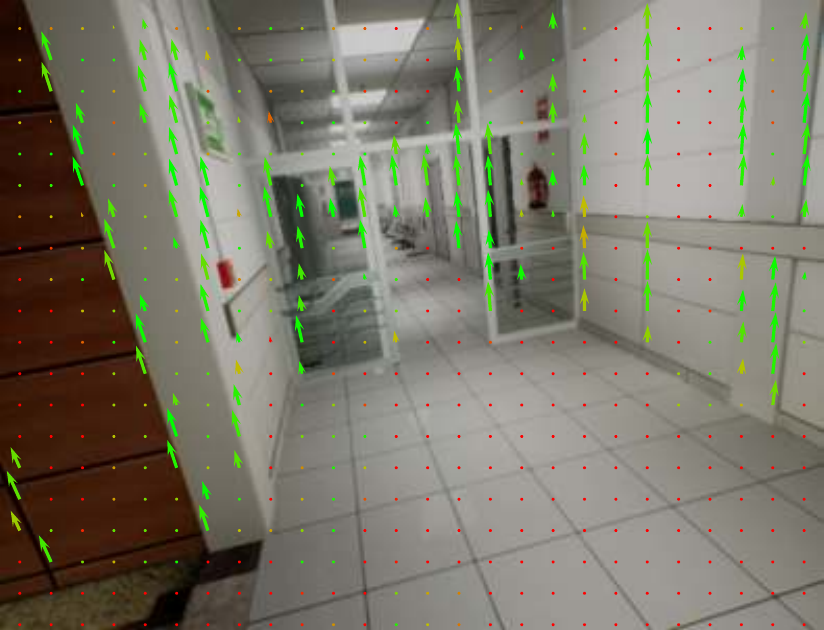}%
    
    \includegraphics[width=\iwidth]{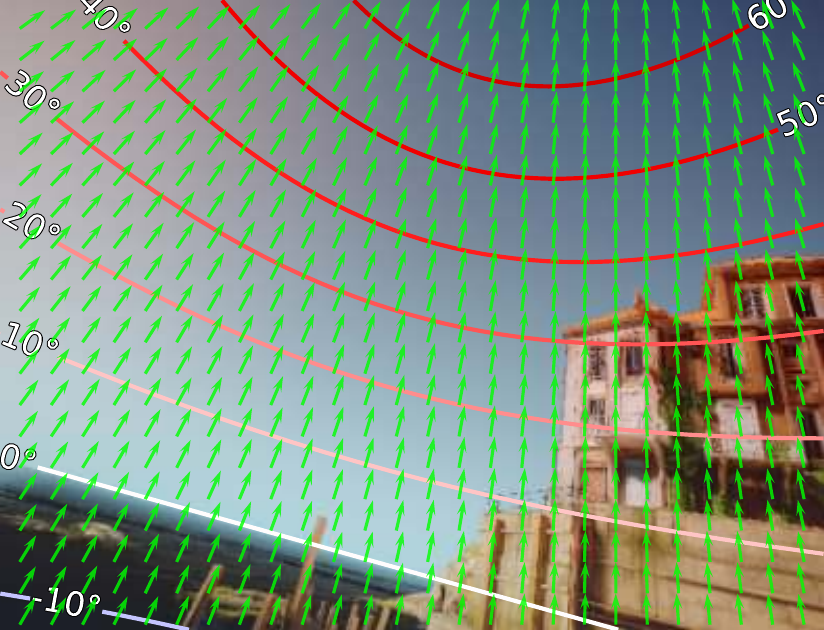}%
    \hspace{\pwidth}%
    \includegraphics[width=\iwidth]{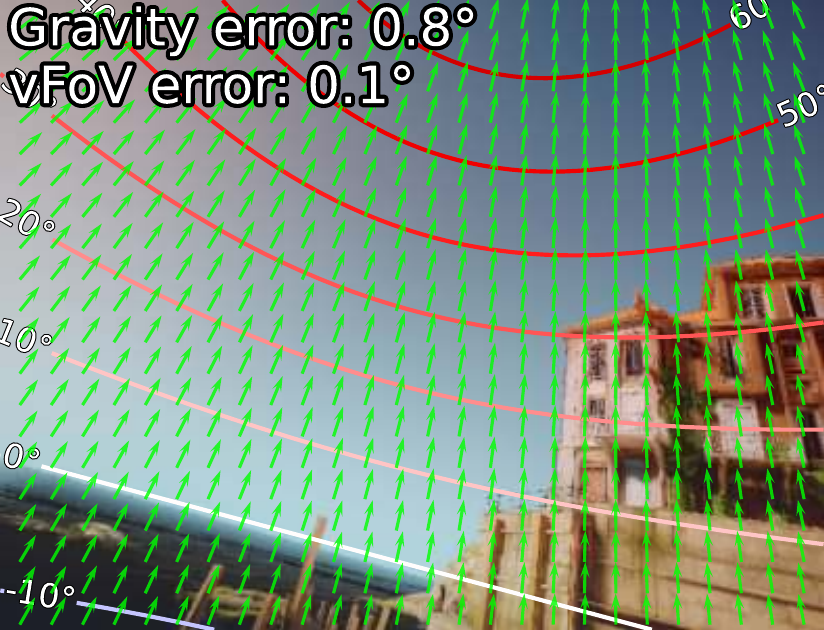}%
    \hspace{\pwidth}%
    \includegraphics[width=\iwidth]{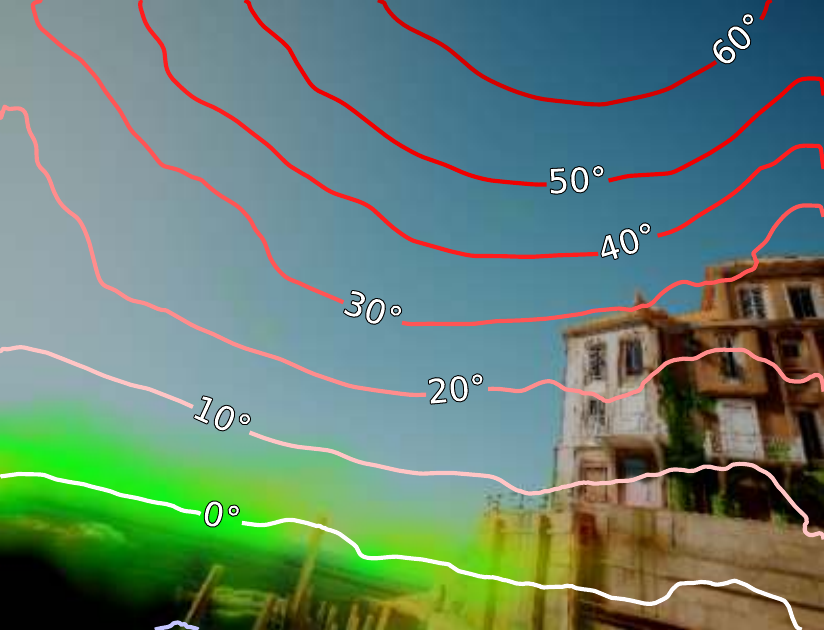}%
    \hspace{\pwidth}%
    \includegraphics[width=\iwidth]{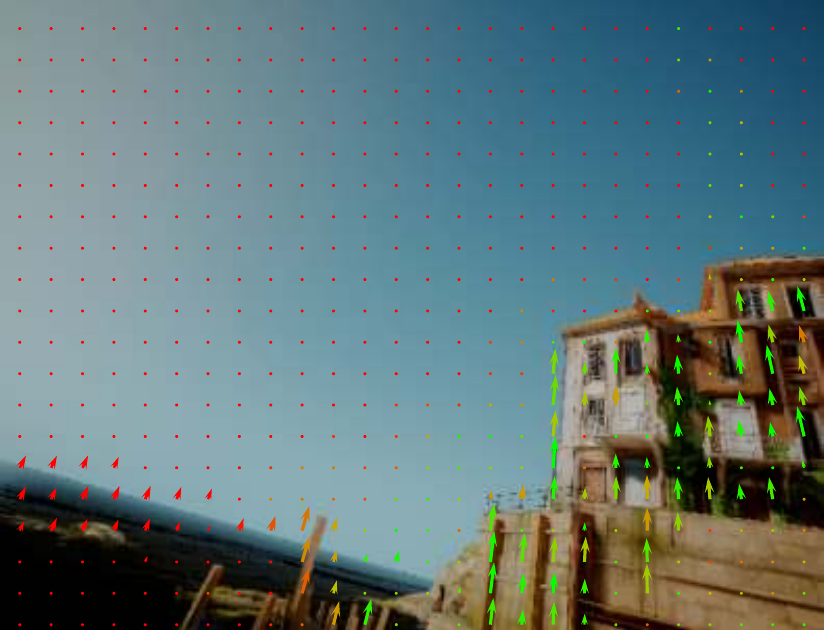}%
    
    \includegraphics[width=\iwidth]{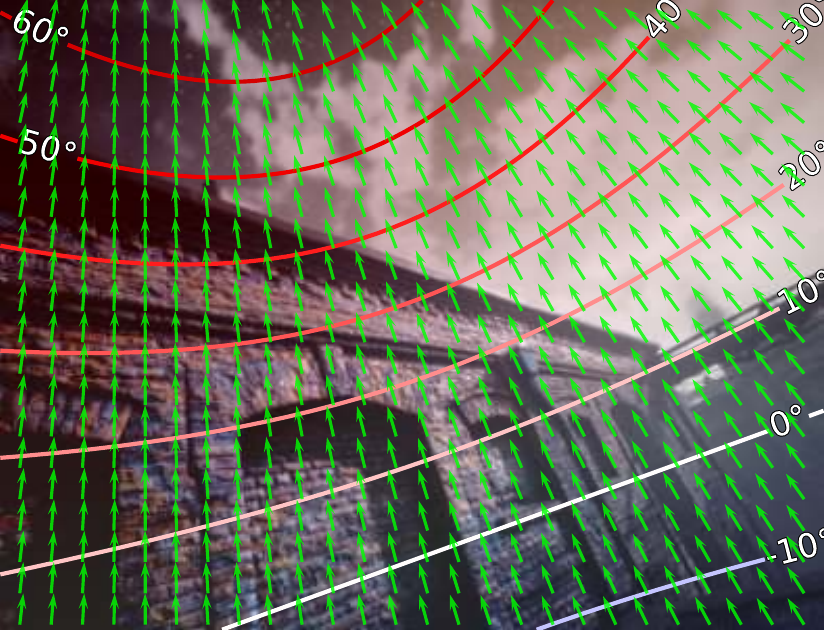}%
    \hspace{\pwidth}%
    \includegraphics[width=\iwidth]{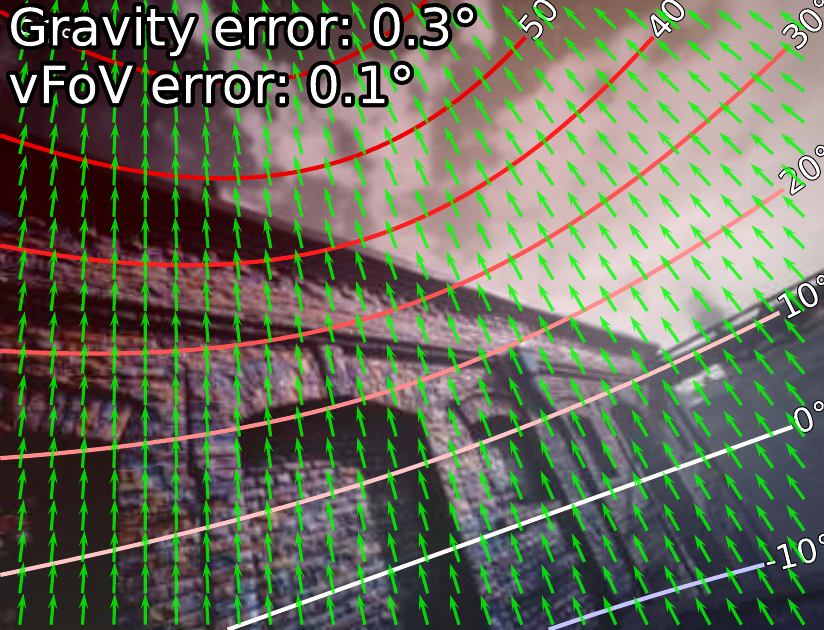}%
    \hspace{\pwidth}%
    \includegraphics[width=\iwidth]{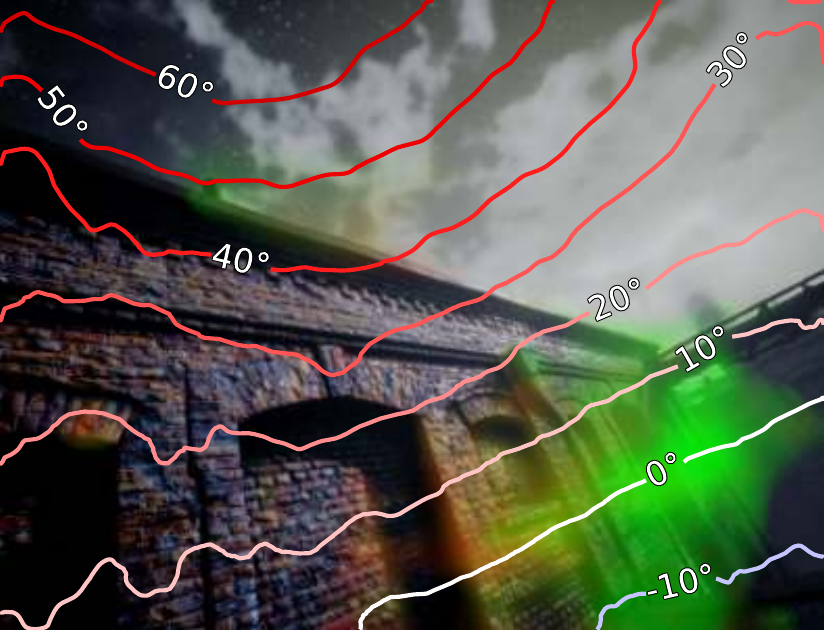}%
    \hspace{\pwidth}%
    \includegraphics[width=\iwidth]{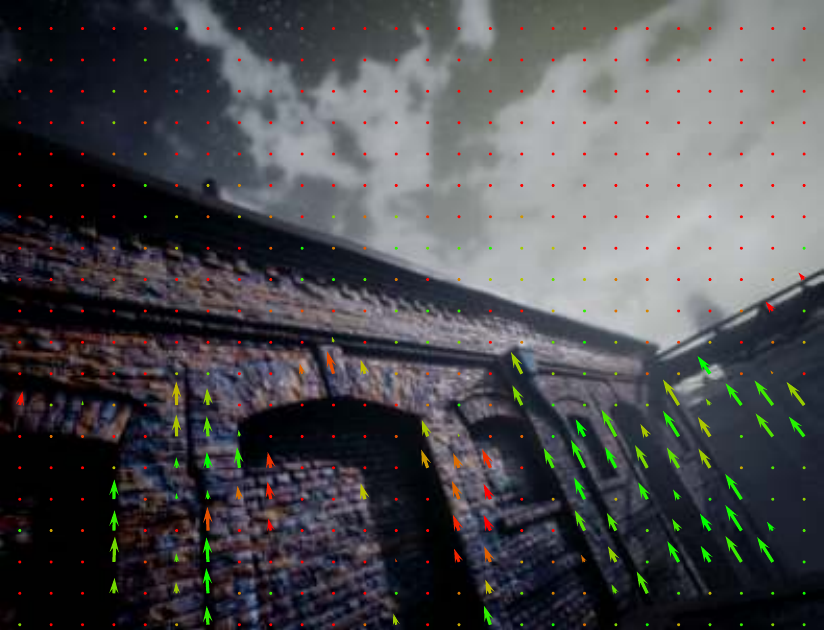}%
    
    \includegraphics[width=\iwidth]{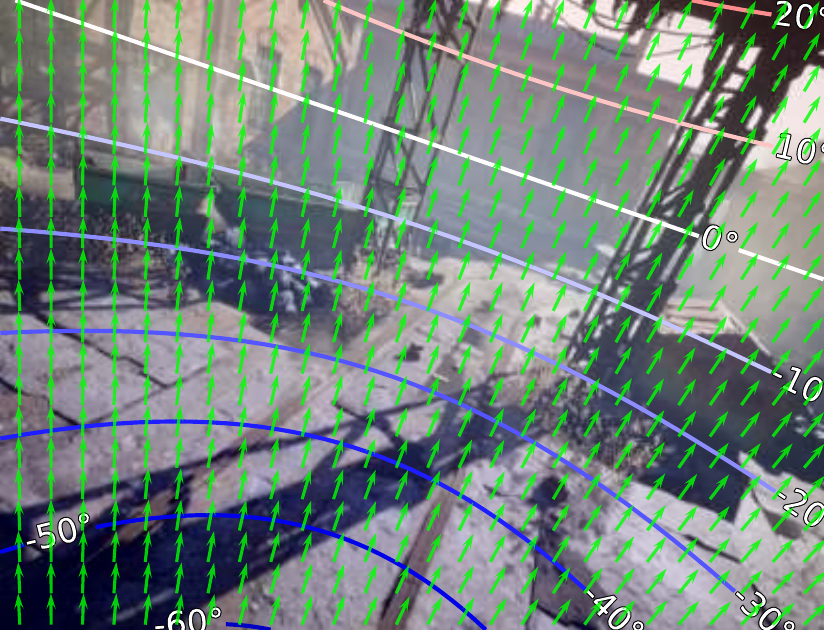}%
    \hspace{\pwidth}%
    \includegraphics[width=\iwidth]{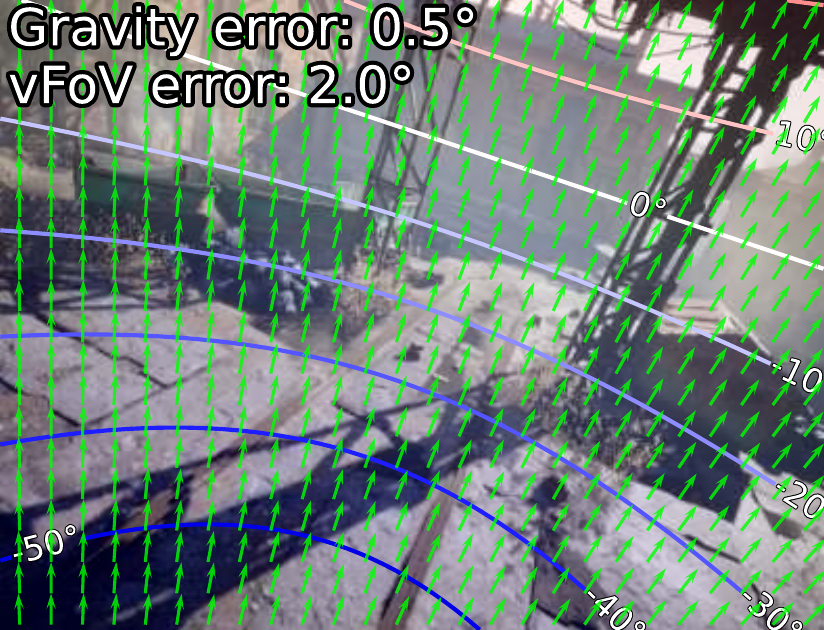}%
    \hspace{\pwidth}%
    \includegraphics[width=\iwidth]{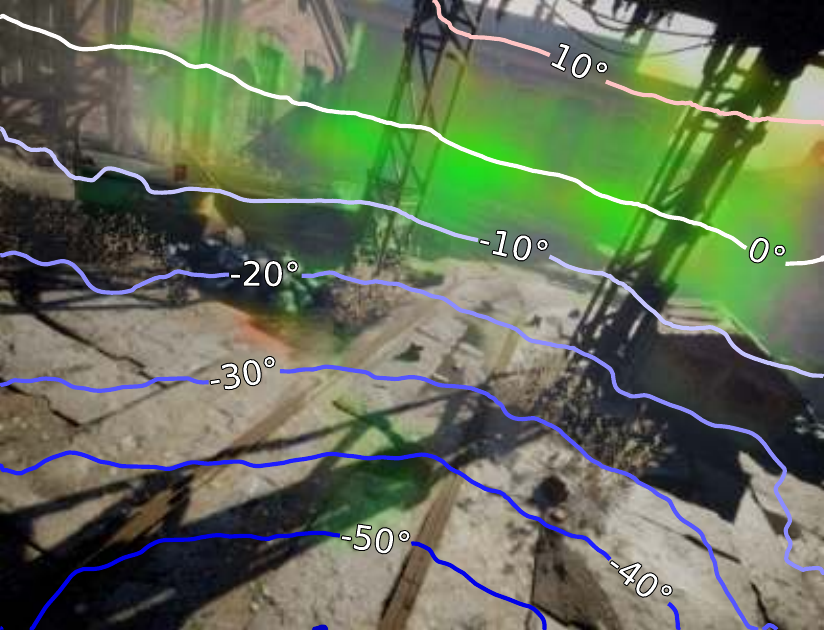}%
    \hspace{\pwidth}%
    \includegraphics[width=\iwidth]{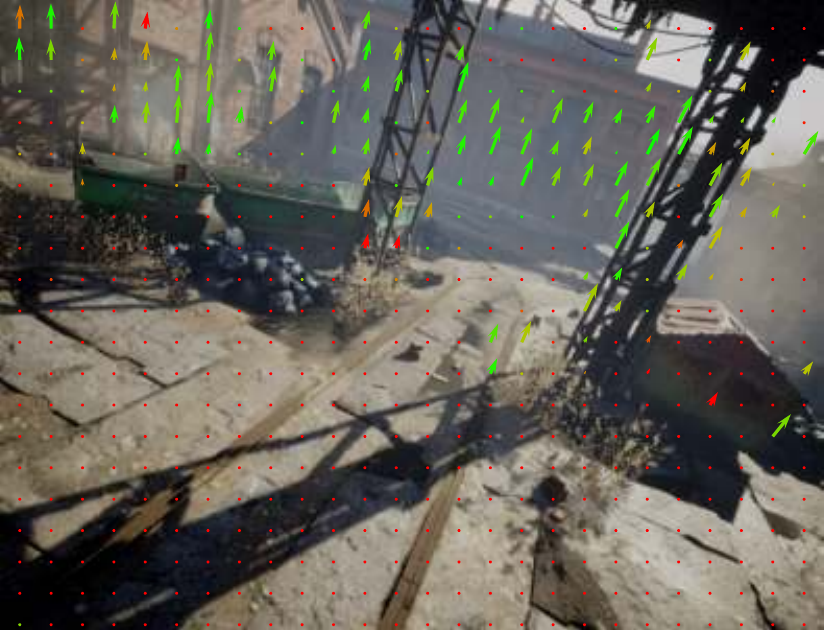}%
    
    \includegraphics[width=\iwidth]{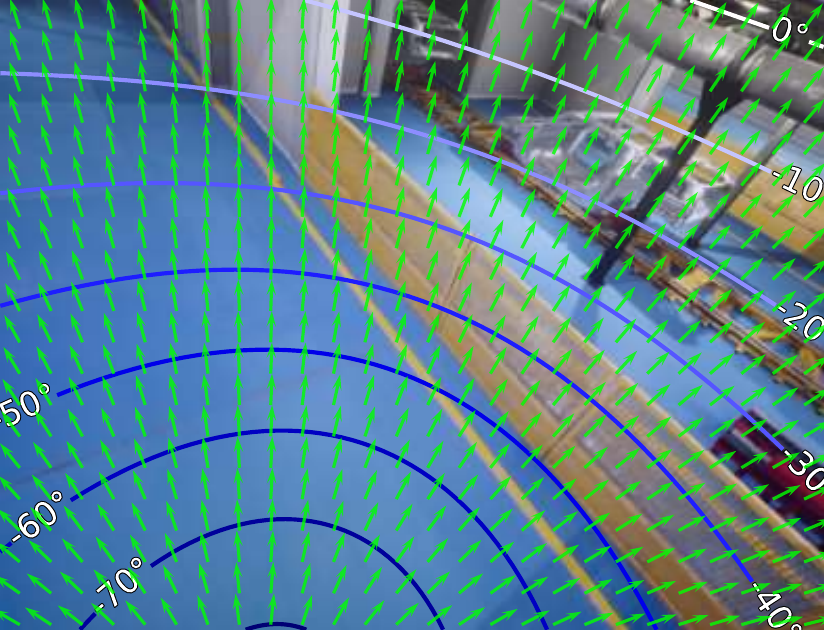}%
    \hspace{\pwidth}%
    \includegraphics[width=\iwidth]{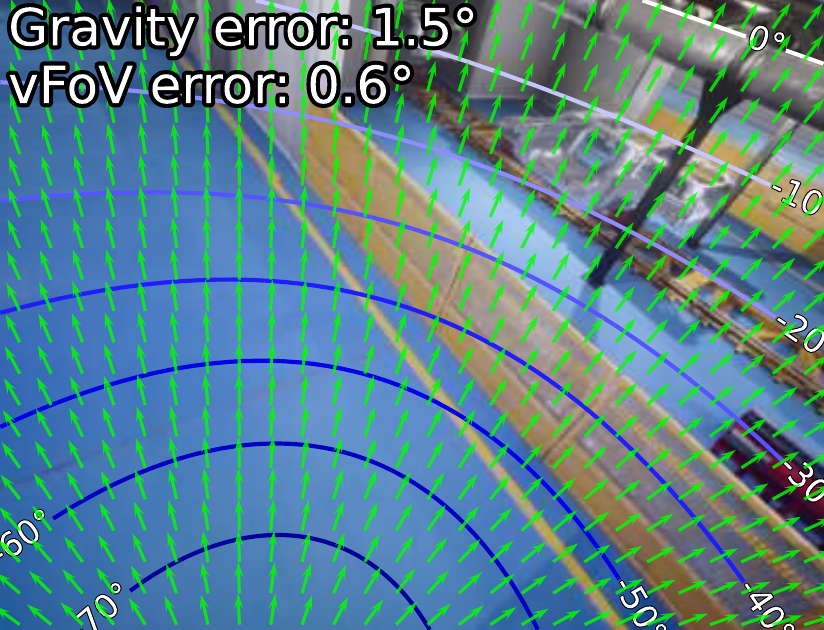}%
    \hspace{\pwidth}%
    \includegraphics[width=\iwidth]{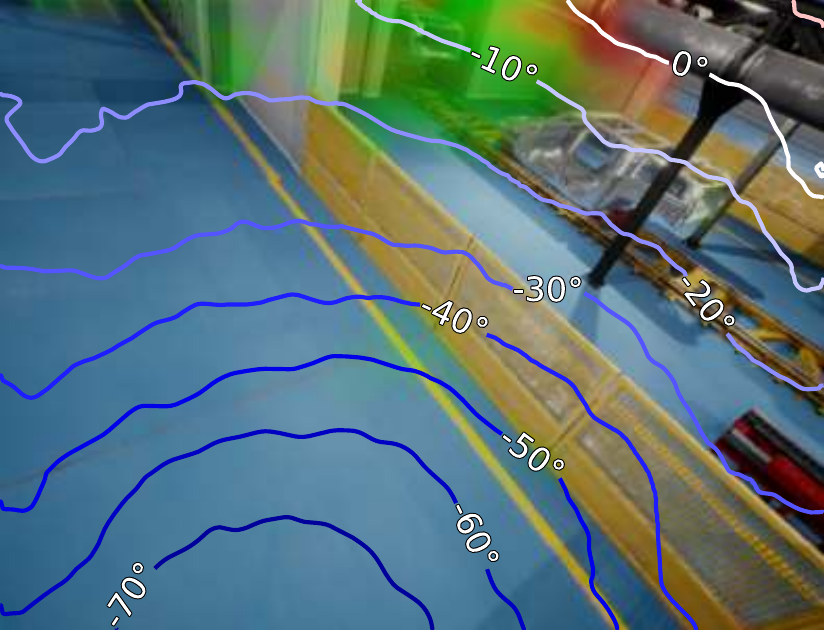}%
    \hspace{\pwidth}%
    \includegraphics[width=\iwidth]{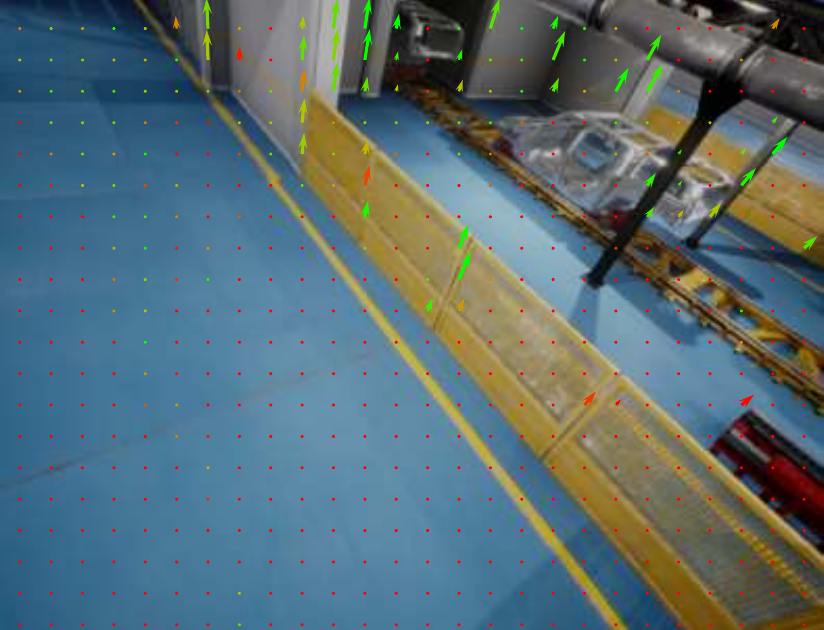}%
    
    \includegraphics[width=\iwidth]{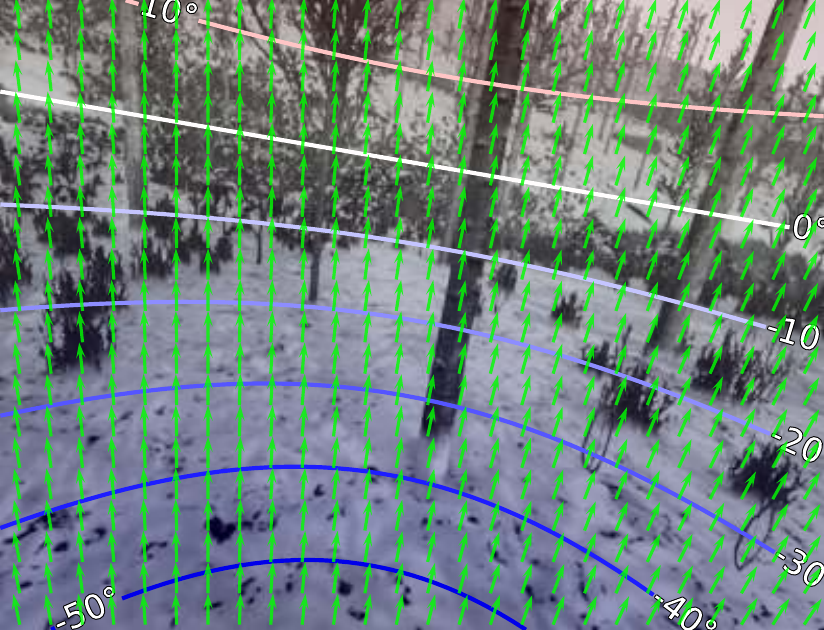}%
    \hspace{\pwidth}%
    \includegraphics[width=\iwidth]{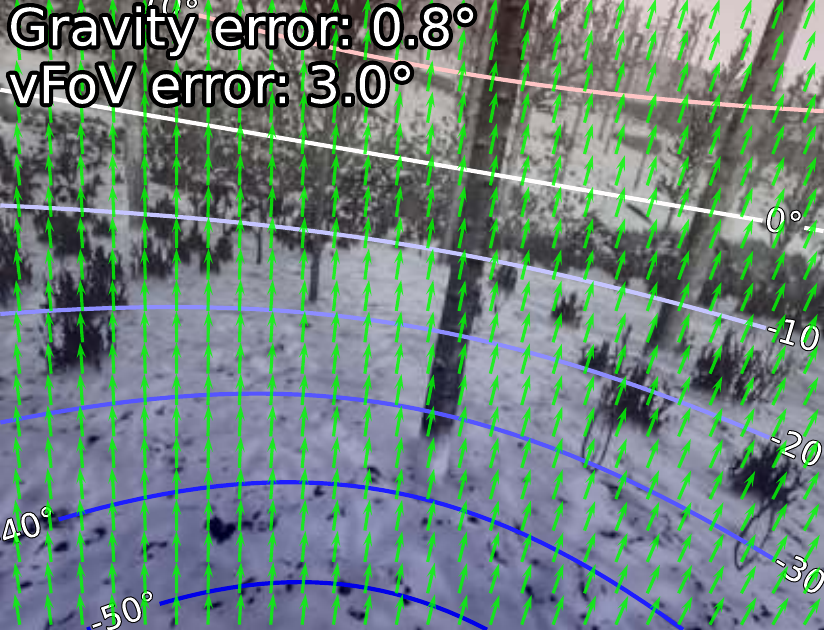}%
    \hspace{\pwidth}%
    \includegraphics[width=\iwidth]{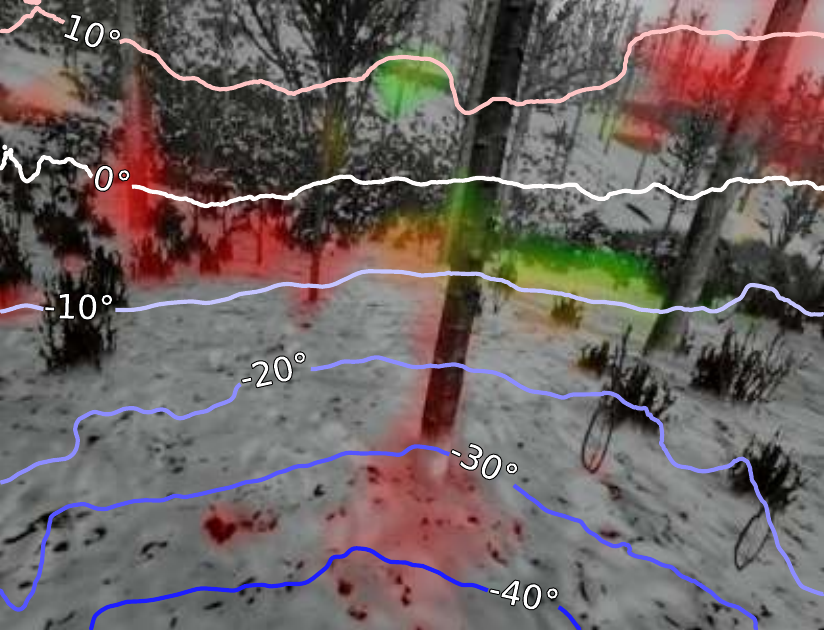}%
    \hspace{\pwidth}%
    \includegraphics[width=\iwidth]{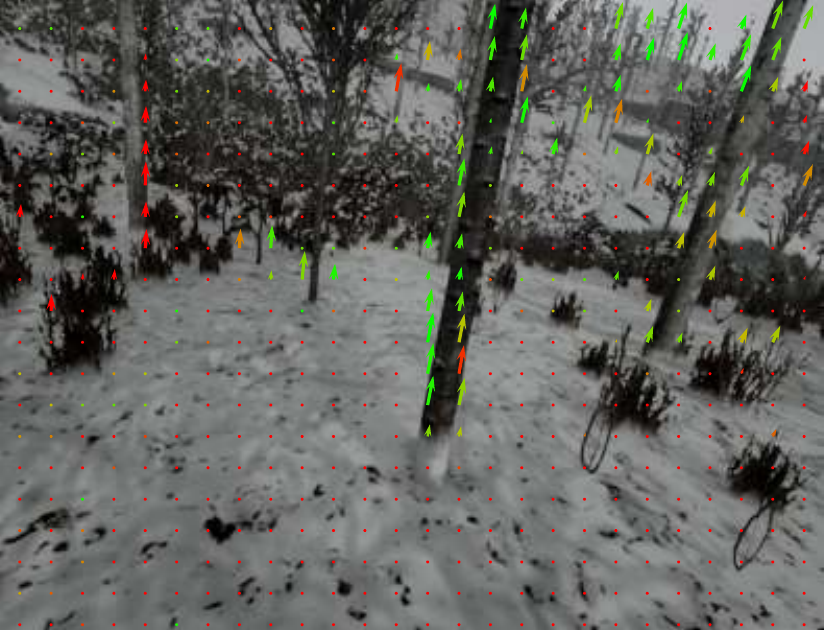}%

    \caption{\textbf{Qualitative examples from the TartanAir dataset~\cite{tartanair2020iros}.}}%
    \label{fig:tartanair}%
\end{figure}

\begin{figure}[p]
    \centering
    \def\ncols{4}
    \setlength{\pwidth}{0.005\linewidth}
    \setlength{\iwidth}{\dimexpr(0.999\linewidth - \ncols\pwidth + \pwidth)/\ncols \relax}
    \setlength{\lamarwidth}{\dimexpr(0.999\linewidth - 8\pwidth + \pwidth)/8 \relax}
    
    \begin{minipage}[b]{\iwidth}
    \centering{\footnotesize a) ground-truth}
    \end{minipage}%
    \hspace{\pwidth}%
    \begin{minipage}[b]{\iwidth}
    \centering{\footnotesize b) final prediction}
    \end{minipage}%
    \hspace{\pwidth}%
    \begin{minipage}[b]{\iwidth}
    \centering{\footnotesize c) observed latitude}
    \end{minipage}%
    \hspace{\pwidth}%
    \begin{minipage}[b]{\iwidth}
    \centering{\footnotesize d) observed up-vect.}
    \end{minipage}%
    
    \includegraphics[width=\iwidth]{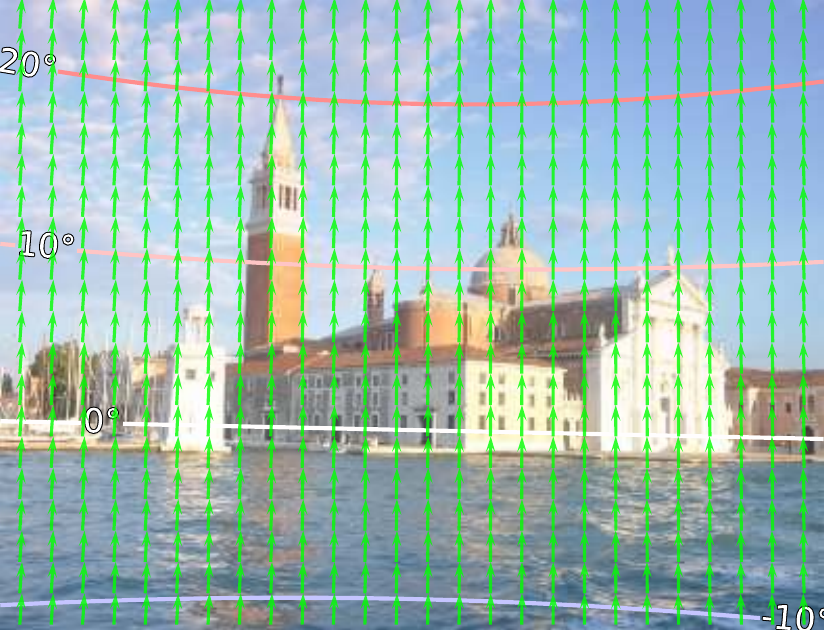}%
    \hspace{\pwidth}%
    \includegraphics[width=\iwidth]{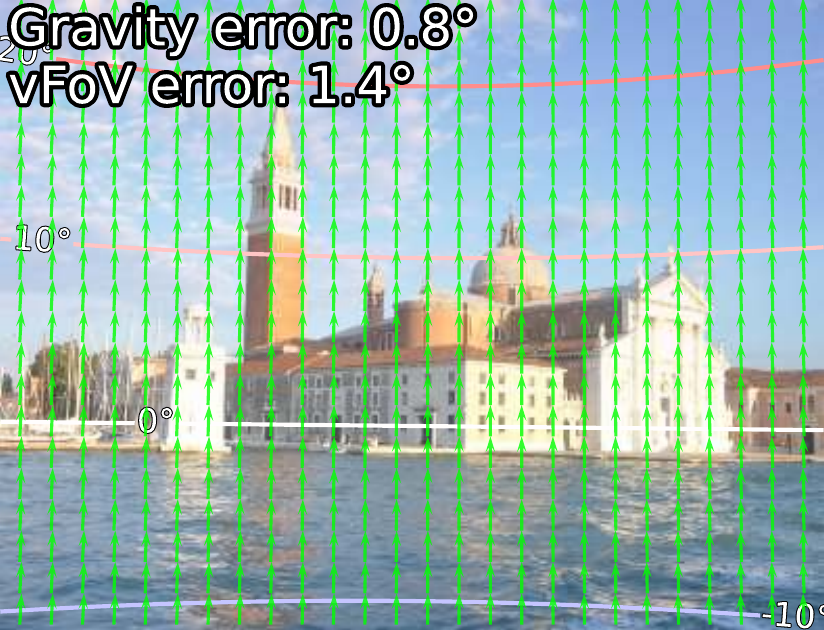}%
    \hspace{\pwidth}%
    \includegraphics[width=\iwidth]{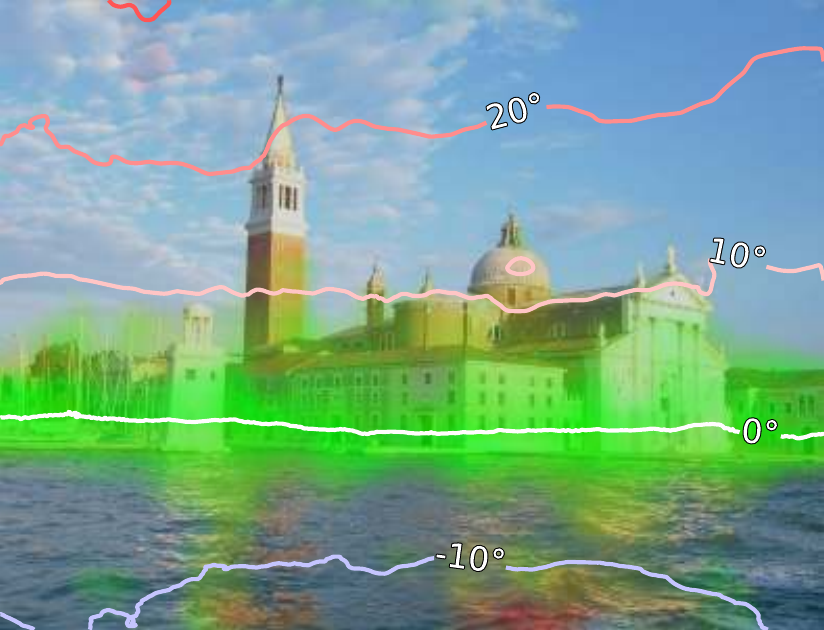}%
    \hspace{\pwidth}%
    \includegraphics[width=\iwidth]{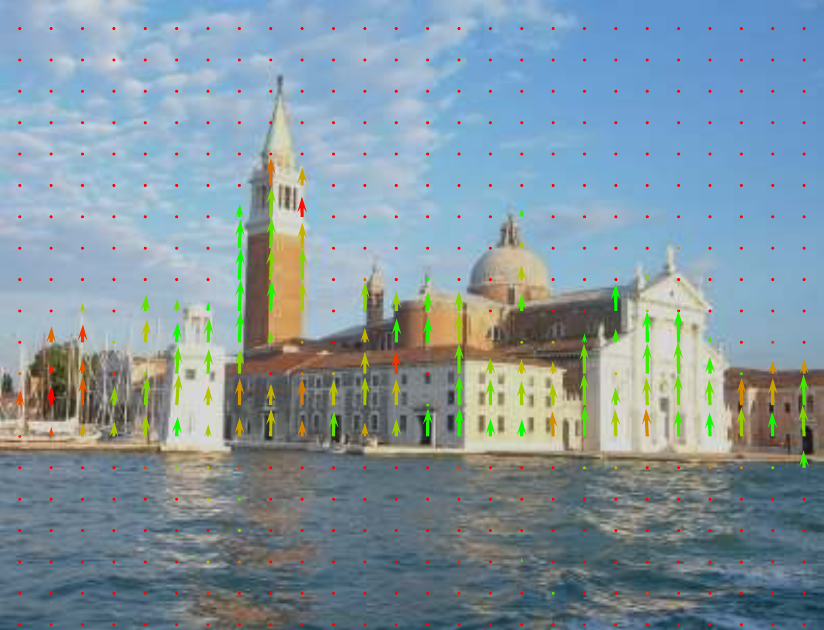}%
    
    \includegraphics[width=\iwidth]{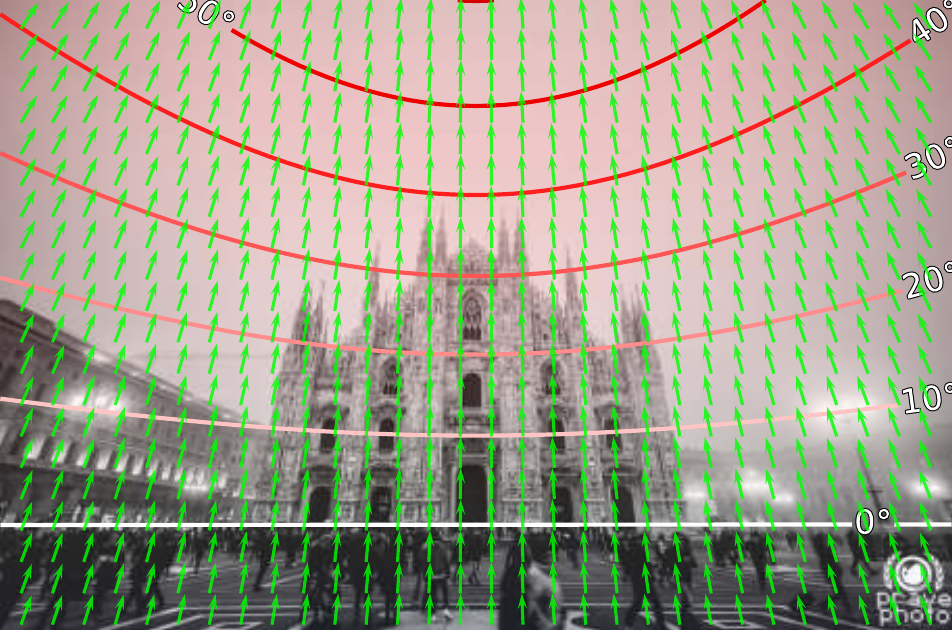}%
    \hspace{\pwidth}%
    \includegraphics[width=\iwidth]{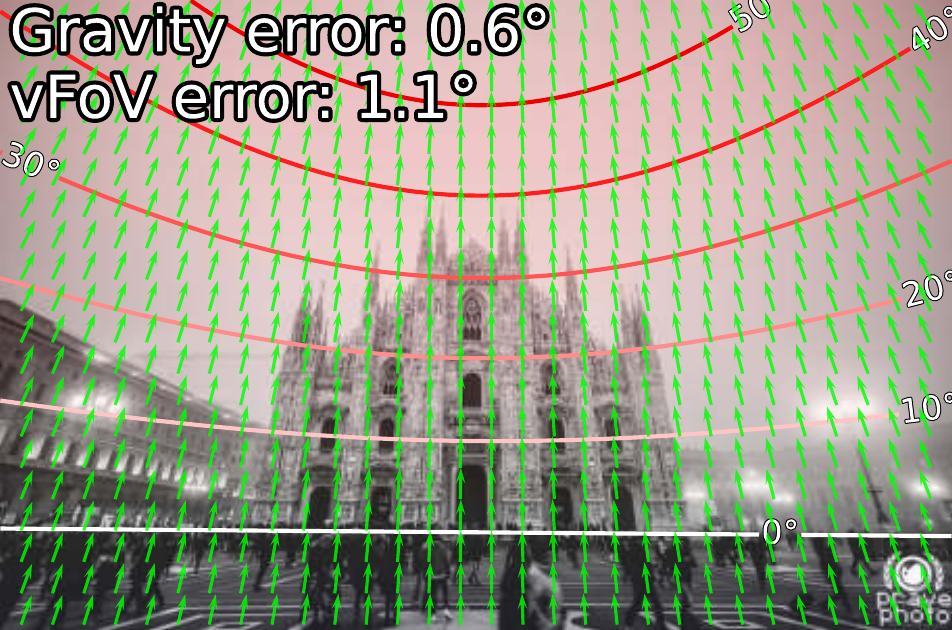}%
    \hspace{\pwidth}%
    \includegraphics[width=\iwidth]{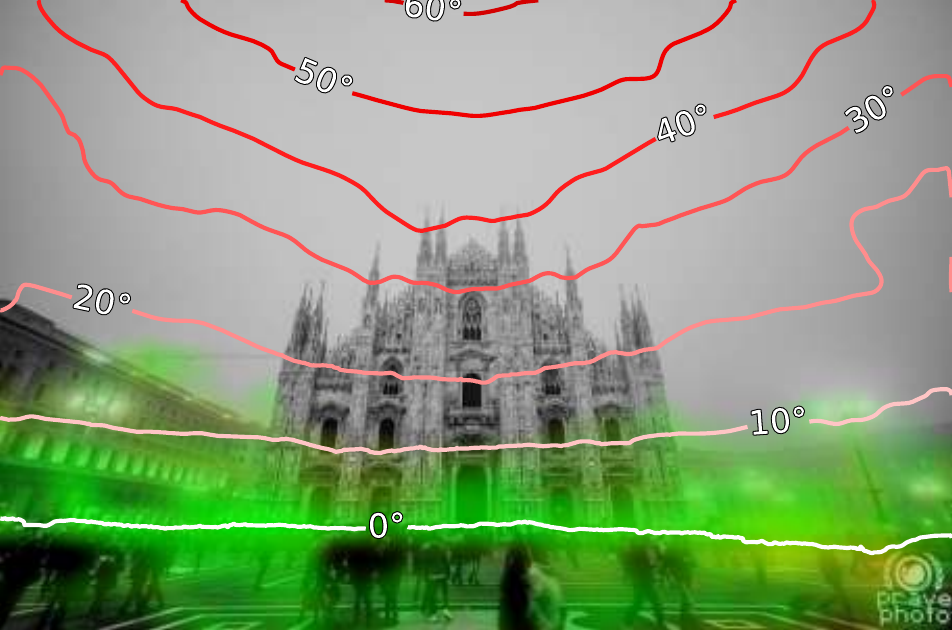}%
    \hspace{\pwidth}%
    \includegraphics[width=\iwidth]{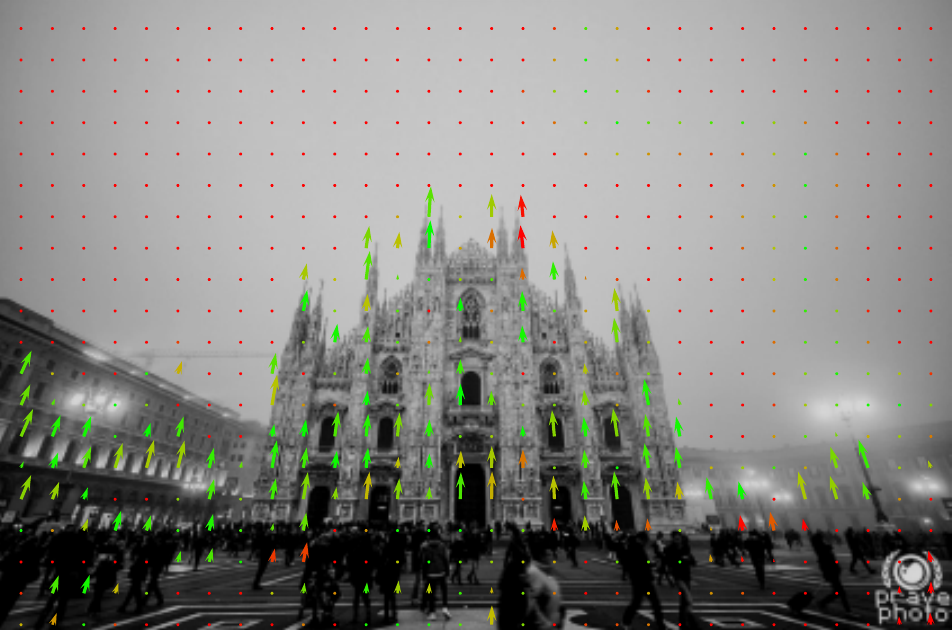}%
    
    \includegraphics[width=\iwidth]{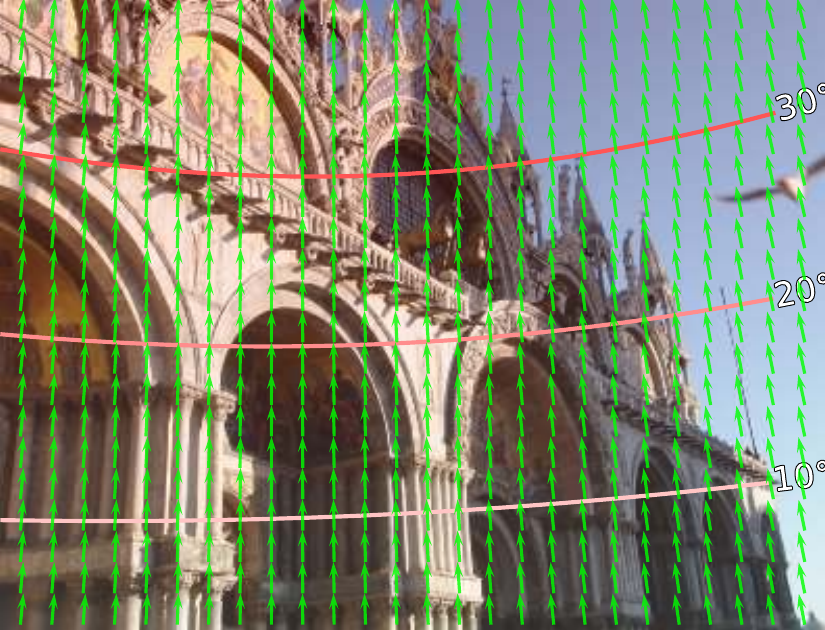}%
    \hspace{\pwidth}%
    \includegraphics[width=\iwidth]{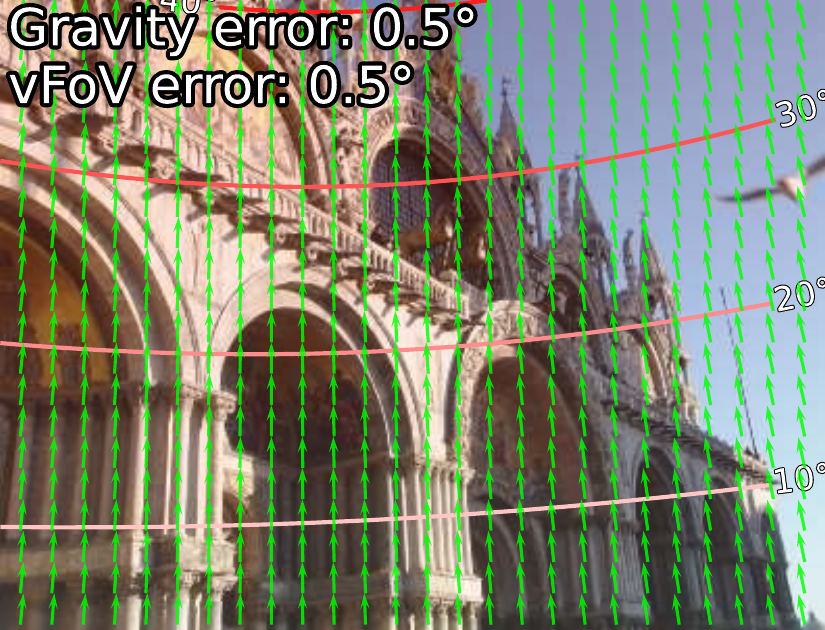}%
    \hspace{\pwidth}%
    \includegraphics[width=\iwidth]{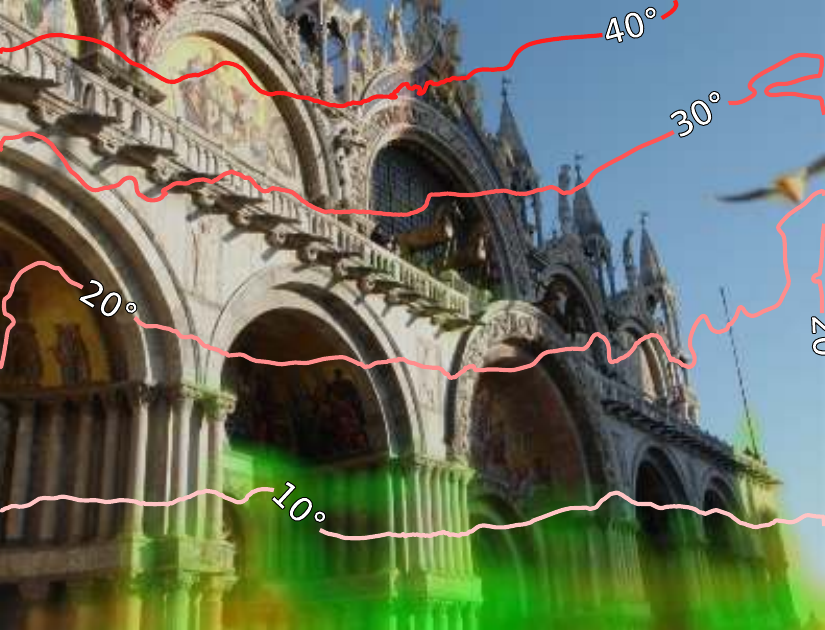}%
    \hspace{\pwidth}%
    \includegraphics[width=\iwidth]{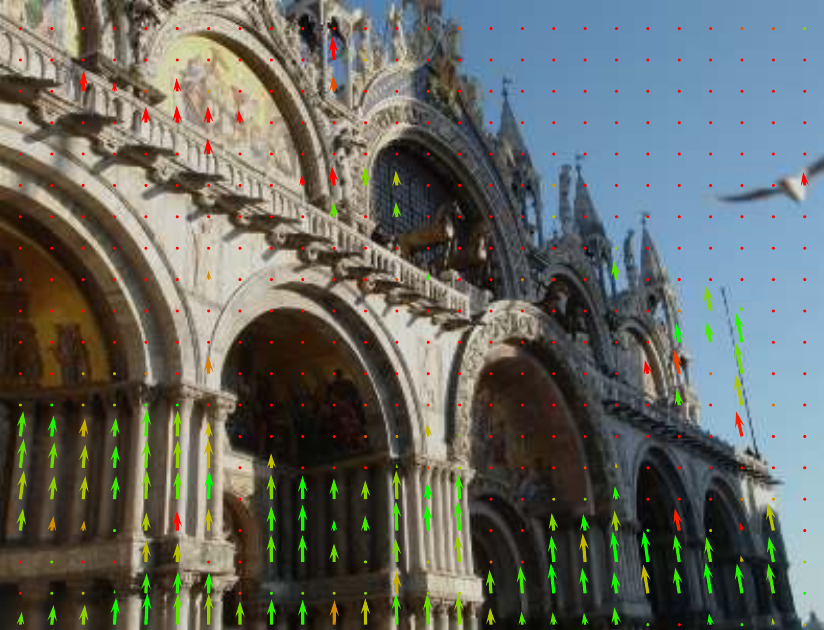}%
    
    \includegraphics[width=\lamarwidth]{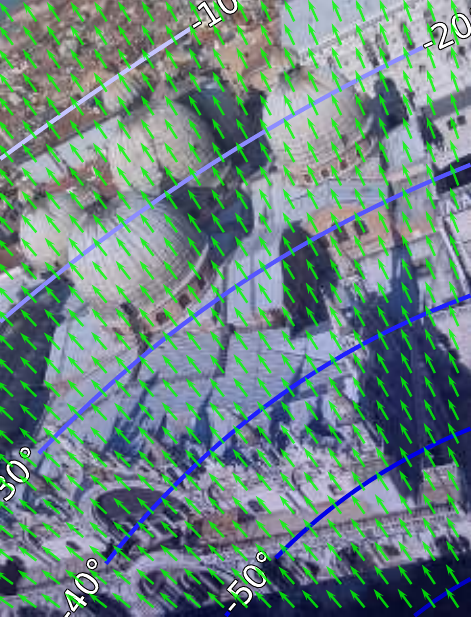}%
    \hspace{\pwidth}%
    \includegraphics[width=\lamarwidth]{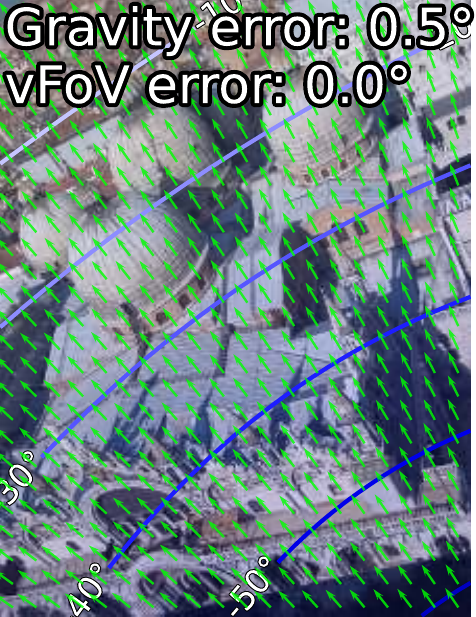}%
    \hspace{\pwidth}%
    \includegraphics[width=\lamarwidth]{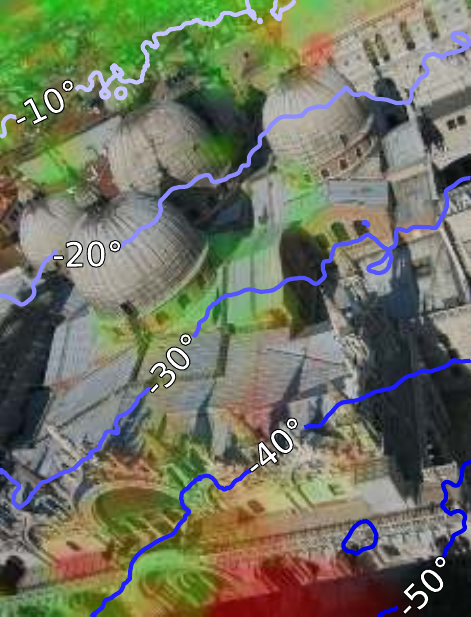}%
    \hspace{\pwidth}%
    \includegraphics[width=\lamarwidth]{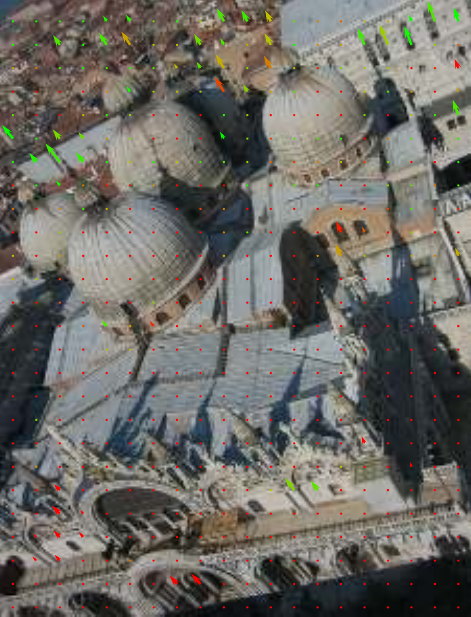}%
    \hspace{\pwidth}%
    \includegraphics[width=\lamarwidth]{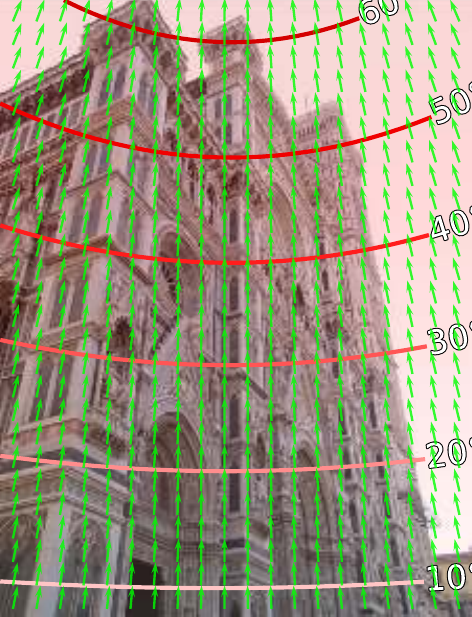}%
    \hspace{\pwidth}%
    \includegraphics[width=\lamarwidth]{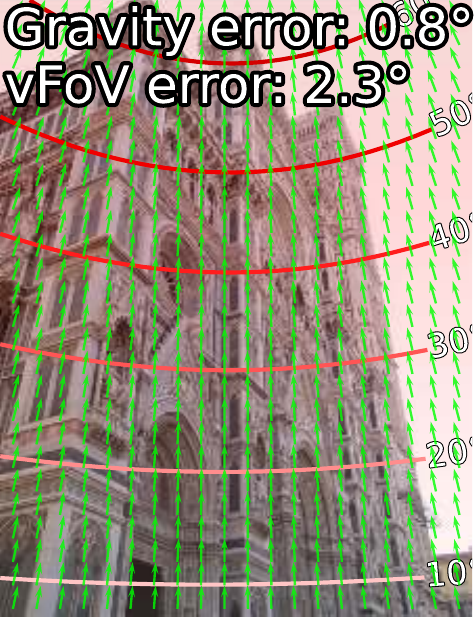}%
    \hspace{\pwidth}%
    \includegraphics[width=\lamarwidth]{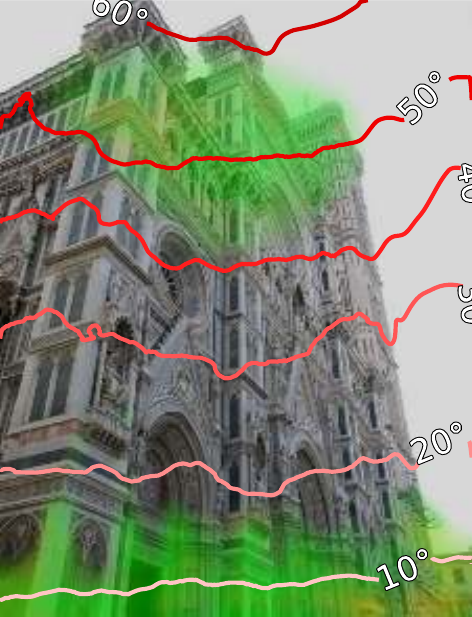}%
    \hspace{\pwidth}%
    \includegraphics[width=\lamarwidth]{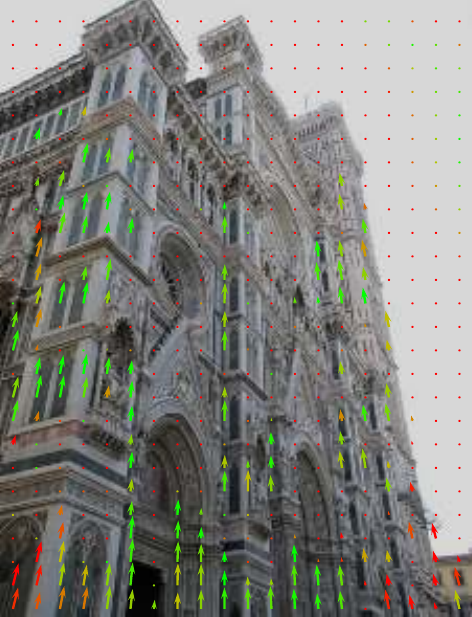}%
    
    \includegraphics[width=\iwidth]{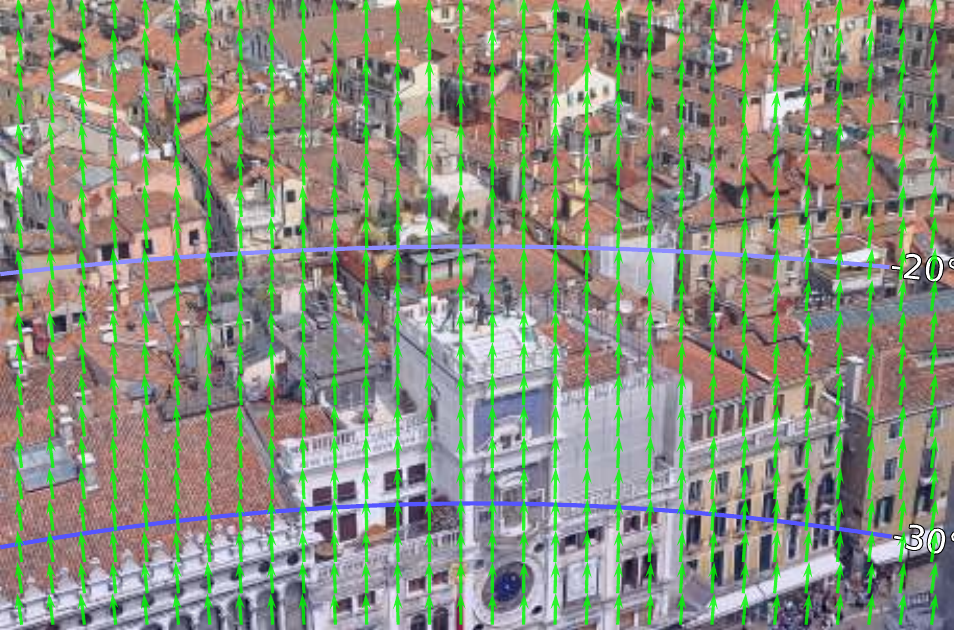}%
    \hspace{\pwidth}%
    \includegraphics[width=\iwidth]{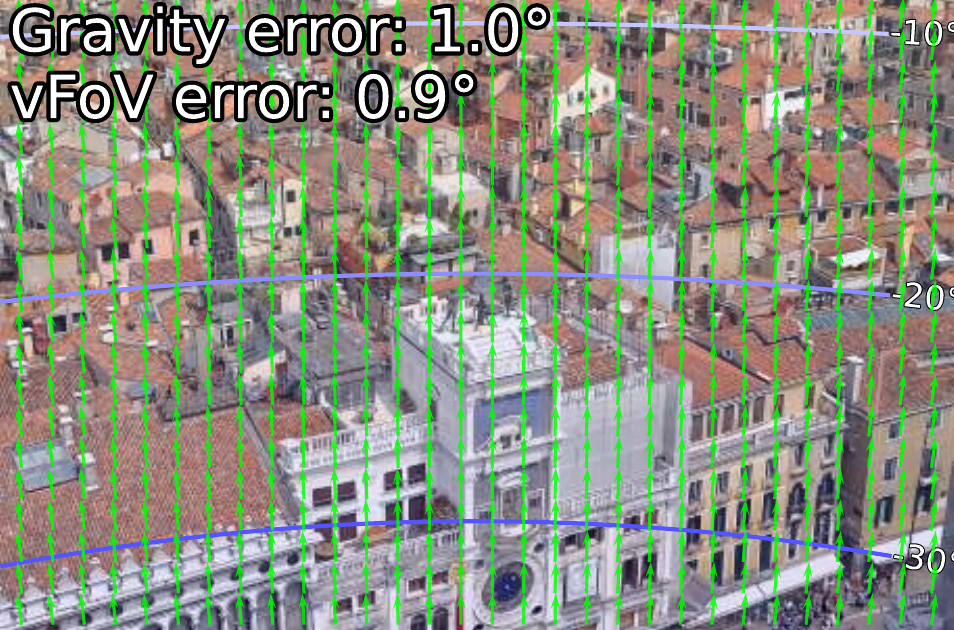}%
    \hspace{\pwidth}%
    \includegraphics[width=\iwidth]{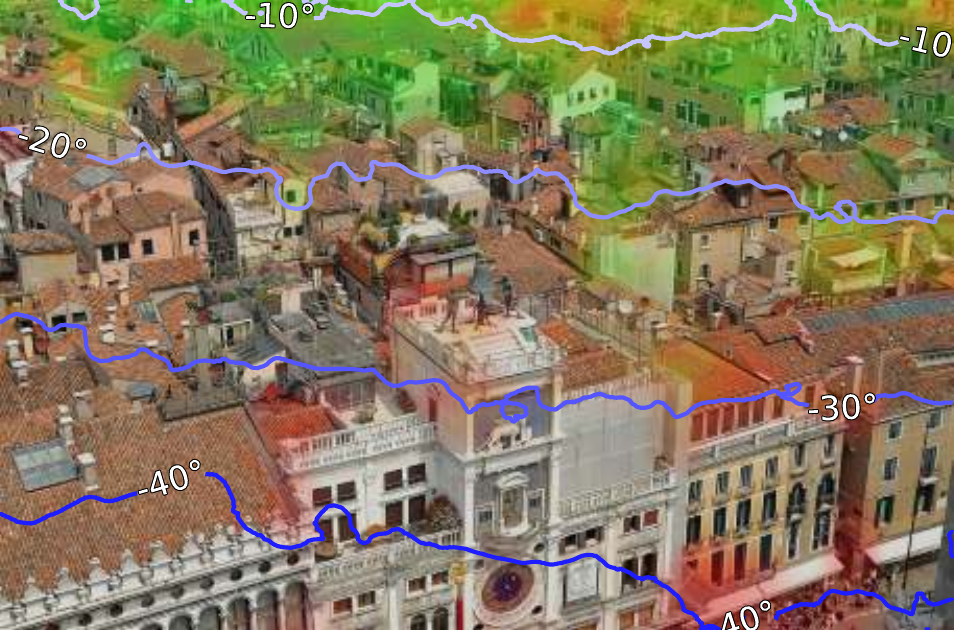}%
    \hspace{\pwidth}%
    \includegraphics[width=\iwidth]{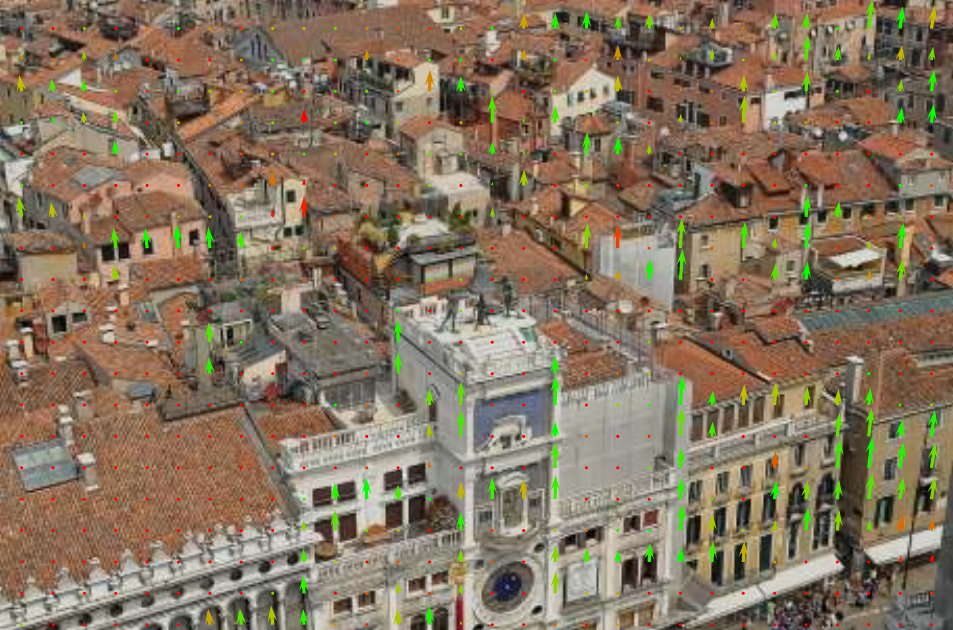}%
    
    \includegraphics[width=\iwidth]{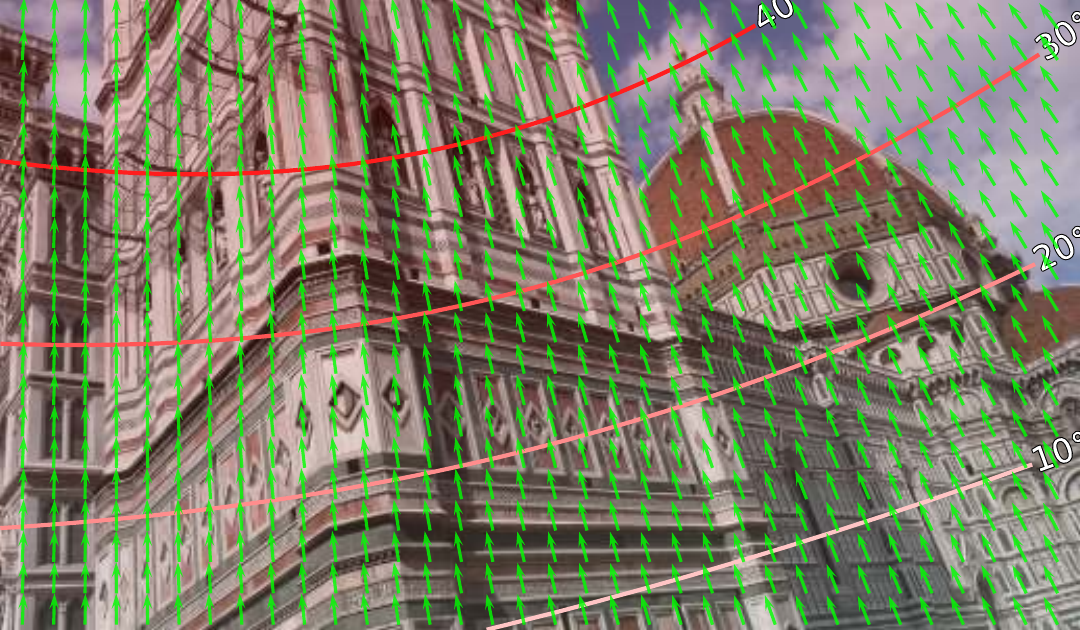}%
    \hspace{\pwidth}%
    \includegraphics[width=\iwidth]{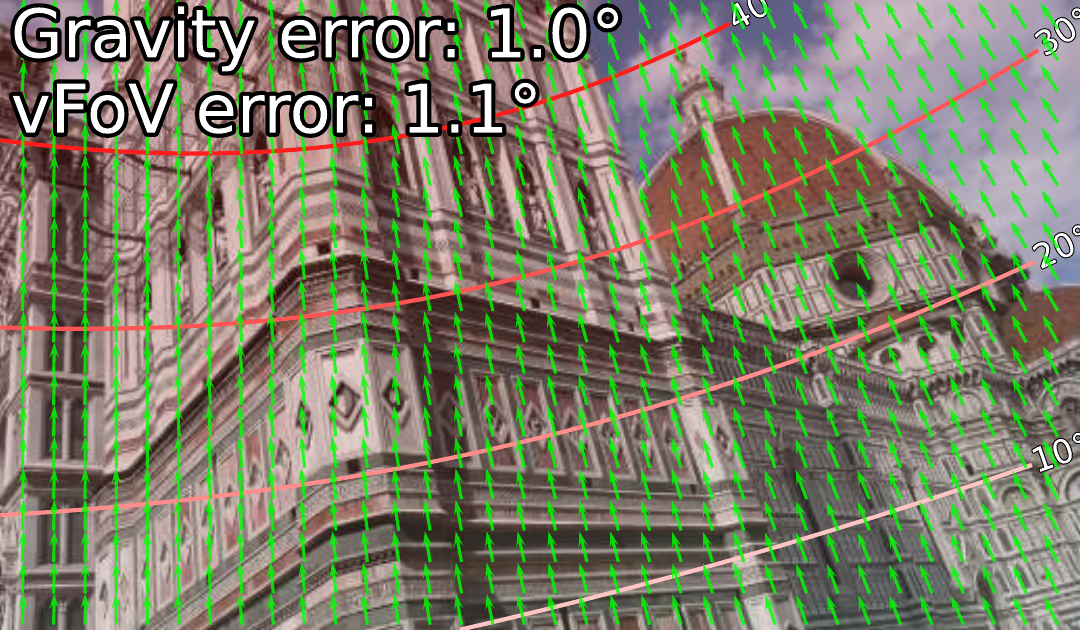}%
    \hspace{\pwidth}%
    \includegraphics[width=\iwidth]{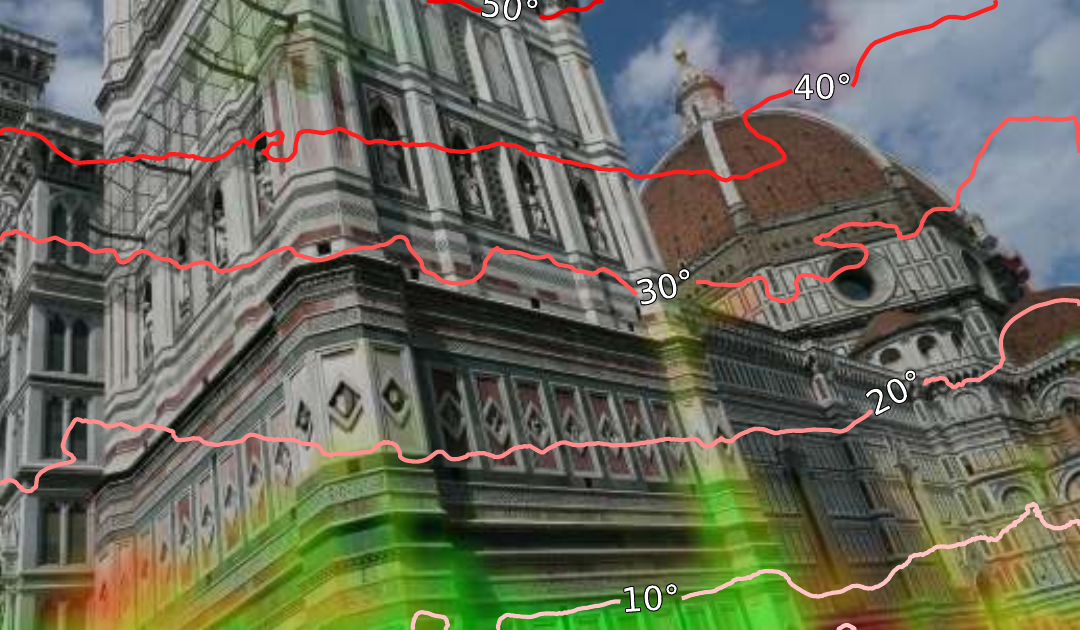}%
    \hspace{\pwidth}%
    \includegraphics[width=\iwidth]{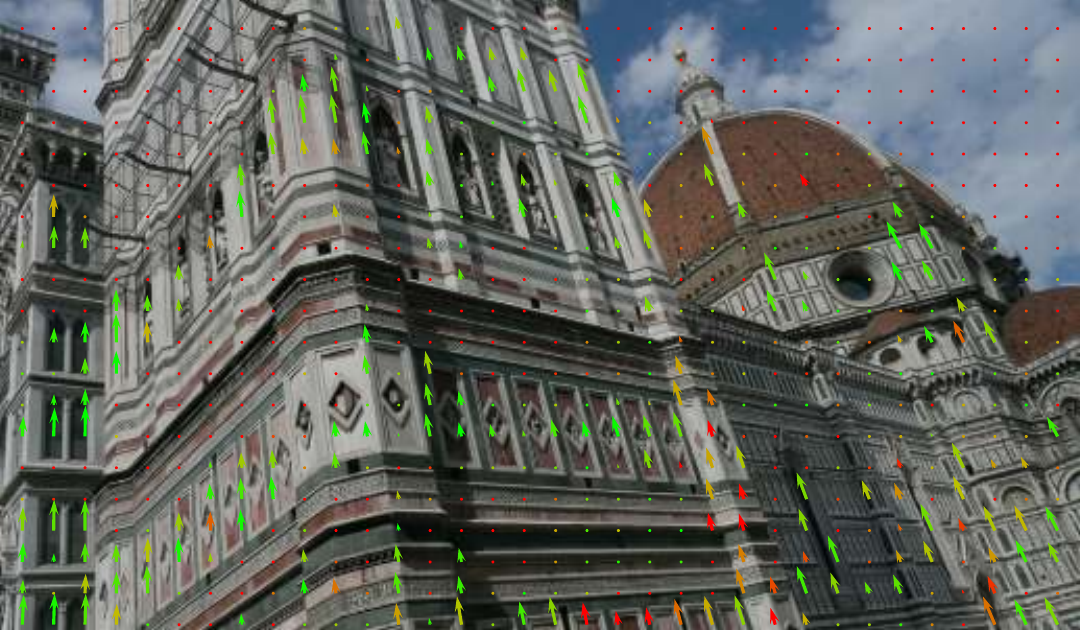}%
    
    \includegraphics[width=\iwidth]{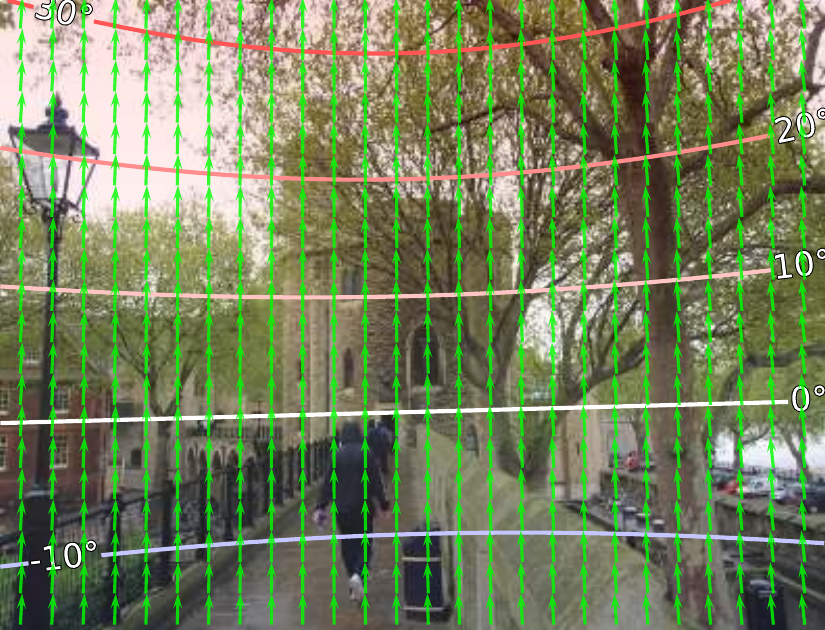}%
    \hspace{\pwidth}%
    \includegraphics[width=\iwidth]{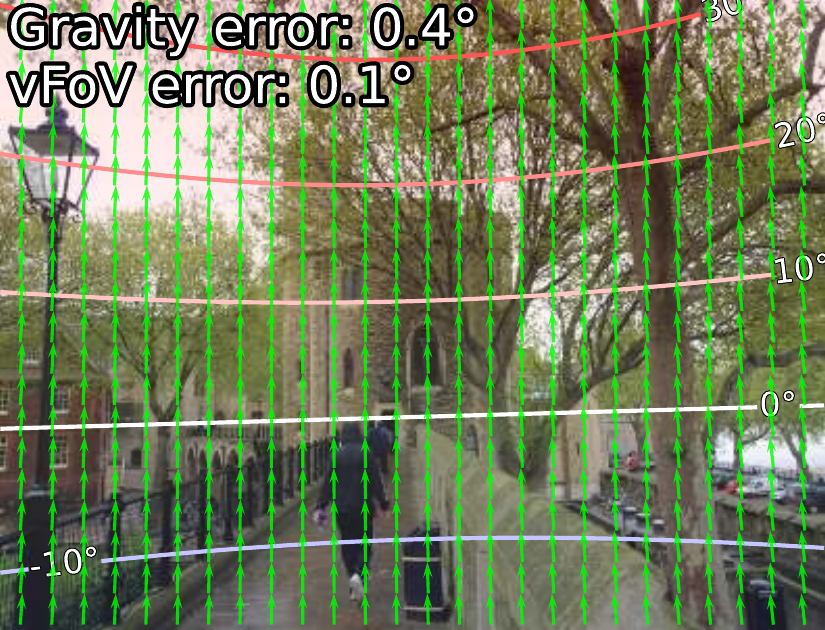}%
    \hspace{\pwidth}%
    \includegraphics[width=\iwidth]{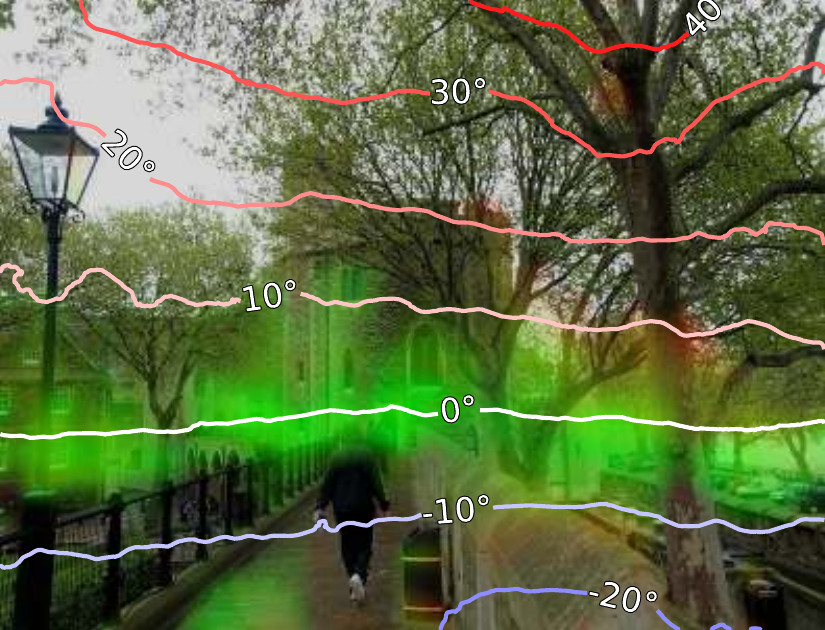}%
    \hspace{\pwidth}%
    \includegraphics[width=\iwidth]{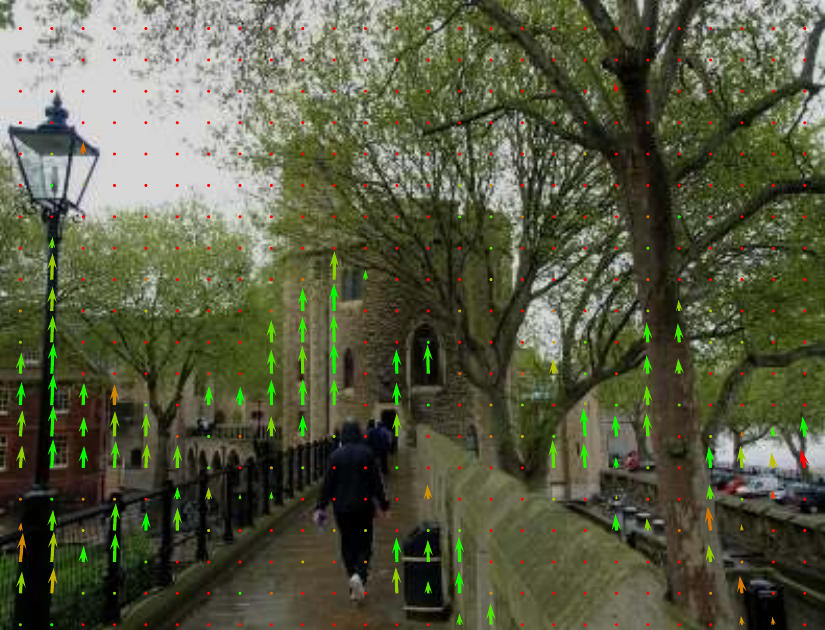}%

    \caption{\textbf{Qualitative examples from the MegaDepth dataset~\cite{li2018megadepth}.}}%
    \label{fig:megadepth2k}%
\end{figure}

\begin{figure}[p]
    \centering
    \def\ncols{4}
    \setlength{\pwidth}{0.005\linewidth}
    \setlength{\iwidth}{\dimexpr(0.999\linewidth - \ncols\pwidth + \pwidth)/\ncols \relax}
    \setlength{\lamarwidth}{\dimexpr(0.999\linewidth - 8\pwidth + \pwidth)/8 \relax}
    
    \begin{minipage}[b]{\iwidth}
    \centering{\footnotesize a) ground-truth}
    \end{minipage}%
    \hspace{\pwidth}%
    \begin{minipage}[b]{\iwidth}
    \centering{\footnotesize b) final prediction}
    \end{minipage}%
    \hspace{\pwidth}%
    \begin{minipage}[b]{\iwidth}
    \centering{\footnotesize c) observed latitude}
    \end{minipage}%
    \hspace{\pwidth}%
    \begin{minipage}[b]{\iwidth}
    \centering{\footnotesize d) observed up-vect.}
    \end{minipage}%
    
    \includegraphics[width=\lamarwidth]{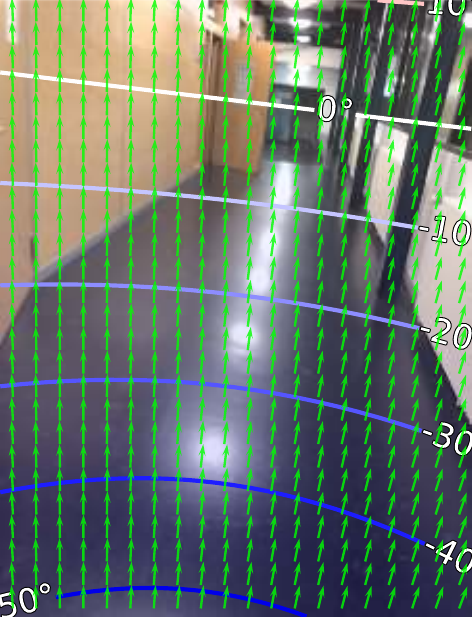}%
    \hspace{\pwidth}%
    \includegraphics[width=\lamarwidth]{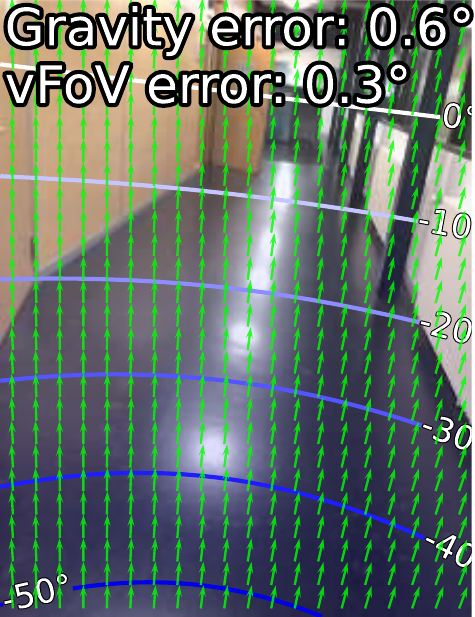}%
    \hspace{\pwidth}%
    \includegraphics[width=\lamarwidth]{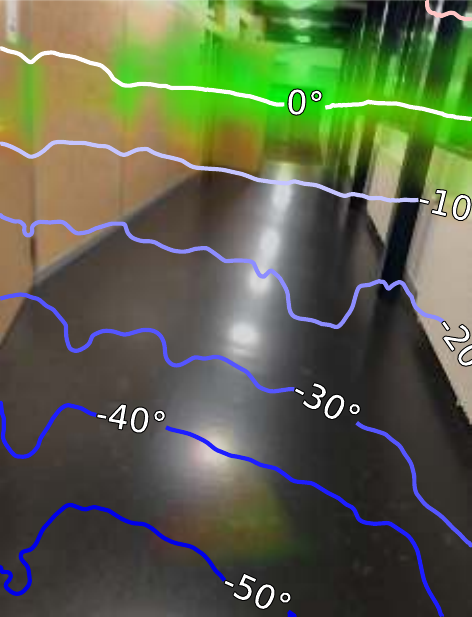}%
    \hspace{\pwidth}%
    \includegraphics[width=\lamarwidth]{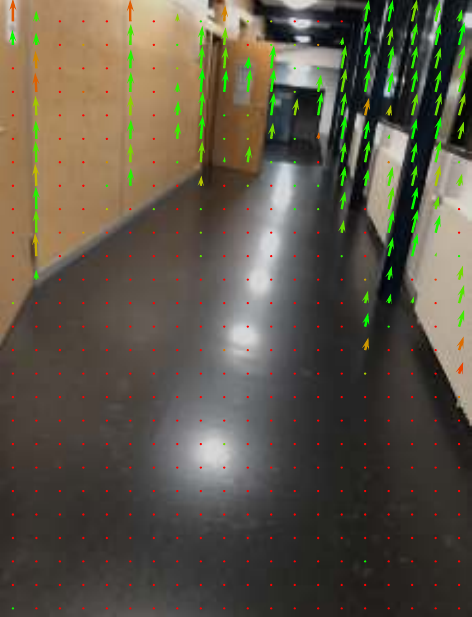}%
    \hspace{\pwidth}%
    \includegraphics[width=\lamarwidth]{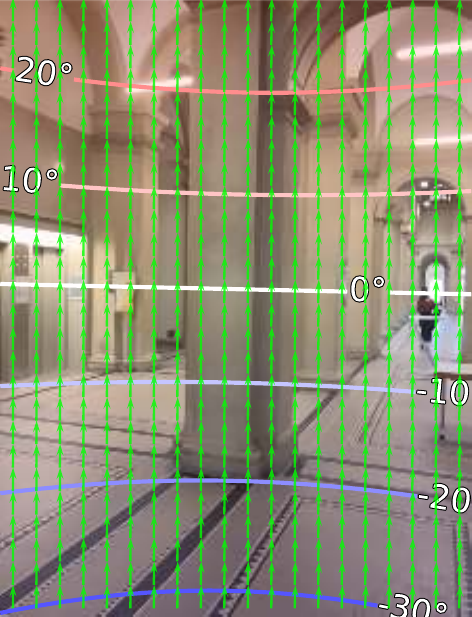}%
    \hspace{\pwidth}%
    \includegraphics[width=\lamarwidth]{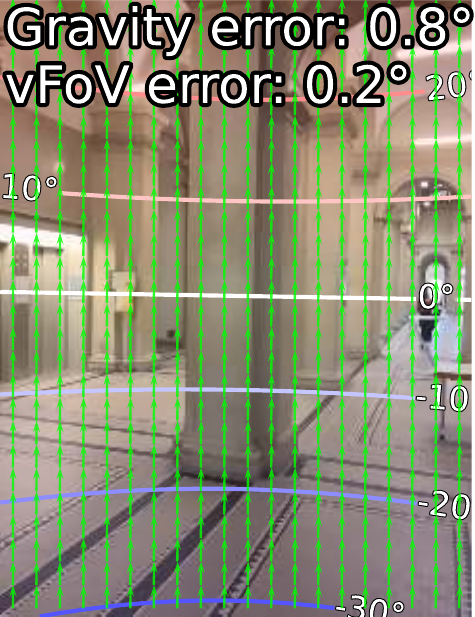}%
    \hspace{\pwidth}%
    \includegraphics[width=\lamarwidth]{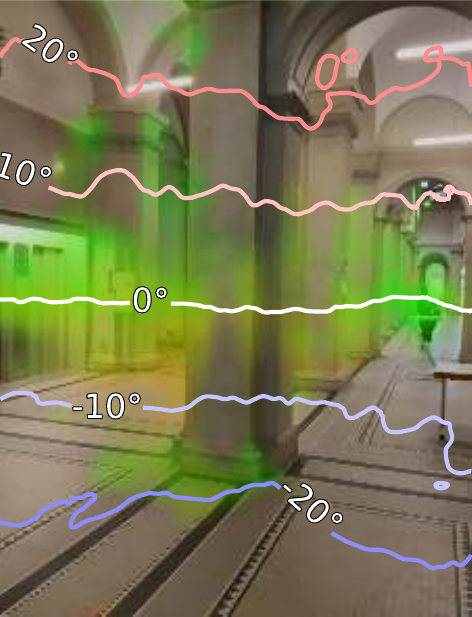}%
    \hspace{\pwidth}%
    \includegraphics[width=\lamarwidth]{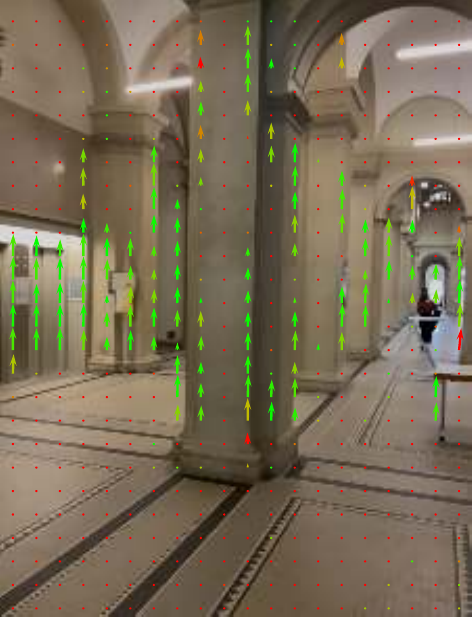}%
    
    \includegraphics[width=\iwidth]{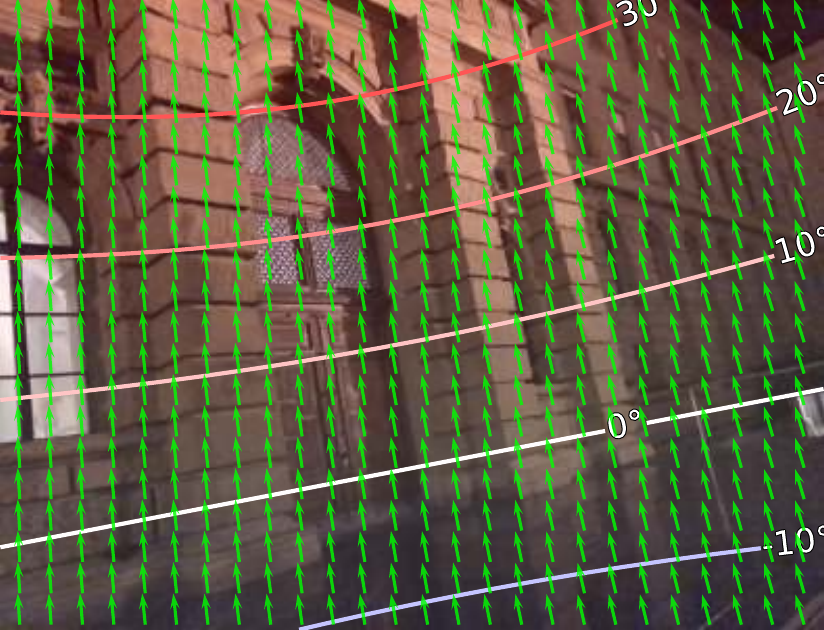}%
    \hspace{\pwidth}%
    \includegraphics[width=\iwidth]{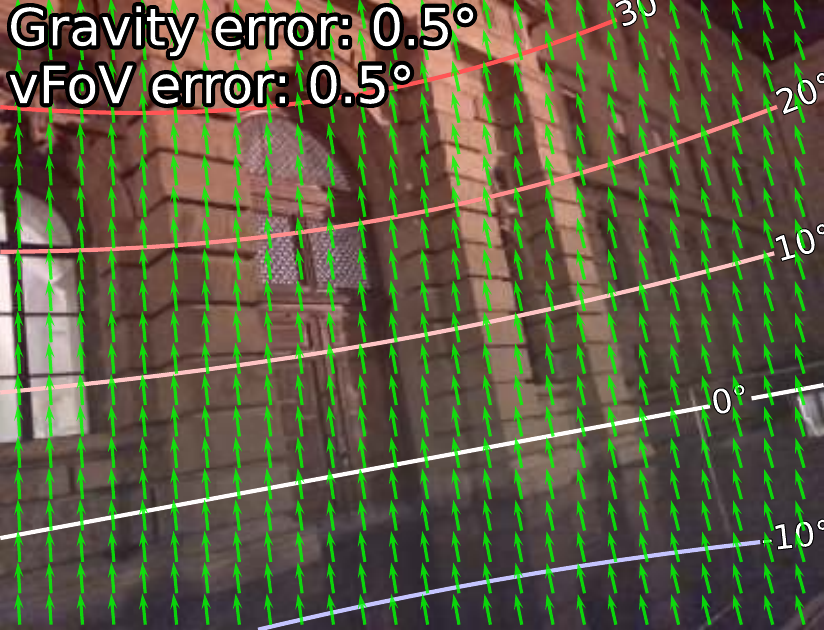}%
    \hspace{\pwidth}%
    \includegraphics[width=\iwidth]{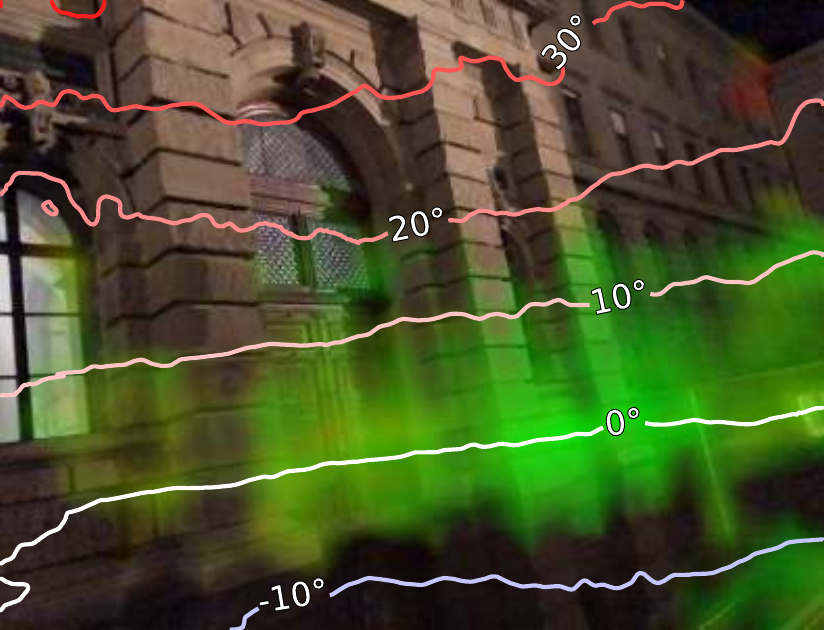}%
    \hspace{\pwidth}%
    \includegraphics[width=\iwidth]{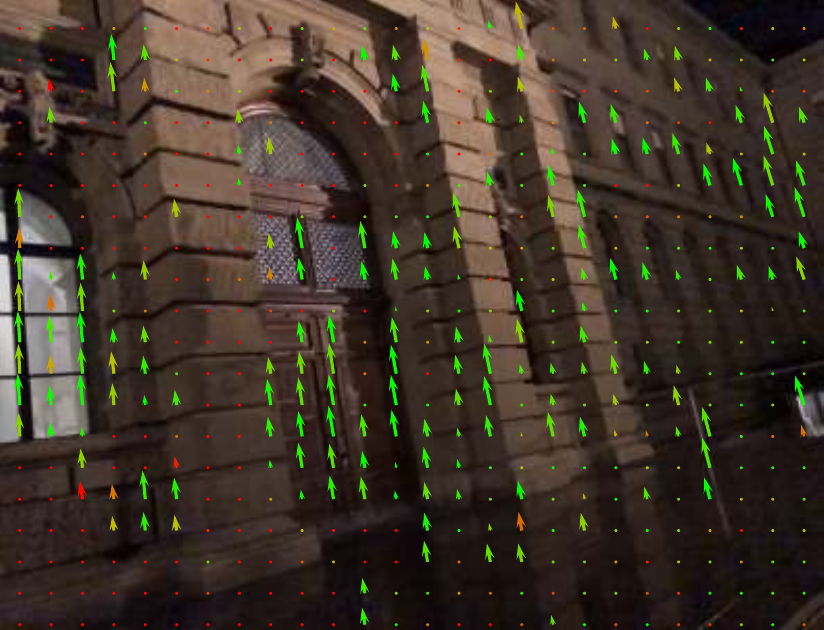}%
    
    \includegraphics[width=\lamarwidth]{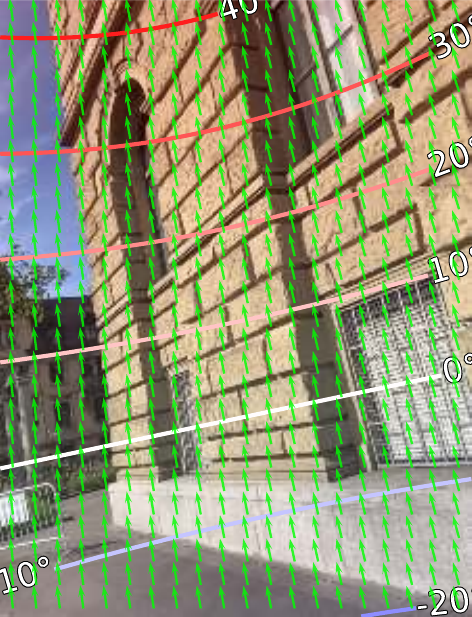}%
    \hspace{\pwidth}%
    \includegraphics[width=\lamarwidth]{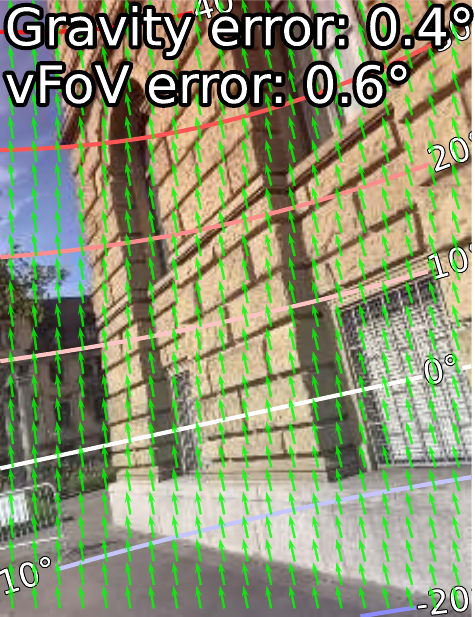}%
    \hspace{\pwidth}%
    \includegraphics[width=\lamarwidth]{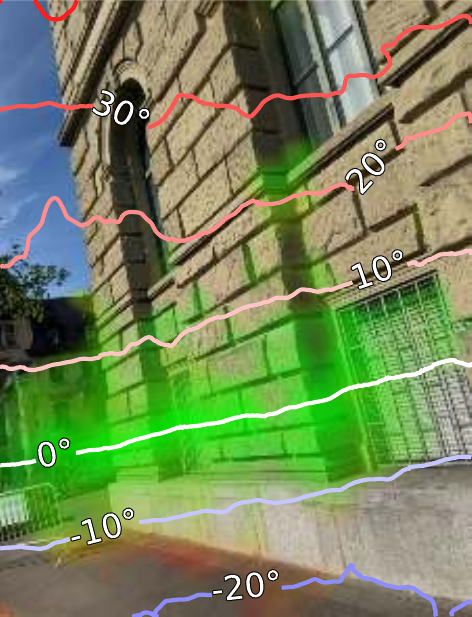}%
    \hspace{\pwidth}%
    \includegraphics[width=\lamarwidth]{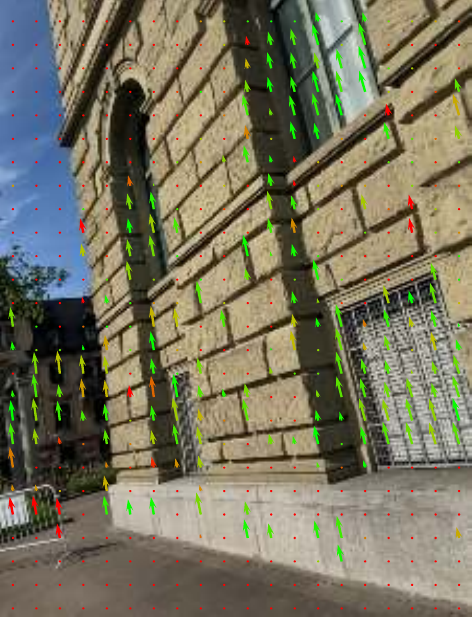}%
    \hspace{\pwidth}%
    \includegraphics[width=\lamarwidth]{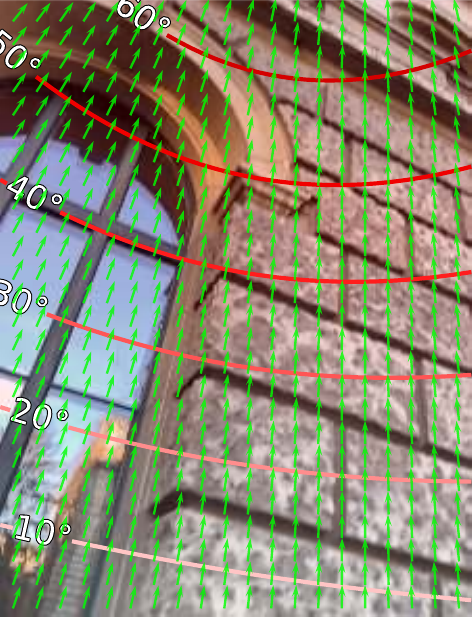}%
    \hspace{\pwidth}%
    \includegraphics[width=\lamarwidth]{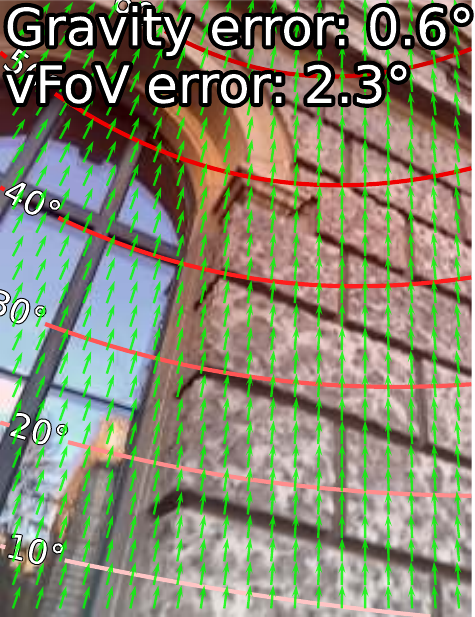}%
    \hspace{\pwidth}%
    \includegraphics[width=\lamarwidth]{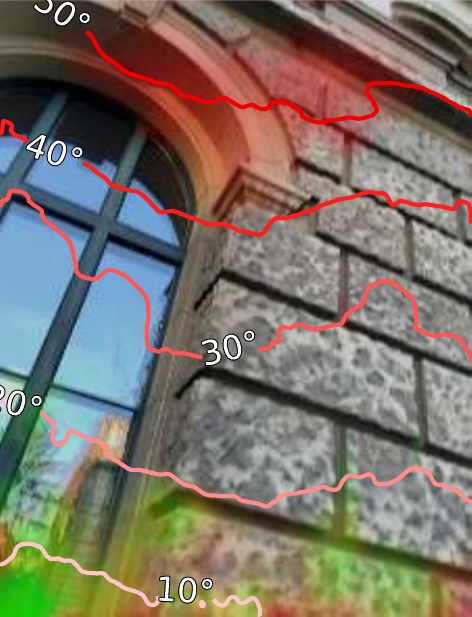}%
    \hspace{\pwidth}%
    \includegraphics[width=\lamarwidth]{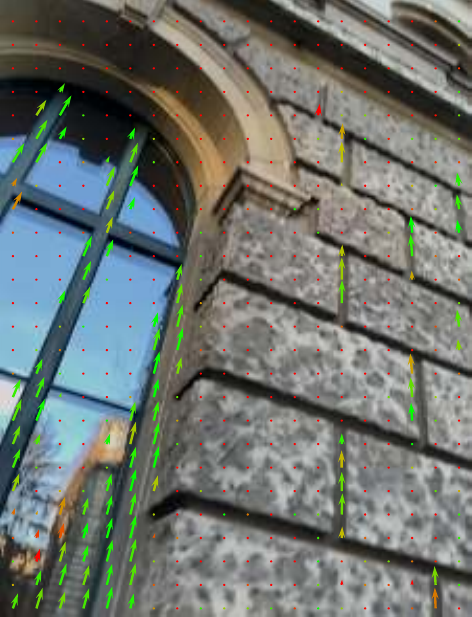}%
    
    \includegraphics[width=\lamarwidth]{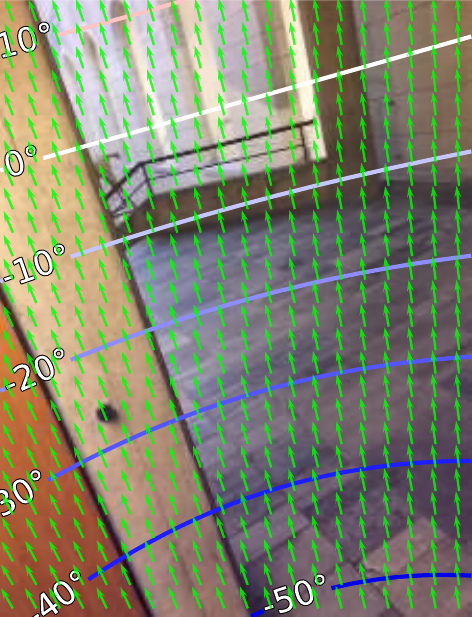}%
    \hspace{\pwidth}%
    \includegraphics[width=\lamarwidth]{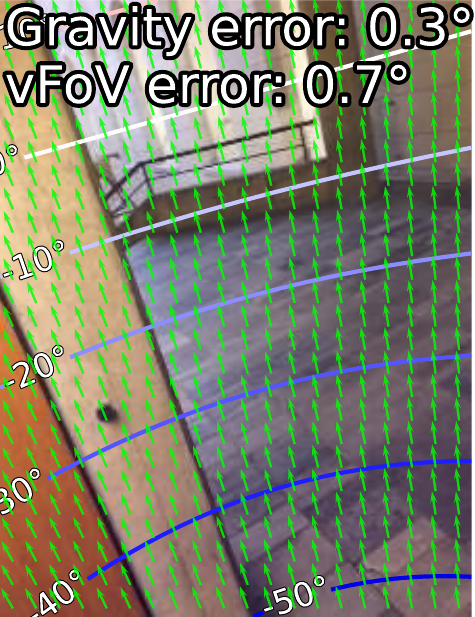}%
    \hspace{\pwidth}%
    \includegraphics[width=\lamarwidth]{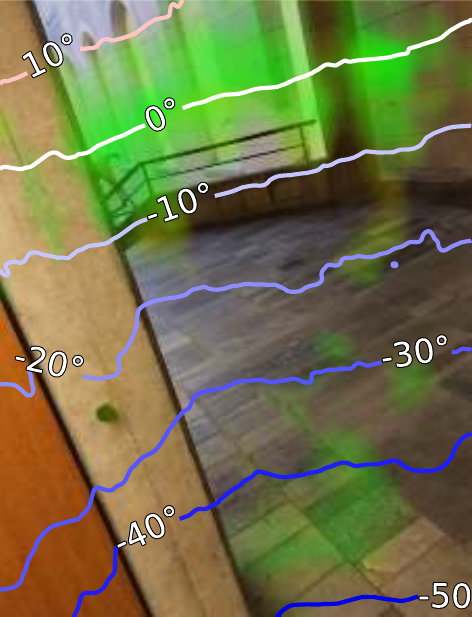}%
    \hspace{\pwidth}%
    \includegraphics[width=\lamarwidth]{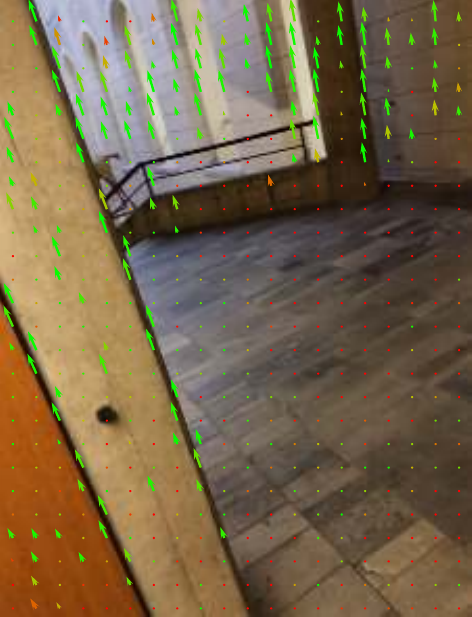}%
    \hspace{\pwidth}%
    \includegraphics[width=\lamarwidth]{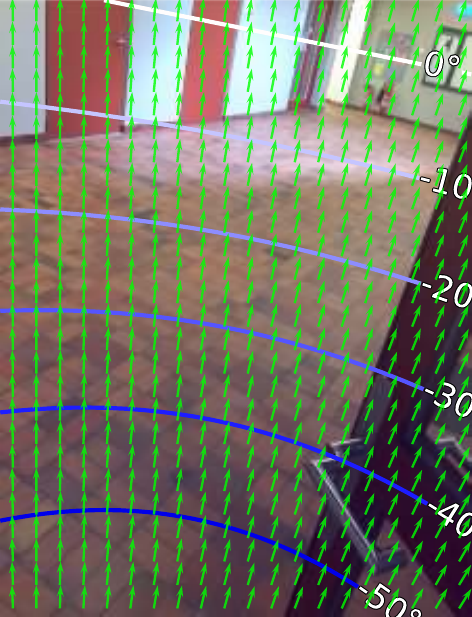}%
    \hspace{\pwidth}%
    \includegraphics[width=\lamarwidth]{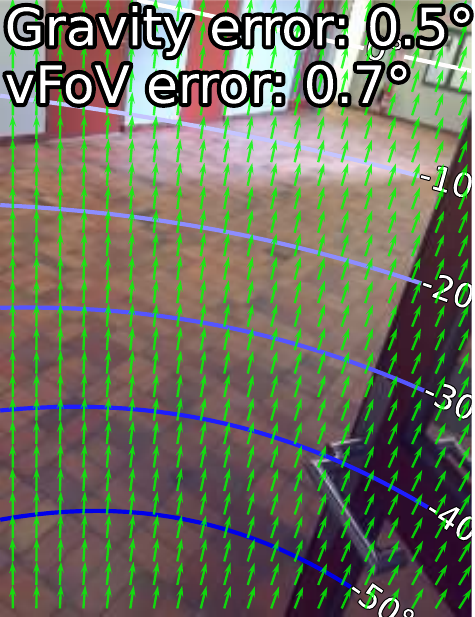}%
    \hspace{\pwidth}%
    \includegraphics[width=\lamarwidth]{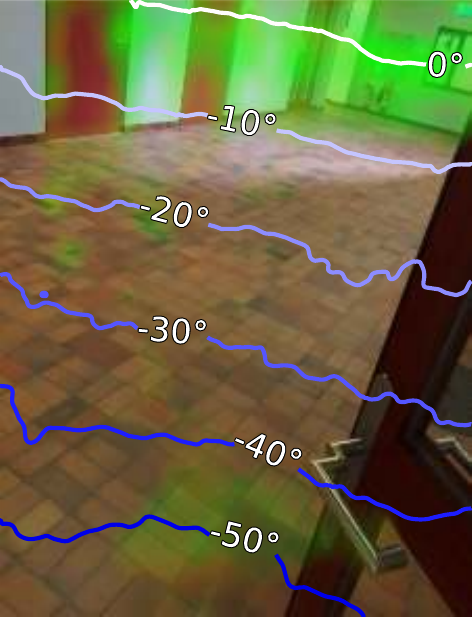}%
    \hspace{\pwidth}%
    \includegraphics[width=\lamarwidth]{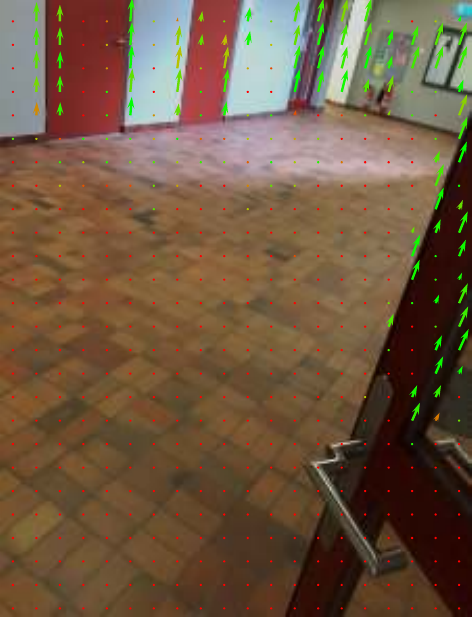}%
    
    \includegraphics[width=\lamarwidth]{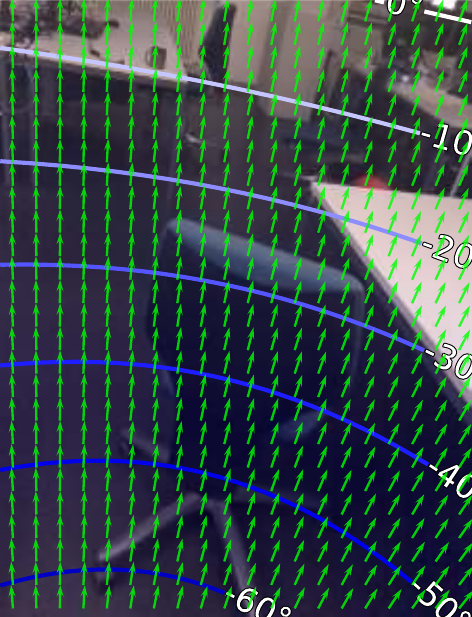}%
    \hspace{\pwidth}%
    \includegraphics[width=\lamarwidth]{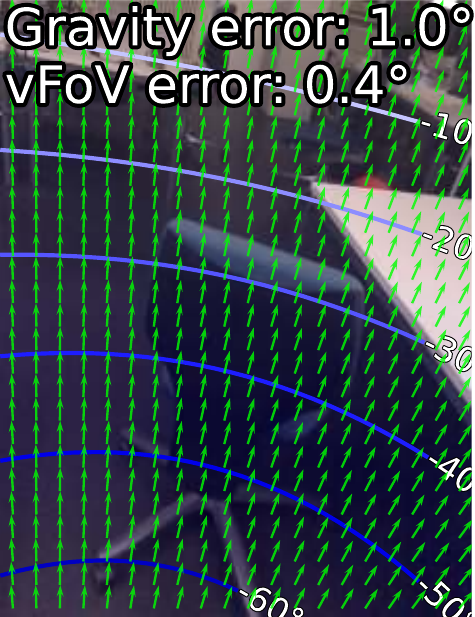}%
    \hspace{\pwidth}%
    \includegraphics[width=\lamarwidth]{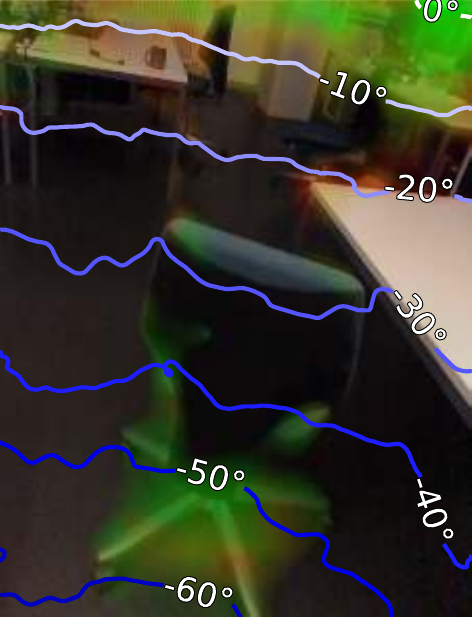}%
    \hspace{\pwidth}%
    \includegraphics[width=\lamarwidth]{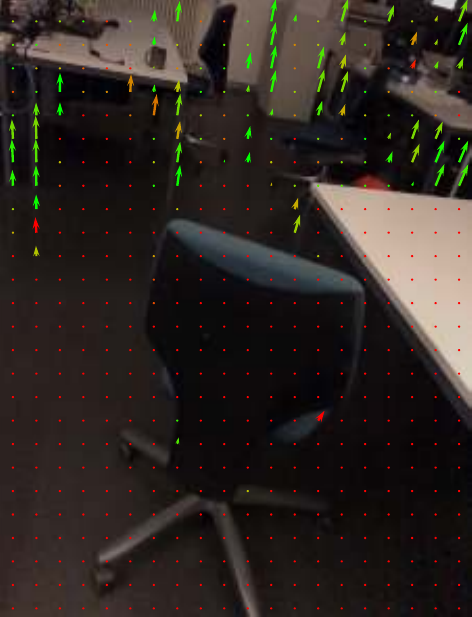}%
    \hspace{\pwidth}%
    \includegraphics[width=\lamarwidth]{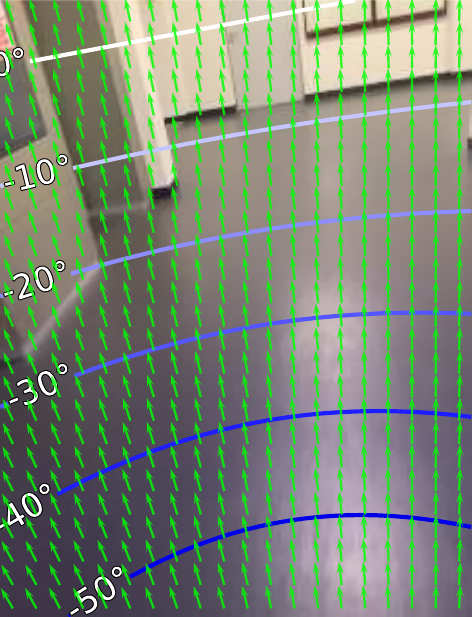}%
    \hspace{\pwidth}%
    \includegraphics[width=\lamarwidth]{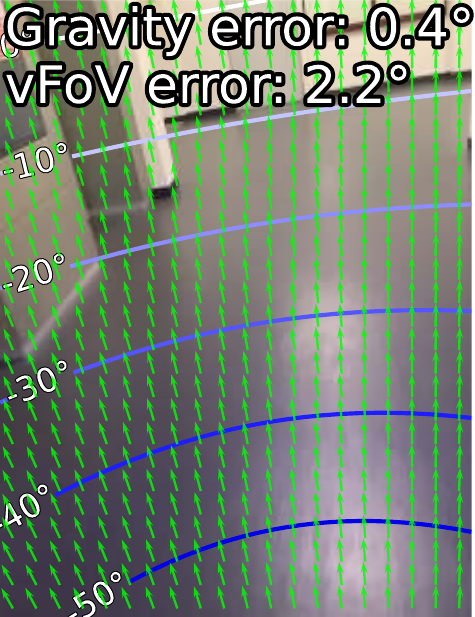}%
    \hspace{\pwidth}%
    \includegraphics[width=\lamarwidth]{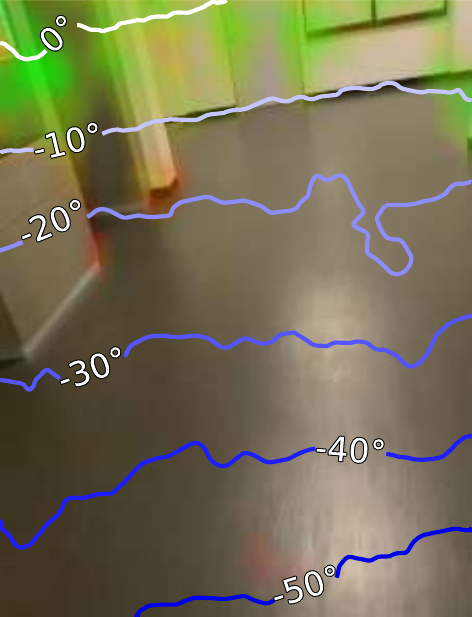}%
    \hspace{\pwidth}%
    \includegraphics[width=\lamarwidth]{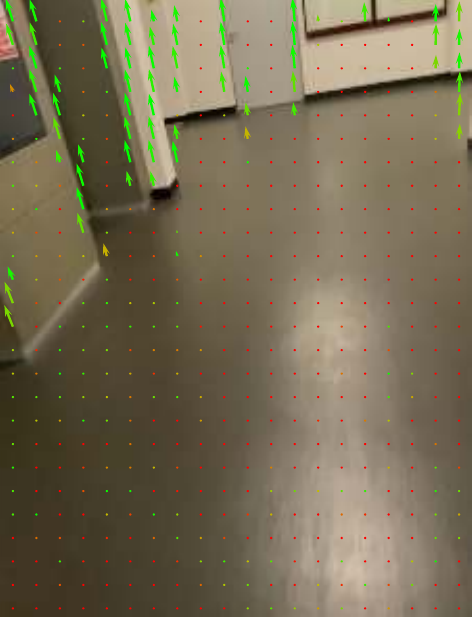}%
    
    \includegraphics[width=\iwidth]{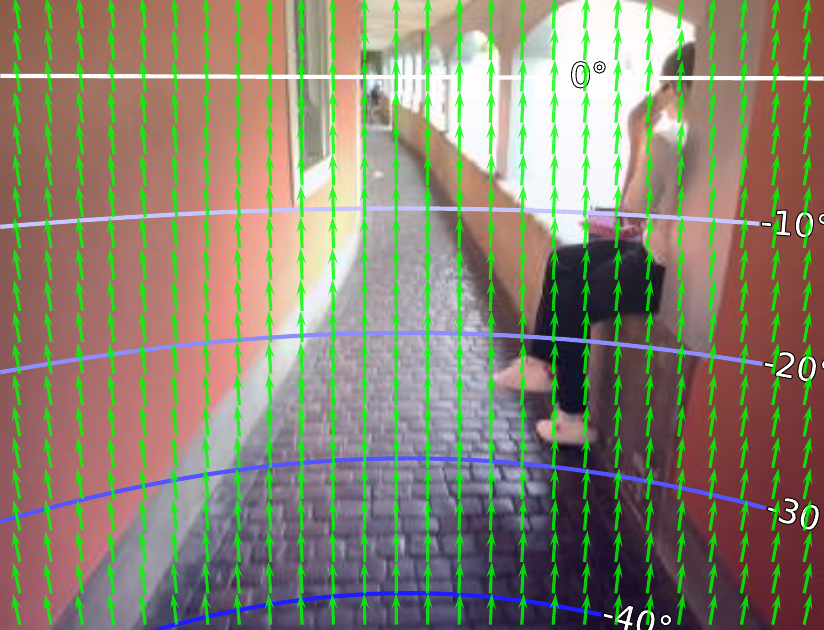}%
    \hspace{\pwidth}%
    \includegraphics[width=\iwidth]{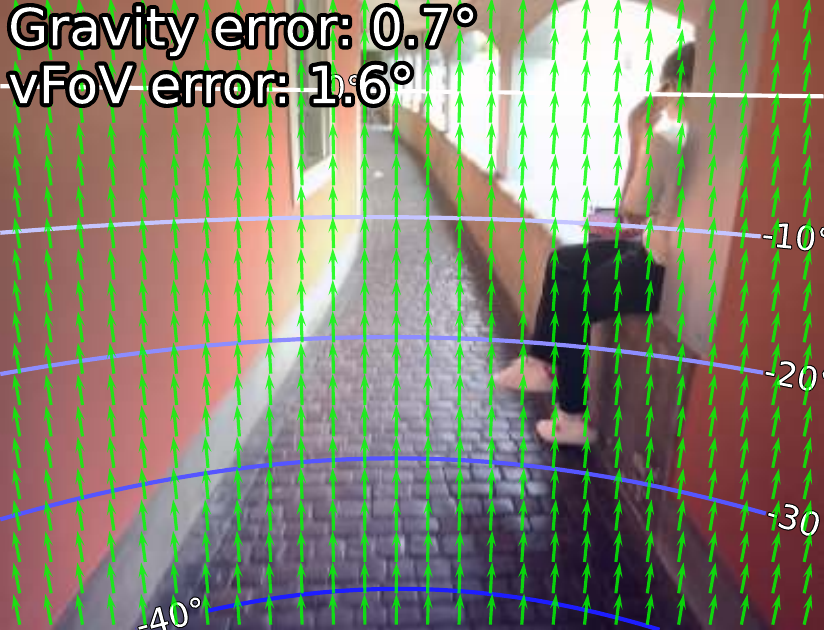}%
    \hspace{\pwidth}%
    \includegraphics[width=\iwidth]{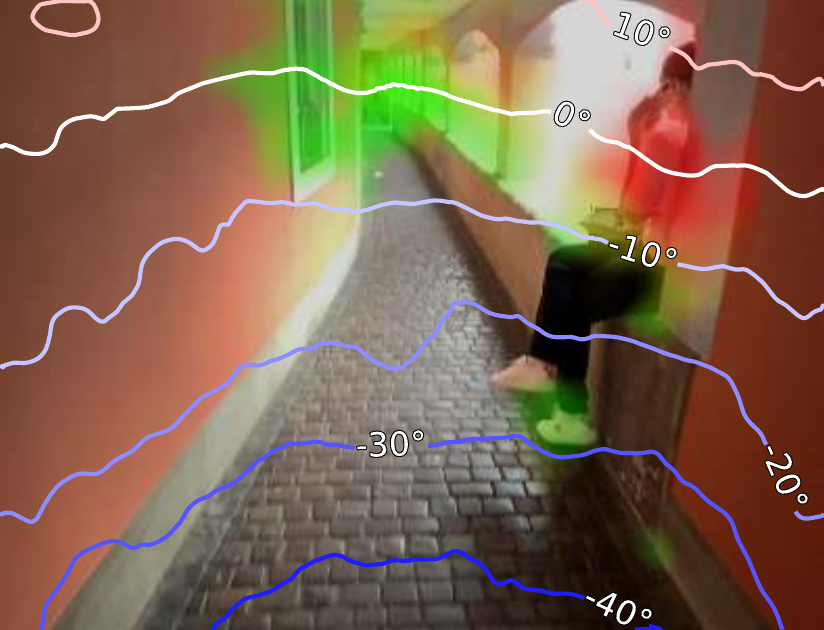}%
    \hspace{\pwidth}%
    \includegraphics[width=\iwidth]{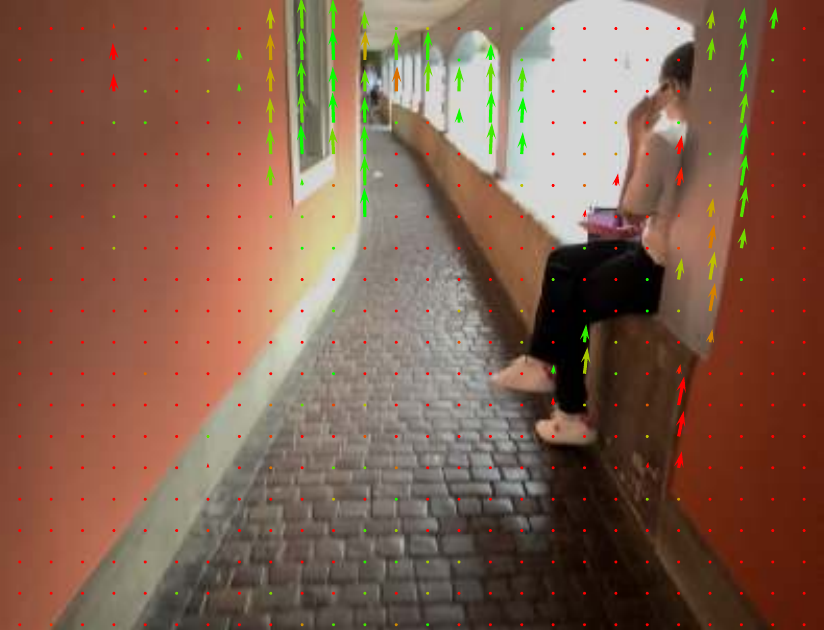}%
    
    \includegraphics[width=\lamarwidth]{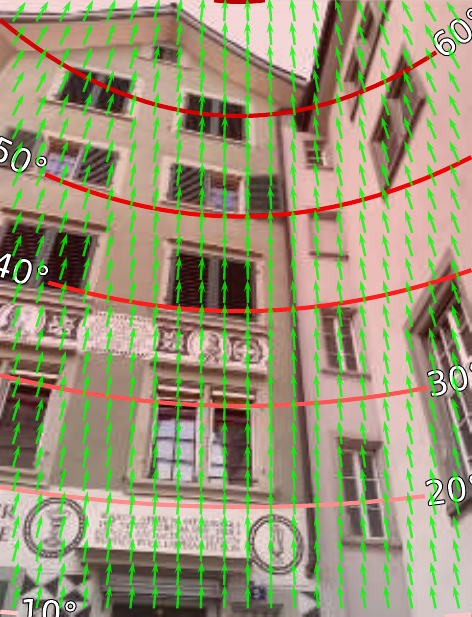}%
    \hspace{\pwidth}%
    \includegraphics[width=\lamarwidth]{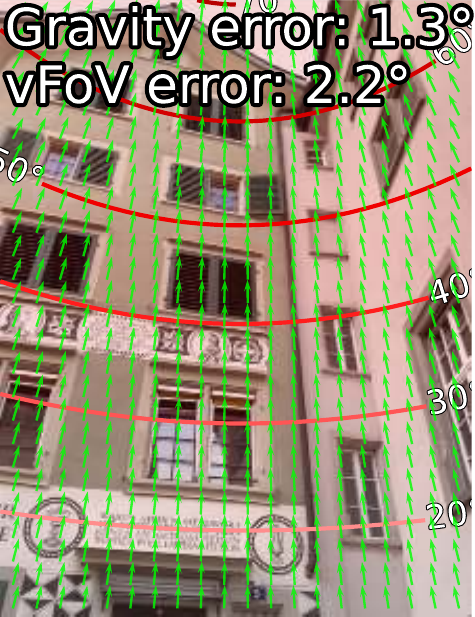}%
    \hspace{\pwidth}%
    \includegraphics[width=\lamarwidth]{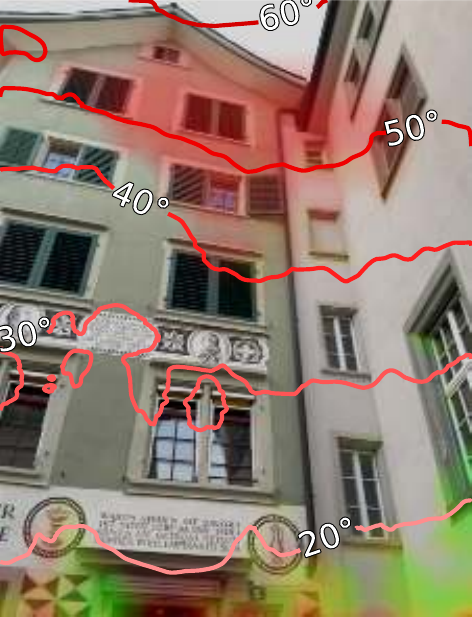}%
    \hspace{\pwidth}%
    \includegraphics[width=\lamarwidth]{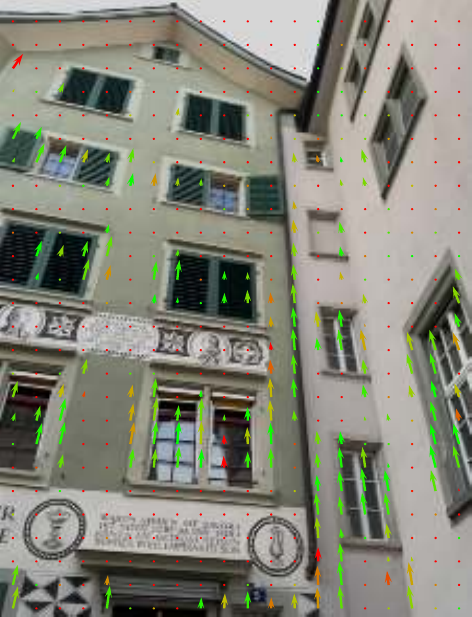}%
    \hspace{\pwidth}%
    \includegraphics[width=\lamarwidth]{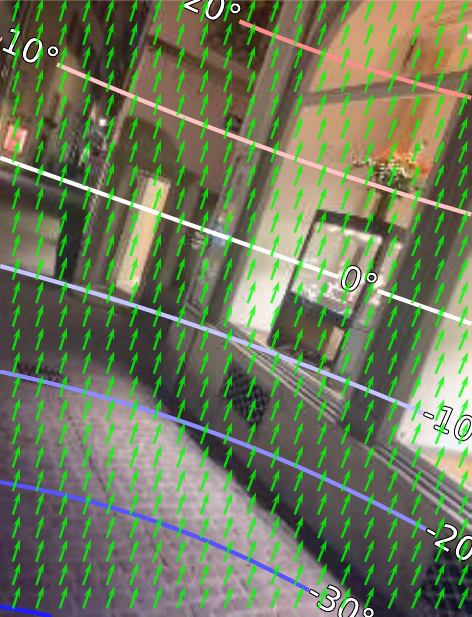}%
    \hspace{\pwidth}%
    \includegraphics[width=\lamarwidth]{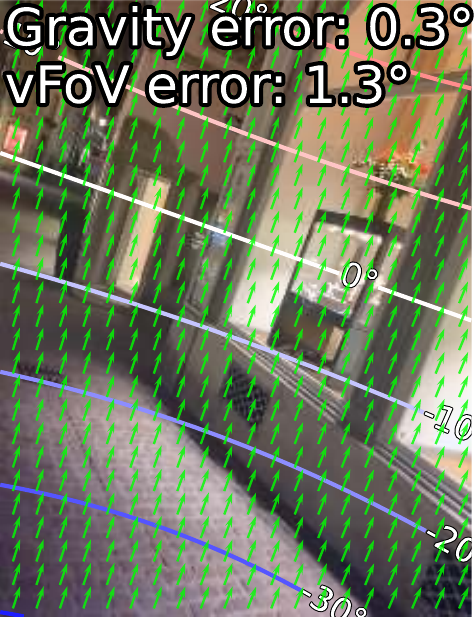}%
    \hspace{\pwidth}%
    \includegraphics[width=\lamarwidth]{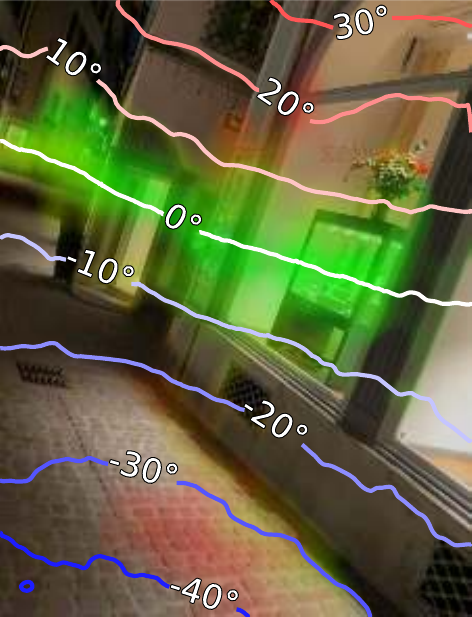}%
    \hspace{\pwidth}%
    \includegraphics[width=\lamarwidth]{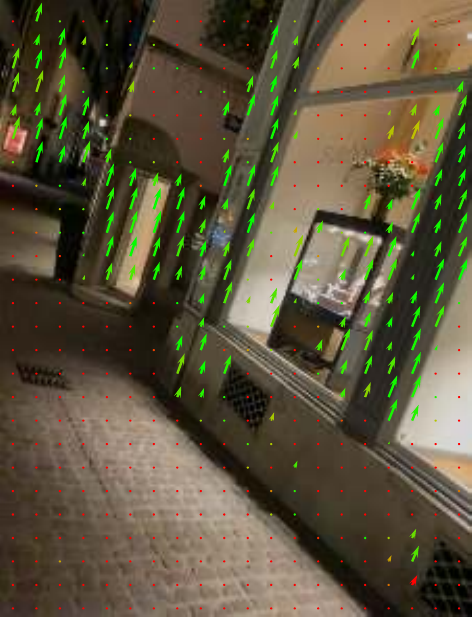}%

    \caption{\textbf{Qualitative examples from the LaMAR dataset~\cite{sarlin2022lamar}.}}%
    \label{fig:lamar2k}%
\end{figure}

\begin{figure}[p]
    \centering
    \def\ncols{4}
    \setlength{\pwidth}{0.005\linewidth}
    \setlength{\iwidth}{\dimexpr(0.999\linewidth - \ncols\pwidth + \pwidth)/\ncols \relax}
    \setlength{\lamarwidth}{\dimexpr(0.999\linewidth - 8\pwidth + \pwidth)/8 \relax}
    
    \begin{minipage}[b]{\iwidth}
    \centering{\footnotesize a) ground-truth}
    \end{minipage}%
    \hspace{\pwidth}%
    \begin{minipage}[b]{\iwidth}
    \centering{\footnotesize b) final prediction}
    \end{minipage}%
    \hspace{\pwidth}%
    \begin{minipage}[b]{\iwidth}
    \centering{\footnotesize c) observed latitude}
    \end{minipage}%
    \hspace{\pwidth}%
    \begin{minipage}[b]{\iwidth}
    \centering{\footnotesize d) observed up-vect.}
    \end{minipage}%
    
    \includegraphics[width=\iwidth]{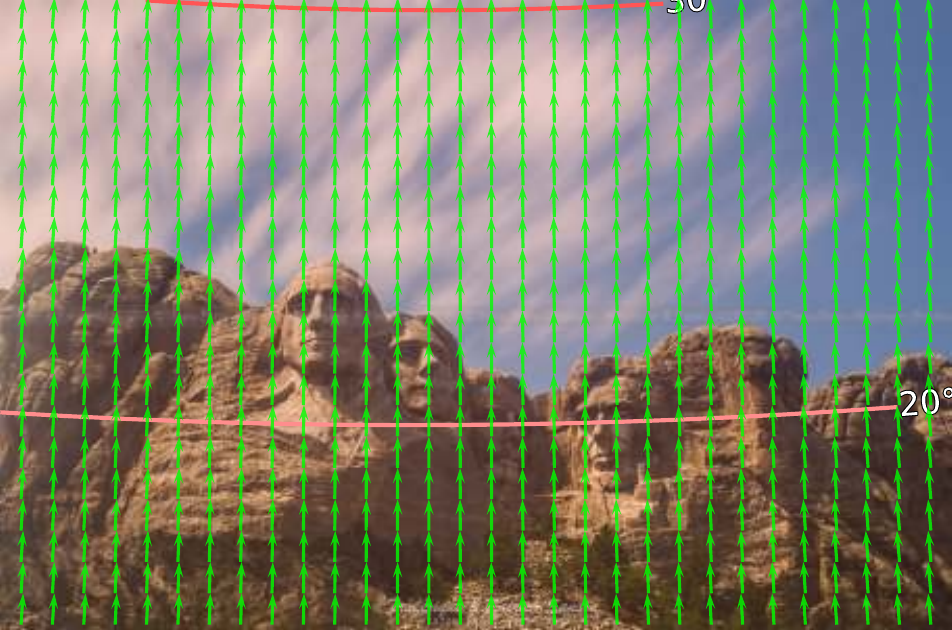}%
    \hspace{\pwidth}%
    \includegraphics[width=\iwidth]{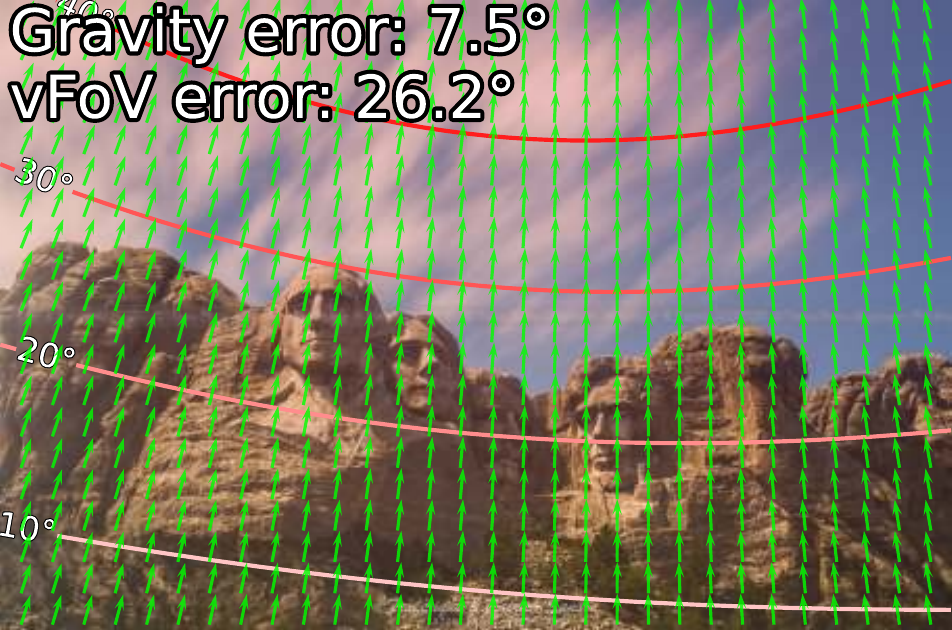}%
    \hspace{\pwidth}%
    \includegraphics[width=\iwidth]{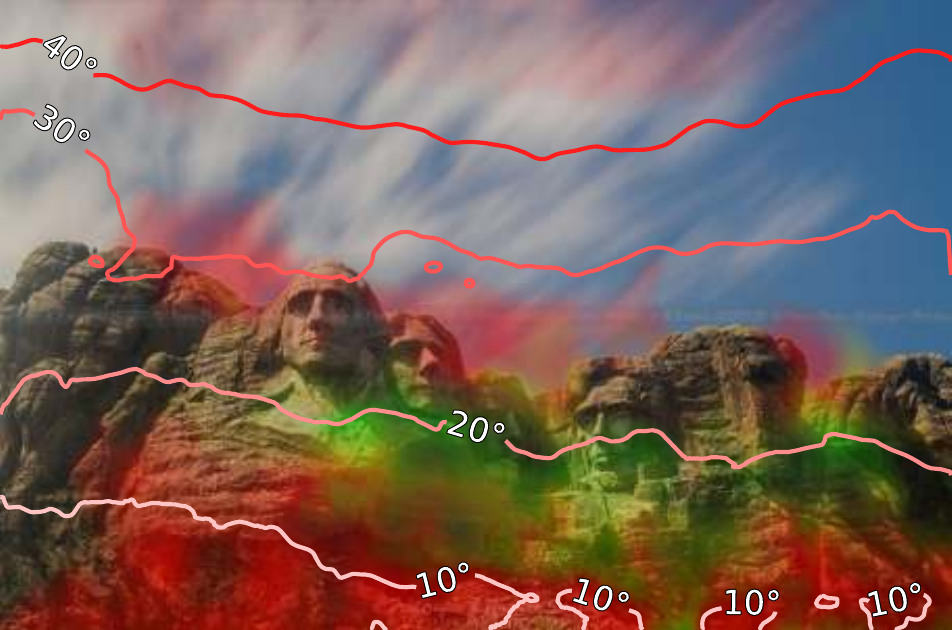}%
    \hspace{\pwidth}%
    \includegraphics[width=\iwidth]{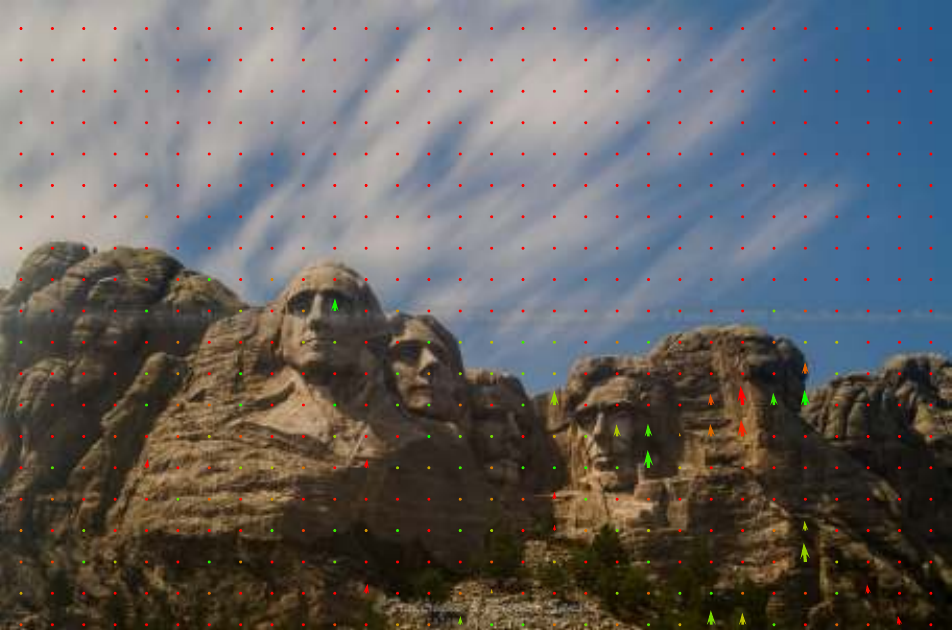}%
    
    \includegraphics[width=\iwidth]{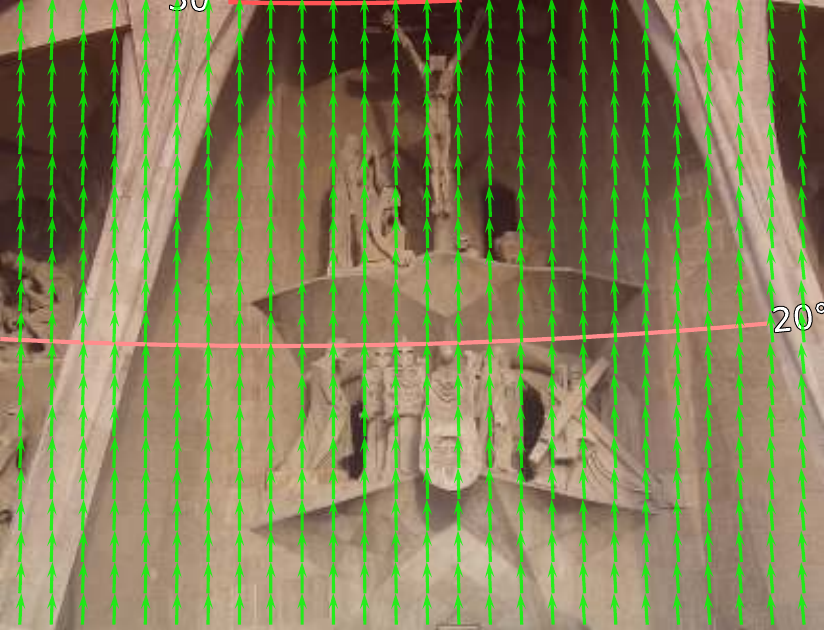}%
    \hspace{\pwidth}%
    \includegraphics[width=\iwidth]{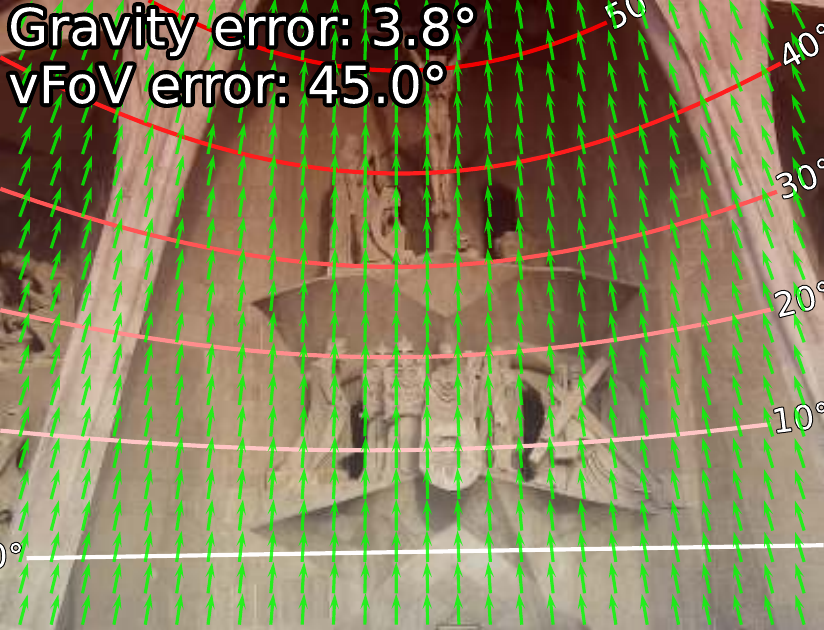}%
    \hspace{\pwidth}%
    \includegraphics[width=\iwidth]{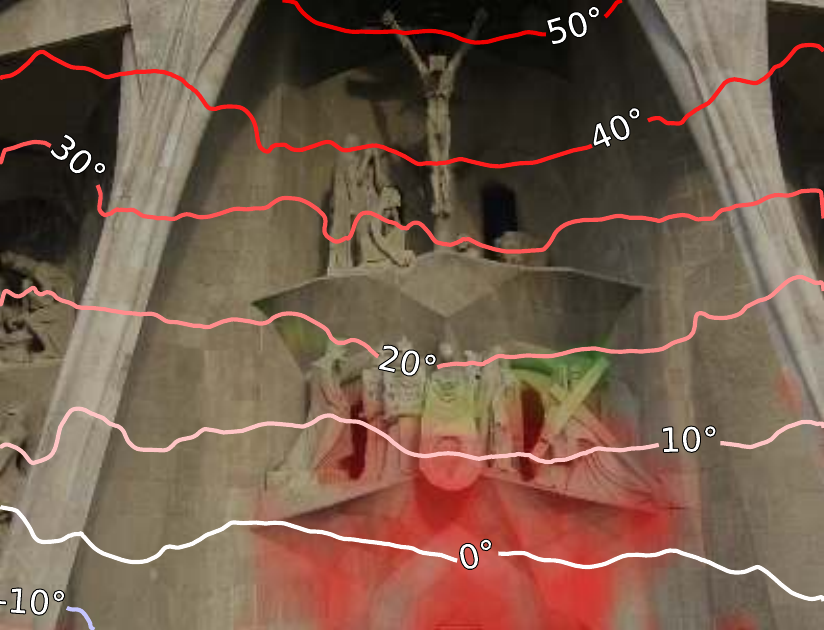}%
    \hspace{\pwidth}%
    \includegraphics[width=\iwidth]{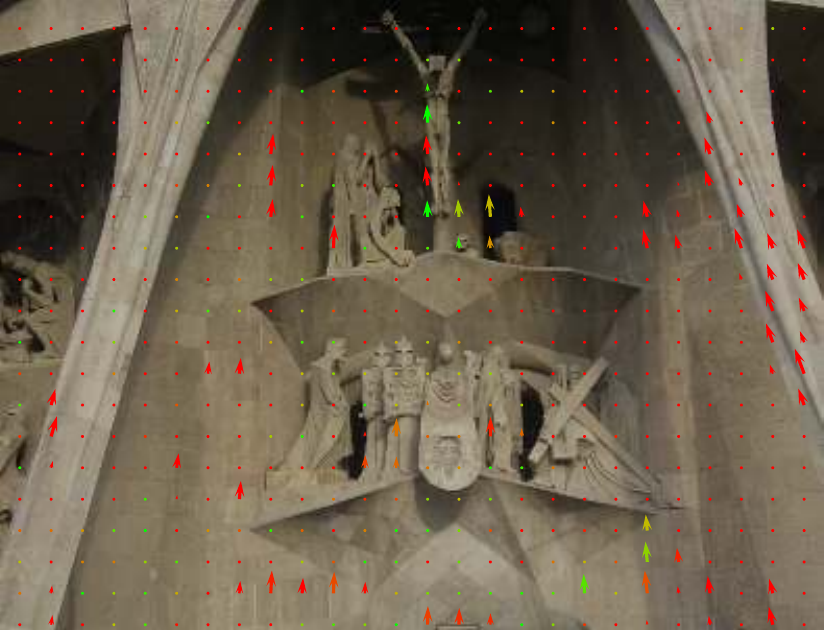}%
    
    \includegraphics[width=\iwidth]{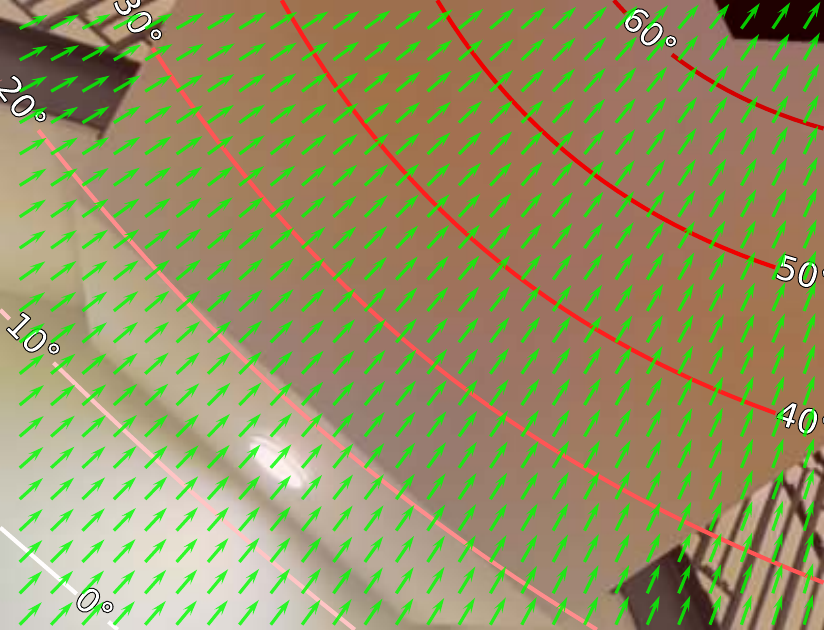}%
    \hspace{\pwidth}%
    \includegraphics[width=\iwidth]{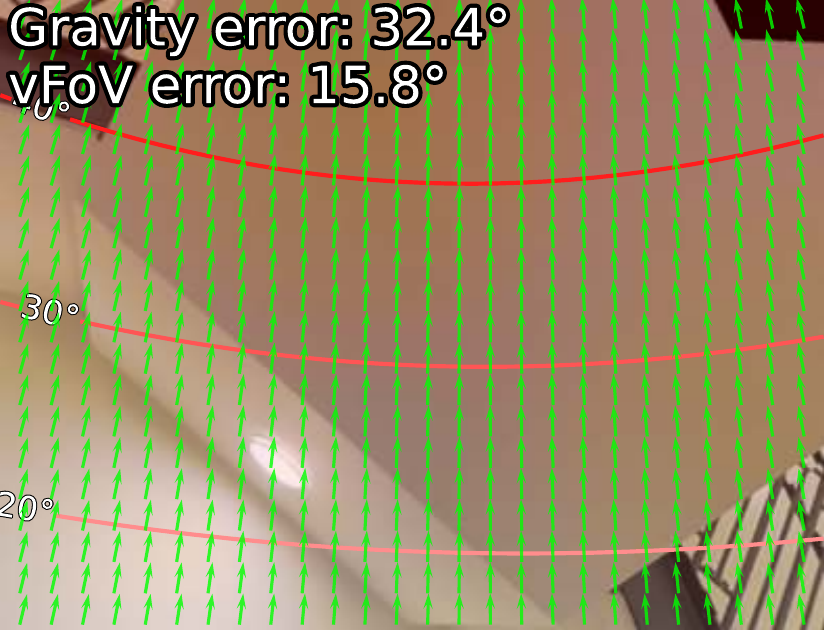}%
    \hspace{\pwidth}%
    \includegraphics[width=\iwidth]{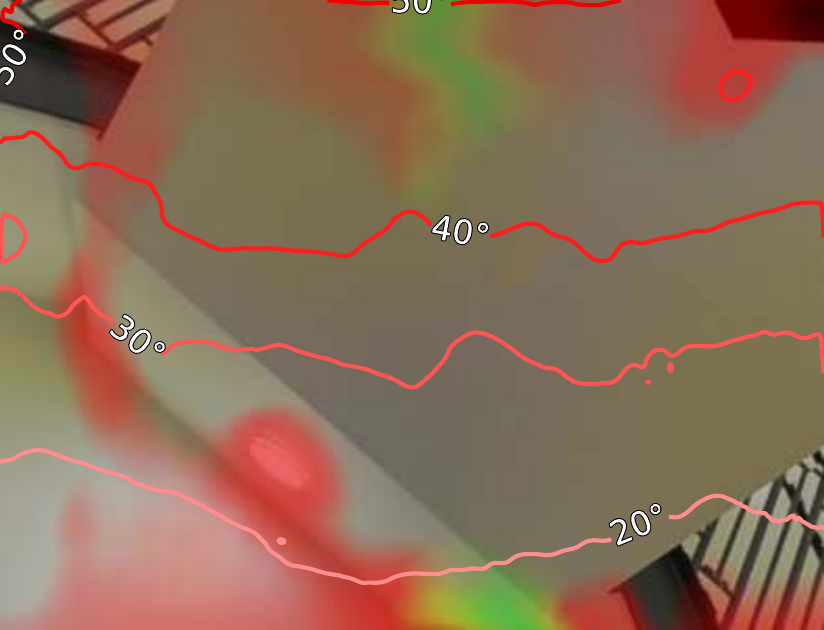}%
    \hspace{\pwidth}%
    \includegraphics[width=\iwidth]{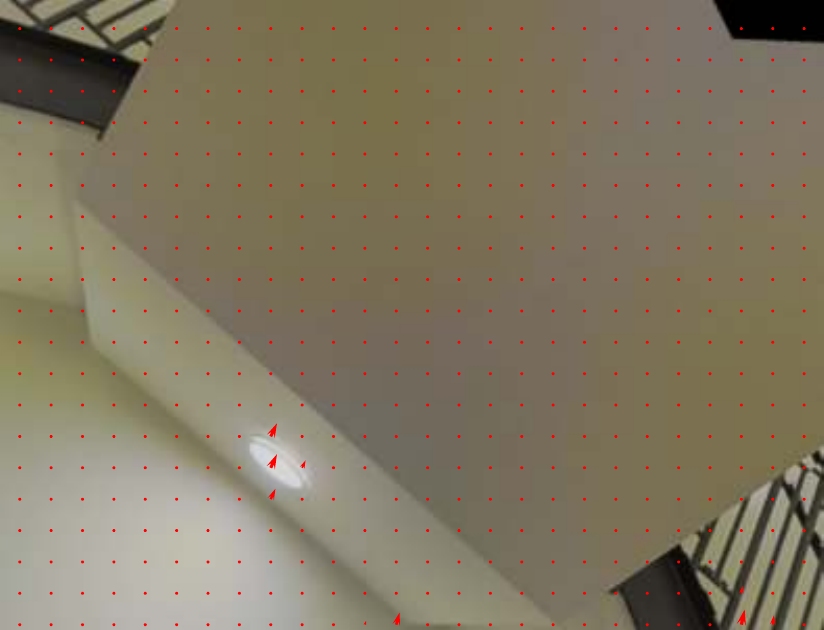}%
    
    \includegraphics[width=\iwidth]{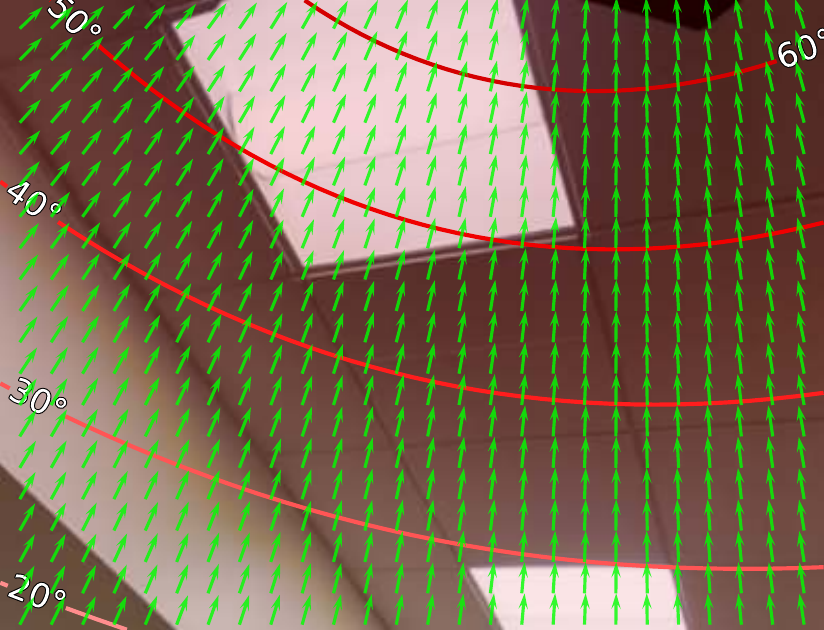}%
    \hspace{\pwidth}%
    \includegraphics[width=\iwidth]{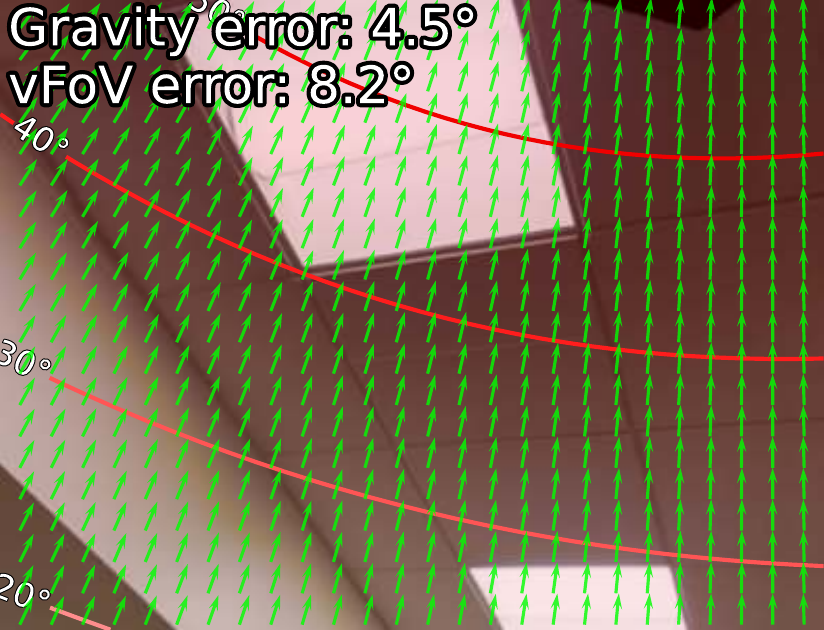}%
    \hspace{\pwidth}%
    \includegraphics[width=\iwidth]{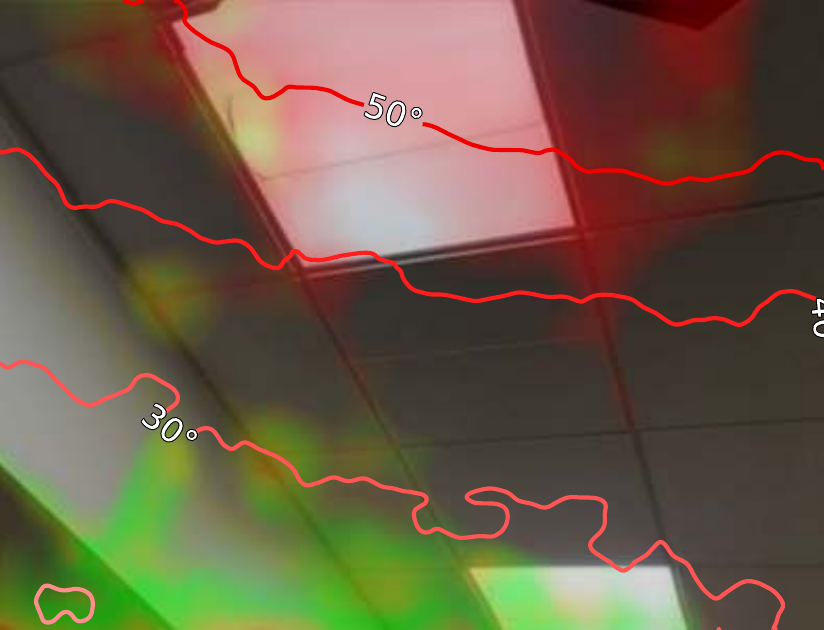}%
    \hspace{\pwidth}%
    \includegraphics[width=\iwidth]{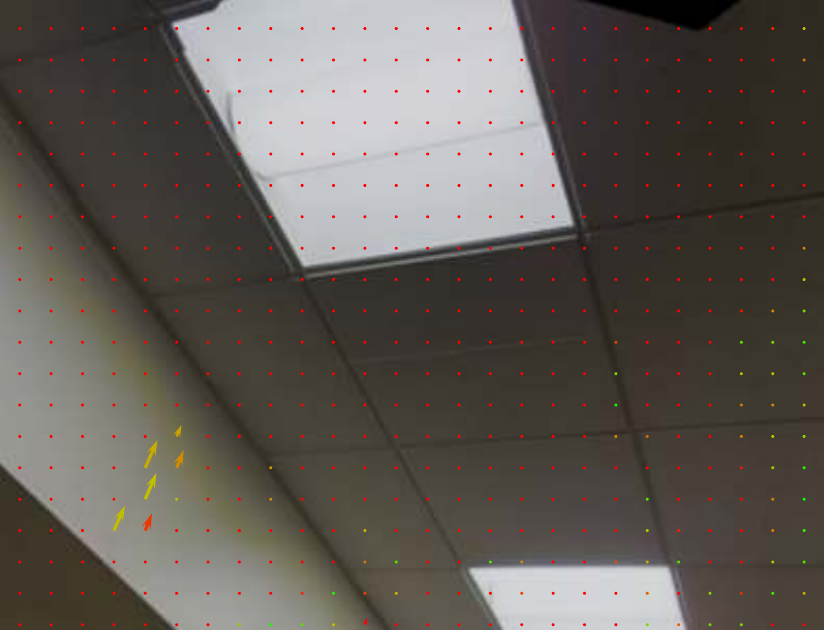}%
    
    \includegraphics[width=\iwidth]{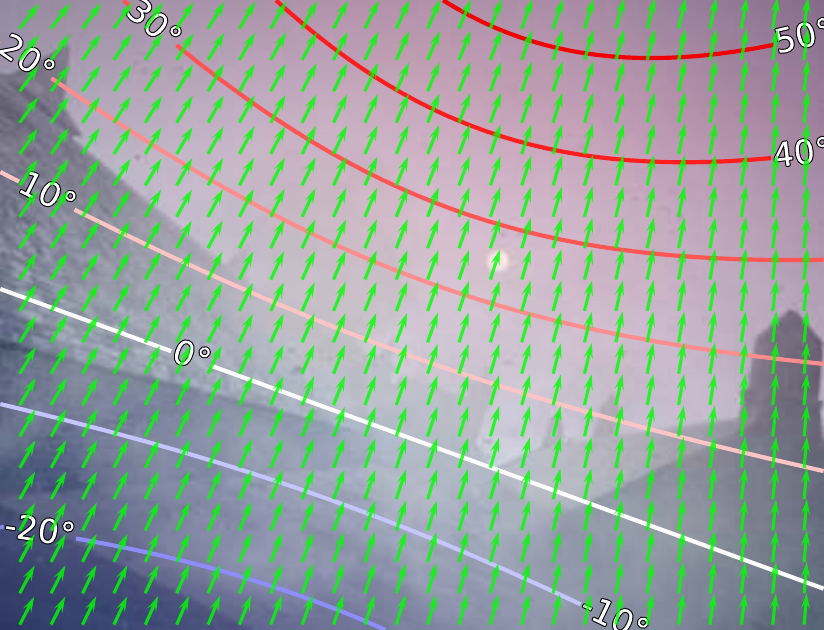}%
    \hspace{\pwidth}%
    \includegraphics[width=\iwidth]{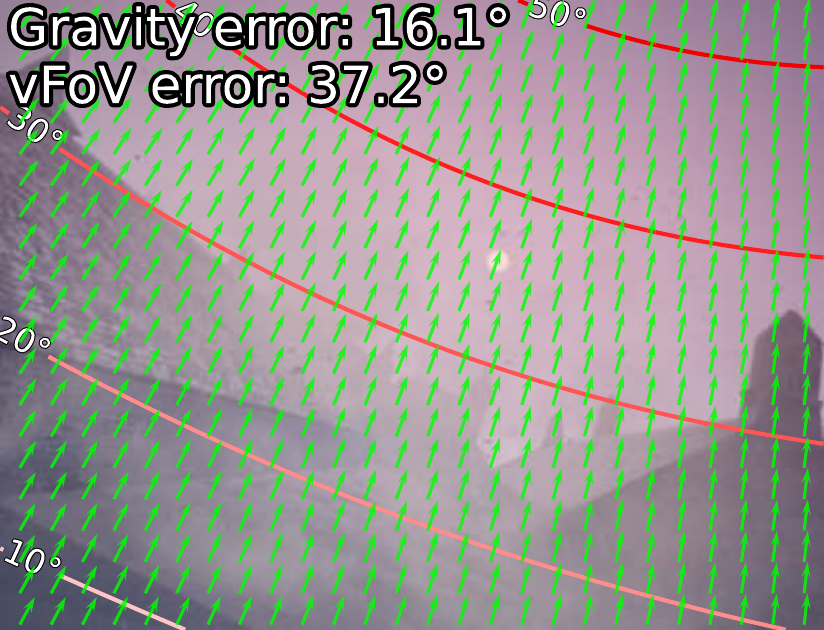}%
    \hspace{\pwidth}%
    \includegraphics[width=\iwidth]{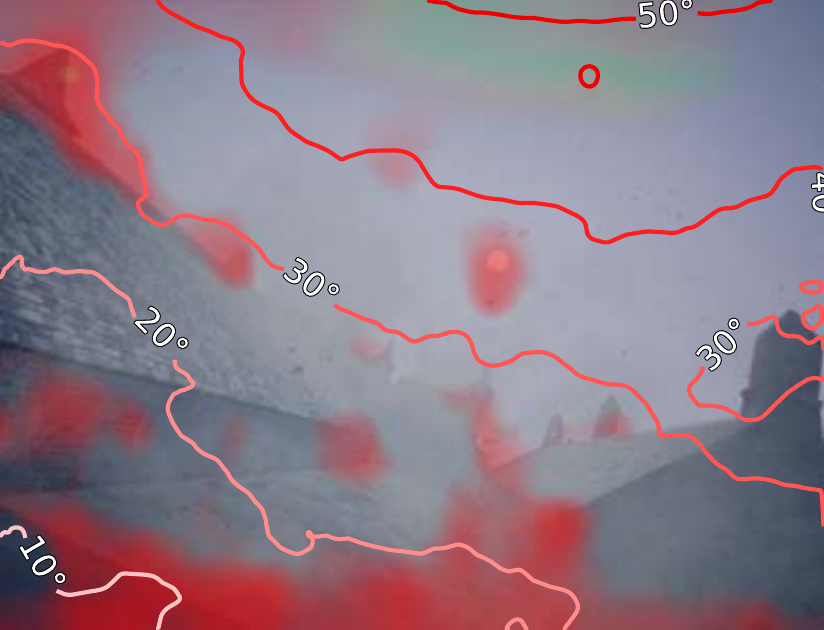}%
    \hspace{\pwidth}%
    \includegraphics[width=\iwidth]{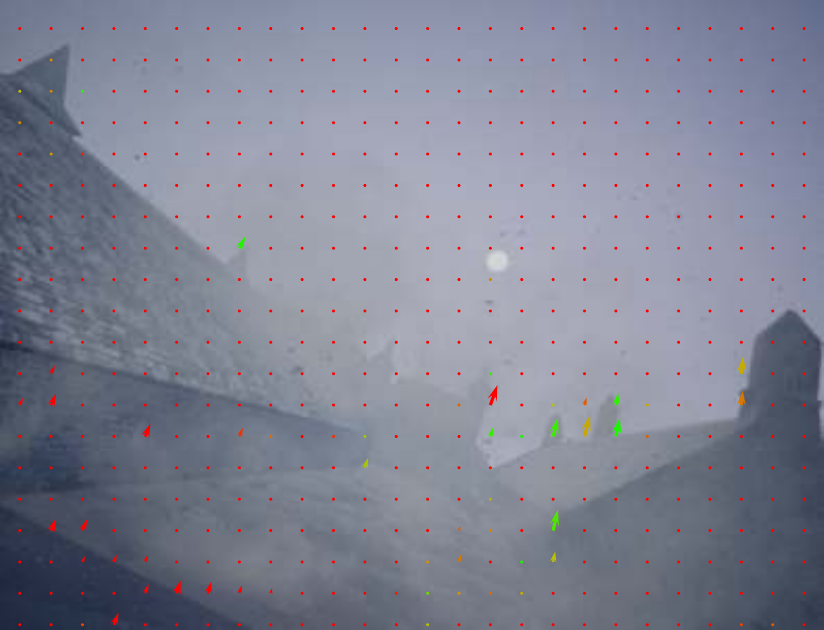}%
    
    \includegraphics[width=\lamarwidth]{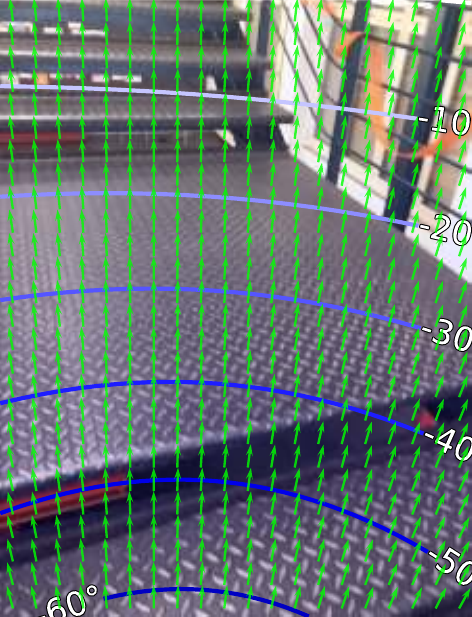}%
    \hspace{\pwidth}%
    \includegraphics[width=\lamarwidth]{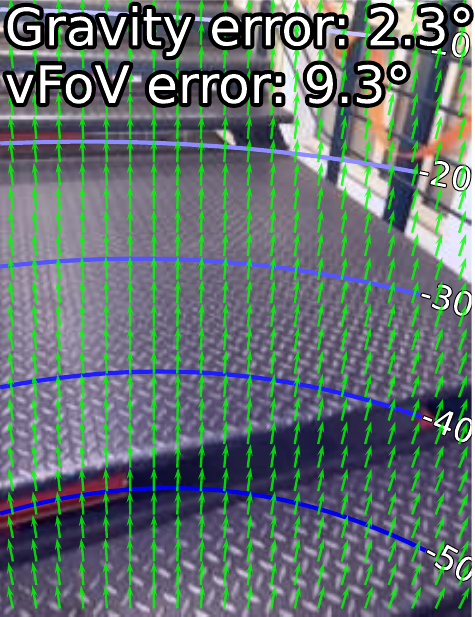}%
    \hspace{\pwidth}%
    \includegraphics[width=\lamarwidth]{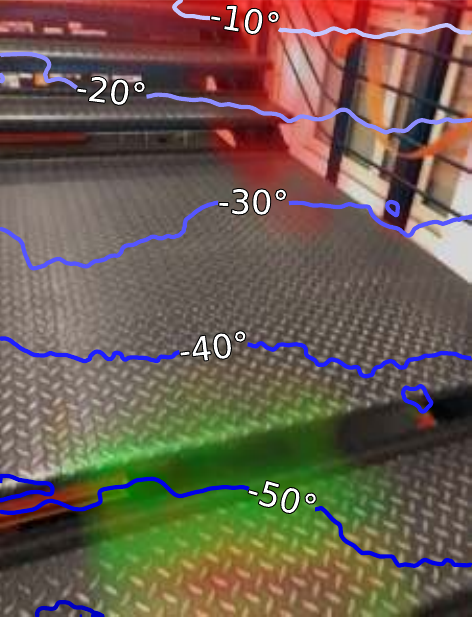}%
    \hspace{\pwidth}%
    \includegraphics[width=\lamarwidth]{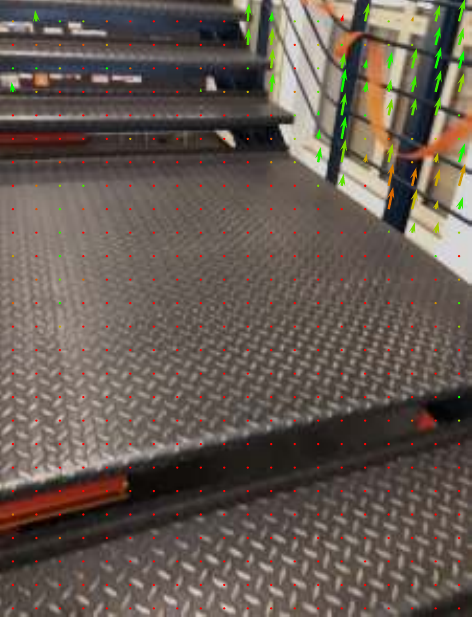}%
    \hspace{\pwidth}%
    \includegraphics[width=\lamarwidth]{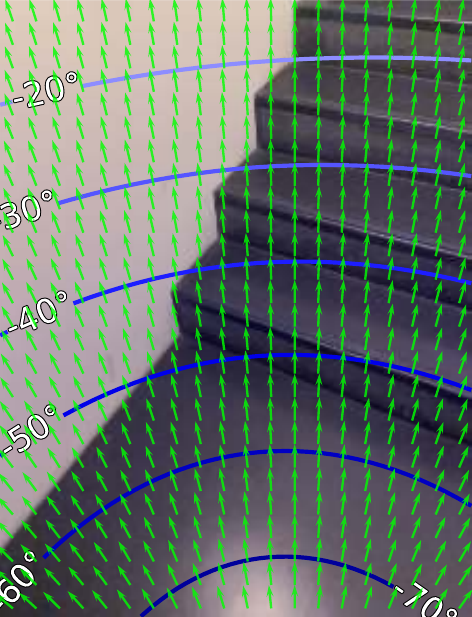}%
    \hspace{\pwidth}%
    \includegraphics[width=\lamarwidth]{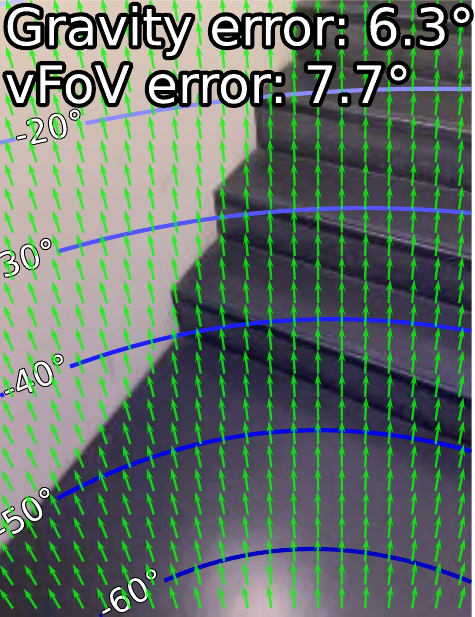}%
    \hspace{\pwidth}%
    \includegraphics[width=\lamarwidth]{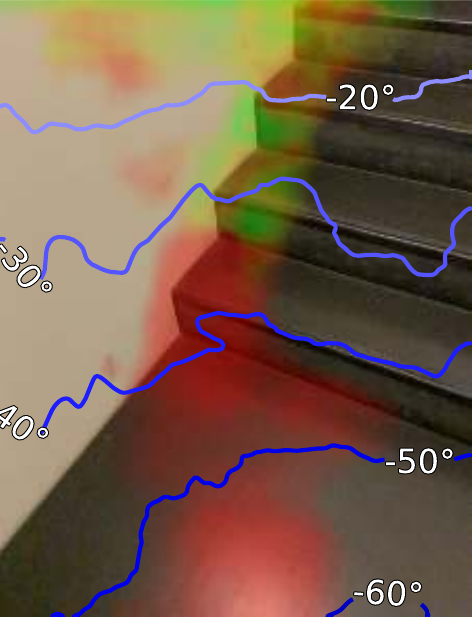}%
    \hspace{\pwidth}%
    \includegraphics[width=\lamarwidth]{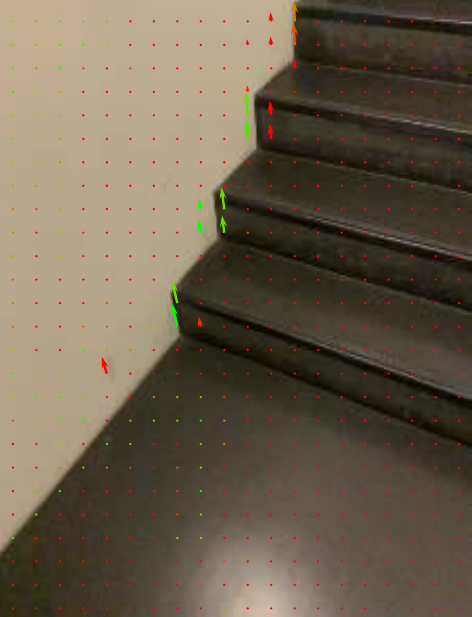}%

    \caption{\textbf{Failure Cases.}
        \ours~struggles on some challenging scenes, might miss important cues and is unable to leverage horizontal lines.
    }%
    \label{fig:failure}%
\end{figure}
\fi

\clearpage
\bibliographystyle{splncs04}
\bibliography{ms}
\end{document}